# Probabilistic Similarity Networks

**GREAT THINKERS**
FR LARRY GIBBS AND SISTER MARY-CATHERINE DEIBEL, EDITORS

*GODEL'S GODHEAD,*
ANNE VON DER LIETH GARDNER, 1987

*MAGNUM OPUS: THE DIGITIZATION OF GOD*
BELL D. CLAPPER, 1991

# Probabilistic Similarity Networks

David E. Heckerman

To Susan

## Contents

## List of Figures

**List of Tables**

# Foreword

For as long as researchers on artificial intelligence in medicine (AIM) have been working to develop high-performance decision-support tools for clinicians, they have been struggling with the question of how best to handle the uncertainty that is inherent in medical diagnosis and therapy planning. In the early 1970s, when the AIM field was getting started, people considered, discussed, and even tried probability theory, but they tended to abandon it because of four major limitations to applying it formally:

1. Inaccuracies due to the perceived practical need to assume conditional independence
2. Practical difficulties with the assessment of large numbers of conditional probabilities
3. Cognitive complexity in modifying or updating large tables of conditional probabilities and their interrelationships, once the numbers had been successfully gathered
4. Computational complexity that resulted if rigorous probabilistic approaches were attempted

Thus, in the early systems, uncertainty was handled by a variety of ad hoc models, of which the certainty-factor model of Mycin (Shortliffe and Buchanan, 1975), the evoking-strength/frequency-weight model of Internist-1 (Miller et al., 1982), and the causal-weighting model of Casnet (Weiss et al., 1978) are perhaps the best known. The uncertain-reasoning issues addressed in such experimental programs were in no way specific to medicine, however. The evolving expert-systems field has consistently found that uncertainty management is a major problem in diverse domains. For example, the Prospector system for geological exploration, an early expert system developed in the 1970s, used a subjective Bayesian model to inspire its inference-network approach to uncertainty management (Duda et al., 1976), but the actual implementation departed sufficiently from classical probability theory that Prospector also can be viewed as an ad hoc adaptation.

In recent years, there has been a resurgence of interest in the use of more formal probabilistic methods to handle uncertainty in large artificial-intelligence (AI) systems. Investigators have concentrated on knowledge acquisition and on Bayesian inference using the belief network (also called knowledge map), which is a graphical representation of uncertain knowledge based on probability theory. What had bothered me about most of that work was its theoretical orientation; no one had built a nontrivial, efficient, and effective system using formal probabilistic methods. Until such a system had been validated, I thought that the concerns of the AI researchers of the 1970s would not have been addressed.

In our laboratory, David Heckerman was the first investigator to question the assumptions that had led AI researchers to develop the ad hoc models (Heckerman, 1985). Heckerman's analyses provided insights into major limitations of these models, encouraging him and other researchers to explore how normative theory could be applied practically within the expert-systems paradigm. Heckerman wished to advance the theory of belief networks, and to show that belief networks could be used to build a large, real-world system that was effective in its decision task. Furthermore, he set out to demonstrate the validity of the approach by undertaking a formal evaluation. Heckerman has accomplished these tasks admirably.

Working with colleagues at the University of Southern California (Bharat Nathwani and Keung-Chi Ng) and at Stanford (Eric Horvitz and Larry Fagan), Heckerman created a medical expert system that used a probabilistic model for its management of uncertainty. This system, known as Pathfinder, assisted pathologists with the interpretation of histologic sections of human tissues—initially, from the lymph nodes. Reasoning had to account for more than 60 diagnoses and more than 100 descriptive findings; workers in the 1970s would not have used a probabilistic technique to handle a problem of this size and complexity. Heckerman demonstrated that, using a probabilistic framework to elicit and encode the knowledge of domain experts, he could construct a useful system.

The large domain of lymph-node pathology helped to motivate Heckerman to develop graphical extensions of the belief network representation that would facilitate the assessment of an immense number of probabilities. One representation, called a similarity network, permits the incremental construction of extremely large belief networks from cognitively manageable subproblems that involve the comparison of two diseases and their distinguishing features. Another representation, called a partition, simplifies the assessment of the conditional probabilities within a belief network. This approach to knowledge acquisition has now been tested extensively by Heckerman and his colleagues in the pathology domain, and has been shown both to be acceptable to experts and to form the basis for high-quality diagnostic advice.

As the book demonstrates, Heckerman is a superb and creative scientist. He brings an unusual educational background to his work. He holds a B.S. in mathematics and physics and a M.S. in physics from UCLA. He decided to enter our training program in medical informatics after he began his training at Stanford Medical School, combining his M.D. and Ph.D studies.

Heckerman's dissertation reflects an impressive medley of theoretical development, practical application, and formal evaluation. I therefore successfully recommended it to our Department of Computer Science for consideration as one of Stanford's two nominations for the Association for Computing Machinery's annual dissertation award. I was delighted when Heckerman was selected as one of the two winners in the national competition for 1990. We in the medical informatics program at Stanford are proud of Heckerman's work and this recognition of its excellence by the computer-science community.

Although Heckerman's work was originally conceived to address specific medical problems, the underlying theory and its formal presentation are clearly domain independent and have great potential for broad applicability. In fact, partitions recently have been used to develop normative expert systems for jet-engine repair and for the diagnosis of efficiency problems in gas turbines that generate electric power. Heckerman's work shows that, by exploiting graphical representations of independence, we can make probability theory a practical tool for managing uncertainty. Heckerman's work has convinced me that there are practical methods for using formal probabilistic approaches in expert systems. He has built an impressive diagnostic tool—Pathfinder—and has rigorously demonstrated its incremental value over earlier approaches. I believe it is a rare dissertation indeed that handles the three areas of theory, implementation, and formal evaluation, as well as Heckerman's does. I therefore commend this volume to you, and await with enthusiasm further explorations of both the theory and the application that you will learn about in these pages.

Edward H. Shortliffe
Stanford University
July, 1991

# Preface

This work describes a new generation of expert systems—called *normative expert systems.* These systems have the potential to provide better decision support than do traditional expert systems in domains where the accurate management of uncertainty is important.

This potential for improvement arises because people, including experts, make mistakes when they make decisions under uncertainty. That is, people often deviate from the rules of decision theory, which provides a set of rational principles or gold standards for how people should behave when reasoning or making decisions under uncertainty. Decision theory includes the rules of probability and the principle that a person should always choose the alternative that maximizes his expected utility.

Traditional expert systems provide decision support by mimicking the recommendations of experts. They do so by managing uncertainty with heuristic or ad hoc methods. Such systems are valuable, because they provide important information to a nonexpert who is confronted with a confusing decision, and because they offer reminders to users who may be stressed or fatigued. Nonetheless, they tend to duplicate the errors made by experts.

In contrast, normative expert systems use decision theory to manage uncertainty. The word "normative" comes from decision analysts and cognitive psychologists who emphasize the importance of distinguishing between *normative behavior*, which is what we do when we follow the gold standards of decision theory, and *descriptive behavior*, which is what we do when unaided by these gold standards. By encoding expert knowledge in a decision-theoretic framework, we can reduce errors in reasoning, and thereby build expert systems that offer recommendations of higher quality.

Normative expert systems have not become commonplace because they have been difficult to build and use. Over the past decade, however, researchers have developed the *influence diagram*, a graphical representation of a decision maker's beliefs, alternatives, and preferences that serves as the knowledge base of a normative expert system. Most people who have seen the representation find it intuitive and easy to use. Consequently, the influence diagram has overcome significantly the barriers to constructing normative expert systems.

Nevertheless, building influence diagrams is not practical for extremely large and complex domains. In this book, I address the difficulties associated with the construction of the probabilistic portion of an influence diagram, called a *knowledge map* or *belief network.* I introduce two representations that facilitate the generation of large knowledge maps. In particular, I introduce the *similarity network,* a tool for building the network structure of a knowledge map, and the *partition,* a tool for assessing the probabilities associated with a knowledge map.

The knowledge map, similarity network, and partition represent graphically a person's judgments about the independence of events. Each of these representations exploit the phenomenon that people can make judgments about the independence of events more easily than they can quantify with probabilities their beliefs that those events will occur. The similarity network and partition, however, can represent more judgements of independence than can a knowledge map. Therefore, experts find it easier to construct knowledge bases using these representations than they do using knowledge maps alone.

The similarity-network and partition representations aided considerably the construction of Pathfinder, a large normative expert system for the diagnosis of lymph-node diseases (the domain contains over 60 diseases and over 100 disease findings). In an early version of the system, I encoded the knowledge of the expert using an erroneous assumption that all disease findings were independent, given each disease. When the expert and I attempted to build a knowledge map for the domain to capture the dependencies among the disease findings, we failed. Using a similarity network, however, we were able to build the knowledge map for the entire domain in approximately 40 hours. Furthermore, the partition representation reduced the number of probability assessments required by the expert from 75,000 to 14,000. Most important, through a comparison procedure based in decision theory, I found that the improvements in diagnostic accuracy afforded by the more sophisticated model of the domain were well worth the additional effort that we had invested in building the revised version of the system.

In this book, I examine in detail the theoretical properties of similarity networks and partitions, and discuss the application of these representations to the construction of Pathfinder. This work suggests strongly that, by identifying specific forms of conditional independence, and by developing representations that exploit these forms of independence for knowledge acquisition, knowledge engineers can construct normative expert systems for domains of larger scope and greater complexity than the domains previously thought to be amenable to the decision-theoretic approach.

David Heckerman
Univeristy of Southern California
July, 1991

# A Guide for the Reader

This book has been written for readers from backgrounds in various areas, including artificial intelligence, decision analysis, and medical informatics. Chapters 1, 2, and 6 contain the fundamental ideas regarding similarity networks and partitions, and should be read by everyone.

Appendix A contains a discussion of basic concepts from decision theory and a tutorial on knowledge maps and influence diagrams. People should read this appendix before reading the main body of the book if they are unfamiliar with the concept of the joint probability distribution, the principle of maximum expected utility, or the distinctions between Bayesian and frequentist philosophies, between normative and descriptive reasoning, or between decision theory and decision analysis.

Chapter 3 contains a detailed axiomatic characterization of the similarity-network representation. Those readers who are mainly interested in an intuitive understanding of the representations may skip this chapter. Readers who are technically inclined should note that all the major results, and the arguments for those results, are contained in Chapter 3. The more complicated proofs are given in Appendix B.

Chapters 4 and 5 describe the construction and evaluation of Pathfinder. Researchers in the field of medical informatics and others who are interested in learning about the practical application of the similarity-network and partition representations to knowledge acquisition will find these chapters particularly relevant.

# Acknowledgments

My deepest thanks go to Ron Howard and Eric Horvitz. Ron Howard introduced me to the concepts of decision analysis, and has been my advisor both in and out of academics. Eric Horvitz has been my partner in academics as well as a good friend since I came to Stanford nearly a decade ago. I thank them both for innumerable interesting discussions, many of which have benefited this work.

In addition, I thank Ted Shortliffe and Peter Hart for encouraging me to pursue a theory that would have practical benefits, and for providing extremely useful feedback on earlier drafts of this work. Peter Hart also helped me to identify the most important lessons to be learned from this work.

Special thanks are due also to Bharat Nathwani, the Pathfinder expert. The theory of similarity networks grew out of the discussions that he and I had during the early phases of the construction of Pathfinder. He also devoted 4 months of painstaking work to the construction and evaluation of Pathfinder's decision-theoretic model.

Jack Breese, Dan Geiger, Ross Shachter, and Mike Wellman provided useful comments on the technical portions of this work. Most notable, Ross Shachter first suggested that ordinary local knowledge maps should be defined in terms of graph disconnection; and Dan Geiger provided invaluable assistance in the proof of Theorem 3.13.

Keung-Chi Ng, Jaap Suermondt, Marty Chavez, and Mark Fischinger helped me to program SimNet, an implementation of similarity networks and partitions on the Macintosh computer.

Several friends and colleagues assisted me with the medical portions of this book. Harold Lehmann and David Campell provided the medical examples that illustrate many of the ideas in this work; Doyen Nguyen served as the community pathologist in the formal evaluation of Pathfinder; Larry Diamond also assisted with the evaluation; and Majorie Bernstein-Singer helped Bharat Nathwani and me to build the Pathfinder utility model.

This research also benefited from discussions with Jeff Bradshaw, Mike Hendrickson, David Matheson, Jim Smith, Pat Suppes, and the Bayesian mafia at Stanford: Ingo Beinlich, Marty Chavez, Greg Cooper, Steve Downs, Brad Farr, Eddie Herskovits, Holly Jimison, Harold Lehmann, Curt Langlotz, Geoff Rutledge, and Jaap Suermondt.

Lyn Dupré meticulously edited the entire manuscript. In the process, she managed to teach me how to communicate effectively and to enjoy writing.

The National Cancer Institute, under Grant RO1CA51729-01A1, the National Library of Medicine, under Grant RO1LM04529, and the National Science Foundation, under Grant IRI-8703710, provided the financial support that made this book possible.

Finally, I especially want to thank my mom and dad for a lifetime of love, care, and support.

# 1 Introduction

*One is almost tempted to say that quite apart from its intellectual mission, theory is the most practical thing imaginable....*

—*Ludwig Boltzmann* (Broda, 1983, page 104)

Over the last 2 decades, decision analysts have been using decision theory in conjunction with a collection of knowledge representations and heuristic techniques to provide clarity of action to individuals and groups who are confused about important decisions. Decision theory includes probability theory (sometimes referred to as subjective probability theory or Bayesian probability theory) and the maximum expected utility principle, which states that a decision maker should choose the alternative that maximizes his expected utility. Perhaps the most significant virtue of decision theory is that it is *normative*. That is, the theory provides a set of gold standards for how people wish they could behave when allocating scarce resources under uncertainty. It is well known that people often do not behave in accordance with these gold standards (Edwards, ed., 1956; Tversky and Kahneman, 1974; Kahneman et al., 1982). Thus, a decision analysis can save millions of dollars or even lives and suffering when the stakes are high.

For nearly the same period of time, knowledge engineers have been using a set of representations and heuristic techniques developed by researchers in artificial intelligence to build *expert systems*: computer programs that bring to bear the knowledge of an expert or group of experts on a class of decisions or *domain* (e.g., Mycin and Prospector are expert systems for the diagnosis and treatment of bacterial infections, and for site selection for mineral exploration, respectively (Shortliffe, 1976; Duda et al., 1976)). Decision theory and decision-analytic techniques have rarely been used in the construction of such systems. Although some artificial-intelligence researchers have avoided a decision-analytic approach on theoretical grounds (Shafer, 1986; Zadeh, 1986), most workers have turned to alternative approaches for decision making because they believe that the normative approach to constructing expert systems for large, real-world domains is impractical (Shortliffe and Buchanan, 1975; Rich, 1983, pages 184–199).

Over the last 6 to 7 years, several researchers working at the boundary of decision analysis and artificial intelligence have been attempting to build *normative expert systems*: expert systems that use a decision-theoretic model as the framework for knowledge representation and inference. From the perspective of decision analysis, such systems could provide decision assistance across a wide range of possible decisions in a given domain. By avoiding the expense of analysts and experts for every confusing and high-stakes decision to be made, a normative expert system could reduce significantly the costs of decision making. From the perspective of artificial intelligence, use of a normative theory as the framework for representing knowledge could improve dramatically the quality of expert knowledge that is delivered to the user of an expert system.[1]

A major breakthrough on the path toward the creation of normative expert systems has been the development of the *influence diagram*, a representation that graphically represents the beliefs, alternatives, and preferences of a decision maker (Howard and Matheson, 1981).[2] The influence diagram is a natural representation for the knowledge base of an expert system. The representation is mathematically precise, yet has a human-oriented qualitative structure that facilitates communication between the expert and a decision model. Moreover, influence diagrams can represent any decision problem.

The influence-diagram representation facilitates the three major facets of expert system development: *knowledge acquisition*, the process of capturing and encoding the knowledge of an expert or experts; *inference*, the generation of recommendations or relevant information based on user input and the expert knowledge; and *explanation*, the process of communicating such recommendations or relevant information to the user. Influence diagrams simplify knowledge acquisition, because we can use them to represent graphically assertions of conditional dependence and independence *before* we need to consider assessments of probabilities or utilities. We can use these assertions of conditional independence to decompose the assessment of a joint probability distribution into a collection of independent assessments of manageable size. Such decomposition helps us to focus attention during knowledge acquisition, and to decrease the size of the construction task. In addition, we can use the assertions of conditional independence in an influence diagram to increase the computational efficiency of decision-theoretic inference. Specifically, researchers have developed exact and approximate inference algorithms that exploit assertions of conditional independence in an influence diagram to avoid direct computations on the joint probability distribution associated with that diagram (Shachter, 1986; Henrion, 1986; Pearl, 1988; Lauritzen and Spiegelhalter, 1988; Cooper, 1990a). Furthermore, we can use the graphical representation of conditional independence to generate cogent explanations to the builders and users of normative expert systems (Pearl, 1988, Chapters 5 and 10). Given these features of the influence-diagram representation, it is not surprising that several normative expert systems have been constructed using the representation (Spiegelhalter and Knill-Jones, 1984; Cooper, 1984; Reggia and Perricone, 1985; Henrion and Cooley, 1987; Olesen et al., 1989; Beinlich et al., 1989).

[1] See Appendix A for a discussion of this point.

[2] Readers unfamiliar with the influence-diagram representation should read Appendix A before reading the body of this book.

In this book, I address the pragmatic aspects of capturing and representing knowledge for normative expert systems. *I show that by identifying specific forms of conditional independence, and by developing representations that exploit these forms of independence for knowledge acquisition, knowledge engineers can construct normative expert systems for domains of larger scope and greater complexity than the domains previously thought to be amenable to the decision-theoretic approach.* In particular, I introduce two graphical extensions to the influence-diagram representation called *similarity networks* and *partitions*. A similarity network is a tool for constructing an influence diagram, whereas a partition is a tool for assessing the probabilities associated with an influence diagram.[3] Both representations facilitate the development of large and complex models by exploiting forms of conditional independence that are not easily represented in an ordinary influence diagram. In this book, I scrutinize these representations and the forms of conditional independence that they embody, and show how their use has made practical the construction of a normative expert system for medicine.

## 1.1 Pathfinder: A Normative Expert System

A normative expert system called Pathfinder provided the primary motivation for the development of the similarity-network and partition representations. Pathfinder assists surgical pathologists with the diagnosis of lymph-node disease (Heckerman et al., 1985; Heckerman et al., 1989b; Heckerman et al., 1990).

The role of the surgical pathologist in lymph-node diagnosis is shown in Figure 1.1. If a patient's physician suspects that a disease process has involved the lymph nodes of his patient, he may remove one or more those nodes for diagnosis. The surgical pathologist examines these samples of the patient's tissue microscopically. Sometimes, the pathologist also incorporates clinical, radiology, and laboratory information, and examines the nodes with expensive tests derived from immunology, microbiology, and cell kinetics research. Based on this examination, the pathologist provides a diagnosis to the patient's physician. That is, the pathologist tells the physician, "the patient has disease $x$." Given this diagnosis, the patient's physician then treats the patient.

The well-being of patients depends strongly on the accuracy of the pathologist's diagnosis. In an extreme case, for example, let us suppose that the patient has Hodgkin's disease, a malignant disease, but that the pathologist makes a diagnosis of mononucleosis, a benign disease that can resemble Hodgkin's disease. In this situation, the patient's chance of death is significantly greater than it would have been had the diagnosis been correct, because he does not receive immediate treatment for his malignancy. In contrast, let us suppose that the patient has mononucleosis, and that the pathologist makes a diagnosis of Hodgkin's disease. In this case, the patient will undergo expensive, painful, and debilitating treatment, to be "cured," only because he never had the malignant disease in the first place.

[3]Some authors use the term influence diagram to refer to both the network representation and the probabilities that underlie the network. In this book, however, I use the term to refer to only the network.

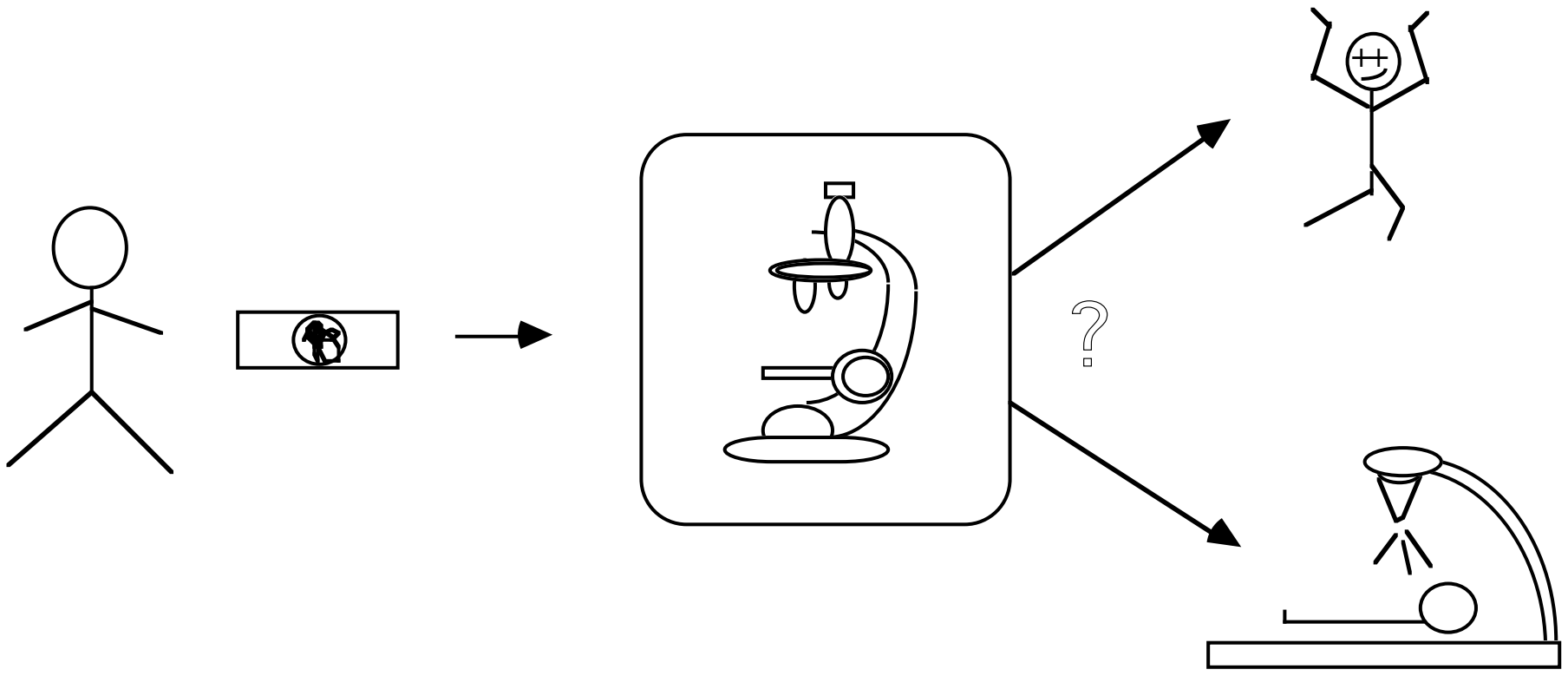

**Figure 1.1:** The role of the pathologist.
Using a microscope and other, more expensive, tests derived from immunology, microbiology, and cell kinetics research, a pathologist examines tissue removed from a patient. Based on this examination and other observations including clinical, radiology, and laboratory information, the pathologist provides a diagnosis to the patient's physician, who then treats the patient in accordance with the diagnosis. The outcome of treatment can be extremely sensitive to the accuracy of the diagnosis rendered by the pathologist.

Unfortunately, pathologists who do not specialize in one or a few types of tissue—the majority of pathologists—often make errors in diagnosis. They are especially likely to make errors when the tissue being examined is from the lymph node (Byrne, 1977; Jones et al., 1977; Coltman et al., 1980; Kim et al., 1982). Several cooperative oncology studies have documented that, although experts in lymph-node pathology show agreement with one another, the diagnoses rendered by a nonspecialist disagree with those made by experts as much as 50 percent of the time (Velez-Garcia et al., 1983).

In summary, the stakes associated with lymph-node diagnosis are high, and there is a significant difference between the accuracy of the nonspecialists and specialists in the field. A normative expert system for this domain is therefore likely to be of benefit to pathologists.

The domain of lymph-node pathology is also an excellent testbed in which to investigate practical issues concerning the construction of normative expert systems. The domain is large by any standard of comparison for expert systems. Over 60 diseases can invade the lymph node (25 benign diseases, 9 Hodgkin's lymphomas, 18 non-Hodgkin's lymphomas, and 10 metastatic diseases). In addition, there are approximately 100 morphologic distinctions or *features* within lymph nodes that can be recognized easily on microscopic examination. Each feature is associated with two or more mutually exclusive and exhaustive *instances*. Also, Pathfinder contains features reflecting clinical, laboratory, immunological, and molecular biological information that is relevant to the diagnosis of lymph-node disease.

### 1.1.1 A Pathfinder Dialog

In rendering a diagnosis, a pathologists (1) identifies and quantifies features; (2) constructs a *differential diagnosis*, a set of diseases consistent with the observations; and (3) decides what additional features to evaluate and what costly tests to employ to narrow the differential diagnosis. He repeats these steps until he has observed all useful features. This procedure, used by diagnosticians in many medical and nonmedical domains, is called the *hypothetico-deductive* approach (Bartlett, 1958; Elstein et al., 1971; Elstein, 1976; Elstein et al., 1978).

Pathfinder uses this same method to assist pathologists with their task of diagnosis. Let us consider a sample dialog between the system and a user of the system illustrated in Figures 1.2 through 1.8. Figure 1.2 shows the initial Pathfinder screen. The FEATURE CATEGORY window displays the categories of features that are known to the system, the OBSERVED FEATURES window displays feature–instance pairs that have been observed by the pathologist, and the DIFFERENTIAL DIAGNOSIS window displays the list of possible diseases and their probabilities. The probabilities in Figure 1.2 are the prior probabilities of disease—the probabilities for disease given only that a patient's node has been removed and is being examined.

If the user selects (double-clicks) the feature category SPHERICAL FEATURES, then Pathfinder displays a list of features for that category, as shown in Figure 1.3. To enter a particular feature, the user double-clicks on that feature, and then selects one of the mutually exclusive and exhaustive instances for that feature. For example, Figure 1.4 shows what happens when the user selects the feature F % AREA (percent area of the lymph-node section that is occupied by follicles). In the figure, a third window appears that lists the instances for this feature: NA (not applicable), 1–10%, 11–50%, 51–75%, 76–90%, and >90%. Figure 1.5 shows the result of selecting the last instance for this feature. In particular, the feature–instance F % AREA: >90% appears in the middle column, and the differential diagnosis is revised, based on this observation.

The user can continue to enter any number of features of his own selection. Figure 1.6 shows the Pathfinder screen after the user has reported that follicles are in a back-to-back arrangement and show prominent polarity. Alternatively, the user can ask the program to recommend additional features for observation. When such a request is made, the program first computes the *value of clairvoyance* for each feature that has not yet been reported to the system.[4] The system then subtracts the cost of observing a feature—including the monetary expense, time delay, and the tedium associated with its observation—from that feature's value of clairvoyance, and displays the most cost-effective features for evaluation. In this case, as shown in Figure 1.7, Pathfinder determines that MONOCYT (monocytoid cells) is the most useful feature for evaluation. Figure 1.8 shows the result of the user reporting that monocytoid cells are prominent. Specifically, the four features observed by the user have narrowed the differential diagnosis to a single disease: the early phase of AIDS.

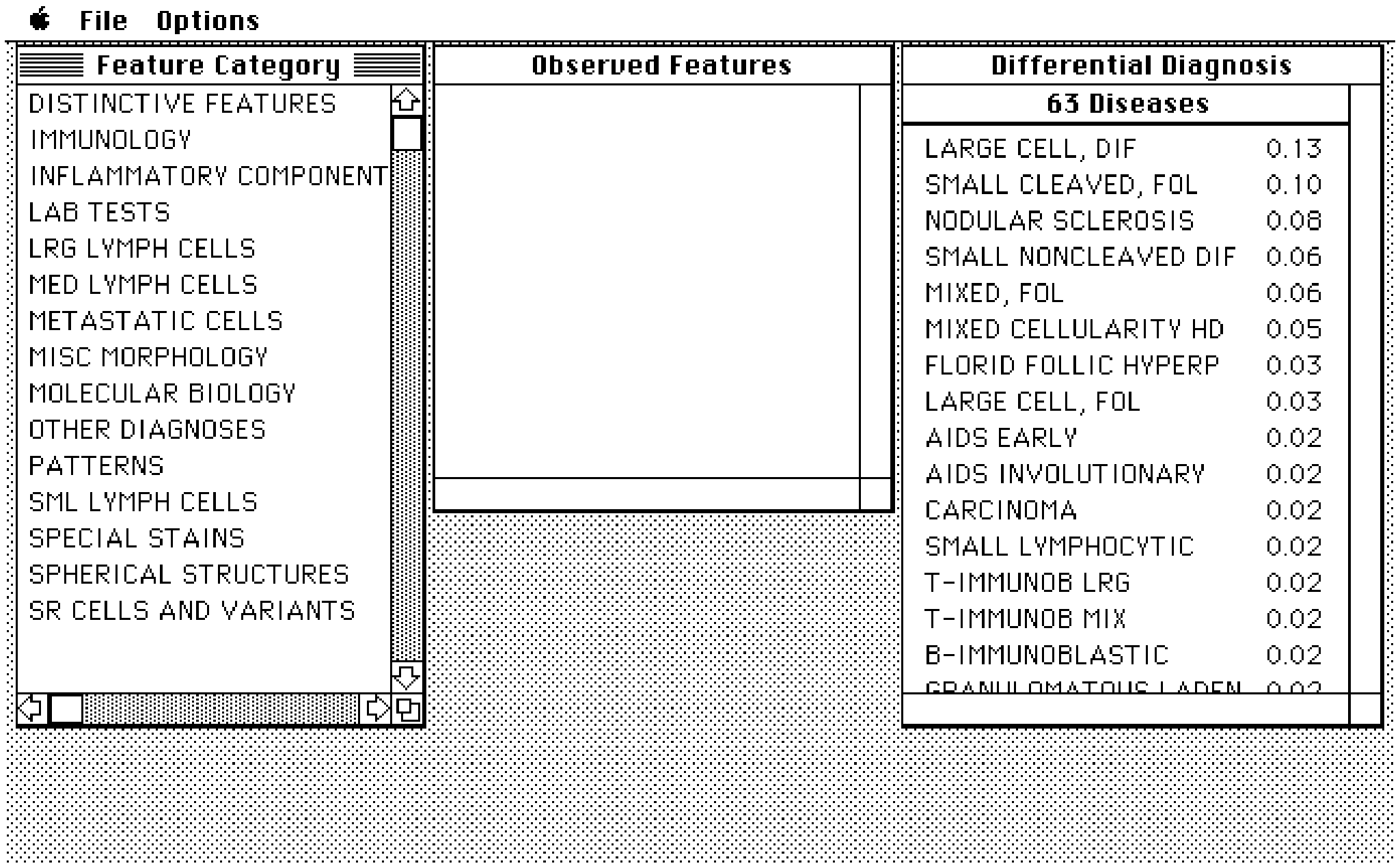

**Figure 1.2:** The Pathfinder interface.
The Macintosh windows display the categories of features, the observed features, and the differential diagnosis. In the differential diagnosis, diseases are ordered by their probabilities. The probability for a disease shown in this figure is the prior probability for that disease—the probability that the disease is present in a patient, given only that the patient's node has been removed and is being examined.

[4]Appendix A contains a general discussion of value-of-clairvoyance computations. Chapter 5 describes the specific decision model used for such computations in Pathfinder. In the computation itself, we assume that the decision maker's preferences approximately satisfy the delta property.

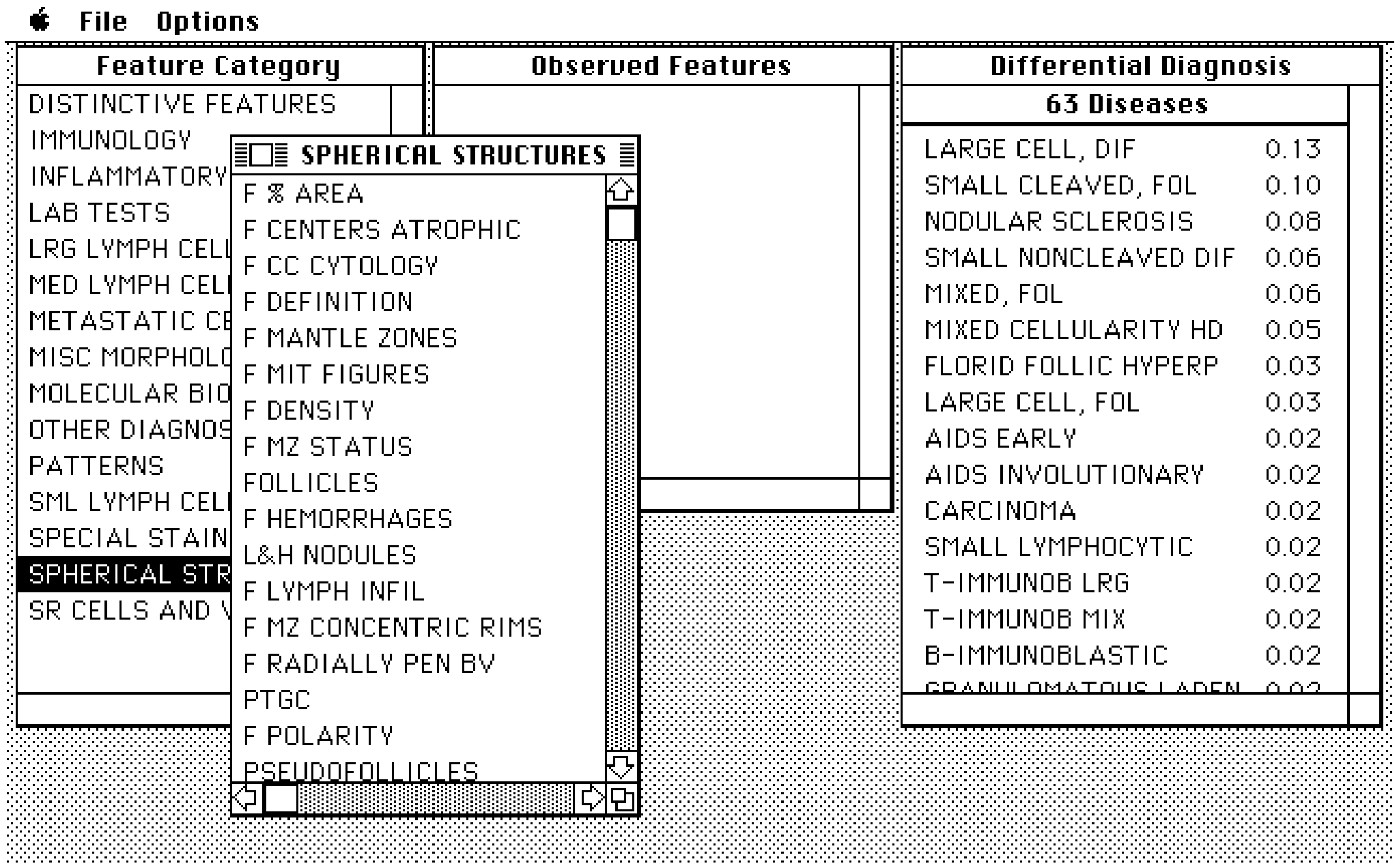

**Figure 1.3:** The selection of a feature category.
The user has selected (double-clicked) the category of features named SPHERICAL STRUCTURES. This action opens a window that contains the list of features for this category.

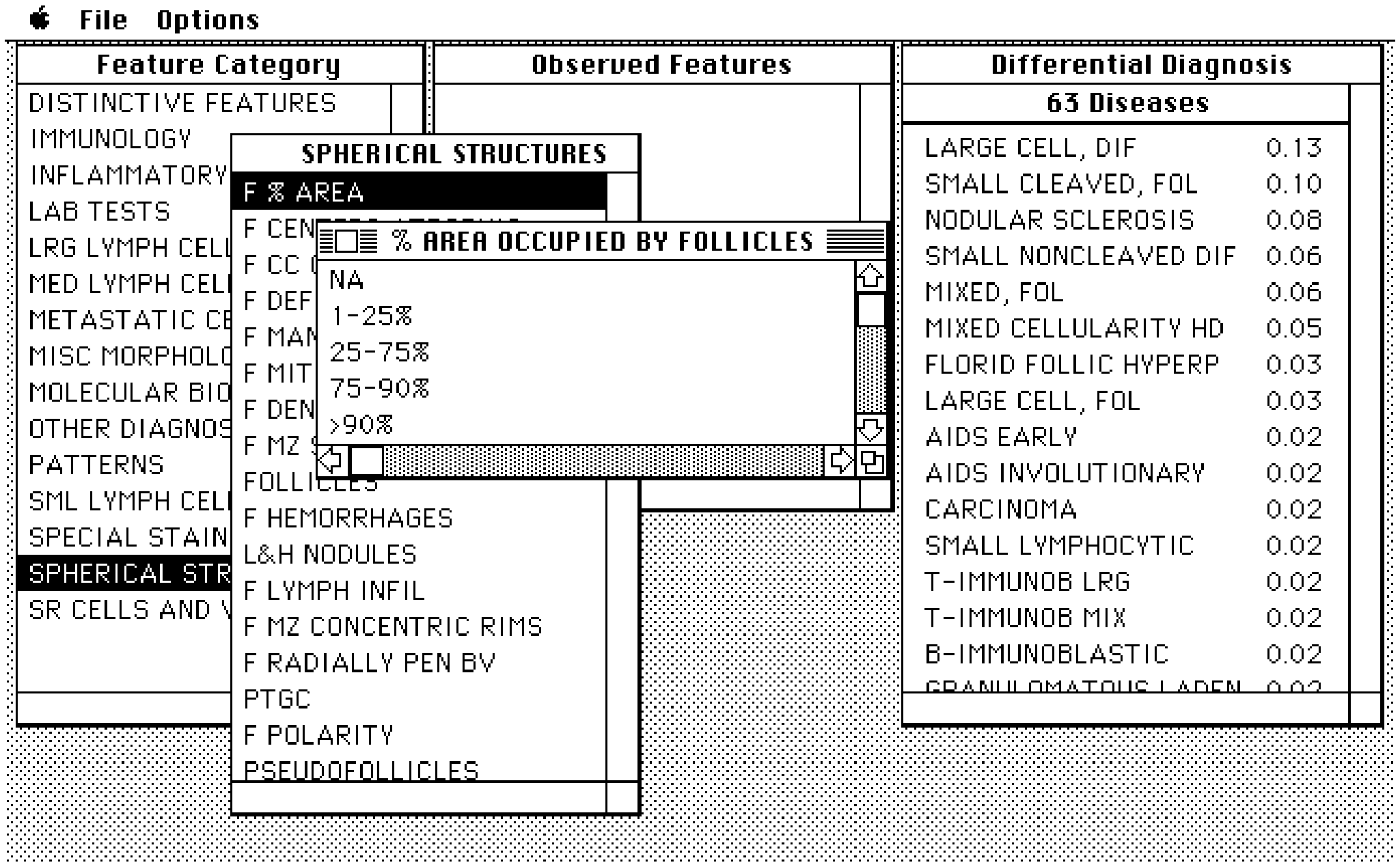

**Figure 1.4:** The selection of a feature.
The user has selected the feature F % AREA. This selection opens a window that contains an expanded version of the feature name and the list of the mutually exclusive and exhaustive instances for the feature.

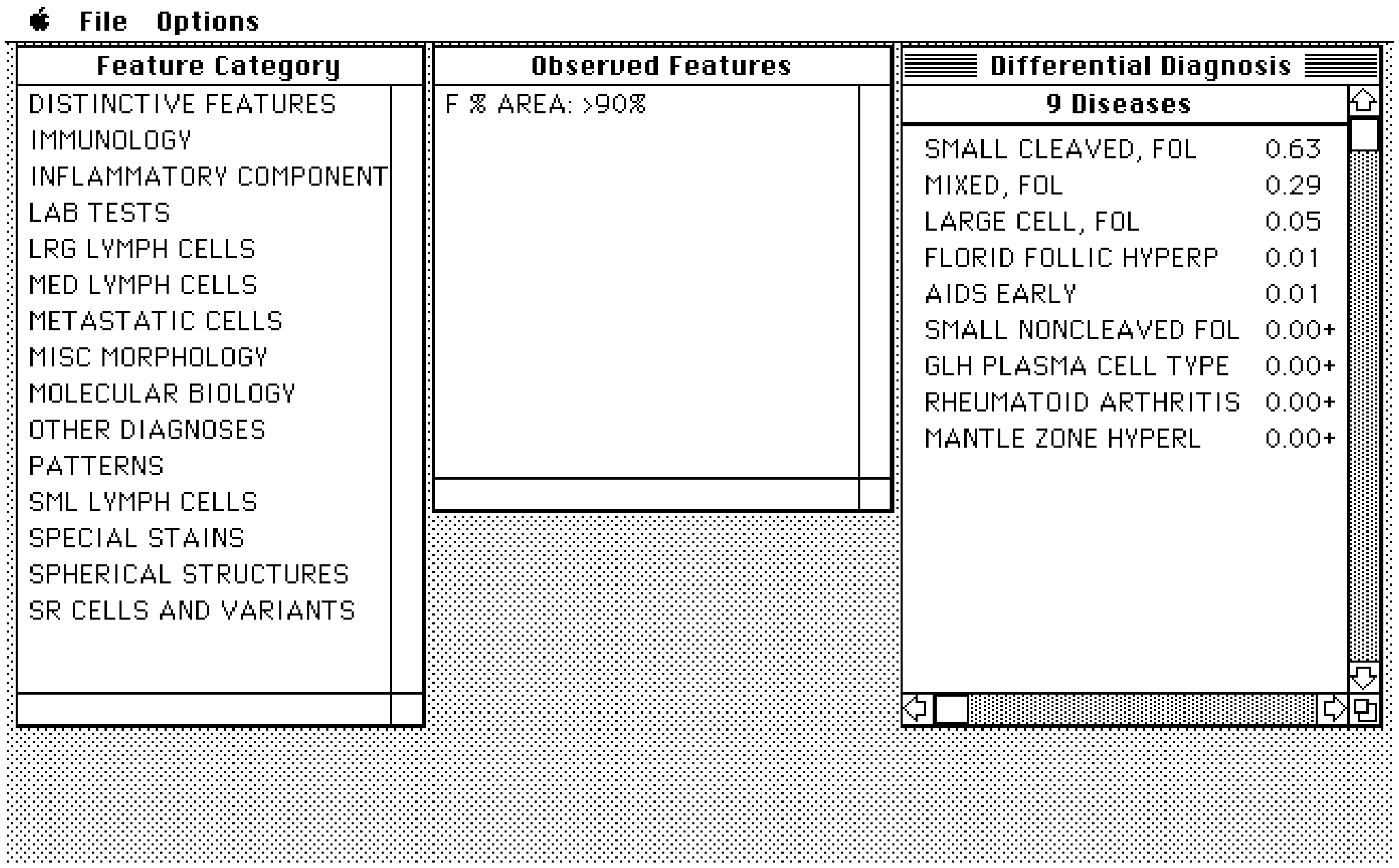

**Figure 1.5:** The entry of a feature–instance pair.
The entered feature–instance pair—F % AREA: >90%—appears in the middle window. Based on this piece of evidence, the program revises the differential diagnosis in the right-most window.

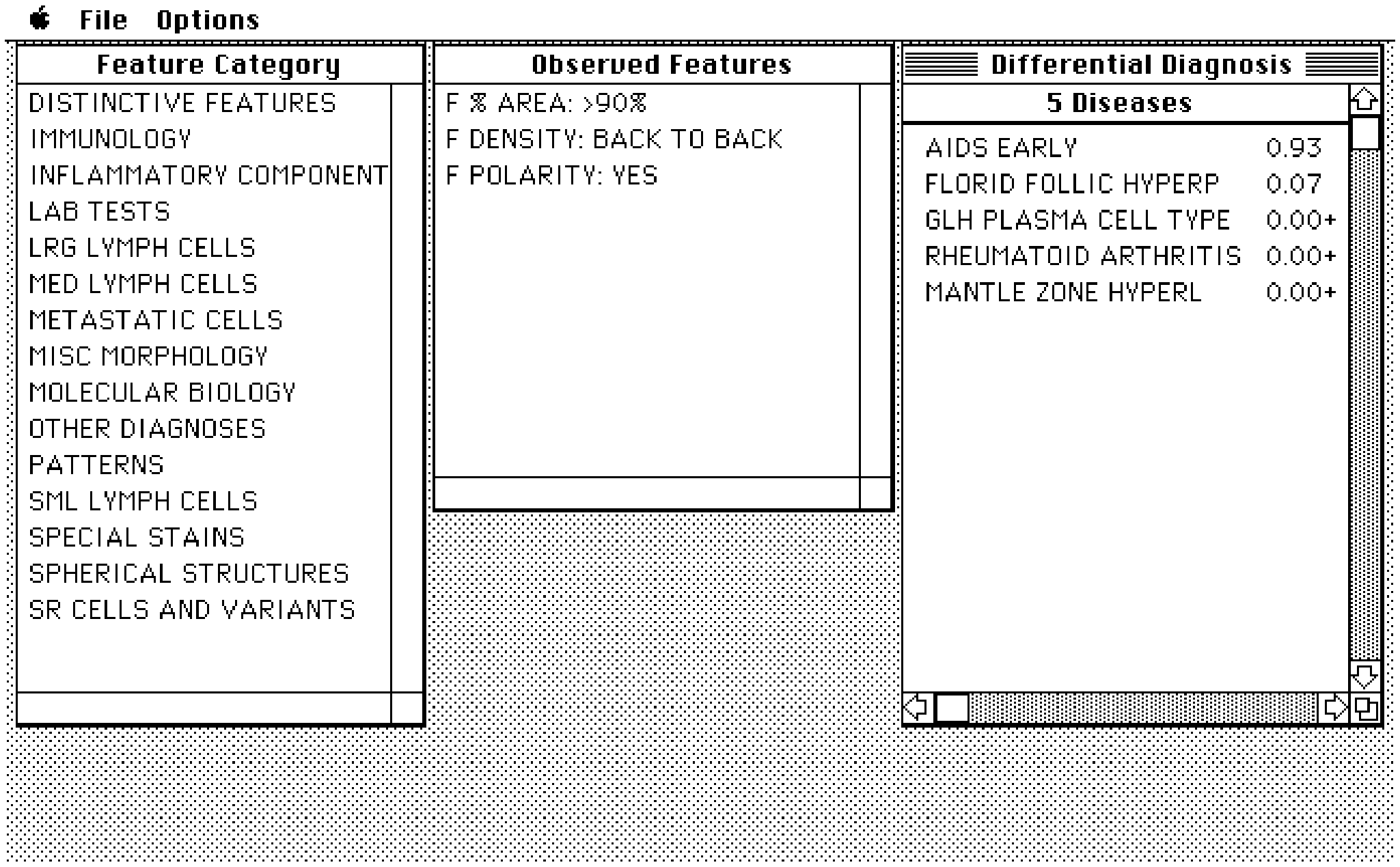

**Figure 1.6:** The entry of two additional feature–instance pairs.
The new feature–instances pairs—F DENSITY: BACK-TO-BACK and F POLARITY: PROMINENT—appear in the middle window. Based on these observations, the program again revises the differential diagnosis.

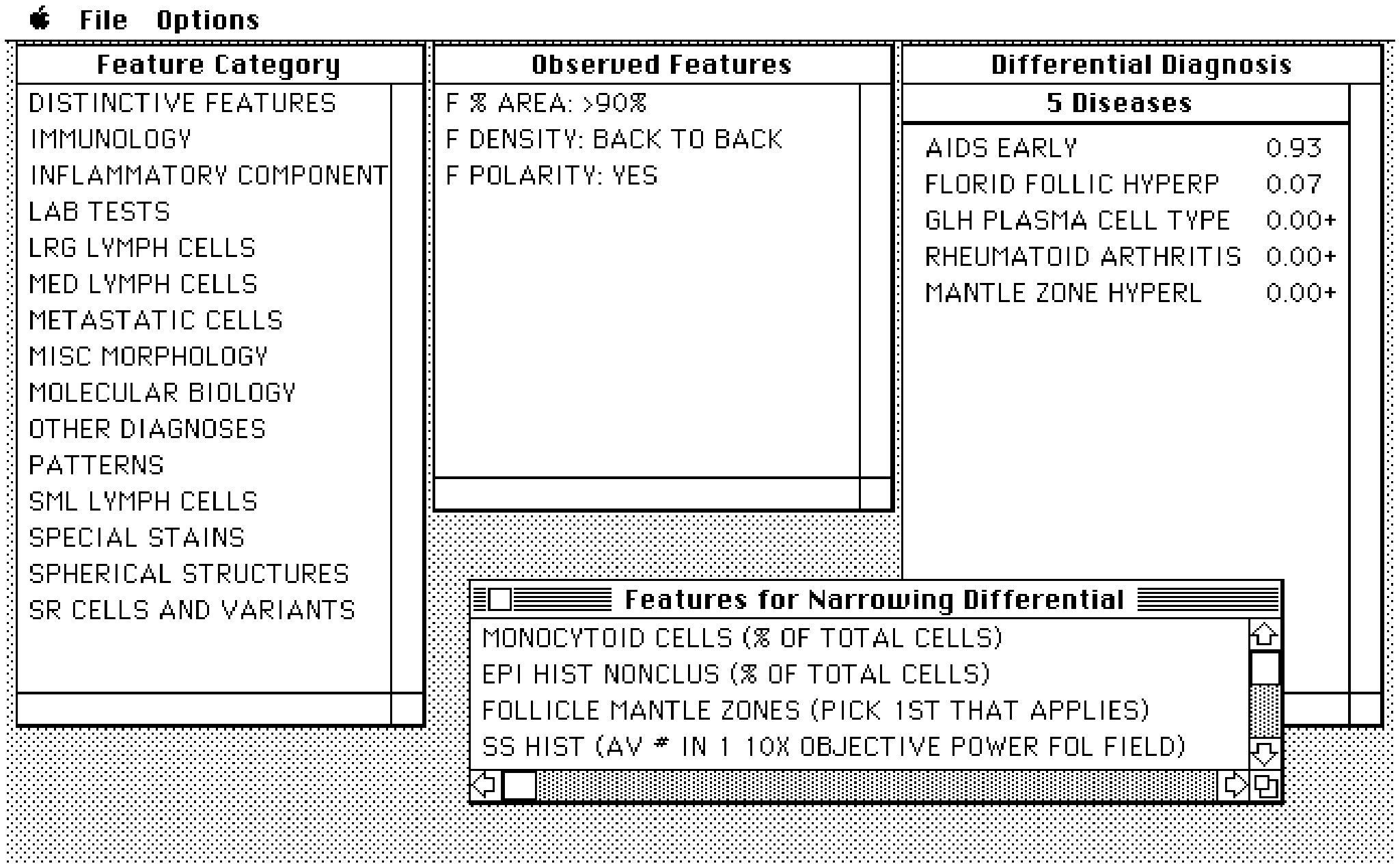

**Figure 1.7:** A request for features to evaluate.
The user has asked Pathfinder to display features that are useful for narrowing the differential diagnosis. The program displays the four most cost-effective features for the pathologist to observe next.

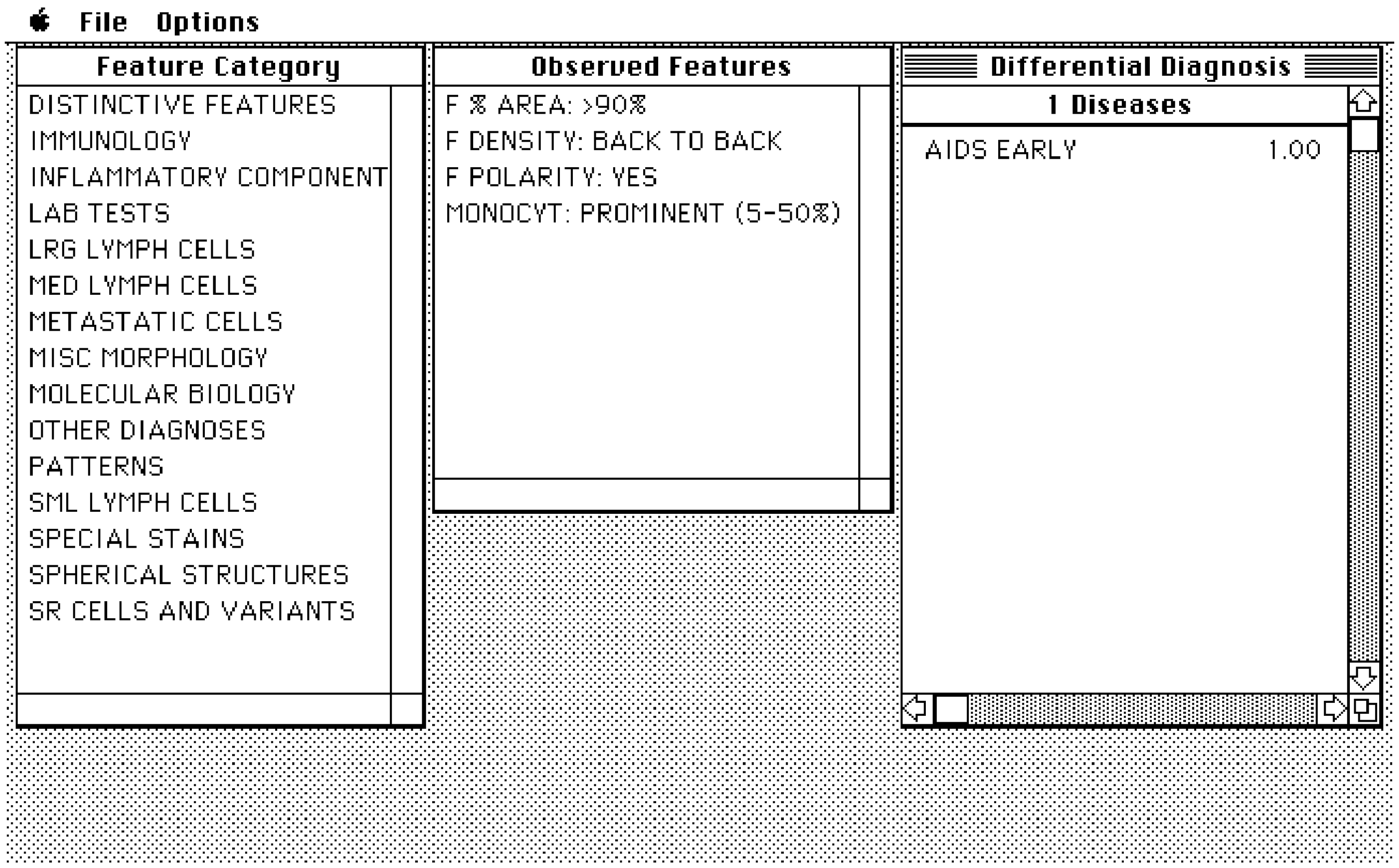

**Figure 1.8:** A diagnosis.
Pathfinder determines that only a single disease—AIDS EARLY (the early phase of AIDS)—is consistent with the four observations shown in the middle window.

Gorry and Barnett introduced the hypothetico-deductive approach to automated systems in 1968 under the name *sequential diagnosis*. The iterative strategy appears in many expert systems that are both normative (Gorry and Barnett, 1968; Gorry et al., 1973; Spiegelhalter and Knill-Jones, 1984; Henrion and Cooley, 1987; Olesen et al., 1989; Beinlich et al., 1989) and nonnormative (Ben-Bassat et al., 1980; Miller et al., 1982).

### 1.1.2 Diagnosis: A Decision

In the patient case that we have considered, Pathfinder can provide the pathologist with a diagnosis, because there is only one disease that is consistent with the patient's findings. The example, however, is atypical in that more than one disease usually remains on the final differential diagnosis. In these situations, Pathfinder would have to make a *decision*—that is, a choice under uncertainty—to render a diagnosis.

In principle, Pathfinder can make such decisions. The system contains a utility model that represents a typical patient's preference for outcomes associated with every combination of disease and diagnosis that can befall a patient. We discuss this model in detail in Chapter 5. Pathfinder uses the utility model in its value-of-clairvoyance computations, and it can also use the model for rendering diagnoses under uncertainty.

Nonetheless, preferences vary among patients. Furthermore, I have observed that, although recommendations for evidence gathering are not sensitive to the Pathfinder utility model, diagnostic recommendations are somewhat sensitive to the model. Consequently, I do not allow Pathfinder to make diagnoses under uncertainty. I hope that this policy will encourage a change in the way pathologists and care-providing physicians communicate. In the short term, I hope that pathologists will begin to express clearly—in the language of probability—uncertainty associated with their observations. In the long term, I hope that each physician who is associated with the care of a patient—including the primary physician, the pathologist, the radiologist, the surgeon, the oncologist, and the radiotherapist—and the patient himself will communicate in decision-theoretic terms to determine the best treatment for that patient. Such communication could take place via a shared decision model embodied in an expanded version of Pathfinder.[5]

### 1.1.3 A Problem with Knowledge Acquisition

Let us concentrate on Pathfinder's knowledge or probabilistic model. The Pathfinder project began in 1983. Early versions of Pathfinder employed a probabilistic model represented by the influence diagram in Figure 1.9(a). Specifically, the expert on the project, Dr. Bharat Nathwani, and I assumed that diseases were mutually exclusive and exhaustive, and that all features were conditionally independent, given disease. The assumption that diseases were mutually exclusive was appropriate, because co-occurring diseases almost always appear in different lymph nodes or in different regions of the same lymph node. Also, the large scope of Pathfinder made reasonable the assumption that the set of diseases was exhaustive. The assumption of global conditional independence, however, was inaccurate. For example, given certain diseases, finding that follicles are abundant in the tissue section increases greatly the chances that sinuses in the interfollicular areas will be partially or completely destroyed. Thus, in 1986, the expert and I attempted to represent explicitly the dependencies among features in the lymph-node domain, by constructing an influence diagram of the form shown in Figure 1.9(b).

[5] For another discussion of this issue, see Shachter and Hendrickson (1990).

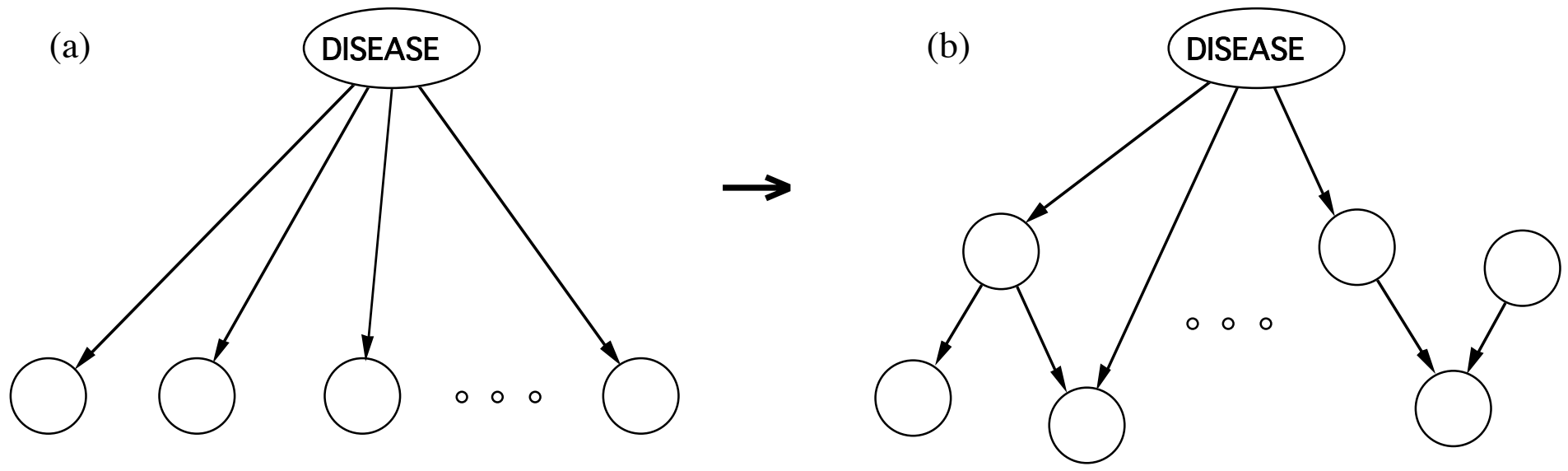

**Figure 1.9:** Two schematic influence diagrams for Pathfinder.
In both influence diagrams, the chance node DISEASE represents a set of mutually exclusive and exhaustive diseases. The features relevant to the diagnosis of disease are represented by the unlabeled chance nodes below DISEASE. In (a), all features are conditionally independent, given DISEASE. In (b), several features are conditionally dependent, given DISEASE. We required a similarity network to construct this more complex influence diagram.

Because of the wide scope of the domain, however, the expert was uncomfortable assessing conditional dependencies among some features. Through many discussions with the expert, I identified the source of his difficulties with the construction of the Pathfinder influence diagram, and developed the similarity-network representation to overcome these difficulties. Given this representation, the expert was able to construct the influence diagram for Pathfinder shown in Figure 1.10. In addition, using some of the insights that motivated the creation of similarity networks, I developed the partition representation to facilitate the probability assessment for the influence diagram. This representation decreased the number of probability assessments required to construct a joint distribution for the lymph-node domain by more than a factor of five. As we shall see, the similarity-network and partition representations produced a new version of Pathfinder whose diagnostic accuracy was superior to that of previous versions.

In Chapters 4 we examine the details of the construction of this influence diagram; in Chapter 5, we discuss the formal evaluation of diagnostic accuracy.

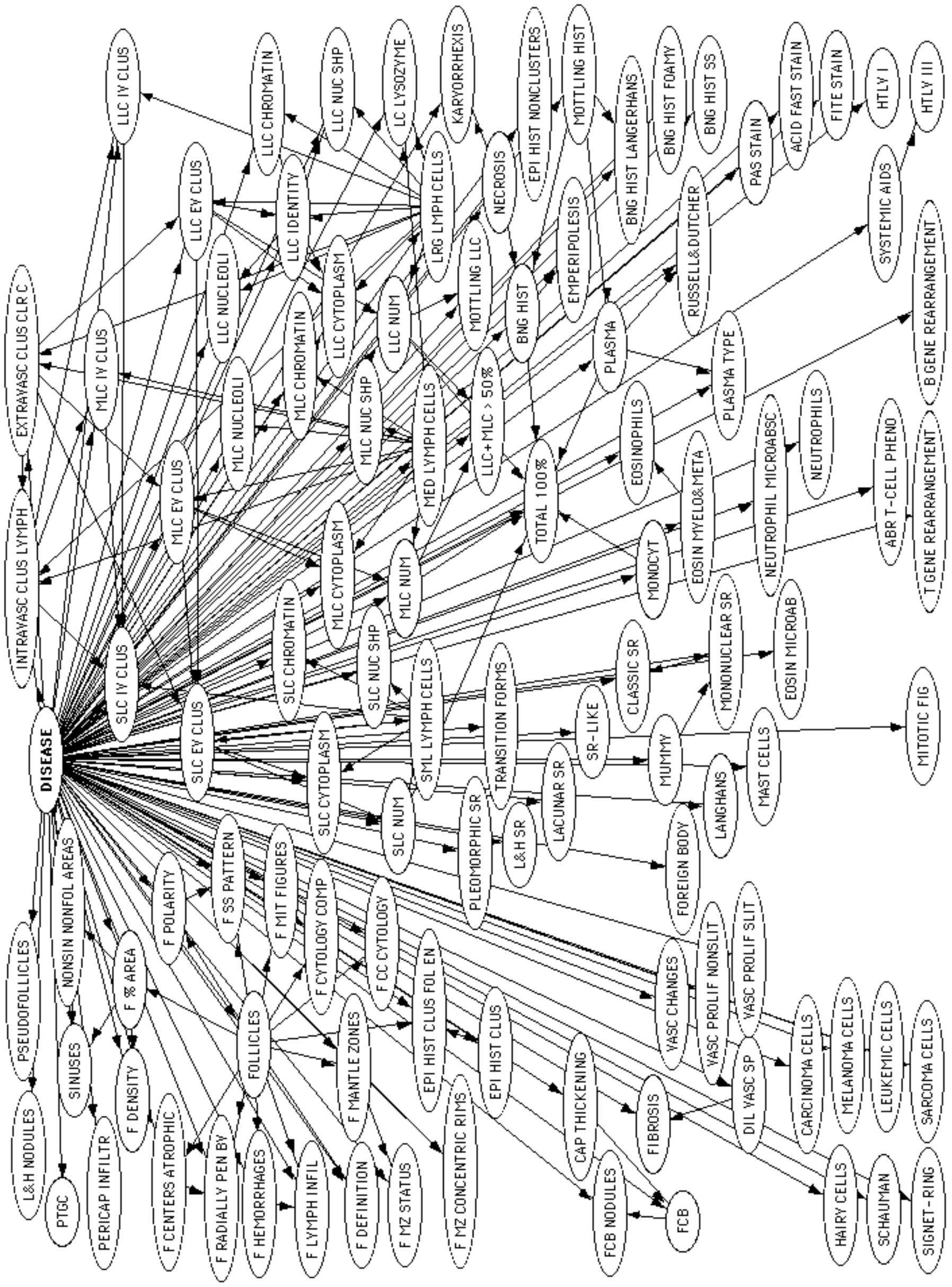

**Figure 1.10:** The Pathfinder influence diagram.
The influence diagram represents over 100 features that are relevant to diagnosis (Appendix C contains a list of the feature abbreviations). The node DISEASE contains over 60 lymph-node diseases.

## 1.2 Similarity Networks and Partitions

In general, influence diagrams can represent the alternatives, preferences, and beliefs of a decision maker. Influence diagrams that represent only the beliefs of a decision maker—that is, influence diagrams that contain only chance nodes and informational arcs—are called *knowledge maps* (Howard, 1989a). Other names for knowledge maps include *belief networks* (Pearl, 1986), *Bayesian networks* (Pearl, 1988), and *probabilistic influence diagrams* (Shachter, 1990). The Pathfinder influence diagram in Figure 1.10, for example, is a knowledge map.

The Pathfinder knowledge map has a special form. In particular, the chance node DISEASE contains many mutually exclusive and exhaustive diseases. In addition, this node conditions many other nodes, but is itself not conditioned by any nodes. Knowledge maps of this form are seen commonly in problems of diagnosis in which a single disease or *fault* is present.

A similarity network is a tool for building large and complex knowledge maps that have this special form. The disease node or—more generally—the *distinguished node* is the center of attention for the construction of a similarity network. The components of a similarity network include a *similarity graph* and a collection of *local knowledge maps*. Each node in a similarity graph represents an instance of the distinguished node, called a *hypothesis*. Edges in a similarity graph connect hypotheses that are similar or that are likely to be confused with one another by a user of the expert system. A local knowledge map is associated with each edge in the similarity graph. A local knowledge map for the edge between hypotheses $h_i$ and $h_j$ is a knowledge map constructed under the assumption that only $h_i$ and $h_j$ are possible. That is, a local knowledge map for hypotheses $h_i$ and $h_j$ is a knowledge map for discriminating only those two hypotheses. By constructing local knowledge maps, a person can concentrate on one manageable portion of the modeling task at a time.

Given a similarity network, we can construct a *global knowledge map* for the entire domain through simple graph manipulations on the local knowledge maps. The construction is *sound* in the sense that we can derive the assertions of conditional independence and dependence in the global knowledge map from the rules of probability and the assertions of conditional independence and dependence in the local knowledge maps. Also, a simple algorithm exists for verifying that the assertions in the local knowledge maps are *consistent.* Thus, the global knowledge map constructed from a similarity network faithfully represents a person's assertions. We say that the global knowledge map is *valid.* In addition, the construction of the global knowledge map is *exhaustive* in the sense that any feature that is relevant to discrimination of the set of hypotheses as a whole will appear in the global knowledge map.

Similarity networks represent two forms of conditional independence called *subset independence* and *hypothesis-specific independence*, neither of which is represented conveniently in a knowledge map. In Chapter 2, we examine these forms of conditional independence, and discuss the problems posed by their representation in knowledge maps. Similarity networks take advantage of these forms of independence to decompose the construction of a knowledge map in much the same way that knowledge maps take advantage of ordinary conditional independence to decompose the construction of a joint probability distribution. In particular, similarity networks exploit people's ability to make judgments of subset independence and hypothesis-specific independence, without assessing the probabilities that underlie such judgments.

We can also use assertions of subset independence and hypothesis-specific independence to simplify the assessment of probabilities associated with a knowledge map. These assertions of conditional independence, as they are represented in a similarity network, simplify assessment somewhat. In Chapter 2, however, we examine the *partition*, a representation that exploits more fully these independencies for assessment. Figure 1.11 summarizes the roles of the partition, similarity network, and knowledge map in the construction of a joint probability distribution.

Although both similarity networks and partitions were designed to simplify the construction and assessment of knowledge maps of the form shown in Figure 1.9(b), we see in Chapter 6 that the representations can be generalized to other problems of diagnosis. For example, we see that the representation can be used in some situations where hypotheses are not mutually exclusive, and where the disease node is conditioned by other nodes.

## 1.3 Historical Background and Contributions

The developments in this book derive from work within the related fields of medical informatics and artificial intelligence and within the discipline of decision analysis. Because these areas of research differ in their methods, language, and philosophy, we examine the historical background and contributions of this work from two perspectives.

### 1.3.1 Medical Informatics and Artificial Intelligence

One large area of research in the field of medical informatics has been the capture, representation, manipulation, and explanation of uncertain knowledge for expert systems. This research has undergone three distinct phases of development (Horvitz, 1986). In the first phase, which began over three decades ago, Ledley and Lusted suggested that probability theory was an appropriate framework for what is now called an expert system (Ledley and Lusted, 1959). Soon after, researchers in the field began experimenting with probabilistic and decision-theoretic expert systems for medical diagnosis and treatment. For example, in 1961, Warner constructed a probabilistic expert system for the diagnosis of congestive heart failure (Warner et al., 1961). In 1968, Gorry and Barnett extended that system to include test and treatment decisions (Gorry and Barnett, 1968). To avoid the complexity of directly acquiring and manipulating a joint distribution for their domains, these researchers made the simplifying assumptions represented by the influence diagram in Figure 1.9(a). In particular, they assumed that diseases were mutually exclusive and exhaustive, that all features were conditionally independent, given the true disease state of the patient, and that a single test or treatment decision was at hand. We refer to this collection of assumptions as the *simple Bayes model*. Using these assumptions, researchers found it relatively easy to implement decision-theoretic expert systems. By 1970, a large number of such programs had been developed (Miller et al., 1977; Wagner et al., 1978).

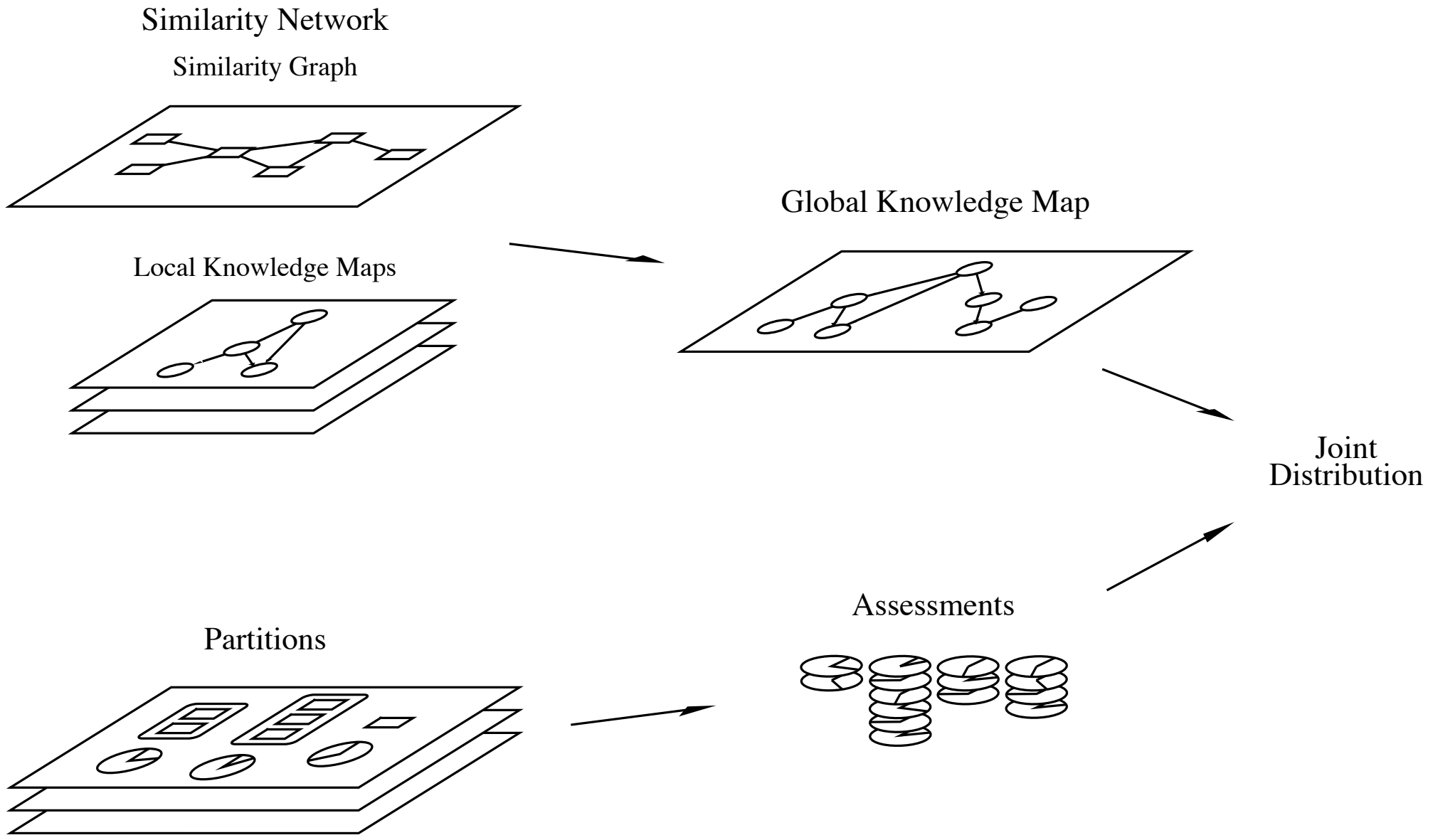

**Figure 1.11:** Decomposition of a joint probability distribution.
A joint distribution can be constructed from a global knowledge map and a set of assessments for each node in the map. The knowledge map itself can be constructed from a similarity network, consisting of a similarity graph and a collection of local knowledge maps. The assessments for each node in the knowledge map can be further decomposed using partitions.

Evaluations of most of these early systems showed that the programs performed well. In fact, the diagnoses rendered by several of them were more accurate than were those made by experienced physicians (de Dombal et al., 1972). Nonetheless, in the early 1970s, researchers began to criticize these systems. They noted that the domains of these programs were small and did not reflect realistic clinical situations. Furthermore, researchers argued that the assumptions of the simple Bayes model would be violated as the domains of these systems were expanded (Gorry, 1973; Shortliffe, 1976). One group of investigators showed that the diagnostic accuracy of an expert system based on the simple Bayes model deteriorated significantly as the number of features in the system increased. These investigators traced the degradation in performance to violations of the conditional-independence assumptions in the simple Bayes model (Fryback, 1978). Another group of researchers showed that the assumption of global conditional independence could be unrealistic in small domains as well (Norusis and Jacquez, 1975).

Researchers at the time thought that they had two alternatives for building expert systems of realistic size within the normative framework: (1) retain the simple Bayes model, or (2) capture and represent a complete joint probability distribution and (possibly) a complete utility model for a given domain. In the latter alternative, they imagined assessing probabilities under the assumption that any combination of diseases was possible, and that any feature could be conditionally dependent on any other set of features. Investigators found neither alternative attractive. Use of the simple Bayes model was unacceptable for the reasons we have discussed, and the assessment and manipulation of a full joint probability distribution was intractable. In addition to the problem of intractability, critics of the normative approach argued that decision-theoretic representations could not easily express the informal, qualitative nature of human reasoning (Gorry, 1973; Szolovits, 1982; Davis, 1982). Most researchers thus abandoned the normative approach in favor of heuristic techniques offered by the emerging discipline of artificial intelligence.

The development of heuristic approaches for use in expert systems dominated much of the research of the 1970s. These approaches generally incorporated expressive knowledge representations often patterned after those that experts seemed to use. The resulting benefits included improvements to explanation, and more sophisticated techniques for knowledge acquisition and inference. For example, several representations allowed investigators to avoid the assumptions of global conditional independence of the simple Bayes model without assessing a complete joint distribution. Examples include *rules* in Mycin (Shortliffe and Buchanan, 1975), *causal networks* in Casnet (Weiss et al., 1978), and *frames* in the Present Illness Program (PIP) (Pauker et al., 1976). Other artificial-intelligence programs addressed the assumption of a single-disease diagnosis. A notable example is the Internist-1 program, which is able to diagnose multiple diseases in a single patient (Miller et al., 1982).

Heuristic representations allowed researchers to construct expert systems for large and complex real-world domains. Nonetheless, in the 1980s, researchers in the fields of medical informatics, artificial intelligence, and decision analysis began to identify significant deficiencies in these representations. For example, heuristic methods were based on approaches used by experts and therefore were vulnerable to the incorporation of undesirable biases in human reasoning. In addition, artificial-intelligence investigators did not axiomatize their methods successfully. Consequently, these investigators were unclear about the meaning of the quantities that they used to represent uncertainty and preference. For example, researchers developed schemes that confused the *absolute* degree of belief for a hypothesis, given evidence for the hypothesis, with the *change* in degree of belief for that hypothesis, given that evidence (Heckerman, 1985; Horvitz and Heckerman, 1985). Also, because researchers did not provide an unambiguous characterization of their representations and inference methods, they found it difficult to modify their approaches when the approaches performed poorly, to extend their approaches to new domains, and to build on the work of other people. Moreover, researchers were often unaware of the implicit assumptions imbedded in their methods. Thus, heuristic-based expert systems were susceptible to unanticipated errors. For example, in 1985, I developed an axiomatization of the certainty-factor (CF) model used by Mycin to manage uncertainty (Shortliffe and Buchanan, 1975). This axiomatization shows that, under certain circumstances, the CF model makes assumptions that are stronger than are those of the simple Bayes model. In particular, the CF model includes the assumption that features are conditionally independent, given each disease, as well as given the *negation* of each disease (Heckerman, 1985). In an empirical study, I showed that the use of these strong assumptions can degrade significantly the diagnostic accuracy of an expert system (Heckerman, 1988).

Overall, researchers began to believe that the more powerful artificial-intelligence representations and inference procedures were compromised by the nonnormative nature of the procedures. This observation set the stage for the third phase of expert-system research. In 1981, decision analysts Howard and Matheson developed the influence-diagram representation (Howard and Matheson, 1981). The representation was easy to use, yet was capable of expressing precisely any inference or decision problem in the normative framework. The representation offered investigators a spectrum of alternatives between using the simple Bayes model and using a complete joint probability distribution. For example, using an influence diagram, an expert could assert that some features were conditionally independent, and that other features were conditionally dependent. In addition, an expert could identify some diseases as being dependent (or mutually exclusive) and others as being independent. In many domains, these assertions could be used by knowledge engineers to make knowledge acquisition and inference tractable. Thus, with the advent of influence diagrams, researchers once again began to construct normative expert systems.

In this book, I argue that we have not yet approached the limits of the application of decision theory to the representation of knowledge. Decision theory is nothing more than a set of constraints that helps us to improve our thinking about important decisions. Within these constraints, I argue, we can build languages that can express almost any rational concept in a tractable manner. The influence diagram, and the extensions of it that I develop in this work, are just the first examples of the normative languages that we can construct. I argue that, to construct normative expert systems for complex domains—even more complex than the domain of Pathfinder—we must identify forms of conditional independence that human experts use to manage the complexity in their domains. We must then develop new representations that can exploit these assertions to facilitate the capture and representation of expert knowledge.

Researchers have argued that this approach to constructing expert systems is impractical for many domains, because a move beyond a simple Bayes model or some other oversimplified Bayesian model might encounter massive interdependencies among distinctions. Indeed, this argument was and still is made by most artificial-intelligence researchers who abandoned the normative approach in the 1970s. I conjecture, however, that cognitive limitations on human experts will constrain the complexity of normative computer-based models for decision making. That is, limitations on human memory and human information-processing capabilities require that an expert impose assertions of conditional independence on his own domain knowledge, whether or not these assertions actually hold (i.e., whether or not large amounts of experimental data would refute these assertions). More important, I believe that, for many domains, we can formulate these expert assertions of conditional independence such that they are self-consistent. The development of the similarity-network and partition representations demonstrates that such a formulation is possible. Thus, I conjecture that, for many domains, we can capture and represent the important details of an expert's knowledge in a coherent, normative framework. The possibility that an expert imposes erroneous assertions of conditional independence on his domain does not present a fundamental problem. If we find that particular assertions of independence are incompatible with available experimental data, we can use the principles of probability theory to update the expert's model for his domain.

Although the identification and exploitation of conditional independence is an important approach by which we can build normative expert systems, other approaches also offer promise. In particular, Wellman has developed a method for representing and reasoning with an incomplete decision-theoretic model. He has shown that, in many cases, unambiguous recommendations for action can be derived from such a model (Wellman, 1986; Wellman, 1988; Wellman, 1990). Also, Horvitz and other investigators are using decision analysis at the metalevel to trade off the benefits of a normative approach with the time and effort required to build, reason with, and comprehend a decision-theoretic model (Horvitz, 1986; Horvitz, 1987; Heckerman and Jimison, 1987; Horvitz, 1988; Heckerman et al., 1989a). Using these approaches and the approach developed here, it is likely that researchers will develop normative experts systems for many real-world domains that deliver valuable information to users.

### 1.3.2 Decision Analysis

Decision analyses are extremely expensive. Typically, a decision analysis requires the participation one or more decision makers, one or more experts, and one or more decision analysts. A person faced with the decision of whether or not to undergo a coronary artery bypass surgery, for example, would have to pay approximately $5,000 in 1990 dollars to buy a decision analysis (Howard, 1990).

In the last 5 years, researchers familiar with principles of both decision analysis and artificial intelligence have developed two approaches in an effort to reduce the high cost of decision analyses. One approach is the normative expert system. Another approach, created by Holtzman, is the *intelligent decision system* (Holtzman, 1989). Although we concentrate on normative expert systems in this work, the similarity-network and partition representations can facilitate the construction of both types of systems. Thus, let us briefly examine intelligent decision systems, and compare them with normative expert systems.

The developers of both expert systems and of intelligent decision systems treat a set of decision problems whose members have a degree of similarity among them as a single unit. As we have discussed, we refer to this unit as a decision class or domain (Holtzman, 1989). Two goals shared by the developers of both types of systems are to capture expert knowledge about a particular decision class, and to deliver this knowledge to many decision makers faced with a decision within that class. Consequently, both types of systems may obviate a decision analysis, or at least reduce the cost of an such analysis. Those researchers who work to build intelligent decision systems, however, have the additional goal of capturing the expertise of decision analysts, and of delivering this expertise to decision makers. In particular, these researchers hope that their systems will help decision makers to structure decision models through automated sensitivity analyses, to provide unbiased assessments of probability and utility for their models, and to explore solved models so that the decision makers may gain insights about their decisions.

The concept of an intelligent decision system is new, and none of the goals of its developers have been realized completely. Nonetheless, Holtzman has constructed an intelligent decision system with results that are promising (Holtzman, 1989). The system, called *Rachel*, advises infertile couples seeking medical assistance. The intelligent decision system helps the couple and their physician derive a recommended course of action from a model that combines the physician's medical knowledge with the couple's knowledge of their preferences and special circumstances.

The architecture of intelligent decision systems, and of Rachel in particular, is a rule-based system. There are two important differences, however, between the architecture of intelligent decision systems and that of rule-based expert systems developed previously. First, an expert system typically contains situation–action rules that encode expert-recommended decisions, given a particular context. On the other hand, an intelligent decision system contains rules that suggest how to construct, modify, or interpret an influence diagram. Figure 1.12 illustrates one of the rules in Rachel's knowledge base. The rule states that, if the patient is in good health, and if a particular surgeon and anesthesiologist perform an internal spermatic vein ligation, then the chance node SURGICAL COMPLICATIONS depends on the decision node SURGERY, as indicated by the probability distributions shown in Figure 1.12. When a decision maker interacts with an intelligent decision system, the system evokes many rules of this form, and thereby constructs an influence diagram for the decision problem. In the process, the decision maker or domain expert may choose to override several rules, or he may provide distinctions, dependencies, probabilities, or utilities that the system needs to complete the decision model.

**IF** The patient is in good health, the internal spermatic vein ligation is performed by Dr. PQR, and the attending anesthesiologist is Dr. STW,

**THEN** The variable SURGICAL COMPLICATIONS can be indirectly assessed by means of the following relation:

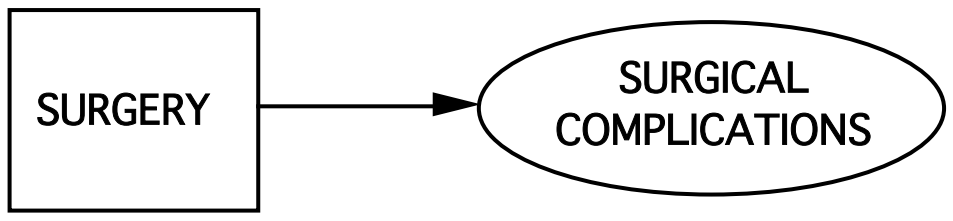

;

with $p($ SURGICAL COMPLICATIONS | SURGERY $, \xi_{+}) = 0.002$
and $p($ SURGICAL COMPLICATIONS | SURGERY $, \xi_{-}) = 0.0.$

**Figure 1.12:** A rule in an intelligent decision system.
The rule relates the probability of postsurgical complications to the track record of the surgical team (Holtzman, 1989, page 142). See the end of this book for a guide to notation.

A second but related difference between intelligent decision systems and rule-based expert systems is that many expert systems contain a mechanism for attaching uncertainty or preference to rules. Most such mechanisms resemble the Mycin CF model, and are necessary because the rules relate situations to action. In intelligent decision systems, however, there is less need to attach uncertainty or preference to the rules, because the rules pertain to the construction of an influence diagram. An intelligent decision system can bring to bear uncertainties and preferences of the decision maker at the time the influence diagram is solved.

If the rules that comprise an intelligent-decision-system knowledge base are consistent, then the procedure for creating influence diagrams from a set of rules has desirable theoretic properties. Breese recently extended the intelligent-decision-systems procedure for constructing influence diagrams, and examined the formal properties of his procedure (Breese, 1987; Breese, 1990). His goal in developing this procedure is somewhat different from that of the developers of intelligent decision systems. In particular, Breese considers domains in which influence diagrams can be constructed automatically from a set of rules in a knowledge base and from a particular decision context. In these domains, a decision maker does not modify or override the rules in the knowledge base. Nonetheless, Breese's method can be applied to the construction of influence diagrams within intelligent decision systems. Breese has shown that, if a set of rules is consistent, and if his procedure produces a directed acyclic influence diagram from that set of rules and a particular decision context, then the chance and decision nodes that do appear in the knowledge base but do not appear in the constructed influence diagram must be irrelevant to the decision problem at hand. Of course, this guarantee fails if the set of rules is not consistent.

Overall, the intelligent decision system is more likely to provide appropriate decision assistance in decision classes where there are events or circumstances that cannot be anticipated, and in decision classes where probabilities or utilities are likely to vary from one person to the next. Not surprisingly, Rachel's domain has these properties. For example, an expert system for infertile couples probably will not anticipate that a particular couple can afford an expert surgeon from a foreign country for a internal spermatic vein ligation. An intelligent decision system containing the rule in Figure 1.12, however, could recognize that an unusual situation has occurred, and request that a knowledgeable agent of the couple provide the system with new probability distributions describing the chances of postsurgical complications. Also, the desirability of having a baby is likely to vary significantly across couples.

Despite these observations, several significant problems are associated with the architecture of intelligent decision systems. First, how does such a system guarantee that a set of rules for a complex decision class is consistent? Checking the consistency of a collection of rules in a logic framework is NP-hard (Garey and Johnson, 1979); the procedure is even more difficult in a decision-theoretic framework. Also, how does a decision maker maintain the consistency of a knowledge base when he adds to or modifies the rules in a knowledge base? He probably will require the assistance of experts knowledgeable about both decision theory and the given decision class. Furthermore, those experts might have to inspect the entire knowledge base to guarantee that their input is consistent with the knowledge currently in the system. This process would be expensive and time consuming.

To address these problems, builders of intelligent decision systems for a given decision class could create a large influence diagram for the entire class. The structure of this influence diagram might not be complete, and the influence diagram might lack some probability distributions and utilities, but the consistency of the rules implicit in the diagram would be guaranteed. In addition, if a decision maker wanted to add to or modify the knowledge base of this intelligent decision system, he could inspect the incomplete influence diagram before and after making those changes, and thereby could maintain the consistency of the knowledge base. Furthermore, if developers of an intelligent decision system constructed the knowledge base of that system with an influence diagram, areas of the knowledge base that are incomplete would be highlighted. These developers probably would build a more complete model than they would if they constructed the knowledge base as separate rules. Thus, the need for a decision maker to intervene during the construction of an influence diagram for his particular problem probably would decrease. In an extreme case of this methodology, where the influence diagram for a given decision class is complete, the intelligent decision system would become a normative expert system.

Almost certainly, there will be decision classes for which we cannot build single coherent influence diagrams. In these cases, we might be able to do nothing better than to construct a knowledge base of possibly inconsistent influence-diagram components. We should keep in mind the advantages of consistency, however, and strive to create intelligent decision systems based on single influence diagrams.

Whether we build normative expert systems or intelligent decision systems, the Pathfinder dilemma illustrates that we need extensions to the influence-diagram representation to facilitate the construction of large and complex influence diagrams. In this book, I demonstrate that we can create such extensions by identifying forms of conditional independence used by experts (and possibly decision analysts) to manage the complexity of their domain, and by creating languages that encode explicitly these forms of independence.

## 1.4 Overview of the Book

In Chapter 2, the similarity-network and partition representations are illustrated by a small real-world example from the domain of medicine. In Chapter 3, a formal theory for similarity networks is developed. I show that the construction of the global knowledge map from a similarity network consisting of a connected similarity graph and a set of local knowledge maps is both *sound* and *exhaustive*. In addition, a simple algorithm for testing the consistency of a set of local maps is developed. In Chapter 4, I describe highlights of the construction and assessment of the knowledge map for Pathfinder using similarity networks and partitions. In Chapter 5, the diagnostic accuracy of the version of Pathfinder developed with the similarity-network and partition representations is compared to that of the version of Pathfinder in which all features were assumed to be conditionally independent. Finally, in Chapter 6, I describe extensions to the theory, and discuss conclusions that we can derive from this work.

# 2 Similarity Networks and Partitions: A Simple Example

In this chapter, we use the similarity-network and partition representations to construct and assess a knowledge map for a small medical expert system. The purpose of this exercise is to illustrate the basic concepts and techniques underlying these representations, and to demonstrate some of the advantages of their use. Because the example is small, however, the full power of these representations for simplifying knowledge acquisition cannot be demonstrated. In Chapter 4, we examine highlights of this knowledge-acquisition approach as it is applied to Pathfinder. There, the power of these representations is illustrated more fully.

The medical example that we examine is real, but it has been simplified for purposes of presentation. Dr. Harold Lehmann served as the expert for the domain. The figures in the chapter were generated by SimNet, an implementation on the Macintosh computer of the similarity-network and partition representations.

Throughout the example and the remainder of this book, I will distinguish between the construction of a knowledge map, similarity network, or partition by a *person* and the construction of these representations by an *algorithm*. In particular, the terms *to compose* and *to construct* will refer to situations where a person and an algorithm generate a representation, respectively.

## 2.1 Similarity Networks: The Construction of a Knowledge Map

Suppose a patient between 5 and 18 years of age comes to an emergency room complaining of severe sore throat. A knowledge map for this situation is illustrated in Figure 2.1. The chance node DISEASE represents the causes of sore throat: VIRAL PHARYNGITIS, STREP THROAT, MONONUCLEOSIS, TONSILLAR CELLULITIS, and PERITONSILLAR ABSCESS. We assume that these diseases are mutually exclusive and exhaustive. The remaining nodes represent evidence relevant to the diagnosis of the patient's disease. We now discuss how to construct this knowledge map using a similarity network.

### 2.1.1 Composition of a Similarity Network

The focus for the composition of the similarity network is the *distinguished node* or *distinguished variable*. For medical domains, the distinguished variable represents a set of mutually exclusive and exhaustive diseases. In general, we refer to the mutually exclusive exhaustive instances of this variable as *hypotheses*.

A similarity network consists of a *similarity graph* and a collection of *local knowledge maps*. To compose a similarity graph, we first compose the similarity graph. The nodes in the similarity graph correspond to hypotheses of the distinguished node. Informally, the edges in the similarity graph connect hypotheses that are similar. We shall discuss soon the precise meaning of edges in a similarity graph. The similarity graph for sore throat is shown in Figure 2.2.

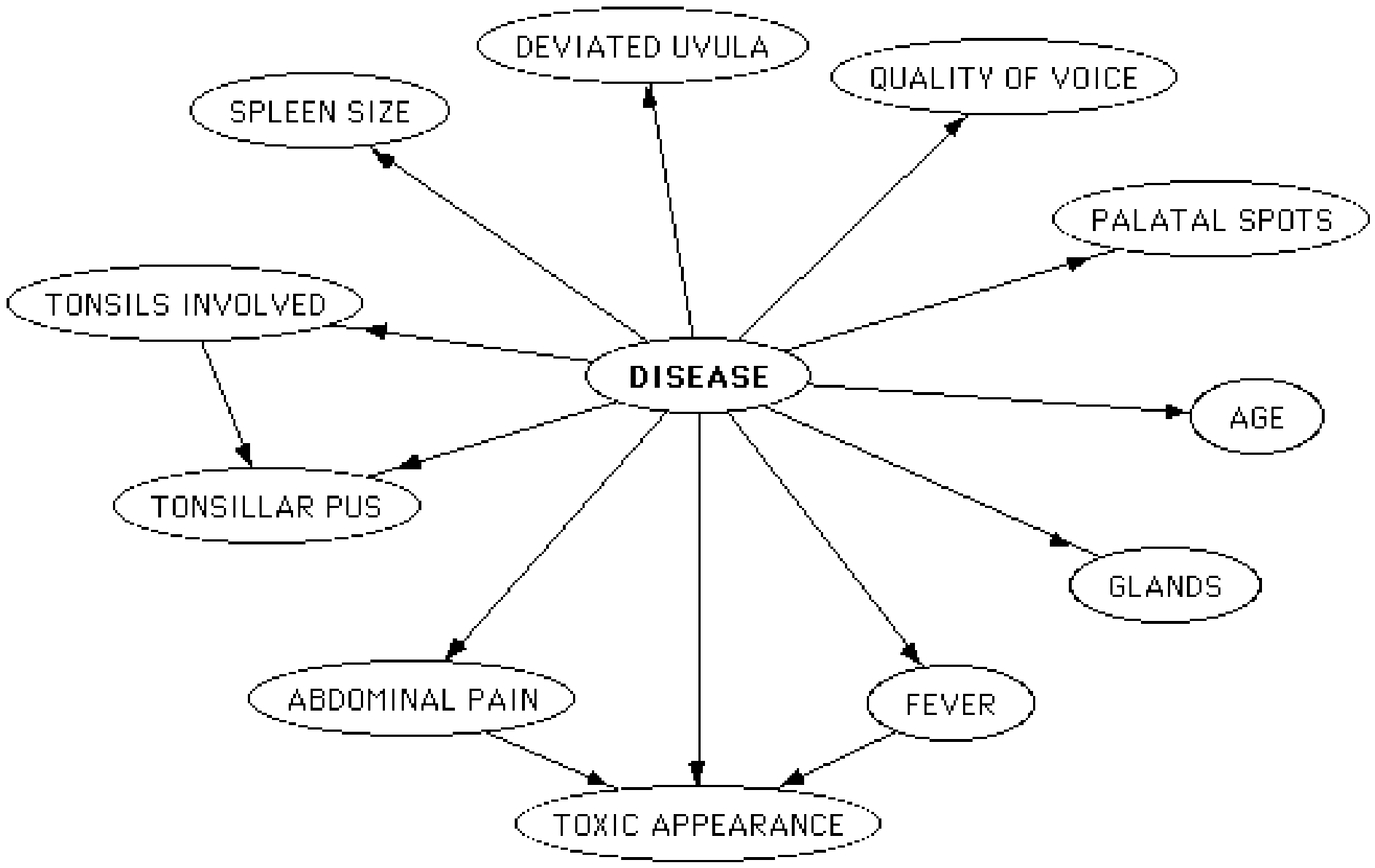

**Figure 2.1:** A knowledge map for sore throat.
This knowledge map describes the diagnostic dilemma for a patient between 5 and 18 years of age who comes to an emergency room with a severe sore throat. The node DISEASE represents the mutually exclusive and exhaustive causes of sore throat: VIRAL PHARYNGITIS, STREP THROAT, MONONUCLEOSIS, TONSILLAR CELLULITIS, and PERITONSILLAR ABSCESS. This node is the focus for the composition of a similarity graph. The remaining nodes represent evidence relevant to the diagnosis of the patient's disease.

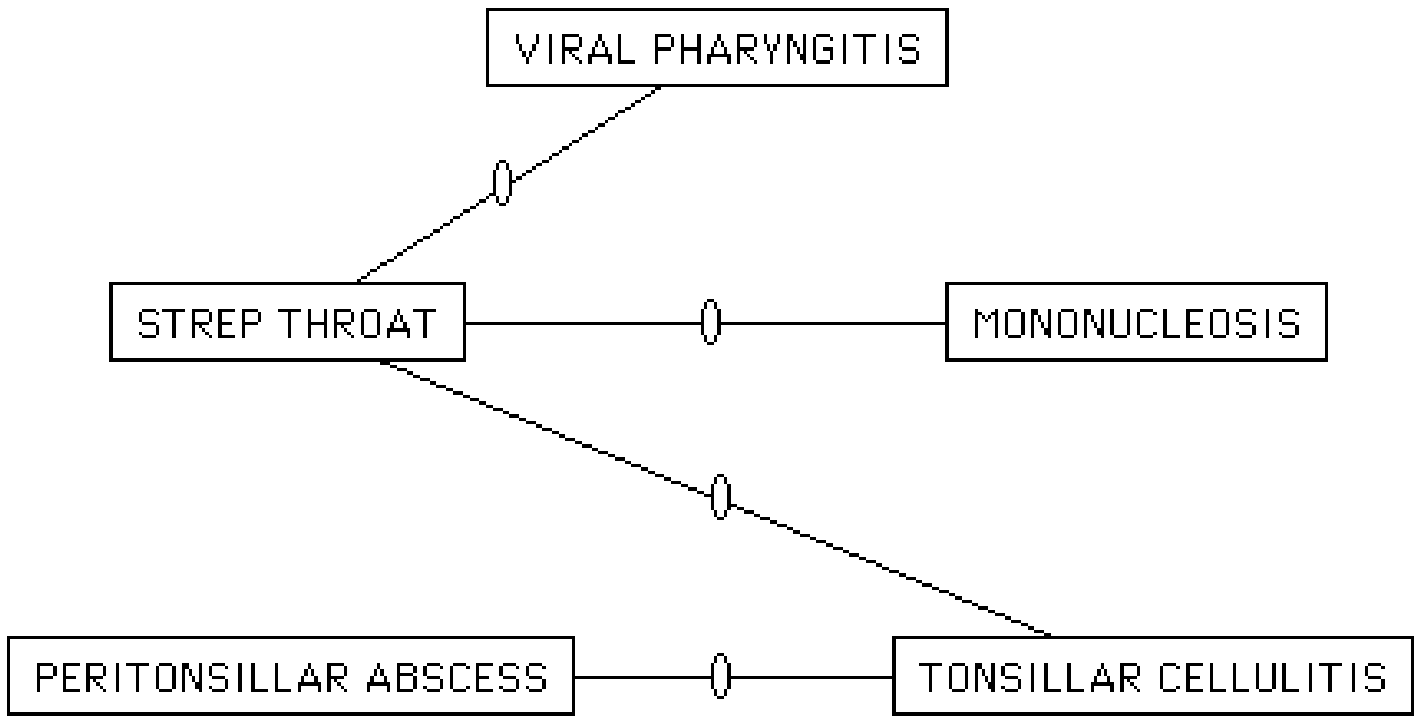

**Figure 2.2:** A similarity graph for sore throat.
The nodes in the graph represent the possible causes of sore throat. Undirected edges connect diseases that are similar. Although this graph is a tree (i.e., there is exactly one path between any two nodes in the graph), in general, similarity graphs can contain cycles.

Next, we compose a local knowledge map for each pair of hypotheses that is connected in the similarity graph. To compose a local knowledge map for the hypothesis pair $h_i$ and $h_j$, we imagine that one of these two hypotheses is true. Given this supposition, we compose a knowledge map consisting of the distinguished node—whose instances are restricted to $h_i$ and $h_j$—and those *nondistinguished nodes* that are relevant to the discrimination of these hypotheses. Formally, we omit a node from the local knowledge map if and only if the node would be disconnected from the distinguished node (i.e., there would be no path between the node and the distinguished node) if we included it in the map.

Figure 2.3 shows the local knowledge map for the edge between TONSILLAR CELLULITIS and PERITONSILLAR ABSCESS in the similarity graph. The node at the top of the knowledge map represents the distinguished variable restricted to these two diseases. The nondistinguished nodes in the local map represent the features or disease findings that are relevant to the discrimination of the diseases TONSILLAR CELLULITIS and PERITONSILLAR ABSCESS. Notice that there are no arcs among these nondistinguished nodes. The missing arcs represent the assertion that, given that the patient has either TONSILLAR CELLULITIS or PERITONSILLAR ABSCESS, all findings in the map are independent. Also note that there are fewer findings in this local map than in the knowledge map for the entire domain (Figure 2.1). This observation tends to be true, in general, because the diseases associated with local knowledge maps are similar.

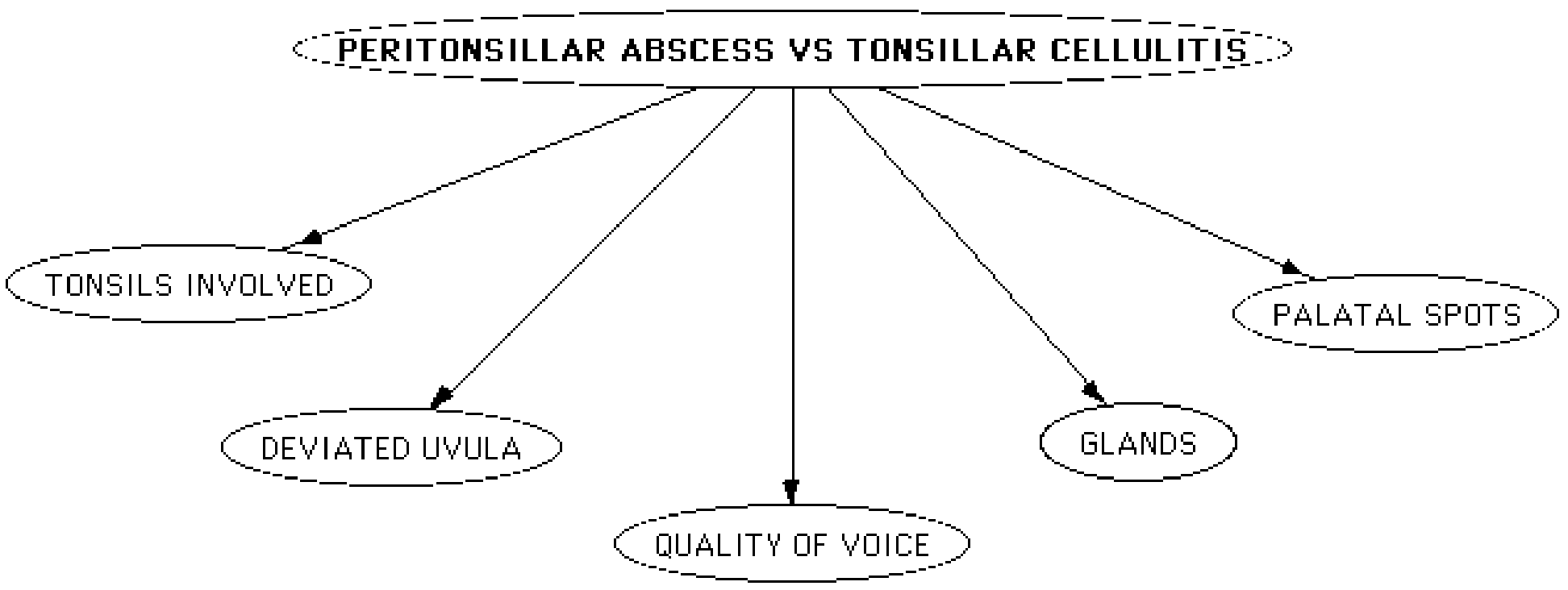

**Figure 2.3:** A local knowledge map.
In this knowledge map for the edge between PERITONSILLAR ABSCESS and TONSILLAR CELLULITIS in the similarity graph, the features exhibit mutual conditional independence. Only findings that are relevant to the discrimination of the two diseases are included in the map. Because the two diseases are similar, the knowledge map contains fewer nodes than does the knowledge map for the entire domain.

Figure 2.4 shows the local knowledge map for the edge between STREP THROAT and VIRAL PHARYNGITIS in the similarity graph. Again, the map contains fewer features than does the knowledge map for the sore-throat domain as a whole. Now, however, some of the disease findings are conditionally dependent. The arc from TONSILS INVOLVED to TONSILLAR PUS reflects the expert's assertion that the probability of seeing pus on a patient's tonsils depends on whether the disease involves one tonsil, both tonsils, or neither tonsil, even when the patient's disease is known. The arcs from FEVER and ABDOMINAL PAIN to TOXIC APPEARANCE reflect the observation that a patient is more likely to present with a toxic appearance if the patient has abdominal pain or a high fever, even when the patient's disease is known. Although FEVER is relevant to the discrimination of STREP THROAT and VIRAL PHARYNGITIS indirectly through its effect on TOXIC APPEARANCE, the missing arc from the disease node to FEVER represents the assertion that temperature alone is not relevant to the discrimination of the two diseases.

The feature PALATAL SPOTS, among other features, appears in both of the local knowledge maps that we have examined. In the local knowledge map for TONSILLAR CELLULITIS and PERITONSILLAR ABSCESS, the instances of this feature are ABSENT and PRESENT. The same instances are associated with this feature in the local knowledge map for STREP THROAT and VIRAL PHARYNGITIS. In general, the instances associated with a feature in one local knowledge map must be identical to the instances associated with that feature is all other local knowledge maps.

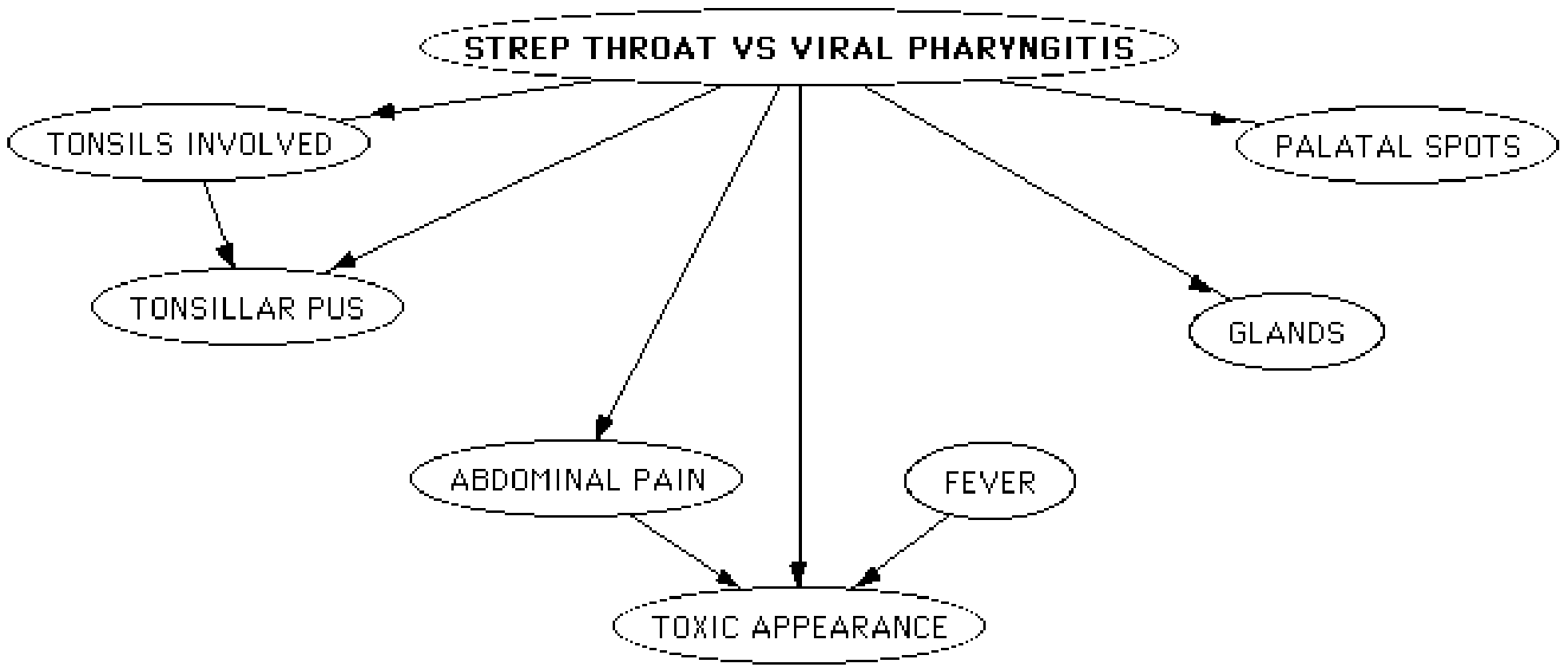

**Figure 2.4:** A local knowledge map with dependencies.
In this knowledge map for the edge between STREP THROAT and VIRAL PHARYNGITIS in the similarity graph, the feature TONSILLAR PUS is conditionally dependent on TONSILS INVOLVED, and the feature TOXIC APPEARANCE is conditionally dependent on ABDOMINAL PAIN and FEVER.

Figure 2.5 shows the local knowledge maps for the remaining two edges in the similarity graph. Again, there are fewer features in each of these maps than there are in the knowledge map for the entire domain.

In SimNet, the local knowledge map for each edge in a similarity graph is accessed via the oval on the edge (see Figure 2.2). Specifically, by clicking on an edge's oval, the user brings up a window in which the local knowledge map associated with that edge can be created or modified.

The formal criteria for drawing edges in a similarity graph are that (1) we connect two diseases only if we can compose a local knowledge map for the disease pair, and (2) the similarity graph must be connected—that is, there must be a path between any two nodes in the graph. There is no formal requirement that connected diseases be similar. As we have seen in this example, however, local knowledge maps for pairs of similar diseases tend to exclude many of the features that distinguish the set of diseases as a whole. Thus, an expert can simplify greatly his task of composing the local knowledge maps by connecting only similar diseases in the similarity graph (provided the graph remains connected). Indeed, in practice, I have found it useful to ask experts to draw a similarity graph using only considerations of similarity; I introduce the concept of a local knowledge map after the similarity graph is composed.

To simplify composition further, an expert may choose not to compose certain local knowledge maps, even if he believes that he can compose them. There is no need to build more knowledge maps than those that are required to create a connected similarity graph. Alternatively, an expert may choose to build a multiply connected similarity graph.[1] As we shall see in Chapter 3, cycles in the similarity graph provide additional opportunities to check the self-consistency of a user's knowledge.

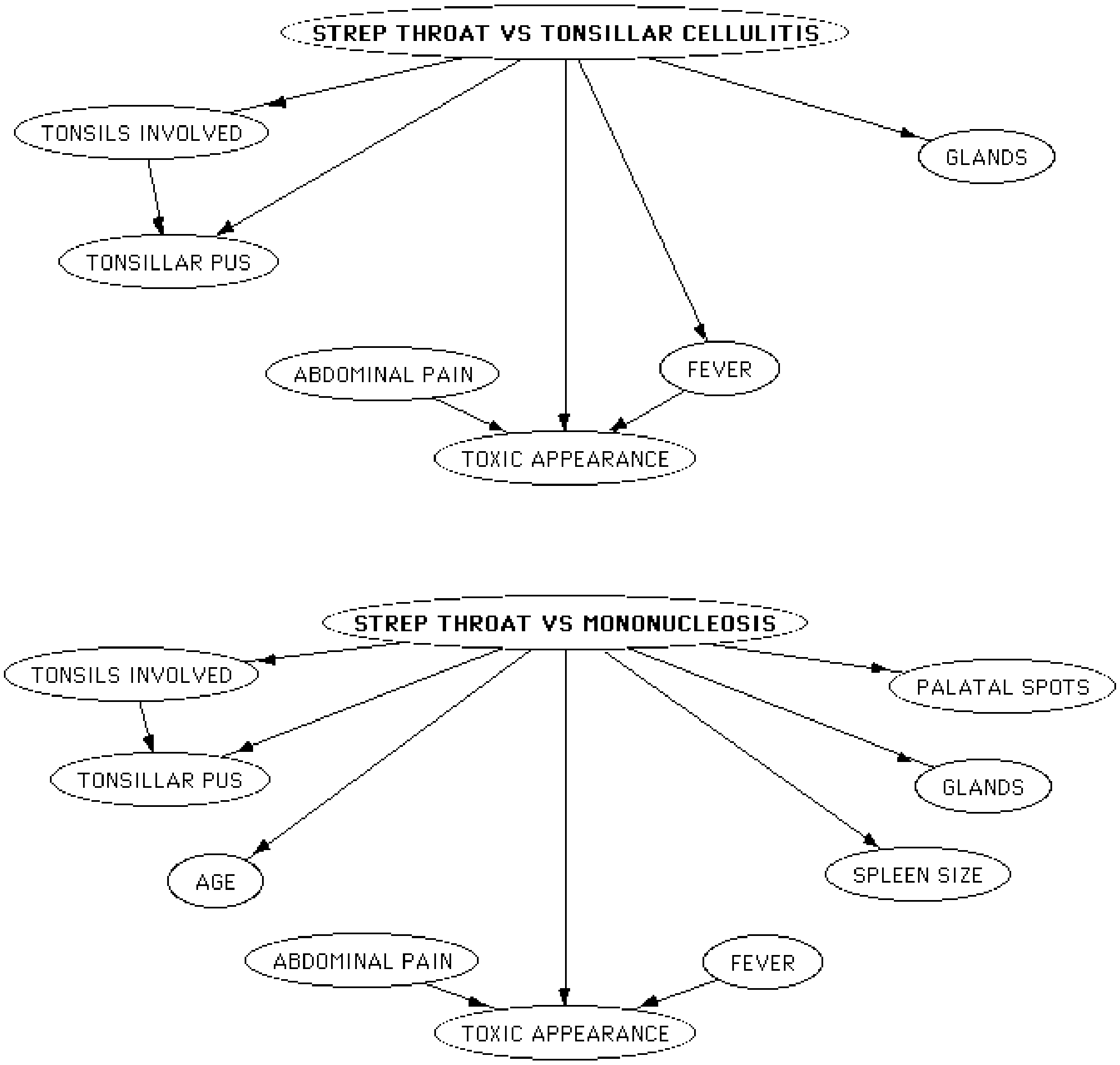

**Figure 2.5:** Local knowledge maps for the other two edges in the similarity graph. The knowledge map contains fewer nodes than does the knowledge map for the entire domain.

Given the similarity network that we have *composed*, we can now *construct* the knowledge map for the full sore-throat problem, called the *global knowledge map.* Specifically, we construct the global knowledge map by forming the *graph union* of the local knowledge maps in the similarity network. The operation of graph union is straightforward. The nodes in the graph union of a set of graphs is the simple union of the nodes in the individual graphs. Similarly, the arcs in the graph union of a set of graphs is the simple union of the arcs in the individual graphs. That is, a node (or arc) appears in the graph union, if and only if there is such a node (or arc) in at least one of the individual graphs. The set of instances associated with a feature in the global knowledge map is the same as the set of instances associated with that feature in each of the local knowledge maps. The set of instances associated with the disease or distinguished node in the global knowledge map is the union of all diseases or hypotheses in the similarity graph.

The global knowledge map for sore throat was shown in Figure 2.1. The node QUALITY OF VOICE, for example, appears in the global knowledge map because it appears in the local knowledge map for PERITONSILLAR ABSCESS and TONSILLAR CELLULITIS. The arc from DISEASE to ABDOMINAL PAIN appears in the global knowledge map because it is present in the local map for STREP THROAT and VIRAL PHARYNGITIS.

### 2.1.2 A Valid Knowledge Map

As we see in Chapter 3, under certain conditions, the construction of the global knowledge map from the similarity network is *sound.* That is, any joint distribution that satisfies the assertions of conditional independence implied by the local knowledge maps also satisfies the assertions of conditional independence implied by the global knowledge map. In addition, we can verify easily that the set of assertions implied by a collection of local knowledge maps is *consistent.* Given these two results, we know that the global knowledge map constructed from a similarity network accurately and coherently reflects the assertions of conditional independence of the person who composes that network. That is, the global knowledge map is *valid.*

These results apply to minimal knowledge maps as well. A *minimal* knowledge map is one in which no arc can be removed without contradicting one or more of the assertions of conditional independence made by the expert. Thus, a minimal knowledge map can represent both assertions of conditional independence and conditional dependence. We can check whether or not a collection of minimal local knowledge maps is consistent. In addition, the construction of a minimal global knowledge map is sound. To be more precise, suppose we have a similarity network in which all the local knowledge maps are minimal. Also, suppose we interpret the global knowledge map constructed from that similarity network to be a minimal knowledge map. Again, under certain conditions, any distribution that satisfies the assertions of conditional dependence implied by the local knowledge maps must also satisfy the assertions of conditional dependence implied by the global knowledge map. That is, if the local knowledge maps are minimal, then the global knowledge map is minimal as well.

[1]A multiply connected graph contains more than one path between some node pairs.

In Chapter 3, we show that the soundness result holds for any similarity network in which the following constraints are satisfied:

1. The instances of the disease node (hypotheses) are mutually exclusive and exhaustive.
2. The similarity graph is connected.
3. The global knowledge map that is equal to the graph union of the local knowledge maps contains no directed cycles.
4. There are no arcs pointing to the distinguished node in any local knowledge map.
5. The joint distribution for the domain is *strictly positive* (i.e., no combination of findings rules out any hypothesis).

In Chapter 4, we examine the effects of each of these constraints on the knowledge-acquisition process within the domain of lymph-node pathology. We see that only constraint 5 detracted from the usefulness of the similarity-network representation for knowledge acquisition within that domain. In Chapters 4 and 6, we discuss how these sufficient conditions for soundness, including condition 5, might be relaxed with additional theoretical work.

### 2.1.3 An Exhaustive Construction

The construction of a global knowledge map from a similarity network is also *exhaustive*. That is, any feature that is relevant to the discrimination of the hypothesis set as a whole must appear in some local knowledge map, and hence in the global knowledge map. In Chapter 3, we prove this result. Together, the soundness, consistency, and exhaustiveness results make the similarity-network representation extremely useful for knowledge acquisition.

### 2.1.4 Advantages of Using Similarity

The concept of similarity does not appear in the conditions for soundness, consistency, or exhaustiveness. Nonetheless, there are important advantages of composing knowledge maps for diseases that are similar. As we discussed in Section 2.1.1, one such advantage is that local knowledge maps for pairs of similar diseases tend to be small. If this fact were the only advantage of similarity networks for constructing a knowledge map, however, there would be no point in using them. If any feature appeared in more than one local knowledge map, we would be duplicating our efforts of composition, regardless of the size of the maps. In this section, we consider another advantage of composing local knowledge maps for pairs of similar diseases that makes the similarity-network representation a valuable tool for constructing knowledge maps.

As mentioned in Section 1.1, the Pathfinder expert could not compose directly the global knowledge map for the lymph-node domain. Specifically, he could not assess dependencies among certain features in the domain. When asked questions of the form

> Given any disease, does observing feature $x$ change your belief that you will observe feature $y$?

the expert sometimes would reply

> I've never thought about these two features at the same time before. Feature $x$ is relevant to the discrimination of a particular set of diseases. Feature $y$, on the other hand, is relevant to the discrimination of a different set of diseases. These two sets of diseases do not overlap, and I never confuse the first set of diseases with the second.

The expert had detailed knowledge about diagnosis in multiple *small worlds* or subsets of similar disease within the lymph-node domain. If two or more features were relevant to the same small world, the expert had no trouble assessing dependencies among those features. If two or more features were not relevant to a common small world, however, he could not evaluate the dependencies among them.

This observation is not that surprising. When an expert pathologist looks at a tissue section under the microscope, he immediately focuses on a relatively small set of diseases. He then expends the majority of his conscious effort looking for features that discriminate among these diseases.[2] Almost by definition, these initial disease sets will consist of diseases that are similar to one another. In fact, many of the experts who experimented with the similarity-network representation independently adopted the following operational definition for similarity when composing the similarity graph:

[2] Recall the hypothetico-deductive approach discussed in Chapter 1.

> Two diseases are similar if and only if they are likely to be confused with each other in practice.

It is reasonable to expect that knowledge about dependencies among features relevant to these small worlds are more available to an expert, because he spends most of his time thinking explicitly about such features and the relationships among them. In extreme cases, such as the situation described in the previous paragraph, knowledge about the discrimination of highly dissimilar diseases may be unavailable to the expert.

The similarity-network representation is an ideal tool for combining knowledge about these small worlds into a coherent whole. Indeed, the representation was developed in direct response to the assessment predicament of the lymph-node expert described two paragraphs earlier. Using a similarity network, an expert can assess dependencies among features that are relevant only to pairs of similar diseases. Given the soundness, consistency, and exhaustiveness results, we can combine the knowledge about such dependencies, recorded in the local knowledge maps, to create a global knowledge map that faithfully represents the assertions of an expert for his domain. This knowledge map, in turn, endows an expert system with the ability to discriminate among any set of diseases, whether they are similar or dissimilar. Consider again, for example, the assessment predicament of the expert. Expressed in terms of a similarity network, the features $x$ and $y$ never appear in the same local knowledge map. Thus, given the procedure for constructing the global knowledge map and the soundness result, we know that $x$ and $y$ must be conditionally independent, given disease. This observation was apparent to neither the expert nor me before the similarity-network representation and its theory were developed.

Even in situations where an expert can compose a global knowledge map, a similarity network should prove useful for knowledge acquisition. Specifically, by composing local knowledge maps for pairs of similar diseases, the expert can use a similarity network to focus his attention on precisely those diagnostic subproblems with which he is familiar. The expert may thereby increase the quality of the knowledge he provides. In constructing the Pathfinder knowledge map, for example, local knowledge maps helped the expert avoid errors of omission in describing features relevant to lymph-node diagnosis. We shall return to this issue in Chapter 4.

### 2.1.5 Soundness and Consistency: Theoretical Considerations

The proof of soundness is closely related to the development of an algorithm for testing the consistency of a similarity network, and both aspects of the theory are complex. In this section, we examine informally the proof of soundness and issues of consistency to make the rigorous treatments in Chapter 3 more understandable.

At first glance, the soundness result may seem trivial. Consider, for example, the similarity network for three hypotheses $h_1$, $h_2$, and $h_3$, and three nondistinguished variables $x$, $y$, and $z$, shown in Figure 2.6(a). The network contains two local knowledge maps for the hypothesis pairs $\{h_1, h_2\}$ and $\{h_2, h_3\}$. In the local knowledge map for $h_1$ vs $h_2$, all nondistinguished variables are connected to the distinguished node, and, therefore, are included in the map. Similarly, all nondistinguished nodes are included in the local knowledge map for $h_2$ and $h_3$.

The global knowledge map constructed from the two local maps in the similarity network is shown in Figure 2.6(b). The global knowledge map asserts that $z$ is independent of $x$ and $y$, given $h$. This assertion is logically implied by the assertions of conditional independence in the local knowledge maps; hence, the construction of the global knowledge map is sound. To see this fact, let us consider the local knowledge map for $h_1$ and $h_2$. From the definition of missing arcs in a knowledge map, we know that $z$ is independent of $x$ and $y$, given $h$ Formally,

$$p\,(z|x_i, y_j, h_k, \{h_1, h_2\}, \xi) = p\,(z|h_k, \{h_1, h_2\}, \xi) \tag{2.1.1}$$

where $x_i$ and $y_j$ range over the instances of variables $x$ and $y$, and where $h_k$ is equal to $h_1$ or $h_2$. The first term refers to the probability distribution over $z$, given $x_i$, $y_j$, and $h_k$, and given the state of knowledge $\{h_1, h_2\}$ and $\xi$. The second term refers to a similar distribution. In both expressions, the symbol $\xi$ denotes the state of knowledge that an expert brings to bear on the global problem. The set $\{h_1, h_2\}$, which conditions both probabilities, denotes the disjunction of $h_1$ and $h_2$. The disjunction appears in both expressions because, by definition, it is part of the state of knowledge of the local knowledge map. Now the hypothesis $h_1$ alone and the hypothesis $h_2$ alone, logically imply the disjunction of $h_1$ and $h_2$. Consequently, we can omit the disjunction from both sides of Equation 2.1.1 to obtain

$$p\,(z|x_i, y_j, h_k, \xi) = p\,(z|h_k, \xi) \tag{2.1.2}$$

where $x_i$, $y_j$, and $h_k$ range over the same instances as in Equation 2.1.1. Thus, $z$ is independent of $x$ and $y$ given both $h_1$ and $h_2$ alone. Similarly, from the local knowledge map for $h_2$ and $h_3$, we can show that $z$ is independent of $x$ and $y$, given $h_2$ and $h_3$ alone. Combining the observations for both local knowledge maps, we see that $z$ is independent of $x$ and $y$ given any hypothesis of $h$.

Furthermore, suppose the local knowledge maps in Figure 2.6(a) are minimal. From the local knowledge map for $h_2$ and $h_3$, we know $x$ and $y$ are dependent, given that $h_2$ or $h_3$ is true. (If $x$ and $y$ were independent, given $h_2$ and given $h_3$, we could remove the arc from $x$ to $y$, contradicting the minimality of the local knowledge map.) It follows that $x$ and $y$ must be dependent, given $h$, and there must be an arc between $x$ and $y$ in the global knowledge map. We can also show that each arc emanating from $h$ must also be present in the global knowledge map. Thus, the soundness result also holds when the knowledge maps as minimal.

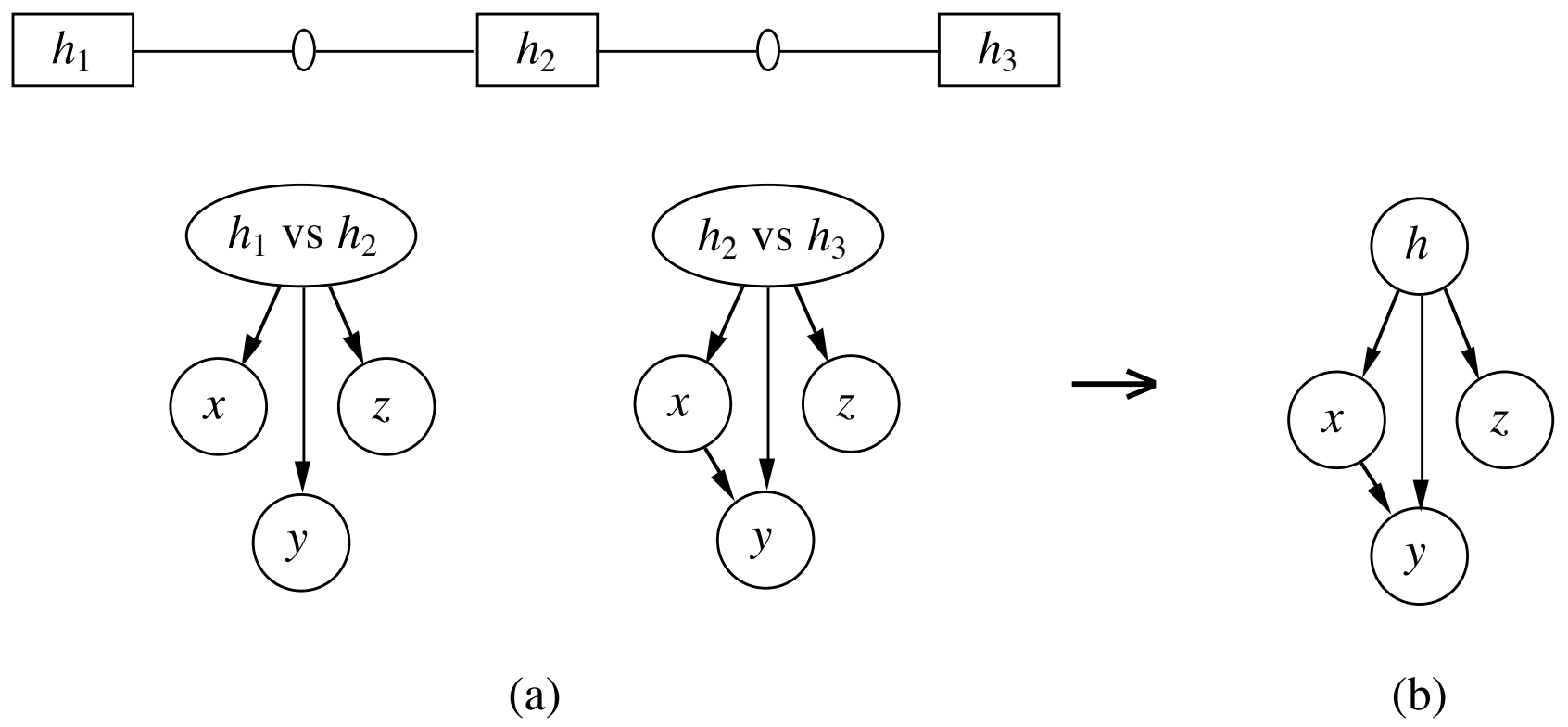

**Figure 2.6:** The construction of a valid global knowledge map.
(a) A similarity network for three hypotheses $h_1$, $h_2$, and $h_3$, and three nondistinguished variables $x$, $y$, and $z$. In the network, there are two local knowledge maps for the hypothesis pairs $\{h_1, h_2\}$ and $\{h_2, h_3\}$. Both local knowledge maps contain all the nondistinguished nodes. (b) The global knowledge map constructed from the similarity network. The assertions of conditional independence in the global knowledge map are logically implied by such assertions of conditional independence in the local knowledge maps. Hence, the construction of the global knowledge map is sound. The construction remains sound when the local knowledge maps and the global knowledge map are minimal. Furthermore, the similarity network is consistent, and thus the global knowledge map is valid.

Finally, consider any probability distribution where (1) $x$, $y$, and $z$ are conditionally independent, given $h_1$ and $h_2$, (2) only $z$ is conditionally independent of $x$ and $y$, given $h_3$, and (3) $x$, $y$, and $z$ are dependent on $h$. This distribution satisfies all the assertions of conditional independence and dependence implied by the two local knowledge maps. Consequently, the similarity network in in Figure 2.6 is consistent, and the global knowledge map in the figure is valid.

In general, if each local knowledge map contains the same set of nondistinguished nodes, the proof of soundness is straightforward. When nodes are omitted from local knowledge maps, however, the proof is not so simple. Consider, for example, the similarity network for three hypotheses and three nondistinguished variables shown in Figure 2.7(a). The local knowledge map for $h_1$ and $h_2$ is identical to the corresponding map in Figure 2.6(a), whereas the the local knowledge map for $h_2$ and $h_3$ is different. In particular, the nodes $x$ and $y$ are disconnected from $h$ in the local knowledge map for $h_2$ and $h_3$. Suppose the local knowledge map for $h_2$ and $h_3$ is minimal. In this case, we know that $x$ and $y$ are dependent, given $h_2$ or given $h_3$. Thus, $x$ and $y$ are dependent, given $h$. However, when we compose this local knowledge map using the procedure described previously, we omit the nodes $x$ and $y$ from the map (indicated by shading in the figure) because these nodes are disconnected from $h$. Consequently, the dependency does not get recorded in the global knowledge map, shown in Figure 2.7(b).

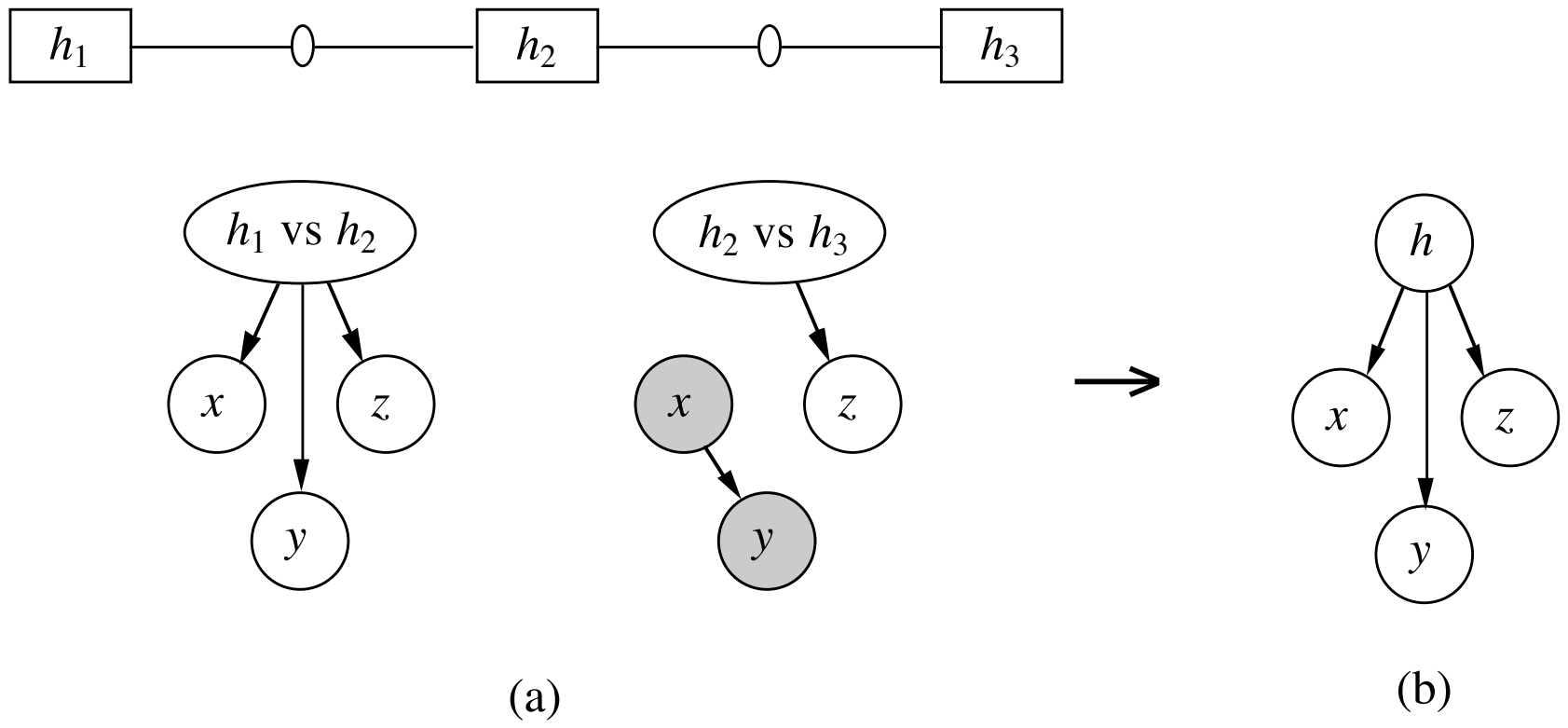

**Figure 2.7:** The construction of an invalid global knowledge map.
(a) A similarity network for three hypotheses and three nondistinguished variables. In the local knowledge map for the hypothesis set $\{h_2, h_3\}$, the nodes $x$ and $y$ (shaded) are not connected to $h$, and are omitted from the local knowledge map. (b) The global knowledge map constructed from the similarity network. If the local knowledge maps are minimal, we know that $x$ and $y$ are dependent given $h_2$ or $h_3$. Because these nodes have been omitted from the local knowledge map, however, this dependency is not recorded in the global knowledge map. This problem occurs because the similarity network is inconsistent.

We could avoid this problem by including all nondistinguished variables in each map. Then, in this example, the dependency between $x$ and $y$ would be registered in the global knowledge map. This alternative, however, would destroy the benefits of the similarity-network representation discussed in Section 2.1.4.

Fortunately, we do not have to abandon the original procedure for composing local knowledge maps to guarantee soundness. As we shall see in the following paragraph, the difficulty in this example does not come from lack of soundness, but rather from the fact that the similarity network in Figure 2.7(a) is inconsistent. That is, there is no joint distribution over the variables $h$, $x$, $y$, and $z$ that satisfies the conditional independence and dependence assertions implied by those local knowledge maps. Thus, the construction is sound, because we can derive any set of dependence and independence assertions from a contradiction, but the global knowledge map is invalid. When we make the similarity network in Figure 2.7 consistent, assuming there is a dependence between $x$ and $y$ in the local knowledge map for $h_2$ and $h_3$, we must add an arc from $x$ to $y$ in the local knowledge map for $h_1$ and $h_2$. Once this is done, the dependency between these nodes is registered appropriately in the global knowledge map. Consequently, the construction remains sound, and the global knowledge map becomes valid.

To see that the similarity network in Figure 2.7 is inconsistent, let us first examine the local knowledge map for $h_2$ and $h_3$, including in this map the nodes $x$ and $y$. From the definition of missing arcs in a knowledge map, we get

$$p\,(y|x_k, h_l, \{h_2, h_3\}, \xi) = p\,(y|x_k, \{h_2, h_3\}, \xi) \tag{2.1.3}$$

where $x_k$ ranges over the possible instances of the variable $x$, and where $h_l$ is equal to $h_2$ or $h_3$. Now $h_2$ and $h_3$ alone logically imply the disjunction of $h_2$ and $h_3$. Therefore, we can remove this disjunction from the left-hand side of Equation 2.1.3 to obtain

$$p\,(y|x_k, h_l, \xi) = p\,(y|x_k, \{h_2, h_3\}, \xi) \tag{2.1.4}$$

where $x_k$ and $h_l$ have the same ranges as in Equation 2.1.3. Also, we have assumed that the local knowledge maps are minimal. Thus, $x$ and $y$ are dependent in the local knowledge map for $h_2$ and $h_3$, and we know that

$$p\,(y|x_i, \{h_2, h_3\}, \xi) \neq p\,(y|x_j, \{h_2, h_3\}, \xi) \tag{2.1.5}$$

for some instances $x_i \neq x_j$. From Equations 2.1.4 and 2.1.5, we get

$$p\,(y|x_i, h_l, \xi) \neq p\,(y|x_j, h_l, \xi) \tag{2.1.6}$$

for $h_l = h_2$ and $h_l = h_3$. Therefore, the local knowledge map for $h_2$ and $h_3$ dictates that $x$ and $y$ are dependent, given both $h_2$ and $h_3$ separately. The local knowledge map for $h_1$ and $h_2$, however, implies that $x$ and $y$ are independent, given $h_1$ and given, $h_2$. We thus obtain the contradiction that $x$ and $y$ are both conditionally independent and dependent, given $h_2$.

In Chapter 3, we see that the situation described in the previous paragraphs holds in general. That is, if a set of minimal local knowledge maps is consistent, and the other constraints discussed in Section 2.1.2 are satisfied, then arcs between nodes that are omitted from one local knowledge map must appear in other local knowledge maps. Thus, the arcs appear in the global knowledge map, and the construction of this map is sound. We can extend this argument to include nonminimal local knowledge maps as well. In addition, we see that we easily can identify and correct inconsistencies in a set of local knowledge maps. Consequently, we can omit nodes from the local knowledge maps, and thereby retain the benefits of similarity networks for knowledge acquisition discussed in Section 2.1.4.

### 2.1.6 Assertions of Asymmetric Conditional Independence

A similarity network derives its power from its ability to represent assertions of conditional independence that are not conveniently represented in an ordinary knowledge map. In fact, a similarity network can represent two specific forms of such conditional independence.

To illustrate the first of these assertion types, let $h_{\subseteq}$ denote a proper subset of the hypotheses of $h$. If $h$ and $x$ are independent, given that one of the elements of $h_{\subseteq}$ is true, we say that $x$ *is not relevant to* $h_{\subseteq}$. Formally, we have the following definition.

**Definition 2.1** *A variable $x$* **is not relevant to** *the set $h_{\subseteq}$, given a state of knowledge $\xi$, if and only if*

$$p\left(h_i|x_j, h_{\subseteq}, \xi\right) = p\left(h_i|h_{\subseteq}, \xi\right) \tag{2.1.7}$$

*for all instances $x_j$ of variable $x$, and for all hypotheses $h_i$ in $h_{\subseteq}$.*

As in the previous section, the set $h_{\subseteq}$, which conditions both probabilities, denotes the disjunction of its elements. We call the form of conditional independence represented by Equation 2.1.7 *subset independence.* Using Bayes' theorem, we can derive an equivalent criterion for subset independence (see Appendix B.1).

**Theorem 2.1** *The feature $x$ is not relevant to the set of hypotheses $h_{\subseteq}$, given a state of knowledge $\xi$, if and only if*

$$p(x|h_i, \xi) = p(x|h_j, \xi) \tag{2.1.8}$$

*for all pairs $h_i, h_j \in h_{\subseteq}$.*

We shall return to Equation 2.1.8 when we discuss probability assessment in the following section.

Now suppose there is no arc from $h$ to $x$ in the local knowledge map for $h_i$ and $h_j$. That is, suppose $x$ and $h$ are independent in the state of information $\{h_i, h_j\}$. By definition, we know that $x$ is not relevant to $\{h_i, h_j\}$. Alternatively, suppose that $x$ is omitted from this local knowledge map. From the definition of local knowledge map, we know that there would be no path from $h$ to $x$, if $x$ were included in the map. Consequently, $x$ and $h$ must be independent in the state of information $\{h_i, h_j\}$. Again, it follows that $x$ is not relevant to $\{h_i, h_j\}$. In either case, using a similarity network, we can represent assertions of subset independence.

To illustrate the second form of conditional independence that we can encode in local knowledge maps, let us consider the similarity network in Figure 2.6. Using arguments similar to those in the previous section, we can show that $x$ and $y$ are dependent given $h_3$, but $x$ and $y$ are independent, given $h_1$ and $h_2$. Thus, a similarity network can represent assertions of conditional independence that are specific to individual hypotheses. We call this form of conditional independence *hypothesis-specific independence*.

Subset independence refers to relationships between the distinguished node and nondistinguished nodes. In contrast, hypothesis-specific independence refers to relationships among nondistinguished nodes. Nonetheless, both forms of independence are closely related in that they are *asymmetric*. In general, an assertion of conditional independence is asymmetric if it holds for only some instances of its variables. Assertions of subset independence and hypothesis-specific independence, in particular, hold for only proper subsets of the distinguished node.

As mentioned previously, we cannot easily encode assertions of asymmetric conditional independence in an ordinary knowledge map. In Section 2.2.4, we discuss this observation in detail.

## 2.2 Partitions: The Assessment of a Knowledge Map

Once the global knowledge map has been constructed, there are several alternative techniques for assessing the map. One approach is to assess directly conditional distributions for each variable in the global knowledge map. This approach is straightforward, but it does not take advantage of additional assertions of conditional independence represented in the similarity network.

In this section, we examine how we can use assertions of subset independence to simplify assessment. In Section 3.10, we discuss how we can exploit both subset independence and hypothesis-specific independence to facilitate assessment. We postpone the latter discussion because it requires some of the technical machinery that will be developed in Chapter 3.

### 2.2.1 Use of Similarity Networks for Assessment

There are two approaches available in SimNet for assessing a knowledge map that exploit assertions of subset independence. In one approach, only those assertions of subset independence represented by nodes missing from the local knowledge maps are employed. In another method, judgments of subset independence embodied by partitions are used. The first approach is illustrated in Figure 2.8 for the feature QUALITY OF VOICE. The rounded rectangle labeled with the feature name contains the mutually exclusive and exhaustive instances of the feature: NORMAL and MUFFLED. The two numbers under each disease are the probability distribution for the feature given that disease. For example, the probability that QUALITY OF VOICE is NORMAL, given STREP THROAT, is 0.9.

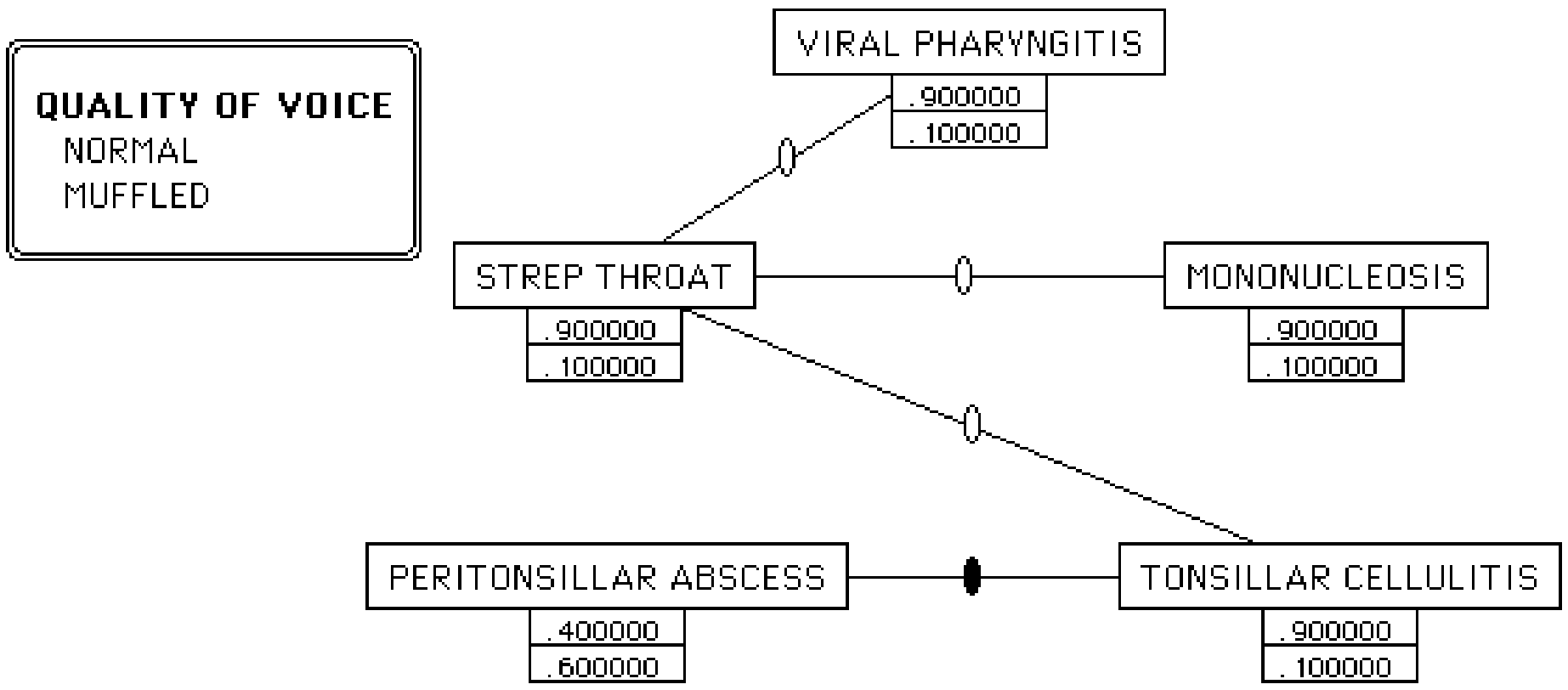

**Figure 2.8:** Probability assessment using a similarity network.
The probability distributions for the feature QUALITY OF VOICE given disease are shown. The rounded rectangle labeled with the feature name contains the mutually exclusive and exhaustive instances of the feature: NORMAL and MUFFLED. The numbers below each disease node are the probability distribution for QUALITY OF VOICE given that disease. The white ovals on the edges reflect the fact that the feature is absent in the corresponding local knowledge map. Conversely, the black oval signifies that the feature is present in the local knowledge map. Distributions bordering an edge with a white oval must be equal.

The black oval on the edge between PERITONSILLAR ABSCESS and TONSILLAR CELLULITIS reflects the fact that the feature QUALITY OF VOICE is present in the local knowledge map for the disease pair. Conversely, the white ovals on the remaining edges represent the fact that this feature is absent from the other local knowledge maps. As shown in the figure, when a feature is omitted from a local knowledge map, the conditional probability distributions on either side of an edge are equal. This observation follows from Theorem 2.1 and the fact that any feature omitted from a local knowledge map cannot be relevant to the two diseases associated with that map. Consequently, for the feature QUALITY OF VOICE, we need to assess probability distributions given only PERITONSILLAR ABSCESS and TONSILLAR CELLULITIS. SimNet automatically propagates the probability distribution for TONSILLAR CELLULITIS throughout the remainder of the similarity graph.

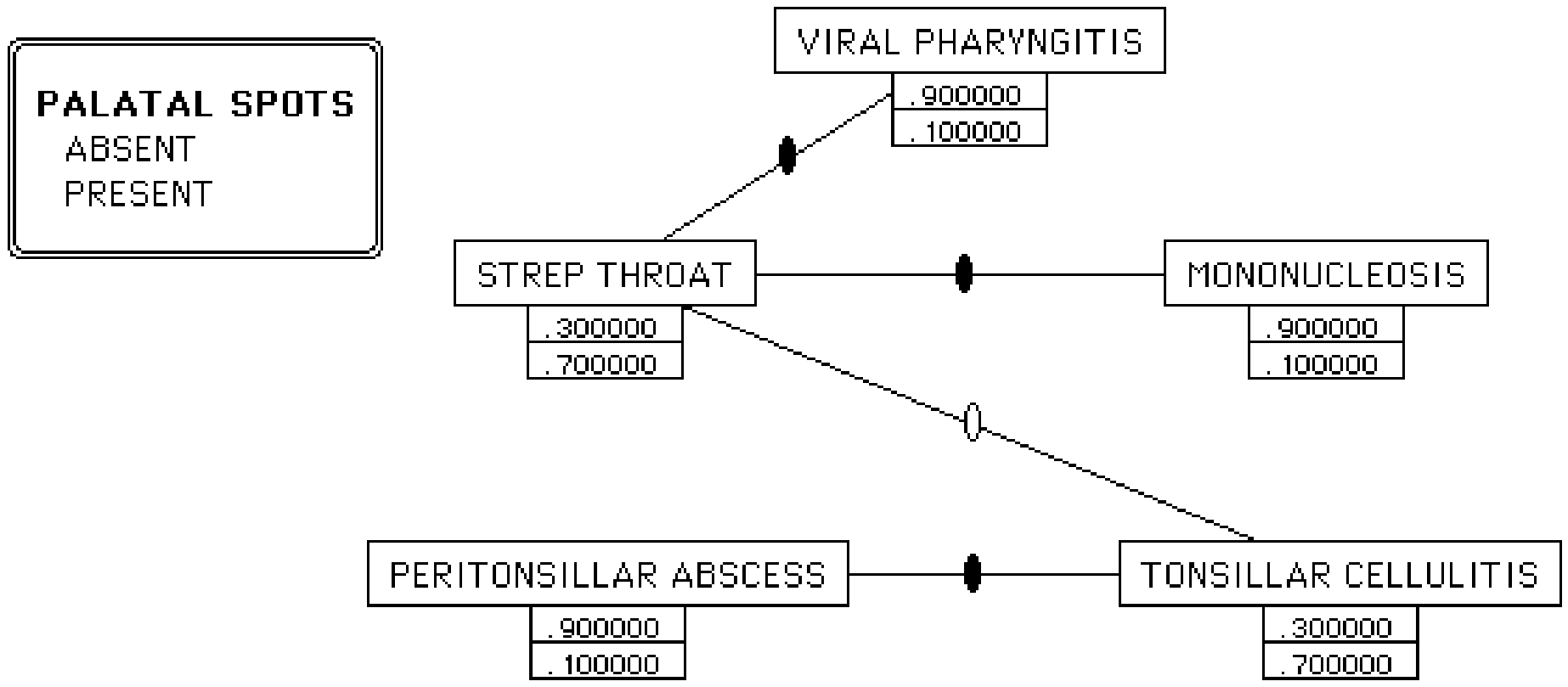

**Figure 2.9:** Hidden equivalence in a similarity network.
The assessment of the feature PALATAL SPOTS is shown. Although the distributions for VIRAL PHARYNGITIS, MONONUCLEOSIS, and PERITONSILLAR ABSCESS are equal, these equalities are hidden until the actual assessments are made, because there are no edges that directly connect any pair of these diseases in the similarity graph.

### 2.2.2 Use of Partitions for Assessment

A problem with this approach to assessment is illustrated in Figure 2.9. Specifically, the probability distributions for the feature PALATAL SPOTS given VIRAL PHARYNGITIS, MONONUCLEOSIS, and PERITONSILLAR ABSCESS are equal. Because we did not connect these diseases in the similarity graph, however, the equality of these distributions remains hidden until we assess the actual probabilities.

We can remedy this difficulty by composing a local knowledge map for every pair of diseases. For domains with more than just a few diseases or hypotheses, however, this alternative is impractical. Alternatively, we can compose a *partition* of the hypotheses for each nondistinguished variable to be assessed. In composing a partition, we place each hypothesis or instance of the distinguished variable into one and only one set. We place two or more hypotheses in the same set only if the nondistinguished variable associated with the partition *is not relevant to* those hypotheses in the set (see Definition 2.1). After composing the partition for a given nondistinguished variable, we assesses probability distributions for the variable given each hypothesis. By Theorem 2.1, however, we need to assess only one distribution for each set in the partition.

A partition for the feature PALATAL SPOTS is shown in Figure 2.10. In this partition, the possible causes of sore throat are divided into two groups: those diseases in which palatal spots are likely to be seen (STREP THROAT and TONSILLAR CELLULITIS), and those diseases for which palatal spots are not likely to be seen (MONONUCLEOSIS, PERITONSILLAR ABSCESS, and VIRAL PHARYNGITIS). The partition reflects the assertions that (1) the feature PALATAL SPOTS *is not relevant to* the disease pair STREP THROAT and TONSILLAR CELLULITIS, and (2) the feature is *not relevant to* the disease triplet MONONUCLEOSIS, PERITONSILLAR ABSCESS, and VIRAL PHARYNGITIS. Consequently, we need to assess only two probability distributions. These distributions, shown below the hypothesis sets in Figure 2.10, are the same as those shown in Figure 2.9. By using this partition, however, we uncover equalities among the distributions for PALATAL SPOTS before we assess probabilities; we thereby avoid the assessment of three additional distributions.

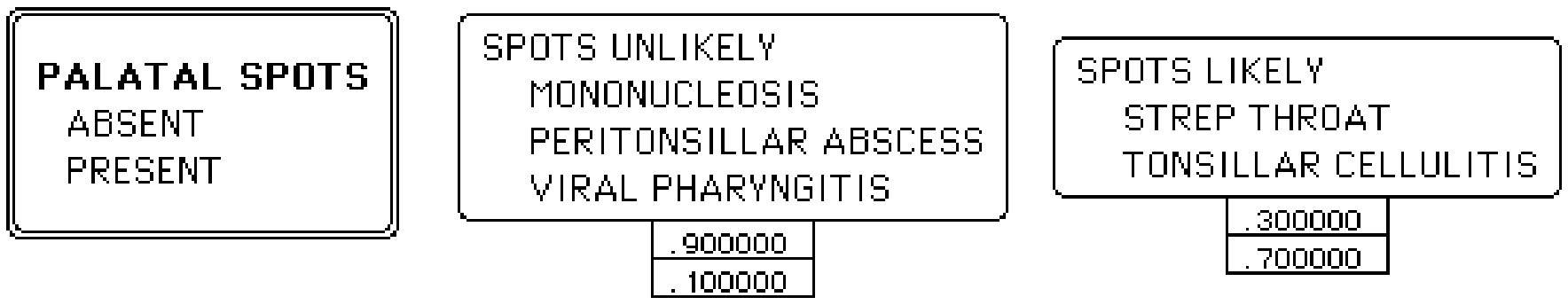

**Figure 2.10:** Assessment of probabilities using a partition.
The partition contains two sets of diseases labeled SPOTS UNLIKELY and SPOTS LIKELY. The partition reflects the assertion that PALATAL SPOTS is relevant to neither the diseases STREP THROAT and TONSILLAR CELLULITIS, nor the diseases MONONUCLEOSIS, PERITONSILLAR ABSCESS, and VIRAL PHARYNGITIS. Consequently, only two probability distributions are assessed.

When a feature is dependent on other features, we can compose a partition for each instance of the set of conditioning features.[3] Figure 2.11 illustrates this approach for the feature TOXIC APPEARANCE, which is conditioned by the features FEVER and ABDOMINAL PAIN. In the figure, partitions and assessments for three of the six instances of the feature's conditioning variables are shown. The three partitions in the figure correspond to the cases where a patient has no abdominal pain and a mild fever, no abdominal pain and a high fever, and abdominal pain and a high fever. In the first and third partition, the feature TOXIC APPEARANCE is relevant to neither PERITONSILLAR ABSCESS and TONSILLAR CELLULITIS, which are two localized diseases (diseases that tend to affect only a small area of the throat—usually one or more tonsils), nor MONONUCLEOSIS and STREP THROAT, which are two diseases that tend to affect a large area of the throat as well as other organs in the body. In the second partition, the feature is not relevant to the two localized diseases, but is relevant to the remaining diseases.

[3]In general, an instance of a set of variables is an assignment of an instance to each variable in that set. Thus, if variable $x$ has instances $x_1$ and $x_2$, and variable $y$ has instances $y_1$ and $y_2$, then the pairs $(x_1, y_1)$, $(x_1, y_2)$, $(x_2, y_1)$, $(x_2, y_2)$ comprise all instances of the set $\{x, y\}$.

In general, partitions are better able to express assertions of subset independence for assessment than are similarity networks. To understand this point, consider all possible similarity networks that an expert might construct for a given domain. The similarity network that can represent the most assertions of subset independence is the one that has a completely connected similarity graph (i.e., a similarity graph in which all hypothesis pairs are connected directly). From the assertions of subset independence encoded in this similarity network, we can derive a single partition for each feature. Specifically, we place hypotheses $h_i$ and $h_j$ in the same set for variable $x$ if and only if the variable $x$ is not present in the local knowledge map for $h_i$ and $h_j$. We cannot, however, derive different partitions for different instances of the features that condition $x$.

### 2.2.3 Partitions and Classification Hierarchies

As we see in Chapter 4, the use of partitions can decrease the time to assess a knowledge map by more than a factor of five. At first, this observation may seem surprising, given that a partition must be composed for each conditioning instance of every feature. In the medical domains that I have investigated, however, two factors have contributed to the efficiency of the approach. First, the task of composing a single partition is straightforward. Apparently, as is the case with assertions of symmetric conditional independence, people find it easy to make judgments of subset independence without assessing the probabilities underlying such judgments.

Second, partitions often are identical or related from one feature to another, and more often are identical or related among the conditioning instances of the same features. For example, two of the partitions for TOXIC APPEARANCE in Figure 2.11 are identical, and these two partitions are closely related to the third. The partition for FEVER, shown in Figure 2.12, is almost identical to the partitions for TOXIC APPEARANCE.

The partitions for FEVER and those for TOXIC APPEARANCE are related in that each represents a slice through the same *classification hierarchy* of diseases. This classification hierarchy is shown in Figure 2.13. In the figure, diseases that cause sore throat are divided into two major categories: diseases that tend to affect a large area of the throat (DIFFUSE DISEASE) and diseases that tend to affect a small area of the throat (LOCALIZED DISEASE). The diseases in the former category are subdivided into those diseases that tend to affect multiple organs (MULTIORGAN DISEASE) and the disease VIRAL PHARYNGITIS, which does not affect multiple organs. The partition for FEVER represents a slice through this hierarchy at the level of abstraction DIFFUSE versus LOCALIZED DISEASE. The first and third partitions for TOXIC APPEARANCE in Figure 2.11 represent the slice through the hierarchy that is one level more specific for diffuse diseases. The second partition for TOXIC APPEARANCE reflects the slice that is the most specific for diffuse diseases.

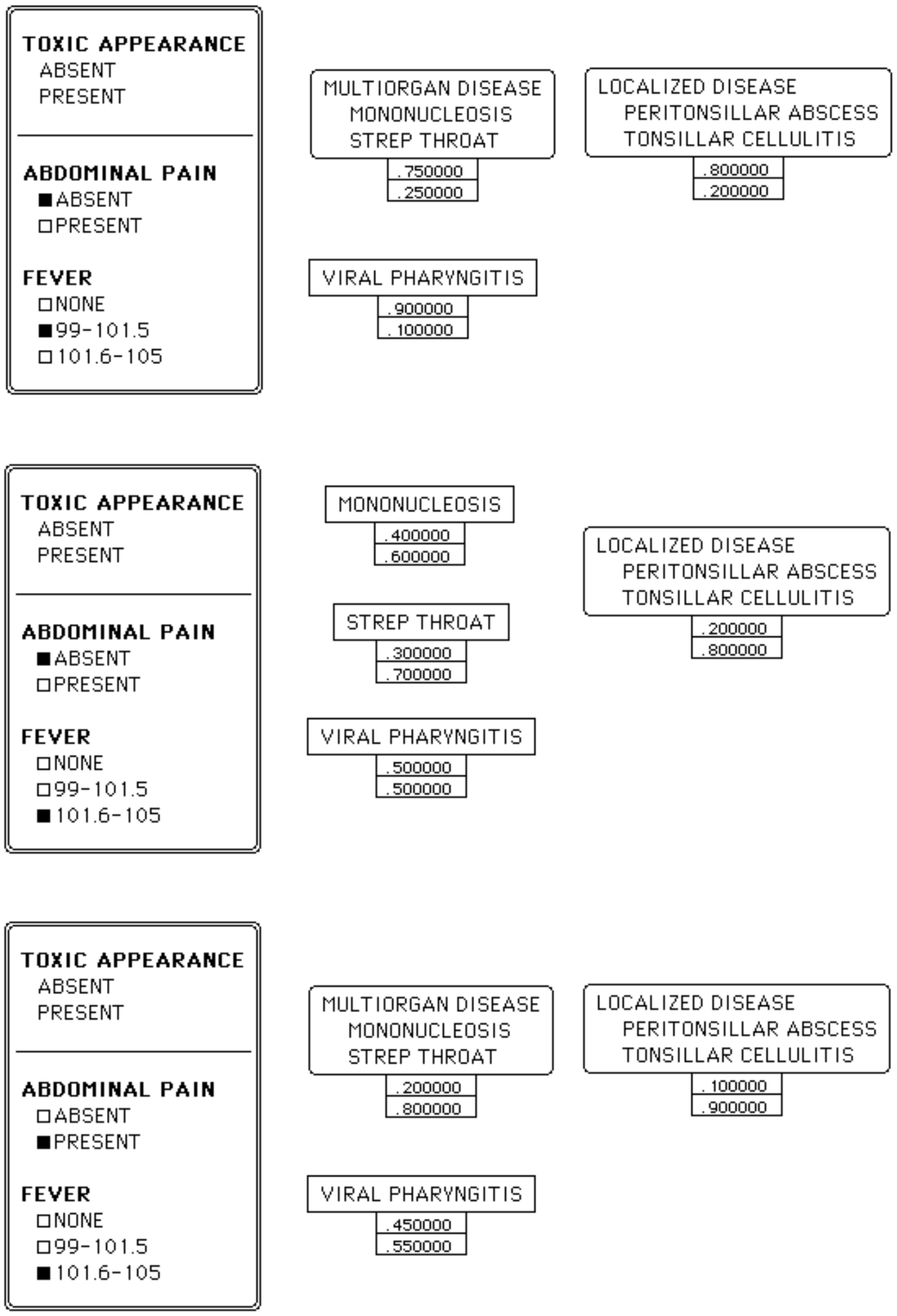

**Figure 2.11:** Assessing dependent features using partitions.
The figure contains partitions and associated assessments for three of the six conditioning instances of the feature TOXIC APPEARANCE. The three partitions correspond to cases where a patient has no abdominal pain and a mild fever, no abdominal pain and a high fever, and abdominal pain and a high fever, respectively. Only assessments for each set in a partition are required. Note that the first and third partitions are identical, and the second is closely related.

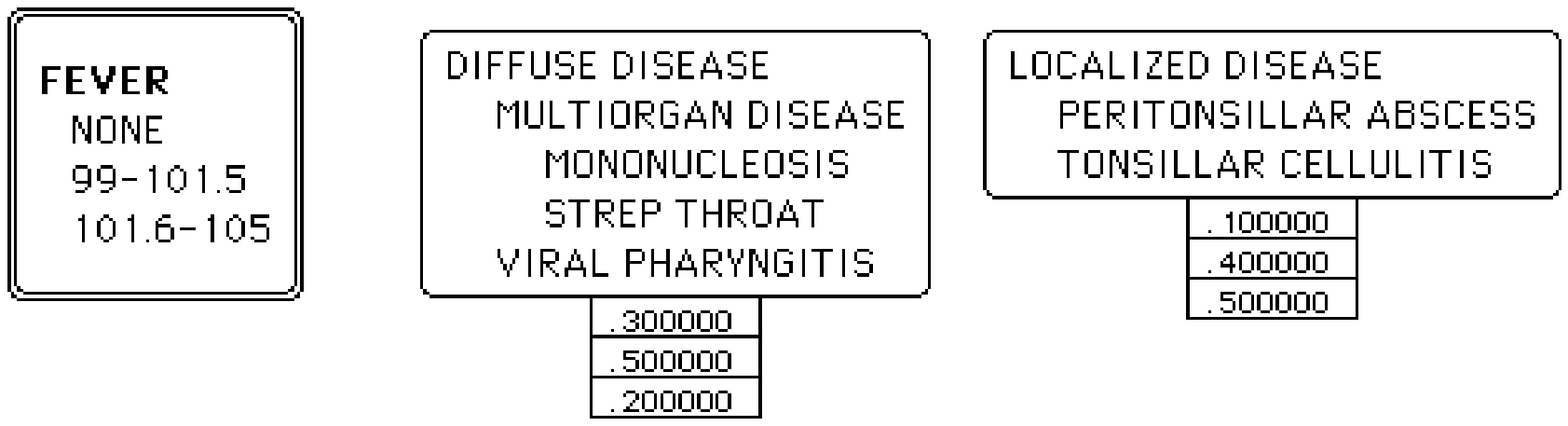

**Figure 2.12:** The partition for FEVER.
The partition is closely related to the partitions for TOXIC APPEARANCE. Note the hierarchical structure of the set DIFFUSE DISEASE. The set contains the disease VIRAL PHARYNGITIS and the set MULTIORGAN DISEASE, which, in turn, contains the diseases MONONUCLEOSIS and STREP THROAT. Hierarchical sets such as this one facilitate the composition of partitions.

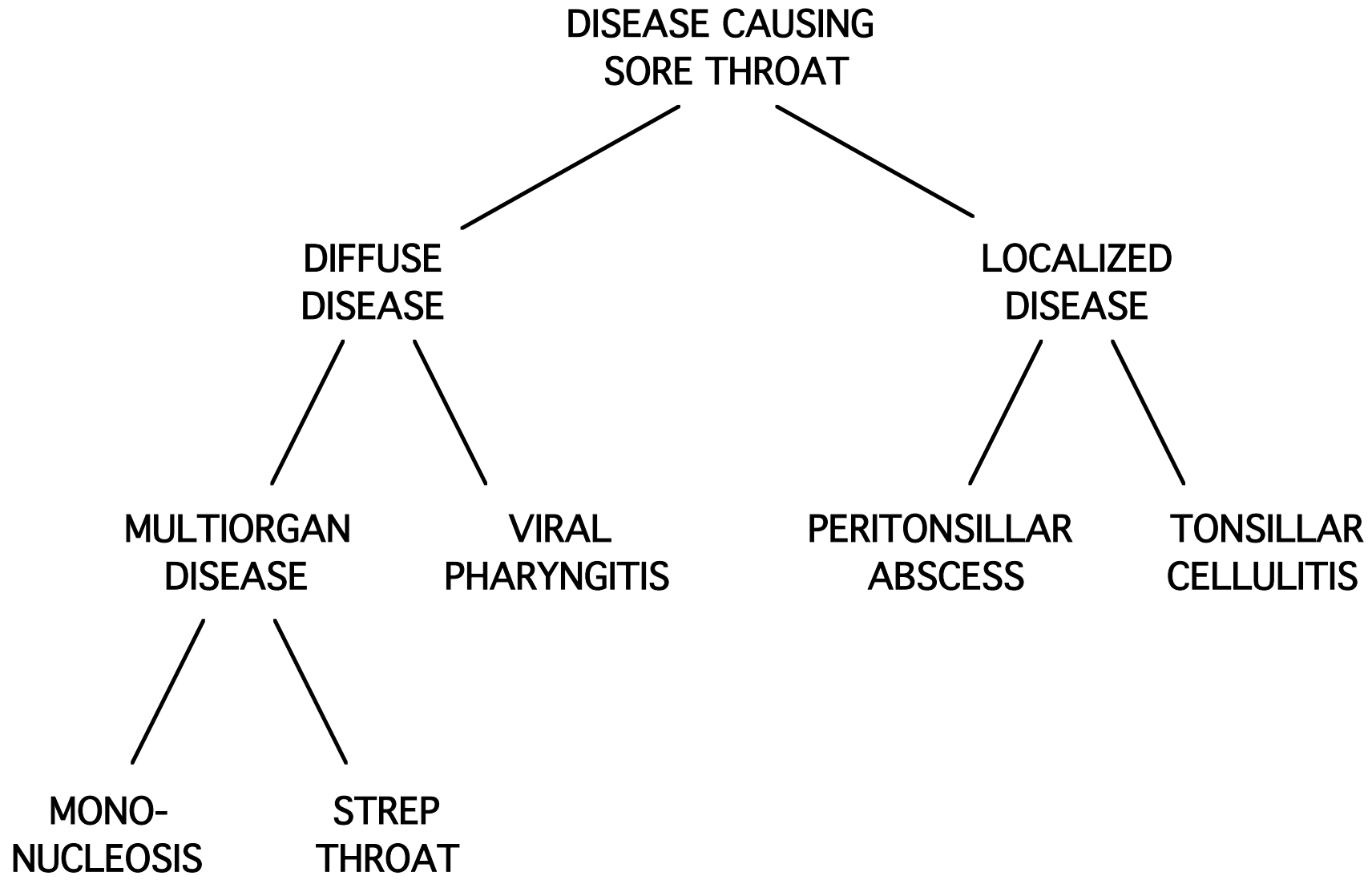

**Figure 2.13:** A classification hierarchy of the diseases causing sore throat.
In the hierarchy, diseases either are or are not localized to one portion of the throat. Two of the diseases that are not localized to one portion of the throat, MONONUCLEOSIS and STREP THROAT, affect multiple organs in the body. The partitions for TOXIC APPEARANCE and FEVER are slices of this hierarchy at different levels of abstraction.

In many domains, experts often fabricate one or more classification hierarchies to describe shared features among groups of hypotheses. In Pathfinder, for example, there are 13 distinct classification hierarchies across the approximately 100 features. In domains where such hierarchies have been explicated, it is likely that partitions will be identical or closely related. When the partitions for two or more features are related, we do not need to create each partition from scratch. Using SimNet, we can copy the structure of a partition for a conditioning instance of a feature, and assign this structure either to the partition of another conditioning instance of the same feature, or to the partition for an entirely different feature. Once the structure of the partition has been copied, we can modify the partition using simple Macintosh-style manipulations to reflect the judgments of subset independence for that feature and conditioning instance. Note that SimNet supports hierarchical set membership within partitions. For example, in Figure 2.12, MONONUCLEOSIS is a member of the set MULTIORGAN DISEASE; this set, in turn, is a member of the set DIFFUSE DISEASE. This feature of the program makes it easy for a user to transform a partition that reflects a slice of a classification hierarchy at one level of abstraction to a partition that represents other levels of abstraction within the hierarchy.

### 2.2.4 Representation of Subset Independence in Ordinary Knowledge Maps

Asymmetrical assertions of conditional independence can be represented in an ordinary knowledge map with deterministic nodes. In this section, we examine the representation of subset independence in a knowledge map, and we discuss the merits of this representation relative to partitions.

Figure 2.14 contains the knowledge map that reflects the judgments of subset independence associated with the partitions for ABDOMINAL PAIN, FEVER, and TOXIC APPEARANCE in the sore-throat example. First, let us consider the feature ABDOMINAL PAIN. In the global knowledge map for the sore-throat problem (Figure 2.1), DISEASE is the only node that conditions ABDOMINAL PAIN. Consequently, there is only one partition for this feature. The sets within the partition for ABDOMINAL PAIN are represented by the instances of the deterministic node PARTITION FOR ABDOMINAL PAIN in the knowledge map. This node is deterministic because the sets in the partition are certain, given the set of possible diseases. The lack of an arc from DISEASE to ABDOMINAL PAIN represents the assertions of subset independence encoded by the partition for that feature. That is, given any set of diseases in the partition, knowing which disease in that set is the cause of the patient's sore throat does not change the probability that the patient will have abdominal pain. These same considerations apply to the feature FEVER.

Now consider the feature TOXIC APPEARANCE. In the global knowledge map, TOXIC APPEARANCE is dependent on both ABDOMINAL PAIN and FEVER. Therefore, the feature has a partition for each instance of these two features. The partitions and the sets within each of these partitions are represented by the deterministic node PARTITIONS FOR TOXIC APPEARANCE. Again, the node is deterministic because the partitions are known with certainty, given instances of DISEASE, ABDOMINAL PAIN, and FEVER. The lack of an arc from DISEASE to TOXIC APPEARANCE represents the assertions of subset independence associated with the partitions for the feature. In particular, given an instance of ABDOMINAL PAIN and FEVER, and a set in the appropriate partition, the probability that the patient's appearance will be toxic is independent of disease.

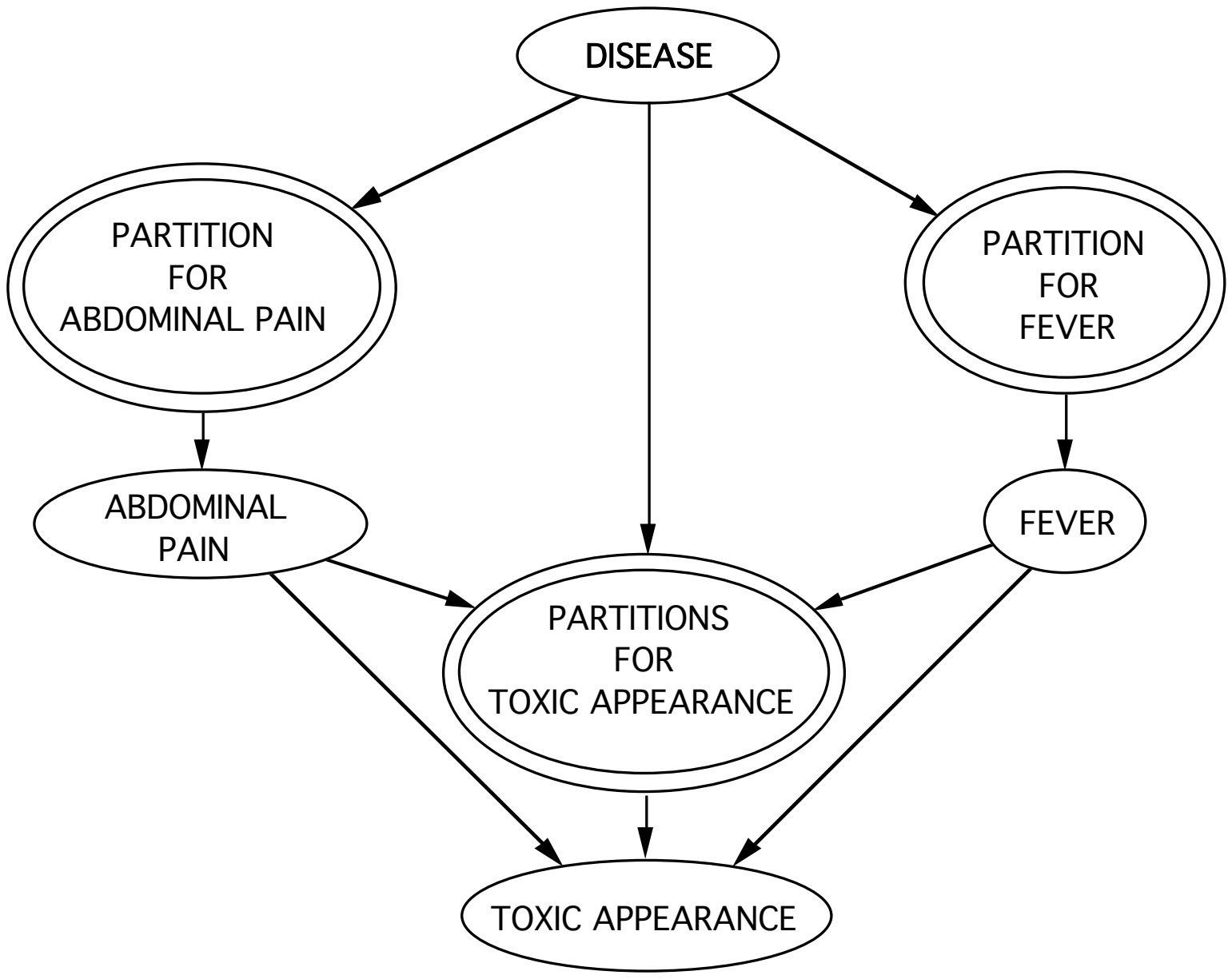

**Figure 2.14:** The representation of partitions in a knowledge map.
This knowledge map encodes the asymmetrical assertions of subset independence that are represented by partitions for the three features ABDOMINAL PAIN, FEVER, and TOXIC APPEARANCE. Because ABDOMINAL PAIN has no conditional predecessors except DISEASE, there is only one partition for the feature. The sets in the partition for ABDOMINAL PAIN are a deterministic function of DISEASE. Furthermore, the probability distribution for ABDOMINAL PAIN is independent of DISEASE, given the elements of the partition for the feature. Similar remarks apply to the feature FEVER. TOXIC APPEARANCE is conditioned both by ABDOMINAL PAIN and by FEVER. Consequently, TOXIC APPEARANCE has a partition for each combination of the instances of these two features. These partitions are a deterministic function of DISEASE, ABDOMINAL PAIN, and FEVER. The probability distributions for TOXIC APPEARANCE are independent of DISEASE, given ABDOMINAL PAIN, FEVER, and the elements of the partition for TOXIC APPEARANCE.

Although this knowledge map accurately encodes the assertions of subset independence for the three features, the representation is cumbersome. The partitions are hidden under the nodes in the knowledge map, and cannot be created or modified with simple graphic manipulations as they are in SimNet. Consequently, the representation of subset independence in ordinary knowledge maps does not facilitate assessment significantly. Similar remarks apply to the representation of hypothesis-specific independence in an ordinary knowledge map.

### 2.2.5 Research Related to Partitions

There is an interesting connection between partitions and the Dempster–Shafer theory of belief. Several leaders in artificial-intelligence research have argued that probability theory is inadequate for reasoning under uncertainty. In providing motivation for the Dempster–Shafer theory, Gordon and Shortliffe argue that probability theory and other theories of uncertainty are inadequate, because they do not allow for the possibility that features or evidence bear on *sets* of diseases or hypotheses (Gordon and Shortliffe, 1985, page 324):

> An advantage of the Dempster–Shafer theory over previous approaches is its ability to model the narrowing of the hypothesis set with accumulation of evidence, a process which characterizes diagnostic reasoning in medicine and expert reasoning in general. An expert uses evidence which may apply not only to single hypotheses but also to sets of hypotheses that together comprise a concept of interest.

Partitions, however, provide an alternative within probability theory for representing evidence that is relevant to sets of hypotheses. In particular, we can interpret the statement that "a piece of evidence applies to only a set of hypotheses" to mean that the evidence *is not relevant to* that set of hypotheses, by Definition 2.1. Furthermore, the probabilistic alternative to Dempster–Shafer theory can be more efficient computationally. Once the probabilities are assessed using partitions, the partitions can be ignored, and the effect of evidence on each hypothesis can be computed separately. Alternatively, the effect of evidence can be accumulated on set intersections of partitions, just as evidence is accumulated within the Dempster–Shafer theory. Here, however, we can accumulate evidence using Bayes' theorem, rather than Dempster's rule of combination. This alternative, when computationally feasible, might be useful for generating cogent explanations for the results of probabilistic inference.

Motivated by the same argument against the adequacy of probability theory, Pearl has developed a representation similar to partitions. In his book, Pearl discusses probability assessments in a domain in which a set of mutually exclusive and exhaustive hypotheses can be organized into what I have been calling a classification hierarchy (Pearl, 1988, pages 333–344). In this context, he examines the statement (my notation):

> Evidence $x$ bears directly on $h_{\subseteq}$, but says nothing about the individual elements of $h_{\subseteq}$.

Pearl interprets this statement to mean

$$p\left(x|h_{\subseteq}, \xi\right) = p\left(x|h_i, \xi\right), \quad h_i \in h_{\subseteq} \tag{2.2.9}$$

$$p\left(x|\bar{h}_{\subseteq}, \xi\right) = p\left(x|h_i, \xi\right), \quad h_i \in \bar{h}_{\subseteq} \tag{2.2.10}$$

where $\bar{h}_{\subseteq}$ is the set of hypotheses in $h$ that are not in $h_{\subseteq}$. Equations 2.2.9 and 2.2.10 imply that $x$ is not relevant to the hypothesis set $h_{\subseteq}$ nor to the hypothesis set $\bar{h}_{\subseteq}$, respectively. Thus, Pearl's method is a special case of the partition approach to assessment in which we compose two-set partitions within a single classification hierarchy.

# 3 Theory of Similarity Networks

In this chapter, a theory for similarity networks is developed. The major results of this work are (1) that the construction of a global knowledge map from a similarity network is sound for strictly positive distributions, (2) that a low-order polynomial algorithm exists for testing the consistency of a similarity network, and (3) that the construction of a global knowledge map from a similarity network is exhaustive. The first two results show that we can construct valid global knowledge maps. Given the third result, we know that any global knowledge map we construct for a given hypothesis set must contain all features that are relevant to the discrimination of that set as a whole.

## 3.1 Background

In this section, we examine the properties of graphs and knowledge maps that we require for the theory of similarity networks. Before introducing additional concepts, however, let us consider some notational conventions. Throughout the chapter, we use a lowercase letter (e.g., $x$, $y$, and $z$) to represent an uncertain variable or a node in a graph that corresponds to that variable. We use an uppercase letter (e.g., $X$, $Y$, $Z$) to represent a set of uncertain variables or a set of nodes in a graph that correspond to that set of variables. In this work, we consider only variables with a finite number of instances; we denote a specific instance of a variable or set of variables by subscripting that variable. For example, $x_i$ refers to the $i$th instance of variable $x$, and $X_i$ refers to the $i$th instance of set $X$. The symbol $\backslash$ denotes set difference. That is, $X \backslash Y$ is the set of variables in $X$ that are not in $Y$. We use $X \backslash x$ as a shorthand for $X \backslash \{x\}$. The expression $p(x_i|X_j, \xi)$ denotes the probability of $x_i$ given $X_j$ assessed by a person with background knowledge $\xi$. When a probabilistic equality holds true for all instances of some variable or set of variables, we omit the universal quantification over that variable or set of variables. For example, the statement

$$p(x|y, \xi) = p(x|\xi)$$

is a shorthand for the statement

$$\forall\ x_i, y_j \quad p(x_i|y_j, \xi) = p(x_i|\xi)$$

In contrast, we omit existential quantifiers associated with probabilistic inequalities. For example, the statement

$$p(x|y, \xi) \neq p(x|\xi)$$

is a shorthand for the statement

$$\exists\ x_i, y_j \quad p(x_i|y_j, \xi) \neq p(x_i|\xi)$$

To avoid confusion, we sometimes express the universal or existential quantifications explicitly. A summary of notation appears at the end of this book.

Next, let us consider some properties of directed and undirected graphs. A *directed graph* consists of a finite collection of nodes and directed arcs. A *node* is a primitive object and a *directed arc* is a line with an arrow on one end connecting two nodes. A *directed path* from a node $x$ to a node $y$ is a sequence of nodes that we can visit in order by moving along the arcs in the direction of the arrows. An *undirected path* is similar to a directed path, except that the direction of arcs is ignored. A *directed (undirected) cycle* is a nontrivial directed (undirected) path that starts and ends with the same node. Two nodes $x$ and $y$ are *connected* if there is some undirected path from $x$ to $y$. A directed graph is *connected* if every pair of nodes is connected. A directed graph is *singly connected* if it contains no undirected cycles, and is *multiply connected* otherwise.

If there is an arc from node $y$ to node $x$, then $y$ is called a *direct* or *conditional predecessor* of node $x$ and $x$ is called a *direct successor* of $y$. In this situation, we also say that $y$ *conditions* $x$. If there is a nontrivial directed path from $x$ to $y$, then $x$ is called a *predecessor* of $y$, and $y$ is called a *successor* of $x$. If there is no nontrivial path from $x$ to $y$, then $x$ is called a *nonsuccessor* of $y$. The terms $C(x)$, $S(x)$, and $\bar{S}(x)$ refer to the conditional predecessors of $x$, the successors of $x$, and the nonsuccessors of $x$, respectively.

An *undirected graph* is similar to a directed graph except that arcs are replaced by undirected *edges*. The definitions for the terms *undirected cycle*, *undirected path*, *connected node pair*, *connected graph*, *singly connected graph*, and *multiply connected graph* in undirected graphs correspond to the definitions for these terms in directed graphs. In addition, we sometimes use $(x, y)$ to represent the undirected edge between nodes $x$ and $y$.

Two or more directed (or undirected) graphs can be combined, provided their nodes are labeled. The *graph union* of graphs $\mathcal{G}_1$ and $\mathcal{G}_2$, denoted $\mathcal{G}_1 \cup \mathcal{G}_2$, is the graph formed by the union of nodes in $\mathcal{G}_1$ and $\mathcal{G}_2$ and the union of arcs (or edges) in $\mathcal{G}_1$ and $\mathcal{G}_2$. That is, a node labeled $x$ is in $\mathcal{G}_1 \cup \mathcal{G}_2$ if and only if a node labeled $x$ is in $\mathcal{G}_1$ or $\mathcal{G}_2$. Also, an arc (or edge) $(x, y)$ is in $\mathcal{G}_1 \cup \mathcal{G}_2$ if and only if the arc (or edge) $(x, y)$ is in either $\mathcal{G}_1$ or $\mathcal{G}_2$.

In addition, we can induce subgraphs on a graph $\mathcal{G}$. Let $N$ be a subset of the nodes in $\mathcal{G}$, and let $E$ be a subset of edges in $\mathcal{G}$. The *node-induced subgraph* of $\mathcal{G}$ given $N$ is the graph containing nodes $N$ and all edges in $\mathcal{G}$ that are bordered by nodes in $N$. The *edge-induced subgraph* of $\mathcal{G}$ given $E$ is the graph containing edges $E$ and all nodes in $\mathcal{G}$ that border edges in $E$.

Let us now examine the knowledge map and some of its fundamental properties.

**Definition 3.1** *A* **knowledge map** *is a directed acyclic graph that captures part of a person's knowledge about some domain. Each node in the knowledge map represents a distinction, variable, or event that may be uncertain. The structure of the graph (i.e., the arcs and missing arcs between nodes) represents the conditional independence assertions*

$$p\left(x|C(x),\xi\right) = p\left(x|\bar{S}(x),\xi\right) \tag{3.1.1}$$

*for all nodes $x$ in the knowledge map. An* **assessed knowledge map** *is a knowledge map in which the distributions $p\left(x|C(x),\xi\right)$, for each node $x$, have been assessed.*

A theorem concerning knowledge maps and conditional independence follows immediately from the definition.

**Theorem 3.1** *For all nodes $x$ in a knowledge map,*

$$p\left(x|C(x),\xi\right) = p\left(x|\bar{S}^{'}(x),\xi\right) \tag{3.1.2}$$

*where $\bar{S}^{'}(x)$ is any subset of $\bar{S}(x)$ that includes $C(x)$.*

**Proof:** Let $S^{''}(x) = \bar{S}^{'}(x)\backslash C(x)$ and $S^{'''}(x) = \bar{S}(x)\backslash \bar{S}^{'}(x)$. The sets $C(x)$, $S^{''}(x)$, and $S^{'''}(x)$ are disjoint and their union is $\bar{S}(x)$. Thus, we can write

$$\begin{aligned} p\left(x|\bar{S}^{'}(x),\xi\right) &= \sum_{S_i^{'''}(x)} p\left(x|C(x),S^{''}(x),S_i^{'''}(x),\xi\right) \quad p\left(S_i^{'''}(x)|C(x),S^{''}(x),\xi\right) \\ &= \sum_{S_i^{'''}(x)} p\left(x|C(x),\xi\right) \quad p\left(S_i^{'''}(x)|C(x),S^{''}(x),\xi\right) \\ &= p\left(x|C(x),\xi\right) \end{aligned}$$

where $\sum_{S_i^{'''}(x)}$ denotes the sum over all instances of $S^{'''}(x)$. The first line is the expansion rule for probabilities (see Appendix A.1.3). The second line follows from the first by the definition of conditional independence in a knowledge map, Equation 3.1.1. The last line follows from the second because $p\left(x|C(x),\xi\right)$ does not depend on $S^{'''}(x)$. □

A knowledge map represents assertions of conditional independence in addition to those assertions delineated in Theorem 3.1. In particular, let $X$, $Y$, and $Z$ be three disjoint subsets of nodes in a knowledge map. We say that $Z$ *d-separates* $X$ from $Y$, if there is no path between a node in $X$ and a node in $Y$ along which (1) every node with converging arcs is in $Z$ or has a successor in $Z$, and (2) every other node is outside $Z$ (see Pearl, 1988, for examples). Pearl (1986) states without proof that, if $Z$ d-separates $X$ from $Y$, then $X$ and $Y$ are conditionally independent, given $Z$. Verma and Pearl (1988) prove this result. In addition, Geiger and Pearl (1988) prove that the d-separation criterion delineates all assertions of conditional independence that we can represent in a knowledge map.[1] Readers who want to gain a thorough understanding of the independence assertions represented by a knowledge map should see Pearl (1988, Chapter 3). For the development here, Theorem 3.1 is adequate.

Let us now consider the relationship between a knowledge map and the joint probability distribution over the variables in that knowledge map. Given a set of $n$ uncertain variables and background knowledge $\xi$, an *expansion* of the joint distribution for the variables is one of the $n!$ factorizations of the joint distribution that we can write using the product rule. For example,

$$p(x, y, z|\xi) = p(x|\xi) \; p(y|x, \xi) \; p(z|x, y, \xi) \tag{3.1.3}$$

$$p(x, y, z|\xi) = p(y|\xi) \; p(z|y, \xi) \; p(x|y, z, \xi) \tag{3.1.4}$$

are two expansions for the set of variables $\{x, y, z\}$. Any expansion of a set of variables induces a total ordering on the variables, which we call an *expansion order* and denote $\leq_E$. We write $x \leq_E y$ if and only if variable $x$ occurs in the conditioning events of variable $y$. The set containing all nodes less than $x$ in the expansion order is denoted $\leq_E(x)$. A knowledge map for a set of variables also defines an ordering, called $\leq_C$, over those variables. In particular, we say that $x \leq_C y$ if and only if there is a (possibly empty) directed path from $x$ to $y$ in the graph. Since the graph contains no directed cycles, $\leq_C$ is a partial ordering. We say that an expansion order $\leq_E$ is *consistent* with $\leq_C$ if $x \leq_C y$ implies $x \leq_E y$. Given these definitions, we can now use Theorem 3.1 to derive a fundamental property of knowledge maps, which was first stated by Howard and Matheson (1981).

**Theorem 3.2** *An assessed knowledge map determines a unique joint distribution over its variables.*

**Proof:** Let $\leq_E$ be any expansion order that is consistent with the partial ordering $\leq_C$ associated with a given knowledge map. From the definition of expansion order, we have

$$p(X|\xi) = \prod_{x \in X} p(x| \leq_E(x), \xi) \tag{3.1.5}$$

[1] This result applies only to knowledge maps that do not contain deterministic nodes. Geiger et al. (1990) discuss a modified criterion, called D-separation, for knowledge maps with deterministic nodes.

where $X$ is the set of all variables in the knowledge map and $p(X|\xi)$ denotes the joint distribution over the variables. By definition of $\leq_C$, it follows that $\leq_E(x)$ is a subset of $\bar{S}(x)$ that includes $C(x)$. Therefore, using Theorem 3.1, we can rewrite Equation 3.1.5 as

$$p(X|\xi) = \prod_{x \in X} p(x|C(x), \xi) \tag{3.1.6}$$

The terms in Equation 3.1.6 are exactly the distributions associated with the assessed knowledge map. □

In light of this proof, we see that a knowledge map implicitly restricts the possible expansions that can be used to construct the joint distribution over the variables in the knowledge map directly from the distributions associated with the variables. In particular, only those expansion orders that are consistent with $\leq_C$ can be used. For the sake of brevity, when an expansion order is consistent with the partial order $\leq_C$ of a given knowledge map, we shall say the the expansion order is *consistent with the knowledge map* itself.

It is often cumbersome to verify Equation 3.1.1 for every node in a knowledge map. The following theorem shows that such comprehensive testing usually is not necessary.

**Theorem 3.3** *Let $\leq_E$ be some expansion order that is consistent with a knowledge map. Then*

$$p(x|C(x), \xi) = p\left(x|\bar{S}(x), \xi\right), \quad \text{for all } x \tag{3.1.7}$$

*if and only if*

$$p(x|C(x), \xi) = p(x| \leq_E(x), \xi), \quad \text{for all } x \tag{3.1.8}$$

**Proof:** Equation 3.1.8 follows from Equation 3.1.7 by Theorem 3.1. For a proof of the converse, see Verma and Pearl (1988) or Shachter (1990). □

The theorem says that, to verify the assertions of conditional independence implied by a knowledge map, a person needs only to check that, for *some* expansion order, the conditional predecessors of each node $x$ render $x$ independent of all other nodes that precede $x$ in the expansion order. The theorem also simplifies many of the derivations in this chapter.

In the definition of knowledge maps, the *absence* of arcs play a critical role. For example, given any knowledge map, we can always add arcs to the graph and not introduce erroneous assertions of conditional independence. However, it is often useful to know that the arcs in a knowledge map reflect actual dependencies. To address this issue, let us consider the concepts of superfluous arc and minimal knowledge map.

**Definition 3.2** *Given the knowledge map of an individual, an arc in that map is* **superfluous** *if and only if its removal from the map does not introduce conditional independencies that contradict the assertions of the individual. A knowledge map is* **minimal** *if and only if it contains no superfluous arcs.*

Each arc in a minimal knowledge map thus represents assertions of conditional dependence. The following theorem provides a simple procedure for identifying superfluous arcs, and hence a simple procedure for composing a minimal knowledge map.

**Theorem 3.4** *An arc from $x$ to $y$ is superfluous if and only if*

$$p\,(y|C(y),\xi) = p\,(y|C(y)\backslash x,\xi) \tag{3.1.9}$$

*for all instances of $C(y)$.*

**Proof** (only if): In the graph without the arc from $x$ to $y$, $x$ must be a nonsuccessor of $y$. Thus, Equation 3.1.9 follows from Theorem 3.1.

**Proof** (if): Let $\leq_E$ be any expansion order that is consistent with the original knowledge map. We know that

$$p\,(y|C(y),\xi) = p\,(y|\leq_E(y),\xi) \tag{3.1.10}$$

Combining Equations 3.1.9 and 3.1.10, we obtain

$$p\,(y|C(y)\backslash x,\xi) = p\,(y|\leq_E(y),\xi) \tag{3.1.11}$$

Because the assertions of conditional independence

$$p\,(z|C(z),\xi) = p\,(z|\leq_E(z),\xi)$$

for $z \neq y$ are unaffected by the removal of the arc from $x$ to $y$, it follows from Equation 3.1.11 and Theorem 3.3 that the arc from $x$ to $y$ can be removed. □

Because minimal maps contain more information than do nonminimal maps, and because they can be constructed easily from nonminimal maps using the criterion in Theorem 3.4, minimal maps are almost always composed in practice. We shall see that minimal knowledge knowledge maps play an important role in the theory of similarity networks.

Theorem 3.4 says that, if an arc from $x$ to $y$ is nonsuperfluous, then there must be some instance of $C(y)$ such that

$$p\left(y|C(y),\xi\right) \neq p\left(y|C(y)\backslash x,\xi\right) \tag{3.1.12}$$

The following theorem shows that, provided all probabilities in the joint distribution are nonzero, Equation 3.1.12 is also true when the distributions in the equation are conditioned on any additional set of nonsuccessors of $y$.

**Definition 3.3** *A probability distribution is* **strictly positive** *if and only if it contains no probabilities equal to zero.*

**Theorem 3.5** *If an arc from $x$ to $y$ is nonsuperfluous in an assessed knowledge map that has a strictly positive joint distribution, then, for all subsets $\bar{S}^{'}(y) \subseteq \bar{S}(y)$ that do not contain $C(y)$,*

$$p\left(y|C(y),\bar{S}^{'}(y),\xi\right) \neq p\left(y|C(y)\backslash x,\bar{S}^{'}(y),\xi\right) \tag{3.1.13}$$

*for some instance of $C(y) \cup \bar{S}^{'}(y)$.*

**Proof:** Suppose the theorem is false. Then

$$p\left(y|C(y),\bar{S}^{'}(y),\xi\right) = p\left(y|C(y)\backslash x,\bar{S}^{'}(y),\xi\right) \tag{3.1.14}$$

for all subsets $\bar{S}^{'}(y) \subseteq \bar{S}(y)$ that do not contain $C(y)$. Because the knowledge map has a strictly positive distribution, Equation 3.1.14 holds for all instances of $C(y) \cup \bar{S}^{'}(y)$. Expanding $p\left(y|C(y)\backslash x,\xi\right)$ over all instances of $\bar{S}^{'}(y)$, we obtain

$$\begin{aligned}
p\left(y|C(y)\backslash x,\xi\right) &= \sum_{\bar{S}_i^{'}(y)} p\left(y|C(y)\backslash x,\bar{S}_i^{'}(y),\xi\right) \quad p\left(\bar{S}_i^{'}(y)|C(y)\backslash x,\xi\right) \\
&= \sum_{\bar{S}_i^{'}(y)} p\left(y|C(y),\bar{S}_i^{'}(y),\xi\right) \quad p\left(\bar{S}_i^{'}(y)|C(y)\backslash x,\xi\right) \\
&= \sum_{\bar{S}_i^{'}(y)} p\left(y|C(y),\xi\right) \quad p\left(\bar{S}_i^{'}(y)|C(y)\backslash x,\xi\right) \\
&= p\left(y|C(y),\xi\right)
\end{aligned}$$

The first line is just the expansion rule for probabilities, and the second line is obtained from the first line using Equation 3.1.14. Because $\bar{S}^{'}(y)$ contains only nonsuccessors of $y$, the third line follows from the second line and Theorem 3.1. Finally, since $p\left(y|C(y),\xi\right)$ does not depend on $\bar{S}^{'}(y)$, we obtain the last line from the third. The fact that $p\left(y|C(y)\backslash x,\xi\right) = p\left(y|C(y),\xi\right)$ contradicts the assumption that the arc from $x$ to $y$ is nonsuperfluous. □

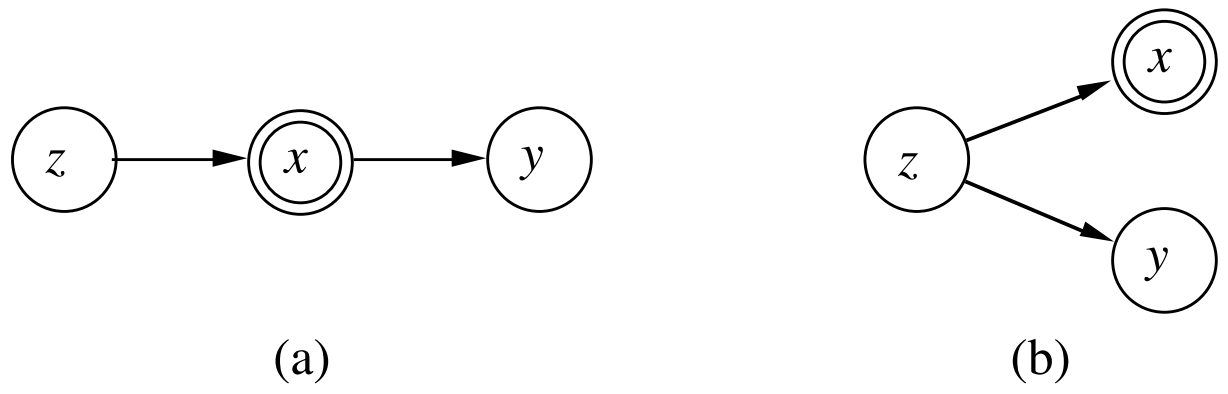

**Figure 3.1:** A counterexample to Theorem 3.5 in the deterministic case.
(a) The variables $x$ and $z$ are logically equivalent, and the arc from $x$ to $y$ is nonsuperfluous. (b) The arc from $x$ to $y$ in (a) can replaced with an arc from $z$ to $y$.

The theorem is false for some distributions that are not strictly positive. Consider, for example, the knowledge map shown in Figure 3.1(a). Suppose that the knowledge map is minimal. In addition, suppose that $x$ and $z$ are logically equivalent, making $z$ a deterministic function of $x$, as shown in the figure. In this case, we can replace the arc from $x$ to $y$ with an arc from $z$ to $y$, as shown in Figure 3.1(b), without destroying the assertions of conditional independence and dependence in the original knowledge map. In particular, we obtain

$$p\,(y|x, z, \xi) = p\,(y|z, \xi)$$

for all instances of the set $\{x, z\}$ that are logically consistent, a contradiction to Theorem 3.5. In general, the proof of the above theorem fails for nonpositive distributions, because we may not be able to replace $p\left(y|C(y)\backslash x, \bar{S}^{'}(y), \xi\right)$ with $p\left(y|C(y), \bar{S}^{'}(y), \xi\right)$, for all instances of $C(y) \cup \bar{S}^{'}(y)$; one or more instances of $x$ may not be logically consistent with some instances of $C(y) \cup \bar{S}^{'}(y)$.

A property of knowledge maps that is important to the development of the theory of similarity networks follows from Theorem 3.5.

**Theorem 3.6** *A strictly positive joint distribution over a set of variables and an expansion order on the variables determines a unique minimal knowledge map.*

**Proof:** Suppose $C'(y) \neq C''(y)$ are two minimal predecessor sets of some variable $y$. Choose a variable $x$ such that $x \in C'(y)$ and $x \notin C''(y)$, and let $C(y) = C'(y) \cup C''(y)$. Because every node in $C(y)$ is a nonsuccessor of $y$, it follows from Theorem 3.5 that

$$p\,(y|C(y)\backslash x, \xi) \neq p\,(y|C(y), \xi)$$

for some instance of $C(y)$. However, since $C''(y) \subseteq C(y) \setminus x$, it follows from Theorem 3.1 that

$$p\,(y|C(y)\backslash x, \xi) = p\,(y|C(y), \xi)$$

for all instances of $C(y)$, a contradiction. □

Pearl (1988, page 119) first proved Theorem 3.6. The theorem is a weak converse of the fundamental theorem for knowledge maps, Theorem 3.2, which says that an assessed knowledge map determines a unique joint distribution over the variables in the map. The knowledge maps in Figures 3.1(a) and (b) are a counterexample to the theorem for nonpositive distributions.

## 3.2 Overview

We are now ready to examine a formal theory for similarity networks. For this development, we shall find it convenient to consider two new varieties of similarity networks, as well as several additional constructions among the various representations. The collection of representations and constructions that we discuss is shown in Figure 3.2.

A *hypothesis-specific similarity network* consists of three component representations, a similarity graph, a hypothesis-specific knowledge map for every hypothesis in the similarity graph, and a relevance set for every edge in the similarity graph. The similarity graph is the same graph that is associated with the similarity networks of Chapter 2. Each node in the graph represents one of a set of mutually exclusive and exhaustive instances or hypotheses of a distinguished variable, denoted $h$. Informally, the edges between nodes in the graph represent judgments of similarity. Formally, the presence of the edge $(h_i, h_j)$ represents an assertion by the network's author that he is willing to construct an ordinary local knowledge map for the pair of hypotheses $h_i$ and $h_j$. The *hypothesis-specific knowledge map* or *hs map* for the hypothesis $h_i$ is a knowledge map of nondistinguished variables under the assumption that $h_i$ is true. Unlike the local maps discussed in Chapter 2, all hs maps must contain the same collection of nondistinguished variables. The *relevance set* for the edge $(h_i, h_j)$ contains assertions of subset independence and subset dependence. We soon consider this set in detail.

A *comprehensive similarity network* contains two component representations, similar to the components of the similarity networks examined in Chapter 2: a similarity graph, and a comprehensive local knowledge map for every edge in the similarity graph. The similarity graph is the same graph associated with hypothesis-specific networks and the similarity networks of the previous chapter. Like the local knowledge map already introduced, the *comprehensive local knowledge map* or *c-local map* for the edge $(h_i, h_j)$ is a knowledge map under the assumption that either $h_i$ or $h_j$ is true. Unlike its counterpart, however, each comprehensive local knowledge map is required to contain all nondistinguished variables.

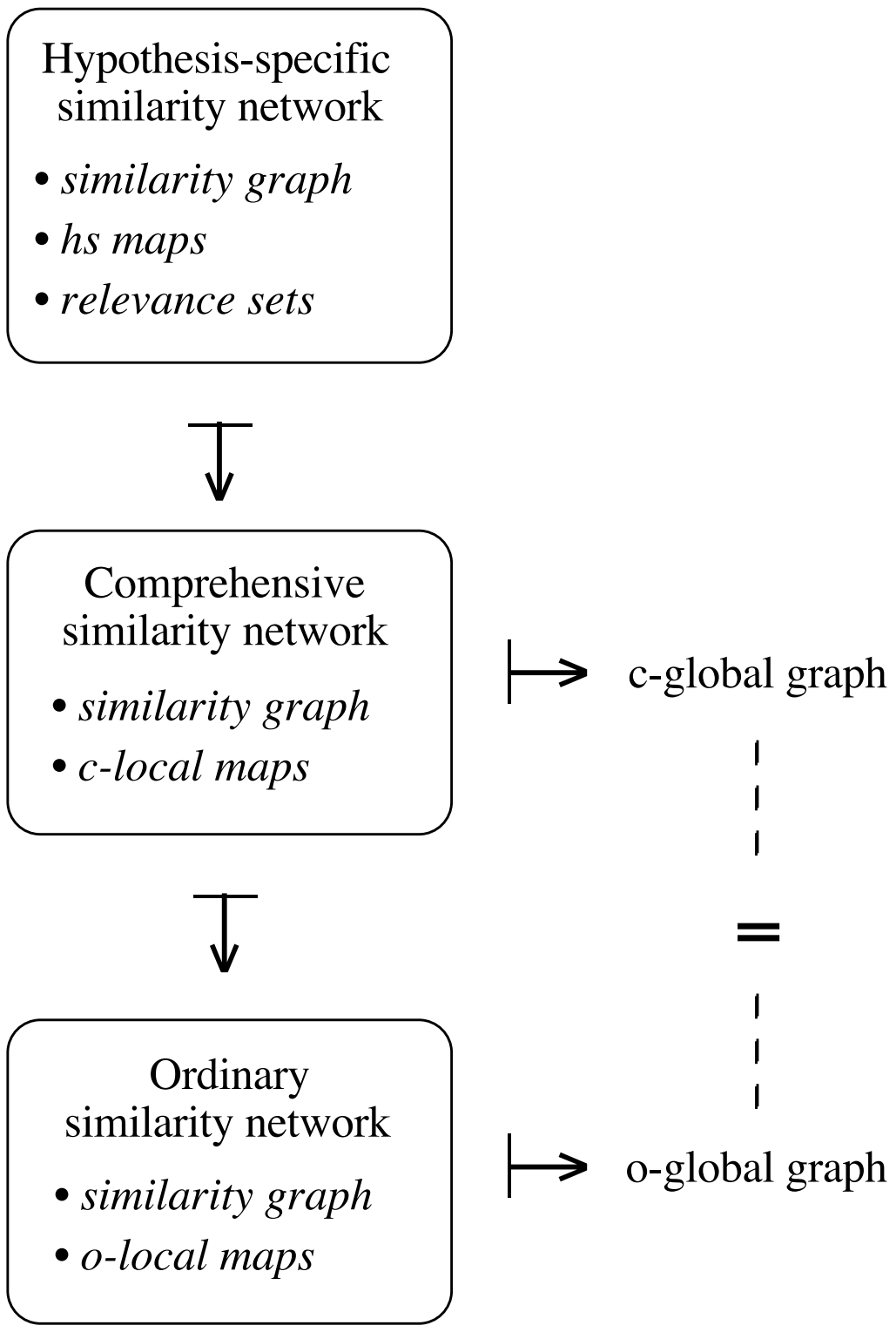

**Figure 3.2:** Hypothesis-specific, comprehensive, and ordinary similarity networks. Hypothesis-specific networks consist of a similarity graph, a relevance set associated with each edge in the similarity graph, and a hypothesis-specific knowledge map (hs map) associated with each node in the similarity graph. This network can be used to construct a comprehensive similarity network that consists of the same similarity graph and a comprehensive local knowledge map (c-local map) for each edge in the similarity graph. The comprehensive similarity network, in turn, can be used to construct a comprehensive global knowledge map (c-global map), which is a knowledge map for the entire domain. The comprehensive similarity network, by definition, also determines an ordinary similarity network. The comprehensive and ordinary networks are identical, except that the knowledge maps associated with the edges in the similarity graph, called ordinary knowledge maps (o-local maps), do not contain nodes that are disconnected from $h$. Finally, the ordinary similarity network can be used to construct an ordinary global knowledge map (o-global map) that, like its comprehensive counterpart, is a knowledge map for the entire domain. Under certain conditions, a c-global map and o-global map constructed in this manner are identical.

The similarity network introduced in Chapter 2 is called an *ordinary similarity network* in this chapter, to distinguish it from the hypothesis-specific and comprehensive varieties. By the same token, the local knowledge maps are called *ordinary local knowledge maps* or *o-local maps*. Formally, ordinary local knowledge maps are defined in terms of their comprehensive counterparts. In particular, the o-local map for the edge $(h_i, h_j)$ in the similarity graph is the node-induced subgraph of the corresponding c-local map that contains only those nodes in the c-local map that are connected to the distinguished variable $h$.

The *ordinary global knowledge map or o-global map* is the global knowledge map described in Chapter 2. As is the case for the similarity network, the map is labeled ordinary to distinguish its counterpart, the *comprehensive global knowledge map or c-global map.*

The symbol $\mapsto$ in Figure 3.2 denotes the construction of one representation from another. The construction of an ordinary global knowledge map from an ordinary similarity network is the main focus of this chapter. As discussed in Chapter 2, we construct the ordinary global knowledge map by forming the graph union of the o-local maps in the ordinary similarity network. The construction of a comprehensive global knowledge map from a comprehensive similarity network is defined analogously. The construction of a comprehensive similarity network from a hypothesis-specific network is somewhat more complicated and will be discused in the following section. Also, we can view the definition of o-local maps in terms of c-local maps as a construction.

In Section 3.3, we formally define these representations and constructions, and illustrate them with examples. In Section 3.4, we define the notion of *soundness* for constructions, and show that the construction of a comprehensive network from a hypothesis-specific network and of a comprehensive knowledge map from a comprehensive similarity network is sound, under the conditions discussed in Section 2.1.2. In Section 3.5, we define what it means for a similarity network to be consistent, and demonstrate that minimal hypothesis-specific and comprehensive similarity networks—networks that contain only minimal knowledge maps—can be inconsistent. We then show that a minimal comprehensive similarity network is consistent if and only if it can be constructed from a consistent, minimal hypothesis-specific similarity network. We use this fact to derive an algorithm for testing the consistency of a comprehensive network. In Section 3.6, we use these consistency results to show that the c-global map and o-global map constructed from a consistent, minimal comprehensive similarity network must be identical. Using this observation, we prove that the construction of the o-global map is sound. In Section 3.7, we extend the algorithm developed in Section 3.5 for testing the consistency of comprehensive networks to ordinary networks. In Section 3.8, we discuss a definition of ordinary local knowledge maps that is more useful to users than is the current definition. In Section 3.9, we show that the construction of an o-global map from an ordinary similarity network is exhaustive. Finally, in Section 3.10, we examine the assessment of probabilities in an ordinary similarity network.

## 3.3 Definitions and Examples

In the following definitions, $h$ is a distinguished uncertain variable with mutually exclusive and exhaustive hypotheses $h_1, h_2, \ldots h_n$, and $Y$ is a set of nondistinguished uncertain variables.

**Definition 3.4** *A* **hypothesis-specific similarity network** *for variable $h$ and $Y$ given background knowledge $\xi$ consists of a similarity graph, a hypothesis-specific knowledge map for each hypothesis of $h$, and a relevance set for each edge in the similarity graph. The* **similarity graph** *is an undirected graph in which the nodes represent the hypotheses of $h$. The presence of the edge $(h_i, h_j)$ means that the author of the network is willing to compose an ordinary local knowledge map for the hypothesis pair. The* **hypothesis-specific knowledge map** *(hs map) for hypothesis $h_i$, denoted $\widehat{h_i}$, is a knowledge map for all variables in $Y$ given background knowledge $h_i \wedge \xi$. The structures of the hs maps are restricted such that the graph union of the $\widehat{h_i}$ cannot contain directed cycles. The conditional predecessors and nonsuccessors of a node $y$ in the hs map $\widehat{h_i}$ are denoted $C^i(y)$ and $\bar{S}^i(y)$, respectively. The* **relevance set** *for the edge $(h_i, h_j)$ in the similarity graph, denoted $\mathcal{R}^{ij}$, contains either a statement of subset independence or subset dependence for each variable $y \in Y$ such that such that the conditional predecessors of $y$ in $\widehat{h_i}$ are the same as those in $\widehat{h_j}$. That is, $\mathcal{R}^{ij}$ contains either the assertion*

$$p\left(y|C^{i/j}(y), h_i, \xi\right) = p\left(y|C^{i/j}(y), h_j, \xi\right) \tag{3.3.15}$$

*or*

$$p\left(y|C^{i/j}(y), h_i, \xi\right) \neq p\left(y|C^{i/j}(y), h_j, \xi\right) \tag{3.3.16}$$

*for every $y \in Y$ such that $C^i(y) = C^j(y) \equiv C^{i/j}(y)$.*

The restriction on the structures of the hs maps guarantees that some expansion order over the variables in the hypothesis-specific network is consistent with every hypothesis-specific knowledge map in the network. This property of hs maps will be used throughout the derivations to follow. By Theorem 2.1, Equation 3.3.15 is the assertion that the variable $y$ is not relevant to the hypothesis pair $\{h_i, h_j\}$, given any instance of the conditional predecessors of $y$ in the hs maps $\widehat{h_i}$ and $\widehat{h_j}$. Conversely, Equation 3.3.16 is the statement that variable $y$ is relevant to $\{h_i, h_j\}$ for some instance of the conditional predecessors.

As we shall see, these assertions determine, in part, whether or not there are arcs from $h$ to the nondistinguished nodes in both comprehensive and ordinary local knowledge maps constructed from the hypothesis-specific similarity network. Note that we do not include in $\mathcal{R}^{ij}$ those assertions of subset independence or dependence for variables $y$, where $C^i(y) \neq C^j(y)$. In Appendix B.2, we see that if these conditional-predecessor sets are not equal, then $y$ must be relevant to $\{h_i, h_j\}$ for some instance of $C^i(y) \cup C^j(y)$, and hence we do not need to include explicitly these assertions in the relevance set.

**Definition 3.5** *A* **comprehensive similarity network** *for $h$ and $Y$ given background knowledge $\xi$ consists of a similarity graph and a comprehensive local knowledge map for each edge in the similarity graph. The* **similarity graph** *is defined exactly as it is for hypothesis-specific similarity networks. The* **comprehensive local knowledge map** *(c-local map) for the edge between $h_i$ and $h_j$ in the similarity graph, denoted $h_i$–$h_j$, is a knowledge map for variable $h$ and all the variables in $Y$ given background knowledge $\{h_i, h_j\} \wedge \xi$. The structures of the c-local maps are restricted such that (1) node $h$ is not the successor of any node, and (2) the graph union of the maps does not contain directed cycles. The conditional predecessors and nonsuccessors of a node $y$ in the c-local map $h_i$–$h_j$ are denoted $C^{ij}(y)$ and $\bar{S}^{ij}(y)$, respectively.*

Restriction 2 corresponds to the constraint imposed on hs maps in Definition 3.4 and provides an analogous guarantee that will be used throughout this development. Specifically, there must be some expansion order over the variables in the comprehensive similarity network that is consistent with every c-local map. We shall examine the importance of restriction 1 after deriving preliminary results concerning soundness.

**Definition 3.6** *An* **ordinary similarity network** *for $h$ and $Y$ given background knowledge $\xi$ consists of a similarity graph and an ordinary local knowledge map for each edge in the similarity graph. The* **similarity graph** *is defined exactly as it is for hypothesis-specific and comprehensive similarity networks. The* **ordinary local knowledge map** *(o-local map) for the edge between $h_i$ and $h_j$, denoted $h_i$–$h_j$, is a knowledge map with background knowledge $\{h_i, h_j\} \wedge \xi$. The map is the node-induced subgraph of the c-local map that includes only nodes connected to $h$ and the node $h$ itself.*

Note that we use the term $h_i$–$h_j$ to denote both a c-local and o-local map. The context in which the term appears will make its meaning unambiguous.

It will sometimes be convenient to use the symbols $\mathcal{HS}$, and $\mathcal{C}$, and $\mathcal{O}$ to denote hypothesis-specific, comprehensive, and ordinary similarity networks, respectively. The following two definitions, analogous to those given for knowledge maps, also will be useful.

**Definition 3.7** *A similarity network (hypothesis-specific, comprehensive, or ordinary) is* **assessed** *if and only if every knowledge map in the network is assessed.*

**Definition 3.8** *A similarity network (hypothesis-specific, comprehensive, or ordinary) is* **minimal** *if and only if every knowledge map in the network is minimal.*

We can use an assessed hypothesis-specific similarity network, in conjunction with a marginal distribution for $h$, to construct a joint distribution for the variables $h$ and $Y$. This fact follows from the definition of hypothesis-specific networks, which requires that we compose an hs map for each hypothesis in such a network. Also, because similarity graphs must be connected, an assessed comprehensive similarity network can be used to construct such a joint distribution. In Section 3.10, we shall see that an assessed ordinary similarity network also determines a unique joint distribution over $h$ and $Y$, even though these networks may contain o-local maps that do not include every nondistinguished variable.

There is a subtle difference between the definitions of the hypothesis-specific and comprehensive networks and the definition of the ordinary similarity network. The first two networks, as defined, can be *composed* or directly constructed by an individual. In contrast, the ordinary similarity network is *constructed* from a comprehensive similarity network. (In practice, of course, the comprehensive similarity network is not composed, and the construction is implicit.) We say that a comprehensive network $\mathcal{C}$ *constructs* an ordinary similarity network $\mathcal{O}$, or that $\mathcal{C}$ is a *constructor* of $\mathcal{O}$. Sometimes, the notational shorthand $\mathcal{C} \mapsto \mathcal{O}$ shall represent the construction.

Each of the representations that we have examined can be *composed* directly. In several cases, however, we also can *construct* one representation from another.

**Construction 3.1** *A comprehensive similarity network $\mathcal{C}$ for $h$ and $Y$ given background knowledge $\xi$ is constructed from a hypothesis-specific similarity network $\mathcal{HS}$ for $h$, $Y$, and $\xi$ as follows:*

- *Copy the similarity network of $\mathcal{HS}$ to $\mathcal{C}$.*
- *For each edge $(h_i, h_j)$ in the similarity graph, construct the c-local map $h_i$–$h_j$ by (1) forming the graph union of $\widehat{h_i}$ and $\widehat{h_j}$, (2) adding the node $h$ to this graph, and (3) adding an arc from $h$ to node $y$ if and only if $C^i(y) \neq C^j(y)$, or $C^i(y) = C^j(y) \equiv C^{i/j}(y)$ and the assertion*

  $$p\left(y|C^{i/j}(y), h_i, \xi\right) \neq p\left(y|C^{i/j}(y), h_j, \xi\right)$$

  *is in the relevance set $\mathcal{R}^{ij}$.*

*We write* $\mathcal{HS}$ **constructs** $\mathcal{C}$*, or* $\mathcal{HS} \mapsto \mathcal{C}$*.*

**Construction 3.2** *A comprehensive global knowledge map (c-global map)* $\mathcal{G}_c$ *is a knowledge map for* $h$ *and* $Y$ *given background knowledge* $\xi$ *that is constructed from a comprehensive similarity network* $\mathcal{C}$ *for* $h$*,* $Y$*, and* $\xi$*. Specifically,* $\mathcal{G}_c$ *is the graph union of all c-local maps in* $\mathcal{C}$*. We write* $\mathcal{C}$ **constructs** $\mathcal{G}_c$*, or* $\mathcal{C} \mapsto \mathcal{G}_c$*. The conditional predecessors and nonsuccessors of a node* $y$ *in the c-global map are denoted* $C^{\mathcal{G}_c}(y)$ *and* $\bar{S}^{\mathcal{G}_c}(y)$*, respectively.*

**Construction 3.3** *An ordinary global knowledge map (o-global map)* $\mathcal{G}_o$ *is a knowledge map for* $h$ *and* $Y$ *given background knowledge* $\xi$ *that is constructed from an ordinary similarity network* $\mathcal{O}$ *for* $h$*,* $Y$*, and* $\xi$*. Specifically,* $\mathcal{G}_o$ *is the graph union of all o-local maps in* $\mathcal{O}$*. We write* $\mathcal{O}$ **constructs** $\mathcal{G}_o$*, or* $\mathcal{O} \mapsto \mathcal{G}_o$*. The conditional predecessors and nonsuccessors of a node* $y$ *in the o-global map are denoted* $C^{\mathcal{G}_o}(y)$ *and* $\bar{S}^{\mathcal{G}_o}(y)$*, respectively.*

Because each of the representations involved in these constructions can be composed directly, it is possible that these constructions are not sound, in the sense that the assertions of conditional independence and dependence implied by a constructed representation may not logically follow the assertions associated with the original representation. As discussed previously, however, one of the major results we examine is that Construction 3.3 is sound, under the conditions discussed in Section 2.1.2. To prove this result, we also prove that Constructions 3.1 and 3.2 are sound under these same conditions.

In the definition of comprehensive similarity networks (Definition 3.5), recall that the graph union of the o-local maps can contain no cycles. However, the graph union of these maps is equivalent to the c-global map. Thus, the restriction in the definition of comprehensive networks guarantees that the c-global map is a legitimate knowledge map, and that any expansion order consistent with the c-global map must be consistent with every o-local map. Also note that we can construct a c-global map indirectly from a hypothesis-specific network, by first constructing a comprehensive network from the hypothesis-specific network, and then constructing a c-global map from the comprehensive network. Again, because the graph union of hs maps cannot contain cycles, the c-global map constructed in this manner is legitimate, and every expansion order consistent with the c-global map must be consistent with every hs map. Although these properties of hypothesis-specific and comprehensive networks are not enough to certify the soundness of the constructions, they will be instrumental in the proofs of soundness.

Figure 3.3 illustrates the various representations and the constructions among them, and depicts the graphical shorthand that we use for these representations in this chapter.

Figure 3.3(a) contains a hypothesis-specific similarity network for three hypotheses $h_1$, $h_2$, and $h_3$, and for three nondistinguished variables $x$, $y$, and $z$. The hypothesis-specific knowledge map for each hypothesis is shown directly under the node representing the hypothesis in the similarity graph. A dashed line around each map serves to define the nodes included in the map. The elements of the relevance sets $\mathcal{R}^{12}$ and $\mathcal{R}^{23}$ are shown above the edges in the similarity graph. Observe that because $C^1(y) \neq C^2(y)$ and $C^2(y) \neq C^3(y)$, there is no assertion involving $y$ in the relevance sets $\mathcal{R}^{12}$ or $\mathcal{R}^{23}$.

The left-hand side of Figure 3.3(b) contains the comprehensive similarity network constructed from the hypothesis-specific network above it. In the figure, the small ovals attached to the edges between hypotheses represent the distinguished node $h$ in each c-local map. Again, a dashed line around each map identifies the nodes belonging to the map. Because there is an arc from $x$ to $y$ in $\widehat{h_1}$, there is a corresponding arc in the c-local map $h_1$–$h_2$. Also, because there is no arc from $z$ to $y$ in either $\widehat{h_1}$ or $\widehat{h_2}$, there is no such arc in $h_1$–$h_2$. There is no arc from $h$ to $z$ in $h_1$–$h_2$ because the assertion "$p\,(z|h_1, \xi) = p\,(z|h_2, \xi)$" is in the relevance set $\mathcal{R}^{12}$. Conversely, there is an arc from $h$ to $x$, because the assertion "$p\,(x|h_1, \xi) \neq p\,(x|h_2, \xi)$" is in the relevance set $\mathcal{R}^{12}$. There is an arc from $h$ to $y$, because the conditional predecessors of $y$ in $\widehat{h_1}$ and $\widehat{h_2}$ are different. The right-hand side of Figure 3.3(b) shows the c-global map constructed from the comprehensive network. The c-global map is the graph union of the c-local maps $h_1$–$h_2$ and $h_2$–$h_3$.

The left-hand side of Figure 3.3(c) contains the ordinary similarity network constructed, by definition, from the comprehensive network in Figure 3.3(b). Because node $z$ is disconnected from $h$ in the c-local map $h_1$–$h_2$, the node is omitted from the corresponding o-local map. Similarly, node $x$ is omitted from the o-local map $h_2$–$h_3$. Because all nodes in an o-local map are connected to $h$, dashed lines surrounding the maps are not needed. The c-global map constructed from the ordinary network is shown on the right-hand side of the figure. Observe that the o-global map is identical to the c-global map.

Notice that the two asymmetric forms of conditional independence represented in a similarity network, subset independence and hypothesis-specific independence, are represented disjointly in only the hypothesis-specific similarity network. In particular, the relevance sets contain assertions of subset independence, and differences among the graph structures of the hs maps reflect hypothesis-specific independence. We exploit this observation in Section 3.10 when we discuss probability assessment in ordinary similarity networks.

Also notice that the two global maps have several constructors. For example, if we add an arc from $z$ to $y$ in the c-local map $h_1$–$h_2$ and an arc from $x$ to $y$ in the c-local map $h_2$–$h_3$, the resulting comprehensive network would construct the same c-global map.

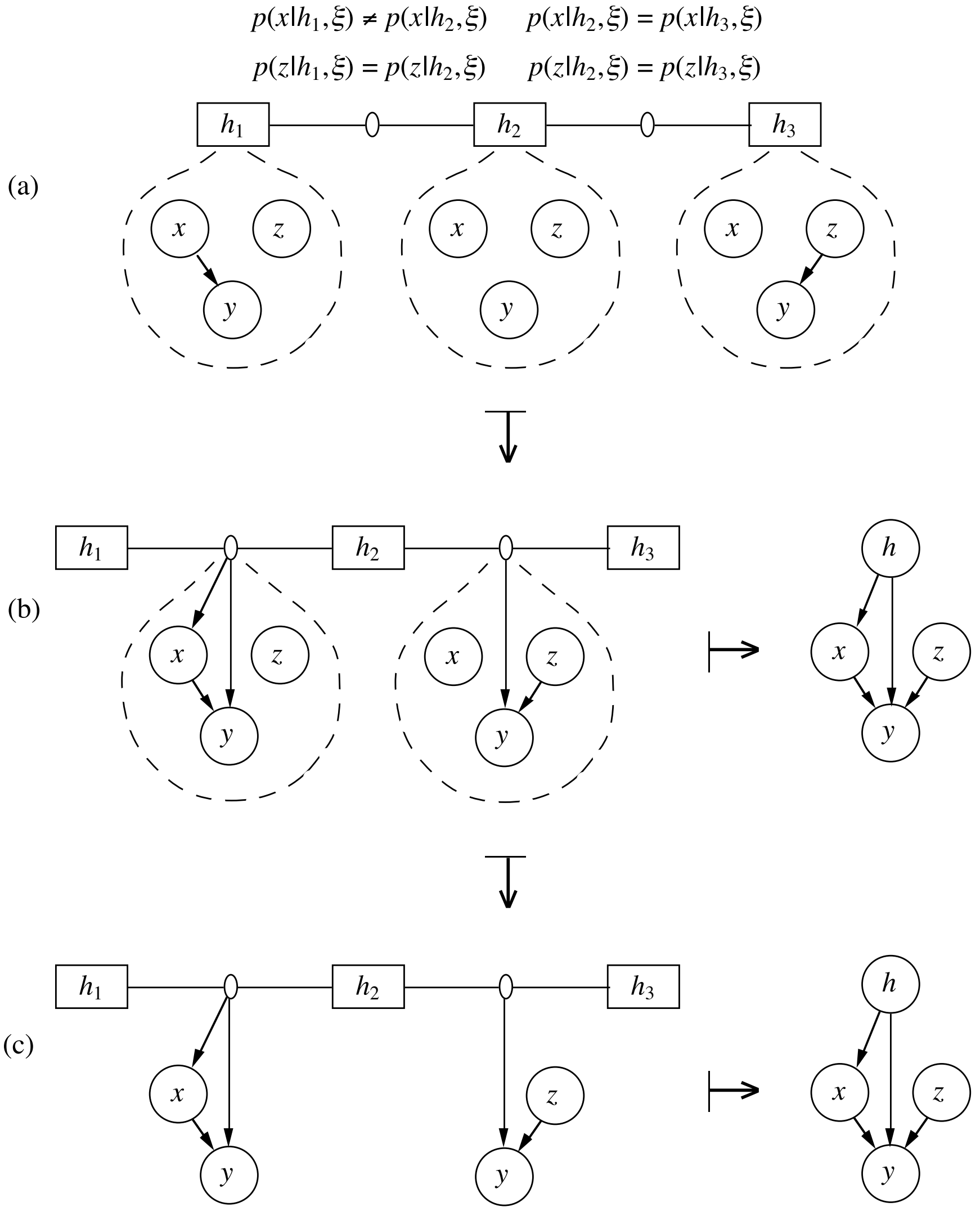

**Figure 3.3:** A hypothesis-specific similarity network and its constructs.
Figure (a) is a hypothesis-specific network for three hypotheses and three nondistinguished nodes. The probabilistic relations above each edge in the similarity graph make up the relevance set for that edge. The left-hand sides of Figures (b) and (c) are the comprehensive and ordinary networks constructed from the network in Figure (a). The right-hand side of Figure (b) is the c-global map constructed from the comprehensive network. The right-hand side of Figure (c) is the o-global map constructed from the ordinary network.

In general, any comprehensive or ordinary similarity network and any comprehensive or ordinary knowledge map can have more than one constructor. In this regard, the following definition will be useful.

**Definition 3.9** *A* **maximal constructor** *of a representation (comprehensive similarity network, ordinary similarity network, comprehensive knowledge map, or ordinary knowledge map) is a constructor of the network that fails to be a constructor when one or more arcs are added to any of its component knowledge maps.*

Also, some comprehensive and ordinary similarity networks have no constructors. We shall examine such situations in detail when we consider the consistency of similarity networks in Sections 3.5 and 3.7.

We have now defined all the representations and constructions among these representations that are necessary to derive the soundness, consistency, and exhaustiveness results that we seek. However, before we proceed with additional technical discussions, it is useful to consider the following informal argument that the construction of the o-global map is sound. Although many subtleties are omitted from the argument, it will serve as a useful guide to the formal development to follow.

First, observe that the knowledge maps in a hypothesis-specific similarity network contain more detailed assertions of conditional independence and dependence than are contained in the knowledge maps in a comprehensive similarity network. Specifically, hs maps contain assertions of independence and dependence, given a state of knowledge where a particular instance of $h$ is known, whereas c-local maps contain such assertions, given a state of knowledge where $h$ is restricted to the disjunction of two instances. Similarly, c-local maps contain more detailed assertions than those embodied by a c-global map. Consequently, constructions $\mathcal{HS}{\mapsto}\mathcal{C}$ (Construction 3.1) and $\mathcal{C}{\mapsto}\mathcal{G}_c$ (Construction 3.2) are sound.

Next, suppose we are given a consistent, minimal comprehensive network. That is, suppose we have a comprehensive network in which the assertions of conditional independence and dependence are satisfied by some joint distribution. It turns out that we can construct this network from some hypothesis-specific network. This observation follows because we can take the expansion order consistent with all c-local maps in the network and use it to compose a collection of hs maps. By the soundness of Construction 3.1 and the fact that an expansion order in conjunction with a strictly positive joint distribution determines a unique knowledge map, the comprehensive network constructed from these hs maps must be equal to the comprehensive network with which we started.

Given this observation, the c-global and o-global map constructed from any consistent, minimal comprehensive similarity network must be identical. To prove this fact, suppose that we are given a consistent, minimal comprehensive network and that there is an arc from nodes $x$ to $y$ in some c-local map $h_i$–$h_j$ in which $x$ and $y$ are disconnected from $h$. We must show that the arc appears in the o-global map constructed from the ordinary counterpart of the comprehensive network, despite the fact that the arc does not appear in the o-local map $h_i$–$h_j$. Now because there is an arc from $x$ to $y$ in the c-local map, there must be such an arc in every pair of hs maps $\widehat{h_i}$ and $\widehat{h_j}$ that can be used to construct the c-local map. (By Construction 3.1, if there is an arc in only one of $\widehat{h_i}$ and $\widehat{h_j}$, there must be an arc from $h$ to $y$ in $h_i$–$h_j$. This situation is impossible, because $y$ is disconnected from $h$.) Thus, there must be an arc from $x$ to $y$ in all c-local maps that are bordered by either $h_i$ or $h_j$ in the similarity graph. Because the similarity graph is connected, when we repeat this argument along every path in the similarity graph emanating from $h_i$ and $h_j$, we find that either $y$ is disconnected from all c-local maps, or there is some c-local map in which $y$ is connected to $h$ and there is an arc from $x$ to $y$. In the former case, $y$ is irrelevant to the diagnosis of $h$ and can be removed from consideration. In the latter case, the arc from $x$ to $y$ will be recorded in the o-global map.

Finally, suppose that we are given a minimal ordinary similarity network. Consider any joint distribution that satisfies the assertions of this network. From this distribution, we can construct a comprehensive similarity network that is both consistent and minimal. It follows from the soundness of Construction 3.2 that the distribution satisfies the assertions of the c-global map constructed from this comprehensive network. However, by the argument in the previous paragraph, this c-global map and the o-global map constructed from the given ordinary network are identical. Thus, the distribution satisfies the assertions of the o-global map and the construction is sound. If we are given a nonminimal ordinary network, we simply remove arcs until the network is minimal, and then apply the preceding argument.

## 3.4 Soundness: Preliminary Results

The three varieties of similarity networks and the two types of knowledge maps each represent assertions of conditional independence and dependence over the variables in the network. In the previous section, we examined four constructions that transform one representation into another. These constructions were defined in terms of the *syntax*, or form, of the representations. In this section, we derive the conditions under which these constructions can preserve the underlying *semantics* of the representations as well.

In the theory of logic, the concepts of semantic and syntactic truth are formally defined. A set of sentences $S_1$ in some formal language *logically implies* another set of sentences $S_2$ in the same language (written $S_1 \models S_2$) if and only if any interpretation of the language that satisfies all sentences in $S_1$ also satisfies all sentences in $S_2$.[2] An interpretation of a language is a translation of the symbols in the language to mathematical or physical objects in the real world. Thus, the concept of logical implication refers to underlying semantics of logical sentences. In contrast, a set of sentences $S_2$ in some formal language is *proved* or *deduced* from another set of sentences $S_1$ in the same language (written $S_1 \vdash S_2$) if and only if there is some series of restricted syntactic manipulations (a proof or deduction) that transforms $S_1$ into $S_2$.

The theory of logic also embodies concepts that relate the notions of syntactic and semantic truth. In particular, if $S_1$ logically implies $S_2$ whenever $S_2$ is proved from $S_1$ for some proof calculus, the calculus is said to be *sound*. Conversely, if $S_2$ can be proved from $S_1$ whenever $S_1$ logically implies $S_2$, then the calculus is said to be *complete*.

These concepts can be mapped to the theory of similarity networks in a natural manner. The formal language of similarity networks is the language of probabilities. The sentences of interest are the assertions of conditional independence and dependence explicitly stated in the irrelevance sets of hypothesis-specific networks and the assertions implicitly determined by the structures of knowledge maps. Such sentences will be called *probability constraints*. The notion of logical implication directly carries over to the theory of similarity networks, provided we restrict interpretations to joint probability distributions.

**Definition 3.10** *Given two sets of probability constraints $K_1$ and $K_2$, $K_1$* **logically implies** *$K_2$, written $K_1 \models_{\mathcal{P}} K_2$, if and only if any joint distribution that satisfies the constraints $K_1$ also satisfies the constraints $K_2$.*

Finally, we can replace the syntactic proof or deduction in the theory of logic with constructions. As a result, we obtain the following definitions of soundness and completeness for each construction.

**Definition 3.11** *A construction $\mapsto$ is* **sound** *if and only if*

$K_1 \mapsto K_2$ implies $K_1 \models_{\mathcal{P}} K_2$

*for any two sets of probability constraints $K_1$ and $K_2$. The construction is* **complete** *if and only if*

$K_1 \models_{\mathcal{P}} K_2$ implies $K_1 \mapsto K_2$

*for any two sets of probability constraints $K_1$ and $K_2$.*

[2] Details of this definition and the definitions to follow have been omitted. For a precise treatment, see Enderton (1972).

Deduction calculi used by mathematicians—for example, unification—are both sound and complete. Thus, $K_1 \models_{\mathcal{P}} K_2$ if and only if the constraints $K_2$ can be deduced from the constraints $K_1$ and the rules of probability using such a calculus. This equivalence will be used throughout the proofs in this chapter. To prove that a construction is sound without this equivalence, we would have to examine every possible network representation that can participate in the construction and examine every joint distribution that can satisfy each representation. Indeed, from a practical standpoint, the definition of soundness (completeness) is that $K_2$ can be proved from $K_1$ and the rules of probability if (only if) $K_2$ is constructed from $K_1$. Definition 3.11 is stated in terms of interpretations to emphasize that the notions of soundness and completeness relate semantics to syntax.

As mentioned previously, strictly positive distributions play an important role in the theory of similarity networks. Consequently, the following definitions will be useful.

**Definition 3.12** *Given two sets of probability constraints $K_1$ and $K_2$, $K_1$* **logically implies** $K_2$ **for strictly positive distributions**, *written $K_1 \models_{\mathcal{P}_+} K_2$, if and only if any strictly positive joint distribution that satisfies the constraints $K_1$ also satisfies the constraints $K_2$.*

**Definition 3.13** *A construction $\mapsto$ is* **sound for strictly positive distributions** *if and only if*

$K_1 \mapsto K_2$ implies $K_1 \models_{\mathcal{P}_+} K_2$

*for any two sets of probability constraints $K_1$ and $K_2$. The construction is* **complete for strictly positive distributions** *if and only if*

$K_1 \models_{\mathcal{P}_+} K_2$ implies $K_1 \mapsto K_2$

*for any two sets of probability constraints $K_1$ and $K_2$.*

Informally, a construction is complete if, whenever one representation contains fewer or less specific probability constraints than another representation, the first can be constructed from the second. Not one of the constructions that we have examined is complete. This observation follows from the fact that assertions of conditional independence in a knowledge map are often insensitive to the direction of arcs, and yet the constructions are constrained to preserve the direction of arcs. We should not confuse completeness with *exhaustiveness*, a weaker property. In Section 3.9, we see that the construction of an o-global map from and ordinary similarity network is exhaustive.

A construction is sound if the constructed representation contains fewer or less specific constraints than its constructor. Under certain conditions, all the constructions have this property, and hence all are sound. For example, the hypothesis-specific network contains assertions of conditional independence and dependence that are specific to *individual* hypotheses in the similarity graph, whereas the comprehensive network, constructed from a hypothesis-specific network, contains assertions specific to only hypothesis *pairs*. Furthermore, the c-global map, constructed from the comprehensive similarity network, contains assertions of conditional independence and dependence that are specific to *no* set of hypotheses. In general, we have the following theorem.

**Theorem 3.7 (Soundness)** *The following constructions are sound for strictly positive distributions:*

*a.* $\mathcal{HS} \mapsto \mathcal{C}$
*b.* $\mathcal{C} \mapsto \mathcal{G}_c$
*c.* $\mathcal{C} \mapsto \mathcal{O}$
*d.* $\mathcal{O} \mapsto \mathcal{G}_o$

*Moreover, each construction is sound when both the constructor and constructed map are minimal.*

The formal proofs of parts a and b are similar and are given in Appendices B.2 and B.3, respectively. The construction of an ordinary similarity network from a comprehensive similarity network is sound (part c) because, by definition, the constraints in an o-local map are a subset of the constraints in its corresponding c-local map. We prove part d after examining the consistency of comprehensive similarity networks in the following section.

Figure 3.4 illustrates three conditions under which the construction $\mathcal{C}\mapsto\mathcal{G}_c$ is not sound. In Figure 3.4(a), the similarity graph of the network is not connected. Specifically, the node representing $h_3$ is connected to neither $h_1$ nor $h_2$. Consequently, if the variable $y$ is relevant to the hypothesis pair $\{h_1, h_3\}$ or to the hypothesis pair $\{h_2, h_3\}$, the c-global map constructed from this network does not record this fact (there is no arc from $h$ to $y$ in the c-global), and the construction is not sound. In Figure 3.4(b), the distinguished node $h$ has nonsuccessors. The conditional independence assertion implied by this c-global map is

$$p(x|y,\xi) = p(x|\xi)$$

However, this assertion cannot be derived from the conditional independence assertions in the c-local maps

$$p(x|y,\{h_1,h_2\},\xi) = p(x|\{h_1,h_2\},\xi)$$

$$p(x|y, \{h_2, h_3\}, \xi) = p(x|\{h_2, h_3\}, \xi)$$

and hence the construction is not sound. In Figure 3.4(c), the underlying joint distribution contains zero probabilities ($x$ is logically determined by $h$). To see that the construction of the c-global map is not sound in this case, suppose $x$ has two instances $x_1$ and $x_2$. In addition, suppose $x$ and $y$ are conditionally independent given $h$, and that

$$p(x_1|h_1, \xi) = 1, \quad p(x_2|h_2, \xi) = 1, \quad p(x_1|h_3, \xi) = 1$$

$$p(y|h_1, \xi) \neq p(y|h_2, \xi) \neq p(y|h_3, \xi) \neq p(y|h_1, \xi)$$

In context $\{h_1, h_2\}$ or $\{h_2, h_3\}$, the variables $h$ and $x$ are logically equivalent. Thus, the c-local maps in the figure accurately reflect the underlying distributions. In the global context, however, knowing $x$ does not render $h$ and $y$ independent. Hence, the construction is not sound. Note that the construction becomes sound if we compose the c-local map $h_1$–$h_3$. In fact, Theorem 3.7(b) holds for nonminimal networks given any distribution, provided every pair of hypotheses in the similarity graph is connected directly. In Section 4.4.1, we discuss the significance of these restrictions on soundness to the construction of the global knowledge map for Pathfinder.

Note that some of the constructions are sound under weaker conditions than those stated in the theorem. For example, the construction $\mathcal{HS}\mapsto\mathcal{C}$ is sound for any joint distribution when the knowledge maps are not required to be minimal. Also, the construction $\mathcal{C}\mapsto\mathcal{O}$ is sound for all distributions for both minimal and nonminimal maps, by definition. However, we are interested primarily in the soundness of the construction $\mathcal{O}\mapsto\mathcal{G}_o$, and the soundness of this construction hinges on the soundness of the construction $\mathcal{C}\mapsto\mathcal{G}_c$. Consequently, we do not pursue these exceptional cases here.

## 3.5 Consistency: Preliminary Results

Let us extend the mapping, introduced in the previous section, from the theory of logic to the theory of similarity networks. In the theory of logic, a set of sentences $S$ is *consistent* if and only if there is some interpretation that satisfies all the sentences in $S$.[3] Restricting interpretations to joint probability distributions over the set of variables in a similarity network, we have the following definition of consistency.

[3]To be more precise, the set of sentences $S$ is said to be *satisfiable* under these conditions. In contrast, consistency is defined in syntactic terms. Specifically, a set of sentences is consistent if and only if it is impossible to prove a sentence and its negation from that set. Nonetheless, the soundness theorem from logic and Gödel's completeness theorem show that the concepts of satisfaction and consistency are equivalent (Enderton, 1972). In this work, we do not distinguish these two concepts.

(a)

(b)

(c)

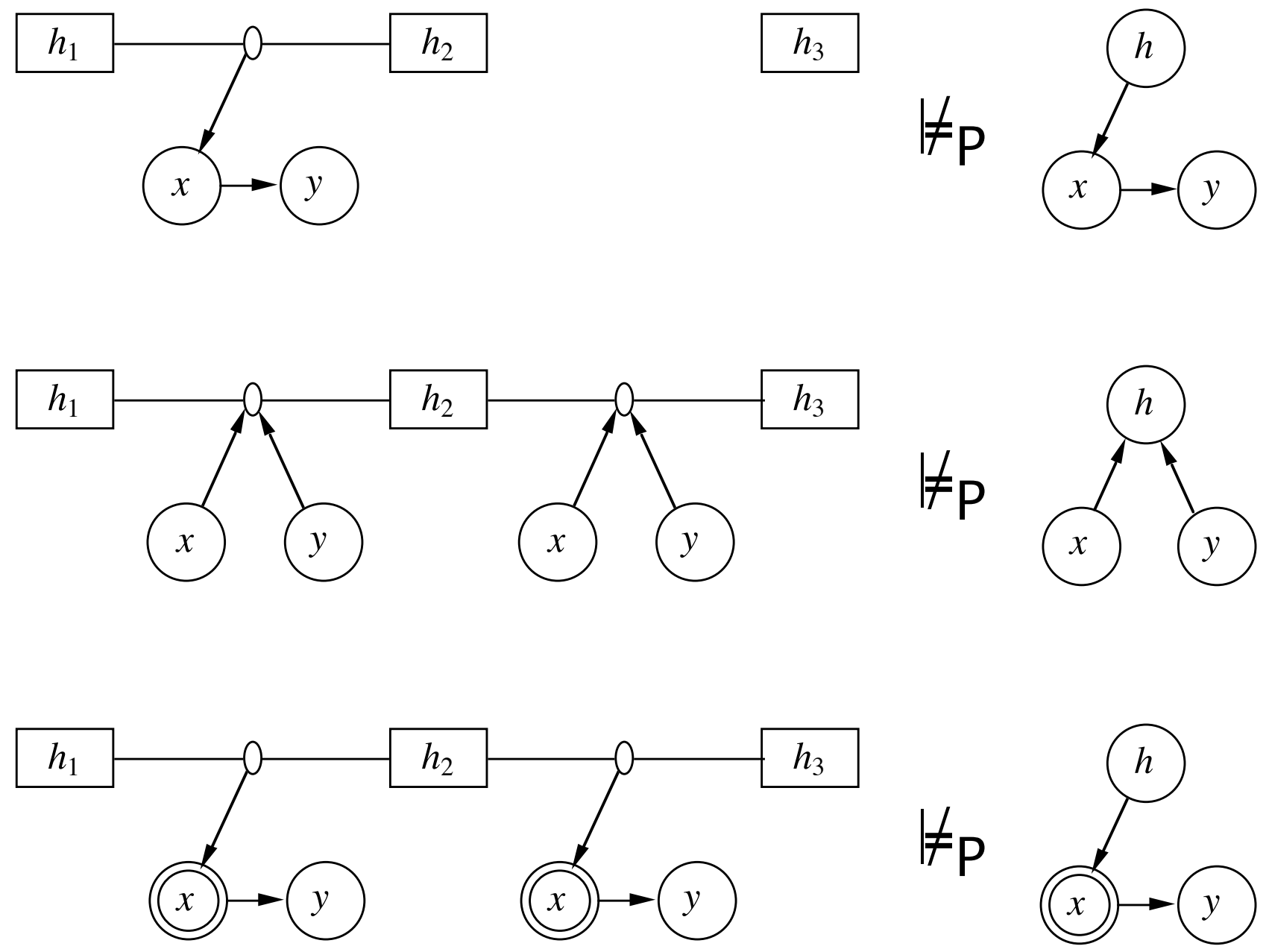

**Figure 3.4:** Exceptions to the soundness of c-global map construction.
The comprehensive similarity network in (a) is not connected. In (b), the comprehensive network contains nodes that condition the distinguished node $h$. The network in (c) contains deterministic relationships. In all three cases, the constraints of the constructor do not logically imply the constraints of the constructed representation. Consequently, each construction is not sound.

**Definition 3.14** *A similarity network (hypothesis-specific, comprehensive, or ordinary) is* **consistent** *if and only if there exists a joint distribution over the variables in the network that satisfies the constraints of each knowledge map in the network.*

A knowledge map is always consistent by design. However, there is no such guarantee for similarity networks. For example, consider the hypothesis-specific similarity network shown in Figure 3.5(a). The relevance sets $\mathcal{R}^{12}$ and $\mathcal{R}^{23}$ contain the assertions

$$p(x|h_1, \xi) = p(x|h_2, \xi)$$

$$p(x|h_2, \xi) = p(x|h_3, \xi)$$

These assertions are contradicted by the assertion

$$p(x|h_1, \xi) \neq p(x|h_3, \xi)$$

in the relevance set $\mathcal{R}^{13}$. Consequently, no joint distribution over $h$ and $x$ can satisfy this network, and the hypothesis-specific network is inconsistent. Also, consider the comprehensive similarity network shown in Figure 3.5(b). If we suppose that the similarity network is minimal, then the c-local map $h_1$–$h_2$ reflects the assertion that $x$ and $y$ are dependent given hypothesis $h_2$ (and $h_1$). However, the c-local map $h_2$–$h_3$ represents the assertion that $x$ and $y$ are independent given hypothesis $h_2$ (and $h_3$). Again, no joint distribution over the distinguished variable and nondistinguished variables can satisfy this network, and the comprehensive network is inconsistent. In Section 3.7, we examine an inconsistent ordinary similarity network.

From a technical standpoint, a similarity network (hypothesis-specific, comprehensive, or ordinary) does not have to be consistent in order for a construction from that network to be sound. In fact, the constructions that we have examined are sound whenever the constructor network is inconsistent, because we can derive any set of propositions from an inconsistent set of constraints. From the standpoint of knowledge acquisition, however, it is not enough to know that a construction from a directly composed set of constraints (e.g., an ordinary similarity network) to another set of constraints (e.g., an o-global map) is sound. To avoid the creation of a model that recommends nonoptimal decisions, we also must guarantee that the original set of constraints is consistent. Consequently, the following definition will be useful.

**Definition 3.15** *A set of probability constraints is* **valid** *for an individual if and only if that set is logically implied by a consistent set of probability constraints asserted directly by the individual.*

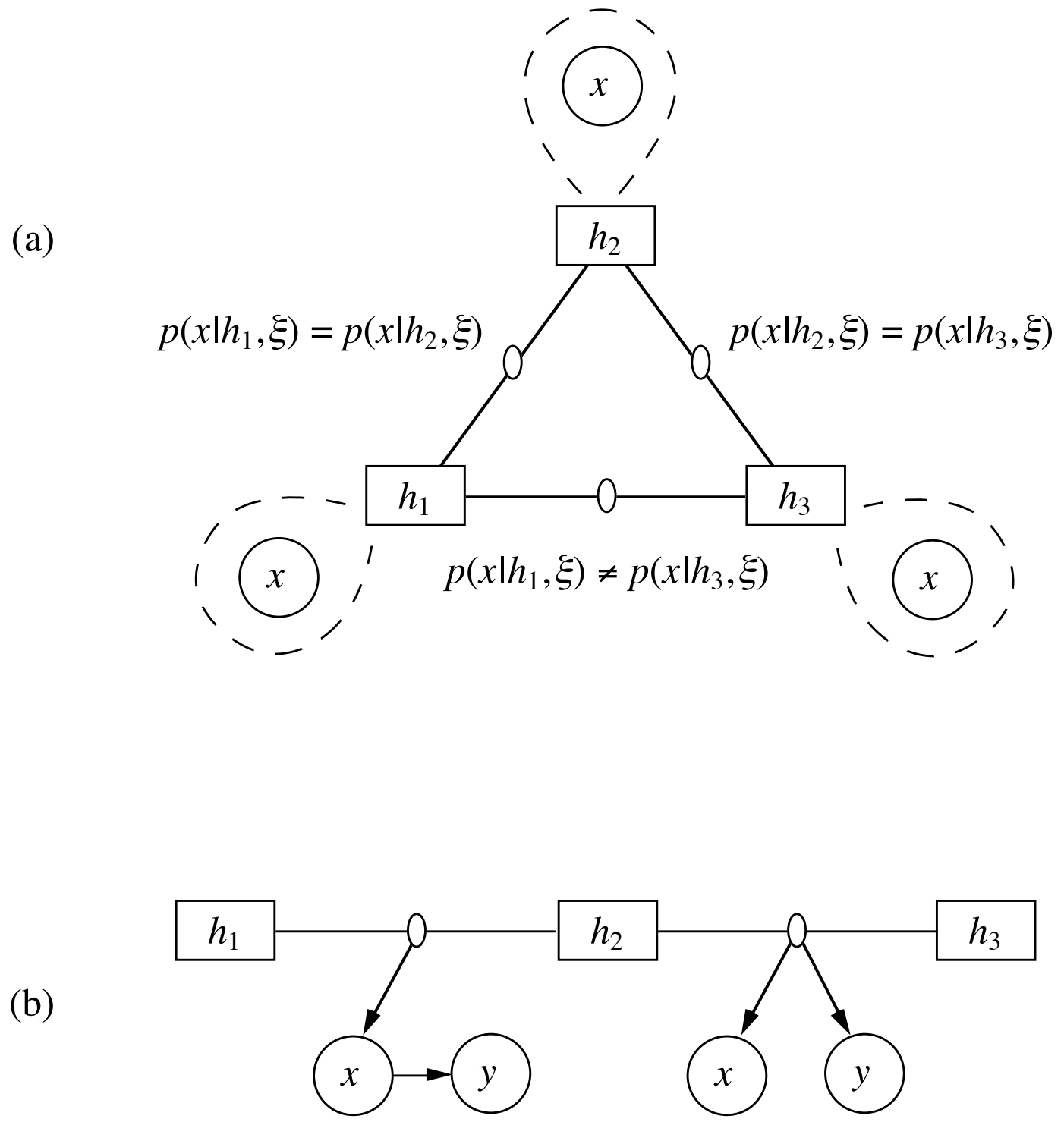

**Figure 3.5:** Inconsistent similarity networks.
The hypothesis-specific similarity network in (a) is inconsistent because it simultaneously asserts $p\,(x|h_1, \xi) = p\,(x|h_2, \xi)$, $p\,(x|h_2, \xi) = p\,(x|h_3, \xi)$, and $p\,(x|h_1, \xi) \neq p\,(x|h_3, \xi)$, which is impossible. The comprehensive similarity network in (b) is inconsistent because the c-local map $h_1$–$h_2$ implies that $x$ and $y$ are dependent, given $h_2$, whereas the c-local map $h_2$–$h_3$ implies that $x$ and $y$ are independent, given $h_2$.

When the author of a set of constraints is implicit, as it usually will be in our discussions, we say simply that the set is valid. Definition 3.15 parallels the definition of validity in logic. In particular, a logic sentence is valid if and only if it can be derived from $\emptyset$, the empty set of sentences. We obtain a correspondence between Definition 3.15 and this definition, if we identify $\emptyset$ with $\xi$, the background knowledge of a person with coherent beliefs.

In the remainder of this section, we develop an algorithm for testing the consistency of a comprehensive similarity network. Given the soundness of the construction $\mathcal{C} \mapsto \mathcal{G}_c$, we thereby show that we can construct valid c-global maps.

Identifying inconsistent hypothesis-specific networks is straightforward. The hs maps in a hypothesis-specific network cannot contradict one another, because the assertions in the hs map $\widehat{h_i}$ are conditioned on the belief that $h_i$ is true, and the hypotheses are mutually exclusive. In addition, the assertions in the relevance sets cannot conflict with one another unless there is a cycle in the similarity graph, as illustrated in Figure 3.5(a). Thus, we have the following theorem.

**Theorem 3.8** *A hypothesis-specific similarity network is consistent if and only if there is no cycle in the similarity graph such that, for any nondistinguished node $y$, the assertion*

$$p\left(y|C^{i/j}(y), h_i, \xi\right) = p\left(y|C^{i/j}(y), h_j, \xi\right) \tag{3.5.17}$$

*is in all but one relevance set $\mathcal{R}^{ij}$ in the cycle.*

**Proof:** See Appendix B.4.

Identifying inconsistent comprehensive similarity networks is only slightly more complicated. First note that all nonminimal comprehensive networks are consistent. This observation follows because nonminimal networks only represent assertions of conditional independence. Hence, any joint distribution in which every variable is independent of all others will satisfy the assertions in the network. As mentioned previously, however, nonminimal knowledge maps and hence nonminimal similarity networks are rarely composed in practice. To understand the situation for minimal comprehensive networks, let us consider the inconsistent comprehensive network in Figure 3.6. (This network is the same network that we saw in Figure 3.5b.) Suppose we were to construct this network from a hypothesis-specific network using the transformation defined in Section 3.3. Because the c-local map $h_1$–$h_2$ contains an arc from $x$ to $y$ and no arc from $h$ to $y$, we must place an arc from $x$ to $y$ in both hs maps $\widehat{h_1}$ and $\widehat{h_2}$. If we did not place such an arc in either hs map, there would be no arc from $x$ to $y$ in the c-local map $h_1$–$h_2$. If we placed such an arc in only one of $\widehat{h_1}$ and $\widehat{h_2}$, the conditional predecessors of $y$ would be different in the two hs maps, resulting in an arc from $h$ to $y$ in $h_1$–$h_2$. However, because there is no arc between $x$ and $y$ in the c-local map $h_2$–$h_3$, we cannot place such an arc from $x$ to $y$ in the hs map $\widehat{h_2}$. Thus, it is impossible to construct the inconsistent comprehensive network from any hypothesis-specific network.

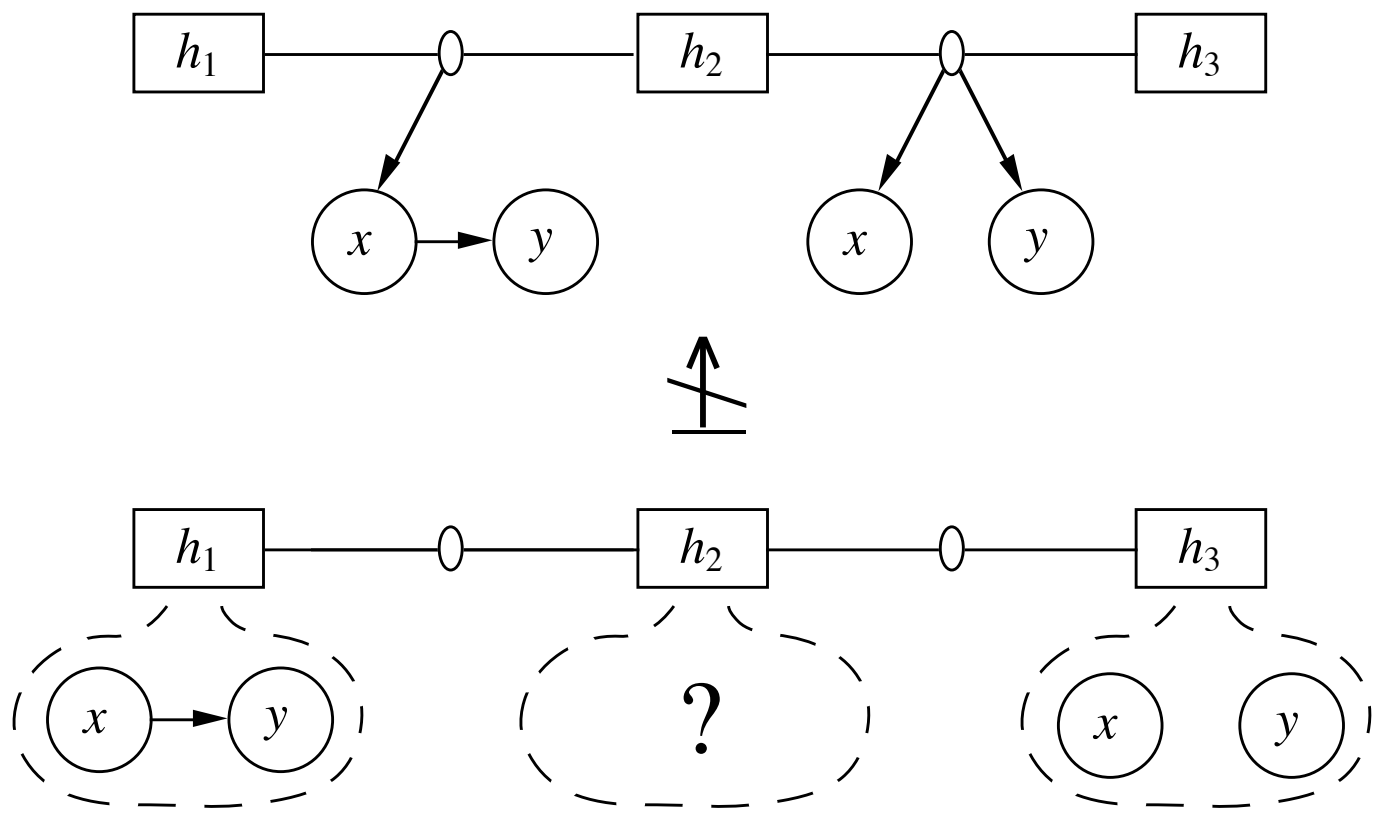

**Figure 3.6:** An inconsistent comprehensive similarity network.
The comprehensive similarity network in the upper half of the figure is inconsistent. Given the rules for constructing comprehensive networks, it cannot be constructed from any hypothesis-specific similarity network.

Also, consider the comprehensive network of Figure 3.7(a). The network is inconsistent because the the c-local maps $h_1$–$h_2$, $h_2$–$h_3$, and $h_1$–$h_3$ represent the assertions

$$p(x|h_1, \xi) = p(x|h_2, \xi)$$

$$p(x|h_2, \xi) = p(x|h_3, \xi)$$

$$p(x|h_1, \xi) \neq p(x|h_3, \xi)$$

respectively. The network can be only constructed from the hypothesis-specific similarity network shown in Figure 3.7(b). This hypothesis-specific network is the same inconsistent network that we saw in Figure 3.5(a).

In general, as described in the following theorem, a comprehensive similarity network is consistent only if it can be constructed from consistent hypothesis-specific similarity network.

**Theorem 3.9** *If the constraints of a minimal comprehensive similarity network are satisfied by some strictly positive joint distribution over the variables in the network (making the network consistent), then the distribution satisfies the constraints of some minimal hypothesis-specific similarity network that is a constructor of that network.*

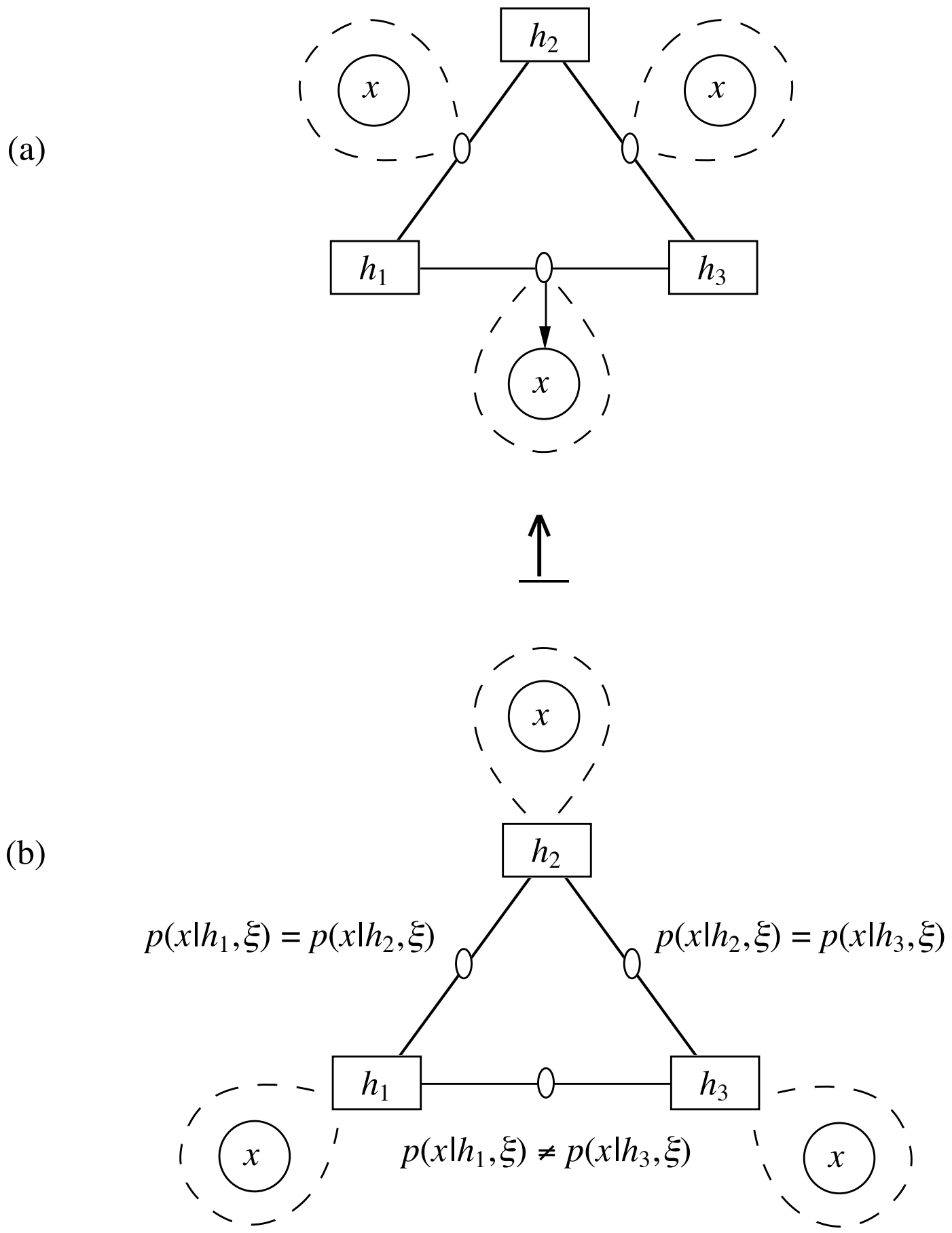

**Figure 3.7:** An inconsistent comprehensive similarity network with cycles. The comprehensive similarity network depicted in (a) can be only constructed from the inconsistent hypothesis-specific similarity network in (b).

**Proof:** Let $\mathcal{C}$ be the given network and let $\leq_E$ be an expansion order such that (1) the order is consistent with the c-global map constructed from $\mathcal{C}$, and (2) the distinguished variable $h$ occurs first in the order. By definition of comprehensive similarity networks, $\leq_E$ is consistent with every c-local map in $\mathcal{C}$. We use this expansion order (excluding $h$) in conjunction with the given distribution to construct a minimal hypothesis-specific similarity network $\mathcal{HS}$, and let $\mathcal{C}'$ be the comprehensive network constructed from $\mathcal{HS}$. By Theorem 3.7(a), the given distribution satisfies the constraints implied by the c-local maps of $\mathcal{C}'$, where $\mathcal{C}'$ is minimal. Also, because all arcs emanate from $h$ in the c-local maps of $\mathcal{C}'$, the expansion order $\leq_E$ must be consistent with $\mathcal{C}'$. It follows from Theorem 3.6 that the maps of each pair must be identical. □

Given Theorem 3.9, we can delineate conditions that are necessary for a minimal comprehensive similarity network to be consistent. The terms $x \longrightarrow y$ and $x \not\longrightarrow y$ are used to abbreviate the statements "arc from $x$ to $y$" and "no arc from $x$ to $y$," respectively.

**Corollary 3.1** *Given a minimal comprehensive similarity network $\mathcal{C}$ that is consistent for some strictly positive distribution, there is a hypothesis-specific similarity network in which the following conditions hold:*

*For each arc from nondistinguished node $x$ to nondistinguished node $y$ in the c-global map constructed from $\mathcal{C}$, and for each c-local map $h_i$–$h_j$ in $\mathcal{C}$,*

*a. If $x \not\longrightarrow y$ in $h_i$–$h_j$, then $x \not\longrightarrow y$ in $\widehat{h_i}$ and $x \not\longrightarrow y$ in $\widehat{h_j}$*

*b. If $x \longrightarrow y$ in $h_i$–$h_j$, then*

*i. If $h \longrightarrow y$, then $x \longrightarrow y$ in $\widehat{h_i}$ or $x \longrightarrow y$ in $\widehat{h_j}$*

*ii. If $h \not\longrightarrow y$, then $x \longrightarrow y$ in $\widehat{h_i}$ and $x \longrightarrow y$ in $\widehat{h_j}$*

**Proof:** Conditions a and b.i follow from Theorem 3.9 and from the procedure for constructing a comprehensive similarity network from a hypothesis-specific network. To prove that condition b.ii must hold, suppose there is no arc from $h$ to $y$ in $h_i$–$h_j$. In this case, $C^i(y) = C^j(y)$. Otherwise, the construction dictates that there is an arc from $h$ to $y$. Thus, there is an arc from $x$ to $y$ in $\widehat{h_i}$ if and only if there is such an arc in $\widehat{h_j}$. Because there is an arc from $x$ to $y$ in $h_i$–$h_j$, there must be a corresponding arc in both hs maps. □

**Corollary 3.2** *A minimal comprehensive similarity network $\mathcal{C}$ is consistent for some strictly positive distribution only if, for every nondistinguished node in the network, there is no cycle in the similarity graph of $\mathcal{C}$ such that $h \longrightarrow y$ is in exactly one c-local map of the cycle.*

**Proof:** The theorem follows directly from Theorem 3.9 and the procedure for constructing a comprehensive similarity network from a hypothesis-specific network. □

The converse of Corollaries 3.1 and 3.2 also holds. That is, if the constraints in the corollaries are satisfied for a given comprehensive network, the network must be consistent. Thus, the constraints are both necessary and sufficient conditions for a comprehensive network to be consistent. The result is proved by way of the following algorithm, which tests whether or not a comprehensive network satisfies the constraints in Corollaries 3.1 and 3.2 and constructs a hypothesis-specific network if the constraints are satisfied.

**Algorithm 3.1 (Consistency, comprehensive networks)**

For every pair of nondistinguished nodes $x$ and $y$ such that $x \longrightarrow y$ in the c-global map constructed from the given network do
> For every c-local map $h_i$–$h_j$ such that $x \not\longrightarrow y$ do
>> Post the constraint "$x \not\longrightarrow y$" on $h_i$ and on $h_j$
>
> For every c-local map $h_i$–$h_j$ such that $x \longrightarrow y$ do
>> If $h \longrightarrow y$ and "$x \not\longrightarrow y$" is posted on $h_i$ and on $h_j$ then
>>> Return "inconsistent"
>>
>> Else if $h \not\longrightarrow y$ and "$x \not\longrightarrow y$" is posted on $h_i$ or on $h_j$ then
>>> Return "inconsistent"
>
> For every hypothesis $h_i$ do
>> If the constraint "$x \not\longrightarrow y$" is not posted on $h_i$ then
>>> Add $x \longrightarrow y$ to $\widehat{h_i}$

For every nondistinguished node $y$ in the c-global map do
> For every c-local map $h_i$–$h_j$ where $h \longrightarrow y$ do
>> From the similarity graph, construct the edge-induced subgraph, $\mathcal{G}$, containing edge $(h_i, h_j)$, and edges $(h_k, h_l)$ such that $h \not\longrightarrow y$ in $h_k$–$h_l$
>>
>> If the edge $(h_i, h_j)$ is in a cycle in $\mathcal{G}$
>>> Return "inconsistent"

For every nondistinguished node $y$ in the c-global map do
> For every c-local map $h_i$–$h_j$ such that $C^i(y) = C^j(y) \equiv C^{i/j}(y)$ do
>> If $h \not\longrightarrow y$ then
>>> Add "$p\left(y|C^{i/j}(y), h_i, \xi\right) = p\left(y|C^{i/j}(y), h_j, \xi\right)$" to $\mathcal{R}^{ij}$
>>
>> Else if $h \longrightarrow y$ then
>>> Add "$p\left(y|C^{i/j}(y), h_i, \xi\right) \neq p\left(y|C^{i/j}(y), h_j, \xi\right)$" to $\mathcal{R}^{ij}$

Return "consistent"

**Theorem 3.10 (Consistency, comprehensive networks)** *Algorithm 3.1 applied to a comprehensive similarity network returns "consistent" if and only if there is a strictly positive joint distribution that makes the network consistent and minimal. Moreover, if Algorithm 3.1 returns "consistent," it generates the hypothesis-specific network that is the maximal constructor of the given network.*

**Proof:** See Appendix B.5.

Figure 3.8(b) illustrates the hypothesis-specific network created by the algorithm applied to the consistent comprehensive network in Figure 3.8(a). Because the arc from $x$ to $y$ is the only arc among nondistinguished nodes in the c-global map constructed from the given comprehensive network, the for-loop beginning at line 1 inspects only this arc on the c-local maps $h_1$–$h_2$ and $h_2$–$h_3$. Also, because there is an arc from $x$ to $y$ on both c-local maps, no constraints of the form "$x \not\to y$" are posted. Consequently, the for-loop of line 1 does not return "inconsistent," and an arc is added to each $\widehat{h_i}$, as shown in Figure 3.8. The for-loop beginning at line 12 does not return "inconsistent," because there are no cycles in the similarity graph. The for-loop beginning at line 17 adds the assertions of relevance and irrelevance necessary to complete the construction of the hypothesis-specific similarity network. Notice that the network created by the algorithm is a maximal constructor of the comprehensive network.

Figure 3.9 shows the results of the algorithm applied to the inconsistent similarity network that we saw in Figures 3.5(b) and 3.6. As in the previous example, the for-loop beginning at line 1 looks for only the arc from $x$ to $y$ on each of the two c-local maps of the comprehensive similarity network. Because there is no such arc on $h_2$–$h_3$, the constraints "$x \not\to y$" are posted on both $h_2$ and $h_3$, as shown in the Figure 3.9. Because there is an arc from $x$ to $y$ in $h_1$–$h_2$ and the constraint "$x \not\to y$" is posted on $h_2$, the condition at line 7 is satisfied and the algorithm returns "inconsistent."

In concluding this section, let us consider the time complexity of Algorithm 3.1. Assume knowledge maps (hypothesis-specific, local, and global) are represented so that the direct predecessors of a given node and the nodes bordering a given arc can be accessed in $O(1)$. Assume that the set of nodes and the set of arcs in each map are stored in hash tables, so that the test to see whether a node or arc is a member of a given map is $O(1)$; further assume that these sets are also stored in linear lists, so that the iteration over nodes and arcs is efficient. In addition, assume that the number of direct predecessors of any node is bounded by some constant, so that the test for equality of two direct predecessor sets is $O(1)$.

Under these assumptions, the time complexity of lines 1 through 11 is $O(al)$, where $a$ is the number of arcs in the c-global map and $l$ is the number of c-local maps (i.e., the number of edges in the similarity graph). The time complexity of lines 12 through 16 is $O(nl(l+hd))$, where $n$ is the number of nodes in the c-global map, $h$ is the number of hypotheses in the similarity graph, and $d$ is the largest degree of any node in the similarity graph.[4] The term $hd$ corresponds to line 15, where we determine whether or not the edge $(h_i, h_j)$ is in a cycle in the subgraph $\mathcal{G}$. To accomplish this task, we remove the edge $(h_i, h_j)$ from $\mathcal{G}$, mark node $h_i$, and mark all neighbors of $h_i$. Next, we repeatedly mark all neighbors of each newly marked node, until no new nodes are marked. There is a path from $h_i$ to $h_j$ in this modified graph, and hence a cycle in $\mathcal{G}$ that includes $(h_i, h_j)$, if and only if node $h_j$ is marked. Finally, the time complexity of lines 17 through 22 is $O(nl)$. Overall, because $n \leq a-1$, $h \leq l-1$, and $d \leq l$, the time complexity of Algorithm 3.1 is $O(al^3)$.

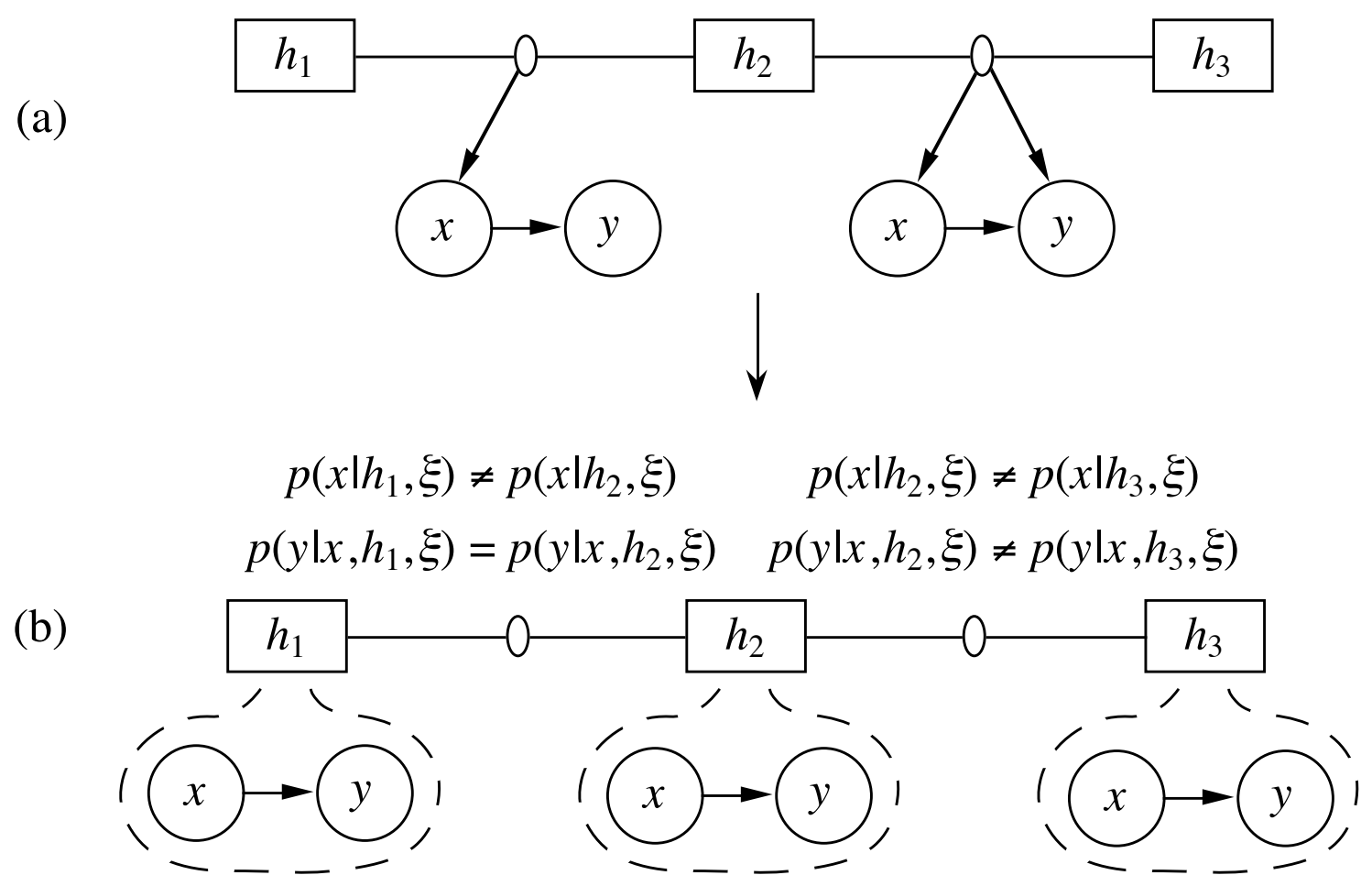

**Figure 3.8:** Algorithm 3.1 applied to a consistent comprehensive network.
Algorithm 3.1 applied to the consistent comprehensive network in (a) produces the hypothesis-specific similarity network in (b). Note that the hypothesis-specific network is a maximal constructor of the comprehensive network.

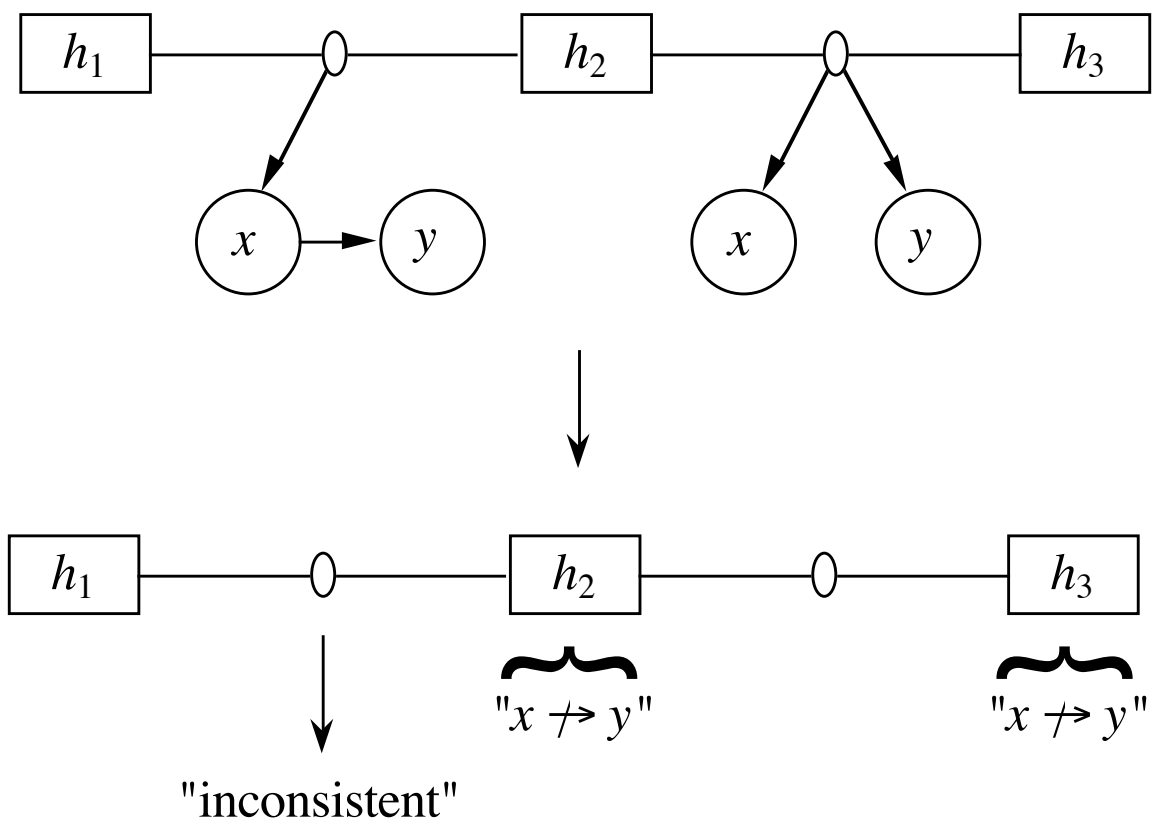

**Figure 3.9:** Algorithm 3.1 applied to an inconsistent comprehensive network.
The figure illustrates the results of applying Algorithm 3.1 to the inconsistent comprehensive network shown in Figures 3.5(b) and 3.6. Since there is no arc from $x$ to $y$ in the c-local map $h_2$–$h_3$, the constraint "$x \not\rightarrow y$" is posted on $h_2$ and $h_3$. The algorithm returns "inconsistent" when it encounters the arc from $x$ to $y$ in the c-local map $h_1$–$h_2$, because there is no arc from $h$ to $y$.

## 3.6 Soundness: Ordinary Similarity Networks

Armed with criteria for evaluating the consistency of a comprehensive similarity network, we can now prove that the construction of the o-global map from an ordinary similarity network is sound for strictly positive distributions. Suppose we are given a minimal comprehensive similarity network that has been certified consistent by Algorithm 3.1. As shown in the following theorem, the comprehensive and ordinary global knowledge maps constructed from this network must be identical.

**Lemma 3.1** *Given a minimal comprehensive similarity network that is consistent for strictly positive distributions, if there is an arc from $x$ to $y$ and no arc from $h$ to $y$ in the c-local map $h_i$–$h_j$, then there will be an arc from $x$ to $y$ in all c-local maps that are bordered by $h_i$ or $h_j$ in the similarity graph.*

**Proof:** Since the network is minimal and consistent for strictly positive distributions, we know from Theorem 3.9 that it can be constructed by some hypothesis-specific similarity network. By Corollary 3.1(b.ii), there must be an arc from $x$ to $y$ in both hs maps $\widehat{h_i}$ and $\widehat{h_j}$ of this network. Hence, by construction, there must be such an arc on each c-local map that is bordered by $h_i$ or $h_j$. $\square$

**Theorem 3.11** *Given a minimal comprehensive similarity network that is consistent for strictly positive distributions, the c-global map and o-global map constructed from the network are identical.*

[4]The degree of a node in an undirected graph is the number of edges that touch that node.

**Proof:** Suppose the theorem is false. In this case, there must be an arc between two nodes—say, from $x$ to $y$—that is present on some c-local map only when both $x$ and $y$ are disconnected from $h$. Starting from this c-local map, traverse the similarity graph until a c-local map where $y$ is connected to $h$ is encountered. Call this c-local map $h_i$–$h_j$. (Because the similarity graph is connected, if no such a map is found, the node $y$ is disconnected from the c-global, and we can ignore the node.) On each c-local map along the path of this traversal, we can apply Lemma 3.1, because $y$ is disconnected from $h$ in the map. It follows that there must be an arc from $x$ to $y$ in $h_i$–$h_j$. However, this result contradicts our original assumption, because $y$ and therefore $x$ are connected to $h$ on $h_i$–$h_j$. □

To make the argument in Theorem 3.11 more concrete, let us apply it to the comprehensive similarity network in Figure 3.10. In the c-local map $h_1$–$h_2$, there is an arc from node $x$ to node $y$, and these nodes are disconnected from $h$. Consequently, we do not copy this arc to the o-global map. However, there is no arc from $h$ to $y$ in $h_1$–$h_2$, so we can apply Lemma 3.1 to conclude that there must be an arc from $x$ to $y$ in the adjoining c-local map $h_2$–$h_3$. In this c-local map, both $x$ and $y$ are connected to $h$, and the arc from $x$ to $y$ is registered in the o-global map.

Given this equality, the soundness of o-global map construction now follows.

**Theorem 3.7(d) (Soundness, o-global map construction)** The construction of an o-global map from an ordinary similarity network is sound for strictly positive distributions. The construction remains sound if both representations are minimal.

**Proof:** First, consider the case where an ordinary similarity network $\mathcal{O}$ is minimal. Suppose some strictly positive joint distribution satisfies the constraints of the network, making $\mathcal{O}$ consistent. (If there is no such distribution, then the constraints of $\mathcal{O}$ imply trivially the constraints of the o-global map constructed from $\mathcal{O}$.) By definition of ordinary similarity networks, the distribution satisfies the constraints of some minimal comprehensive similarity network—say, $\mathcal{C}$. Furthermore, by Theorem 3.7(b), it follows that the joint distribution also satisfies the constraints implied by the c-global map $\mathcal{G}_c$ constructed from $\mathcal{C}$. By Theorem 3.11, however, $\mathcal{G}_c$ and the o-global map constructed from $\mathcal{O}$ are identical. Hence, the joint distribution satisfies the constraints of the o-global map.

If the ordinary network is not minimal, use the joint distribution to remove arcs from the o-local maps until the maps are minimal. Call the new network $\mathcal{O}'$. Applying the argument in the previous paragraph to $\mathcal{O}'$, it follows that the given distribution satisfies the constraints of the o-global map constructed from $\mathcal{O}'$. However, by construction, all arcs in the o-global map of $\mathcal{O}'$ appear in the o-global map of $\mathcal{O}$. Consequently, the assertions of conditional independence implied by the o-global of the original network are satisfied by the joint distribution. □

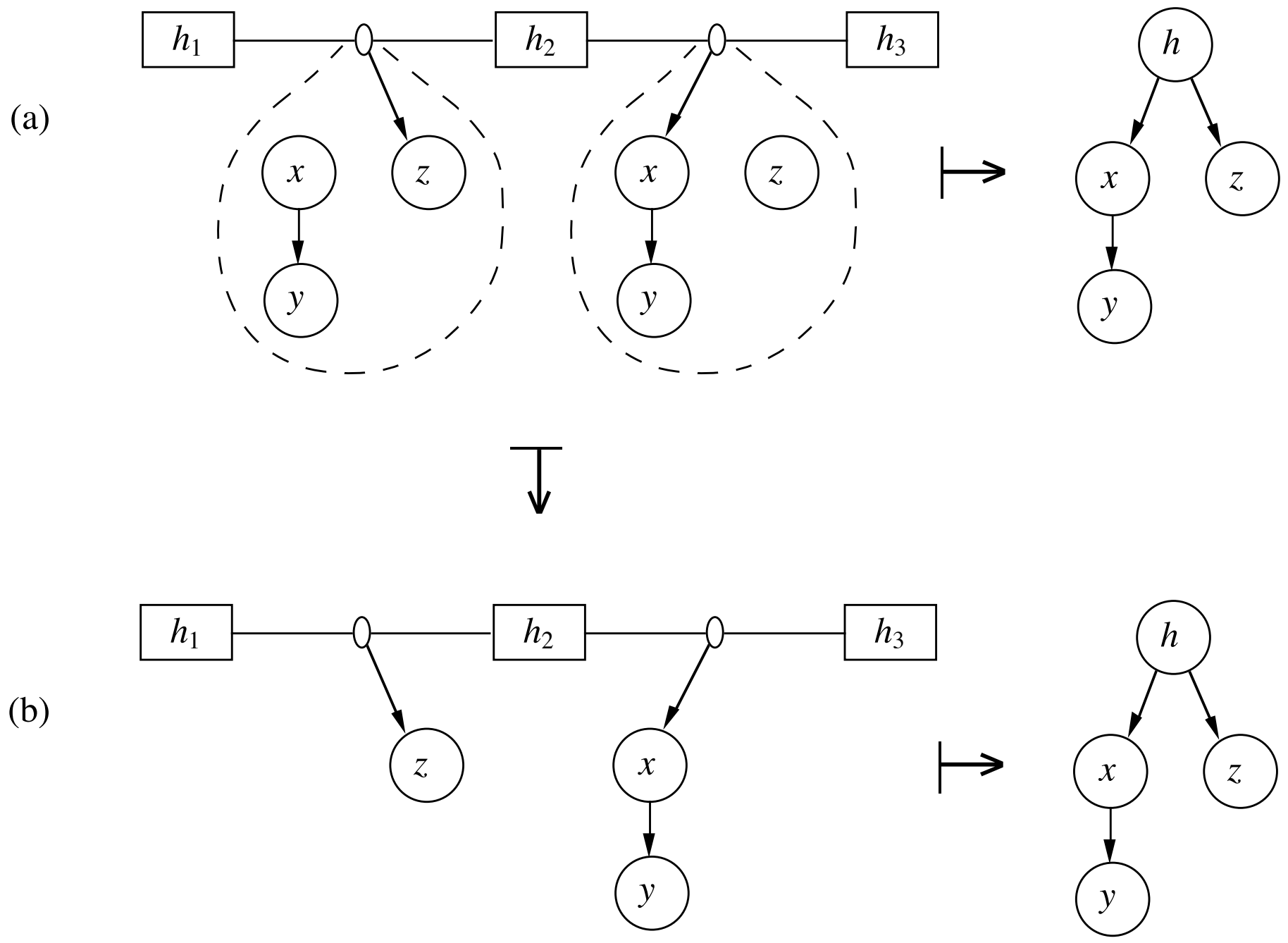

**Figure 3.10:** An example of the equivalence of c-global and o-global maps.
(a) The c-global map on the right is constructed from the comprehensive similarity network on the left. (b) The o-global map on the right is constructed from the ordinary similarity network on the left. Although nodes $x$ and $y$ are missing from the o-local map $h_1$–$h_2$ and node $z$ is not in the o-local map $h_2$–$h_3$, the c-global and o-global maps are identical.

Theorem 3.7(d) is the first of the three major results of this chapter. It says that we do not have to go through the time-consuming process of assessing c-local maps. By constructing o-local maps, ignoring the dependencies among variables disconnected from those maps, we recover a global knowledge map that contains only assertions that can be derived from the c-local maps.

## 3.7 Consistency: Ordinary Similarity Networks

As we have discussed, the soundness result does not carry much force unless we can identify and correct inconsistent ordinary similarity networks. Consequently, we need to extend Algorithm 3.1 to include ordinary networks.

Suppose we are given such a network. For every pair of nondistinguished nodes $x$ and $y$ such that there is an arc from $x$ to $y$ in the o-global map (or, equivalently, in the c-global map) constructed from this network, one of the following three cases must hold in each o-local map $h_i$–$h_j$: (1) both $x$ and $y$ are in the map, (2) only one of $x$ and $y$ are in the map, or (3) neither $x$ nor $y$ are in the map. If case 1 holds, we can directly apply the machinery of Algorithm 3.1 to the nodes $x$ and $y$ and the arc between them. If case 2 holds, we know that there is no arc from $x$ to $y$ in any c-local map from which the o-local map $h_i$–$h_j$ can be constructed. Consequently, we can post the constraint "$x \not\rightarrow y$" on both $h_i$ and $h_j$. If case 3 holds, we cannot tell directly whether there is an arc from $x$ to $y$ in any c-local map from which the o-local map $h_i$–$h_j$ can be constructed. However, if we know that the constraint "$x \not\rightarrow y$" is posted on either $h_i$ or $h_j$, then we know that there cannot be such an arc, and we can post the constraint "$x \not\rightarrow y$" on both $h_i$ and $h_j$. Thus, with the minor modifications to Algorithm 3.1 suggested by these comments, we obtain the following algorithm for testing the consistency of an ordinary similarity network.

**Algorithm 3.2 (Consistency, ordinary networks)**

```
For every pair of nondistinguished nodes x and y such that x ⟶ y in
the o-global map constructed from the given network do
    For every o-local map h_i–h_j such that x -/-> y do
        Post the constraint "x -/-> y" on h_i and on h_j
```

For every o-local map $h_i$–$h_j$ such that only one of $x$ and $y$ is on the map do
  Post the constraint "$x \not\longrightarrow y$" on $h_i$ and on $h_j$

Mark all o-local maps as unvisited
While there is an unvisited o-local map containing neither $x$ nor $y$
such that the constraint "$x \not\longrightarrow y$" is posted on $h_i$ or on $h_j$ do
  Post the constraint "$x \not\longrightarrow y$" on $h_i$ and on $h_j$
  Mark the o-local map as visited

For every o-local map $h_i$–$h_j$ such that $x \longrightarrow y$ do
  If $h \longrightarrow y$ and "$x \not\longrightarrow y$" is posted on $h_i$ and on $h_j$ then
    Return "inconsistent"
  If $h \not\longrightarrow y$ and "$x \not\longrightarrow y$" is posted on $h_i$ or on $h_j$ then
    Return "inconsistent"

For every hypothesis $h_i$ do
  If the constraint "$x \not\longrightarrow y$" is not posted on $h_i$ then
    Add $x \longrightarrow y$ to $\widehat{h_i}$

For every nondistinguished node $y$ in the c-global map do
  For every c-local map $h_i$–$h_j$ where $h \longrightarrow y$ do
    From the similarity graph, construct the edge-induced subgraph, $\mathcal{G}$,
    containing edge $(h_i, h_j)$, and edges $(h_k, h_l)$ such that $h \not\longrightarrow y$ in $h_k$–$h_l$
    If the edge $(h_i, h_j)$ is in a cycle in $\mathcal{G}$
      Return "inconsistent"

For every nondistinguished node $y$ in the c-global map do
  For every c-local map $h_i$–$h_j$ such that $C^i(y) = C^j(y) \equiv C^{i/j}(y)$ do
    If $h \not\longrightarrow y$ then
      Add "$p\left(y|C^{i/j}(y), h_i, \xi\right) = p\left(y|C^{i/j}(y), h_j, \xi\right)$" to $\mathcal{R}^{ij}$
    Else if $h \longrightarrow y$ then
      Add "$p\left(y|C^{i/j}(y), h_i, \xi\right) \neq p\left(y|C^{i/j}(y), h_j, \xi\right)$" to $\mathcal{R}^{ij}$

Return "consistent"

**Theorem 3.12 (Consistency, ordinary similarity networks)** *Algorithm 3.2 applied to an ordinary similarity network returns "consistent" if and only if there is a strictly* *positive distribution that makes the network consistent and minimal. Moreover, if Algorithm 3.2 returns "consistent," it generates the hypothesis-specific network that is the maximal constructor of the given network.*

**Proof:** See Appendix B.6.

Under the assumptions given in Section 3.5, the time complexities of lines 1 through 17, 18 through 22, and 23 through 28 are $O(al^2)$, $O(nl(l+hd))$, and $O(nl)$, respectively. The additional factor of $l$ in the first term is a consequence of the while-loop at line 7 of the algorithm. The loop may iterate $O(l)$ times, and the search for an unvisited o-local map is also $O(l)$. Overall, the time complexity of Algorithm 3.2 is $O(al^3)$—the same time complexity as that of Algorithm 3.1.

Figure 3.11 illustrates the results of Algorithm 3.2 applied to a consistent ordinary similarity network. Because the arc from $x$ to $y$ is the only arc between nondistinguished nodes in the o-global map constructed from the given network, the for-loop beginning at line 1 only looks for this arc on the two o-local maps. Because $y$ is not in $h_1$–$h_2$, the condition in line 4 of the algorithm is met, and the constraint "$x \not\rightarrow y$" is posted on $h_1$ and $h_2$. When the algorithm encounters the arc from $x$ to $y$ on $h_2$–$h_3$, neither the condition at line 11 nor that at line 13 is satisfied, because there is an arc from $h$ to $y$. Consequently, the algorithm returns "consistent." The for-loop at line 23 adds an arc from $x$ to $y$ in the hs map $\widehat{h_3}$, generating the maximal constructor of the ordinary network.

Figure 3.12 illustrates the results of Algorithm 3.2 applied to an inconsistent ordinary similarity network. As in the previous example, the for-loop at line 1 looks for only an arc from $x$ to $y$ on each o-local map. Because $y$ is missing from $h_1$–$h_2$, the condition at line 4 of the algorithm fires and the constraint "$x \not\rightarrow y$" is posted on $h_1$ and $h_2$. Similarly, because $x$ is missing from $h_3$–$h_4$, the constraint "$x \not\rightarrow y$" is posted on $h_3$ and $h_4$. When the algorithm finds the arc from $x$ to $y$ in $h_2$–$h_3$, the condition at line 11 is satisfied, and the algorithm returns "inconsistent."

Algorithm 3.2 and Theorem 3.12 constitute the second major result of this chapter. Given a minimal ordinary network, we can determine in a tractable manner whether or not the network is consistent for strictly positive distributions. The algorithm can easily be extended to assist in the correction of an inconsistent ordinary similarity network. In particular, a modified version of Algorithm 3.2 could return a list of arcs for each o-local map that would cause Algorithm 3.2, as it currently exists, to return "inconsistent." With such a list, it would be a simple matter for the author of the network to resolve any inconsistencies. Once these inconsistencies are corrected, we know, from the soundness result (Theorem 3.7d), that the o-global map constructed from the network is valid.

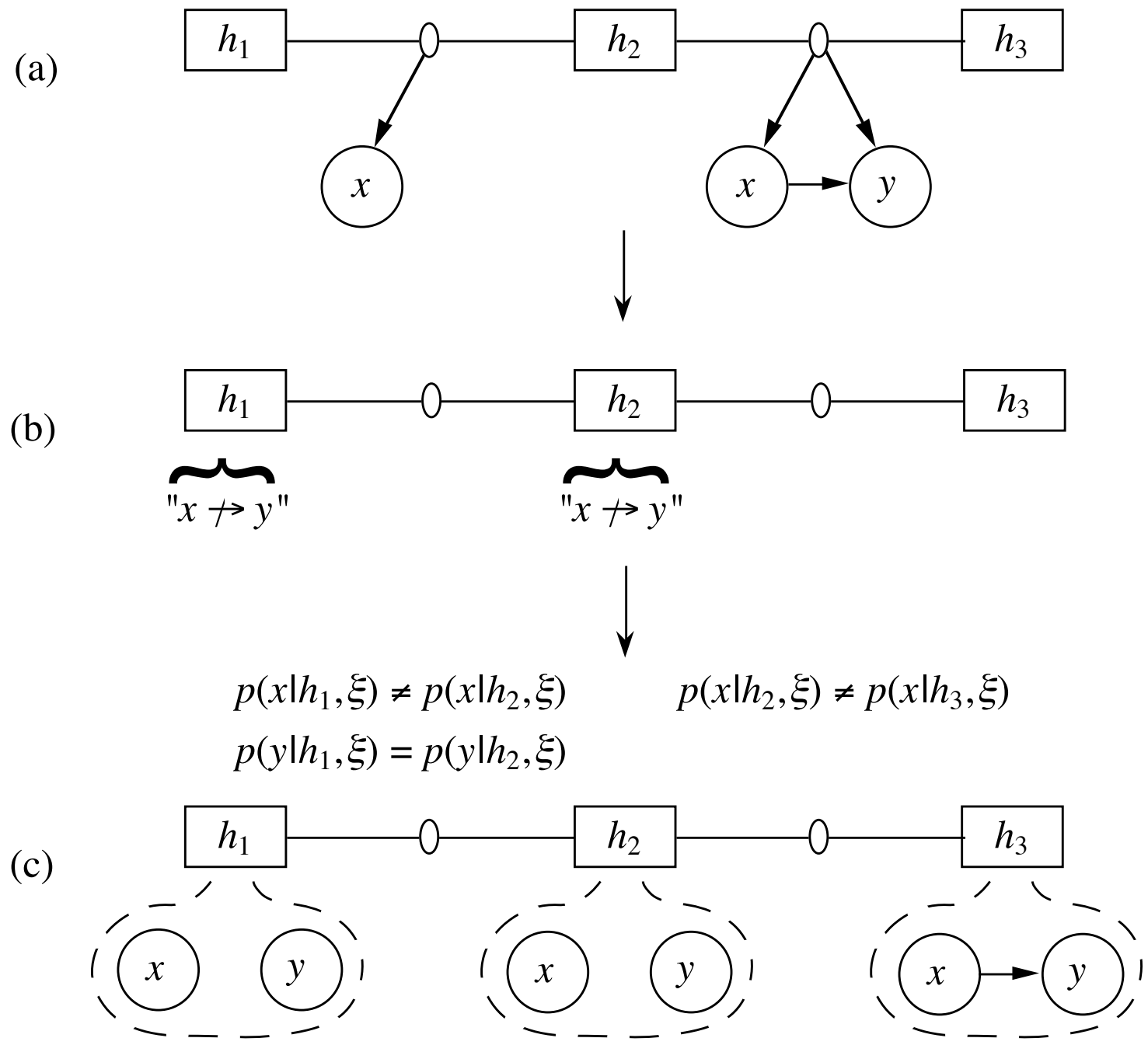

**Figure 3.11:** Algorithm 3.2 applied to a consistent ordinary network.
Algorithm 3.2 is applied to the consistent ordinary similarity network in (a). The constraints posted by the algorithm are shown in (b). Because $x$ and not $y$ is in the o-local map $h_1$–$h_2$, the constraint "$x \not\rightarrow y$" is posted on $h_1$ and $h_2$. Although there is an arc from $x$ to $y$ on the o-local map $h_2$–$h_3$, the algorithm returns "consistent" because there is an arc from $h$ to $y$ in map. The hypothesis-specific network created by the algorithm is shown in (c). Only $\widehat{h_3}$ contains an arc from $x$ to $y$ because $h_3$ is the only hypothesis on which the constraint "$x \not\rightarrow y$" is not posted. The hypothesis-specific network is a maximal constructor of the ordinary network.

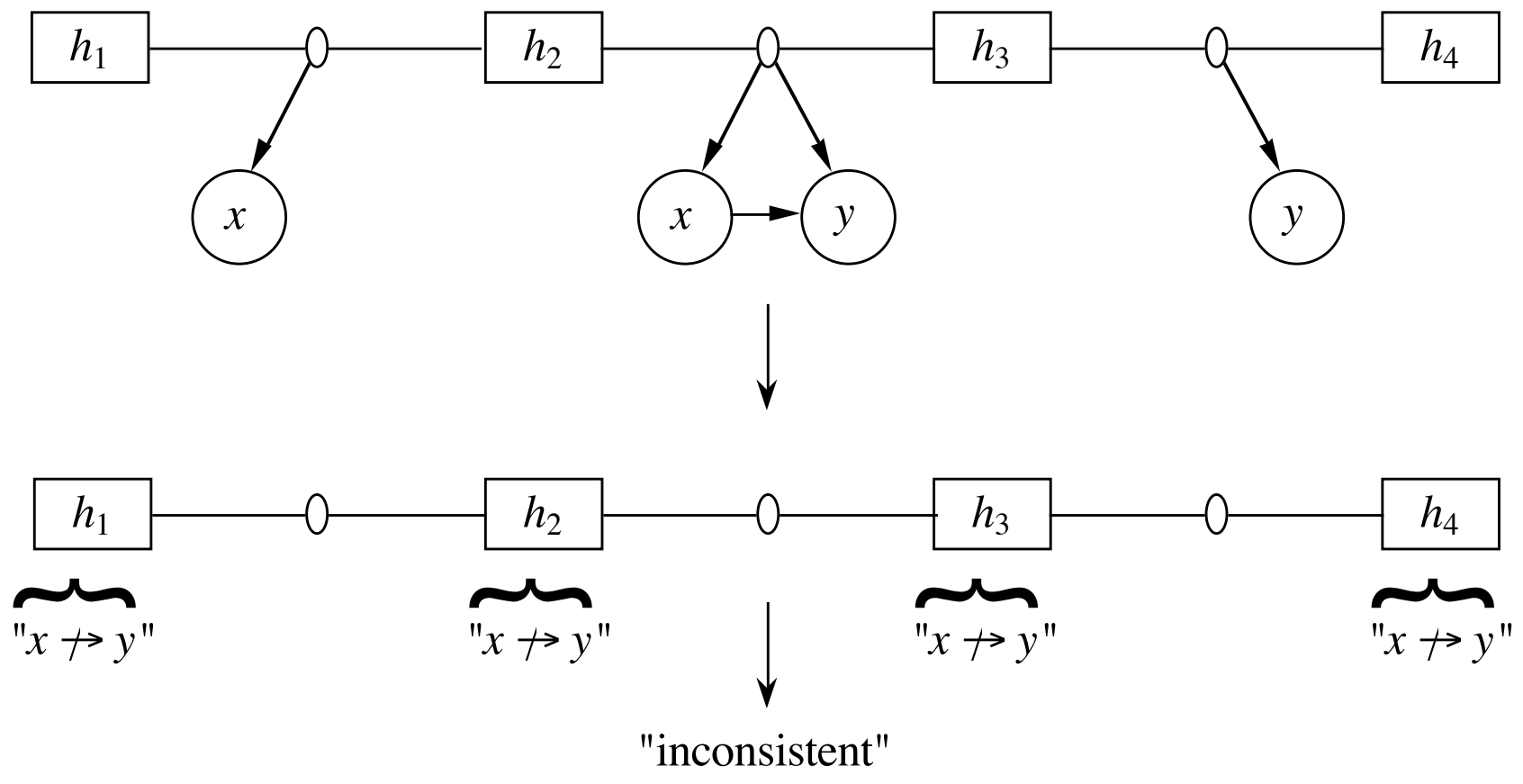

**Figure 3.12:** Algorithm 3.2 applied to an inconsistent ordinary network. Because only one of $x$ and $y$ are in the o-local maps $h_1$–$h_2$ and $h_3$–$h_4$, the constraint "$x \not\rightarrow y$" is posted on all the hs maps. The algorithm returns "inconsistent" when it encounters the arc from $x$ to $y$ on the o-local map $h_2$–$h_3$.

Note that Algorithm 3.2 may incorrectly identify a similarity network as inconsistent if the distributions underlying the network are not strictly positive. For example, consider the similarity network shown in Figure 3.13, in which the two nodes $x$ and $y$ are logically equivalent. In this network, an arc from $x$ to $z$ is equivalent to an arc from $y$ to $z$, and thus the similarity network is consistent. However, Algorithm 3.2 applied to this network returns "inconsistent" because, for example, "$y \not\rightarrow z$" is posted on $\widehat{h_2}$ from $h_1$–$h_2$ and there is an arc from $y$ to $z$ in $h_2$–$h_3$.

## 3.8 Another Definition of Ordinary Similarity Networks

In this section, we examine another definition of ordinary similarity networks that offers several advantages over the definition that we have used so far. Before we can discuss this definition, however, we must consider what it means for two variables $a$ and $b$ to *interact*.

**Definition 3.16** *Let $U$ denote a set of variables, and let $\xi$ denote the background knowledge of some expert. The variables $a, b \in U$* **interact** *in the universe $U$, given $\xi$, if and only if there exists some (possibly empty) subset $U' \subseteq U \setminus \{x, y\}$, such that $a$ and $b$ are conditionally dependent, given some instance of $U'$ and background knowledge $\xi$.*

Thus, whenever two variables $a$ and $b$ do not interact in the universe $U$, we know that $a$ and $b$ are conditionally independent, given all instances of all subsets of $U \setminus \{a, b\}$.

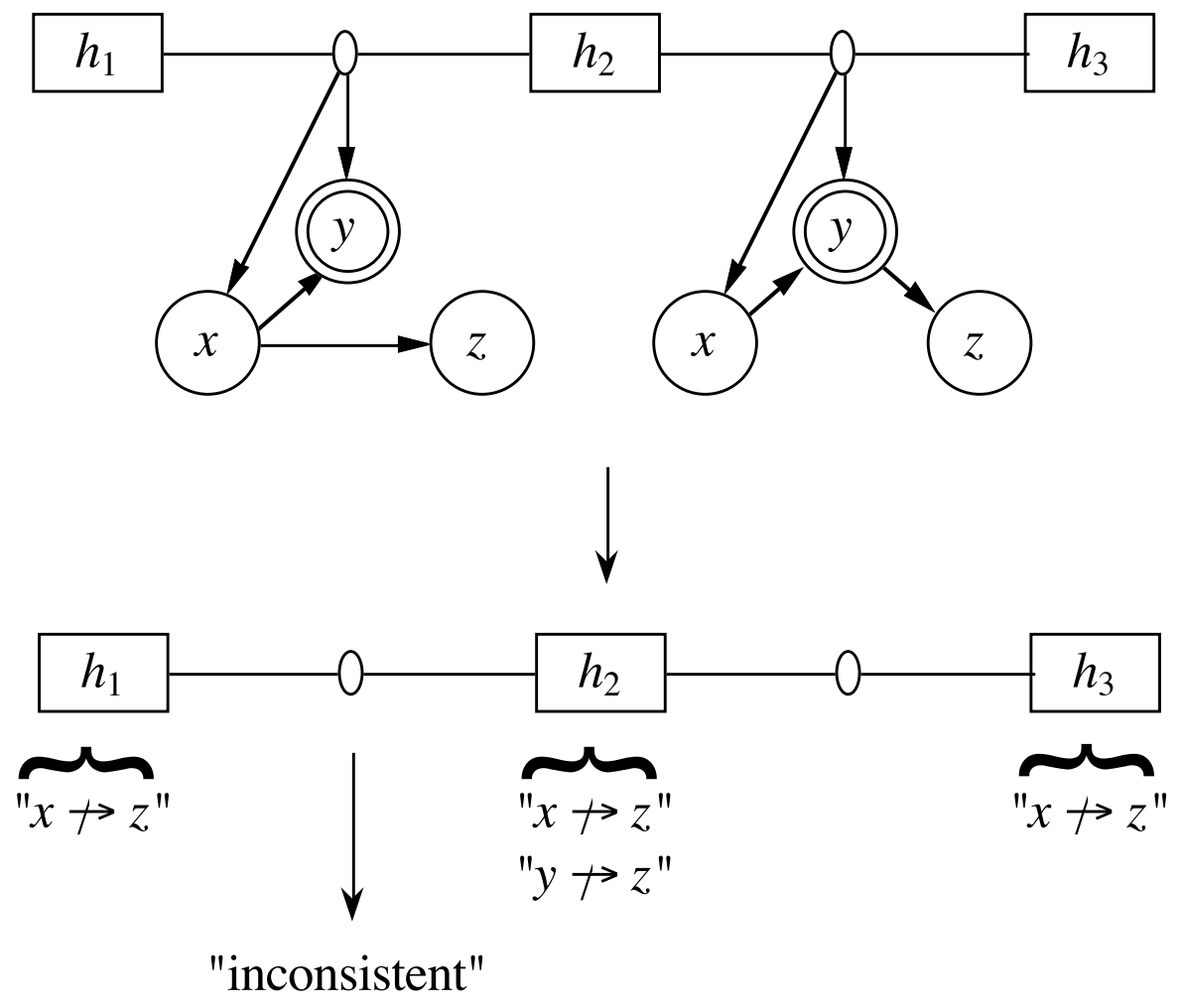

**Figure 3.13:** Algorithm 3.2 applied to a partially deterministic network. If $x$ and $y$ are logically equivalent in this comprehensive similarity network, an arc from $x$ to $z$ is equivalent to an arc from $y$ to $z$. Thus, the network is consistent. However, Algorithm 3.2 applied to this network returns "inconsistent."

In the current definition of ordinary similarity networks, we exclude a node from an o-local knowledge map, if and only if that node is disconnected from the corresponding c-local knowledge map. Alternatively, we can exclude a node from an o-local knowledge map, if and only if that node does not interact with the distinguished node.

**Definition 3.17** *Given a c-local map $h_i$–$h_j$ with nondistinguished nodes $Y$, the o-local map $h_i$–$h_j$ is the node-induced subgraph of the c-local map containing the nodes $y \in Y$ such that $y$ does not interact with $h$ in the universe $Y \cup \{h\}$, given the background knowledge corresponding to the o-local map.*

Recall that $x$ is not relevant to the set $\{h_i, h_j\}$, if and only if $x$ and $h$ are conditionally independent, given the background knowledge corresponding to the o-local map $h_i$–$h_j$ (Definition 2.1). Therefore, in Definition 3.17, we exclude a node from an o-local knowledge map $h_i$–$h_j$ if and only if that node is not relevant to the disease pair $\{h_i, h_j\}$, given any instance of any subset of the remaining nondistinguished nodes.

Using Definition 3.17, we can compose the o-local knowledge map $h_i$–$h_j$ in two phases. In phase 1, we identify those features that are relevant to the hypothesis pair $\{h_i, h_j\}$ in some context. In phase 2, we assess the independence and dependence relationships among those variables. Thus, we can separate the task of composing an o-local knowledge map into that of identifying relevant features and that of assessing dependencies among those features. The lymph-node expert found this separation to be extremely useful, and hence we employed Definition 3.17 in the composition of the Pathfinder similarity network (see Chapter 4).

The Definition 3.17 is not equivalent to the original definition of ordinary similarity networks. It is not difficult to prove that, if two variables are disconnected in a knowledge map, then they do not interact. The converse, however, is not always true. That is, two noninteracting variables may be connected in a knowledge map, even if the knowledge map is minimal. For example, consider the two unrelated events "John has a cold and does not have the flu" and "We see a picture of John driving a red car." We can compose a knowledge map in which nodes representing these two events must be connected by the introduction of an intermediate variable that contains elements of both events. Such a knowledge map is shown in Figure 3.14. In the map, DISEASE represents the event that John either has a cold or has the flu, and PICTURE represents the event that we see or do not see a picture of John driving a red car. The node CAR/FEVER represents the three mutually exclusive and exhaustive possibilities that (1) John owns a white car, (2) John owns a red car and has a fever, and (3) John owns a red car and does not have a fever. There is an arc from DISEASE to CAR/FEVER in the knowledge map because the probability that John has a fever depends on John's disease. There is an arc from CAR/FEVER to PICTURE because the probability that we see the picture depends on the color of John's car. Thus, PICTURE and DISEASE are connected in the map. However, seeing a picture of John driving a red car tells us nothing about John's disease, whether or not we know CAR/FEVER, and so PICTURE and DISEASE do not interact.

The difference between the two definitions propagates through the soundness results proved in this chapter. In particular, it is possible to find examples where the construction of an o-global map from a ordinary similarity network composed using Definition 3.17 is not sound for strictly positive distributions. Fortunately, however, the two definitions of ordinary similarity networks are equivalent for many networks. Dan Geiger and I identified a broad class of knowledge maps for which variables $a$ and $b$ must be disconnected in a minimal knowledge map whenever $a$ and $b$ do not interact. In particular, we proved the following theorem.

**Theorem 3.13** *In a minimal knowledge map containing variables $U$, any two variables $a$ and $b$ are connected in the knowledge map if and only if they interact in the universe $U$, provided the underlying joint probability distribution of the map satisfies a property called* **propositional transitivity**.

The property of propositional transitivity is cumbersome to state (it involves 9 sets of variables), and we shall not examine the property here. For a detailed discussion of this property and the proof of Theorem 3.13, see Geiger and Heckerman (1990). Note, however, that the property is satisfied by at least two types of knowledge maps: those maps in which all variables are binary and the underlying joint distribution is strictly positive, and those maps in which all distributions are normally distributed (Geiger and Heckerman, 1990). Therefore, the two definitions of ordinary similarity networks are equivalent for networks containing knowledge maps of these types.

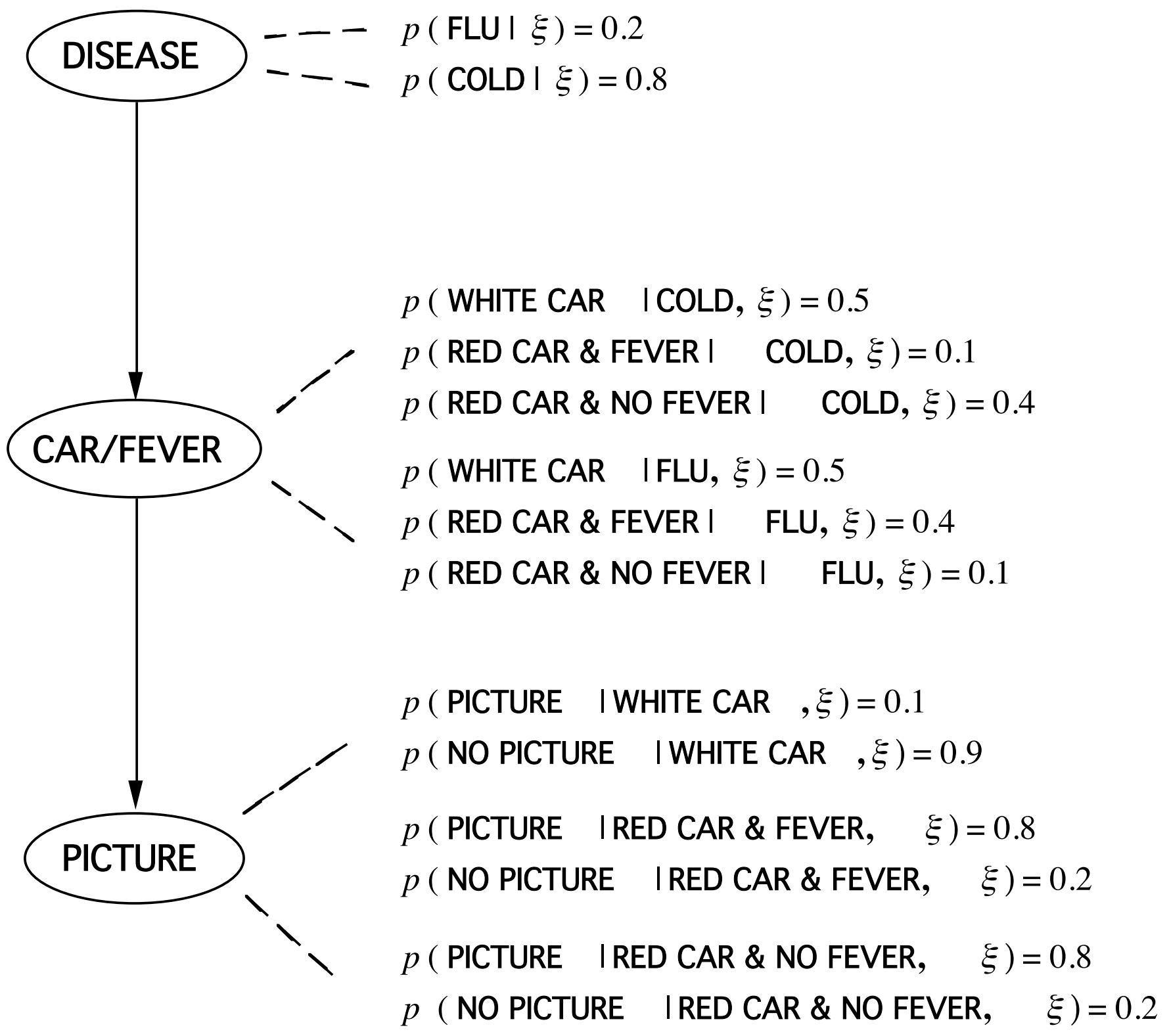

**Figure 3.14:** Connected variables that do not interact.
In the knowledge map, DISEASE represents the event that John either has a cold or has the flu, and PICTURE represents the event that we saw or did not see a picture of John driving a red car. The node CAR/FEVER represents the three mutually exclusive and exhaustive possibilities that (1) John owns a white car, (2) John owns a red car and has a fever, and (3) John owns a red car and does not have a fever. The probability distributions associated with each variable is shown beside the variable. Although PICTURE and DISEASE are connected in the map, PICTURE is not relevant to DISEASE regardless of whether CAR/FEVER is known.

This preliminary work suggests that we can identify other classes of knowledge maps for which the two definitions of ordinary similarity networks are equivalent. Also, it appears that we can develop an algorithm that tests whether or not a joint distribution satisfies propositional transitivity. Given such theoretical results, we shall be able to use Definition 3.17 to construct ordinary similarity networks in most situations, and have a guarantee that the construction of the o-global map is sound.

## 3.9 Proof of Exhaustiveness

In this section, we prove that the construction of an o-global map from an ordinary similarity network is exhaustive. In particular, we show that every variable $x$ that is relevant to the discrimination of $h$ appears in the global knowledge map via some o-local map in the similarity network. This proof constitutes the third major result of this chapter. It demonstrates that an expert does not have to look for additional distinctions for diagnosis in the global context, provided he has described all distinctions that are relevant to the local diagnostic subproblems for his domain.

First, let us formalize the concept of exhaustiveness.

**Definition 3.18** *Let $Y$ denote the set of nondistinguished variables associated with the ordinary similarity network $\mathcal{O}$. Let $x$ be any variable that is not equal to $h$ and that interacts with $h$ in the universe $Y \cup \{h, x\}$. The construction of an o-global map $\mathcal{G}_o$ from $\mathcal{O}$ is* **exhaustive** *if and only if all such nodes $x$ appear in some o-local map of $\mathcal{O}$ (by either of the two definitions of o-local map), and hence appear in $\mathcal{G}_o$.*

In this definition, we assume that the builder of the similarity network is aware of the variable $x$ during construction. More formally, we assume that, if the builder were to construct a comprehensive similarity network, then he would include $x$ in the construction.

We can now prove the desired result.

**Theorem 3.14** *The construction of an o-global map from an ordinary similarity network is exhaustive.*

**Proof:** We prove the contrapositive of the theorem. Let $\mathcal{O}$ be any ordinary similarity network, $\mathcal{G}_o$ be the o-global map constructed from $\mathcal{O}$, and $Y$ be the set of nondistinguished variables in $\mathcal{O}$. Let us suppose that $x$ appears in no o-local map of $\mathcal{O}$. By either definition of o-local maps, we obtain

$$p\,(h|x, Z, \{h_i, h_j\}, \xi) = p\,(h|Z, \{h_i, h_j\}, \xi) \tag{3.9.18}$$

for all $Z \subseteq Y$, and for all $h_i$ and $h_j$ such that $h_i$ and $h_j$ are connected in the similarity graph of $\mathcal{O}$. Following the proof of Theorem 2.1 (see Appendix B.1), Equation 3.9.18 becomes

$$p\,(x|h_i, Z, \xi) = p\,(x|h_j, Z, \xi) \tag{3.9.19}$$

where $Z$, $h_i$, and $h_j$ are defined as they were for Equation 3.9.18. Because the similarity graph is connected, we know that Equation 3.9.19 holds for all combinations of $h_i$ and $h_j$. Thus, $x$ and $h$ do not interact in the universe of $Y \cup \{h, x\}$, by Definition 3.16. □

## 3.10 Use of Similarity Networks for Assessment

As we have discussed previously, a similarity network represents two asymmetric forms of conditional independence that cannot be represented conveniently in a global knowledge map: subset independence and hypothesis-specific independence. In Chapter 2, we saw how subset independence can be used to facilitate the assessment of a knowledge map. In this section, we examine how subset independence and hypothesis-specific independence can be used together to simplify assessment. I should note that, in my experience with similarity networks, the benefits of exploiting hypothesis-specific independence have been minimal. I have found only a handful of cases in which the dependencies among nondistinguished nodes differ across hypotheses. This observation is not surprising, as features for diagnosis in medical domains tend to be defined independently of disease. Thus, features tend to depend on one another in a manner that is not a function of disease. There may be domains, however, in which asymmetrical independence among nondistinguished nodes is important; therefore, a brief examination of this form of conditional independence is provided here.

First, we consider how assertions of subset independence in similarity networks can be used in concert with assertions of hypothesis-specific independence to simplify assessment. We then examine how assertions of subset independence in partitions can be combined with hypothesis-specific independence for assessment.

The first approach is straightforward. Given an ordinary similarity network that has been certified "consistent" by Algorithm 3.2, we assess the hypothesis-specific similarity network generated by the algorithm. These assessments, in conjunction with the marginal distribution for $h$, are sufficient to construct the joint distribution over the variables in the network. We discussed previously that the hypothesis-specific similarity network is the only form of similarity network in which the assertions of subset independence and hypothesis-specific independence are disjoint. In the following procedure for assessment, we take full advantage of this observation.

For all nondistinguished $y$ in the network
  For all $h_i$
    If $y$ is not assessed for $h_i$ then
    Assess $p\left(y|C^i(y), h_i, \xi\right)$
    For all $h_j$
      If the constraint $p\left(y|C^{k/l}(y), h_k, \xi\right) = p\left(y|C^{k/l}(y), h_l, \xi\right)$ is in every
      $\mathcal{R}^{kl}$ along the path from $h_i$ to $h_j$
        Copy $p\left(y|C^i(y), h_i, \xi\right)$ to $p\left(y|C^j(y), h_j, \xi\right)$

If the constraint $p\left(y|C^{k/l}(y), h_k, \xi\right) = p\left(y|C^{k/l}(y), h_l, \xi\right)$ is in every relevance set $\mathcal{R}^{kl}$ along the path from $h_i$ to $h_j$, then, by definition of hypothesis-specific similarity networks, $C^k(y) = C^l(y)$, for every pair of hypotheses $h_l$ and $h_k$ that border the edges along the path. Consequently, $C^i(y) = C^j(y)$, and the last step of the procedure is legitimate. Since the procedure iterates over every variable in the network and every hypothesis $h_i \in h$, it follows that all probabilities required to construct the joint distribution are assessed.

For example, let us apply the procedure to the ordinary similarity network in Figure 3.11. Suppose $x$ is a binary variable with instances $x_+$ and $x_-$. Similarly, suppose $y$ is a binary variable with instances $y_+$ and $y_-$. Figure 3.15 shows the results of a series of hypothetical assessments using the approach. First, we assess the distribution $p\left(x|h_1, \xi\right)$. Because there is an arc from $h$ to $x$ in $h_1$–$h_2$, this distribution is not copied. Similarly, we assess both $p\left(x|h_2, \xi\right)$ and $p\left(x|h_3, \xi\right)$. Next, we assess the distribution $p\left(y|h_1, \xi\right)$. Now, however, there is no arc from $h$ to $y$ in the o-local map $h_1$–$h_2$ ($y$ is not in the map), and hence we copy $p\left(y|h_1, \xi\right)$ to $p\left(y|h_2, \xi\right)$. Finally, we assess $p\left(y|x, h_3, \xi\right)$ for the two instances of $x$.

Let us compare this procedure with a direct assessment of the global knowledge map constructed from the ordinary similarity network of Figure 3.11. If we construct the joint distribution by assessing the global knowledge map directly, all nine distributions in Figure 3.15 need to be assessed. In contrast, the procedure described in the previous paragraph takes advantage of the fact that $x$ and $y$ are independent in $\widehat{h_1}$ and $\widehat{h_2}$. In effect, we need to assess only the distributions $p\left(y|x_-, h_1, \xi\right)$ and $p\left(y|x_-, h_2, \xi\right)$, and then to copy them to $p(y|x_+, h_1, \xi)$ and $p(y|x_+, h_2, \xi)$, respectively. The checks marks in Figure 3.15 reflect the two assessments that we avoid by exploiting the hypothesis-specific independence in the network. In addition, the procedure described in the previous paragraph takes advantage of the fact that $y$ is not relevant to $h_1$ and $h_2$. The asterisks in Figure 3.15 mark the two assessments that are avoided using this information. Since there is one assessment that is avoided on both accounts, only six of the original nine assessments are required using this approach.

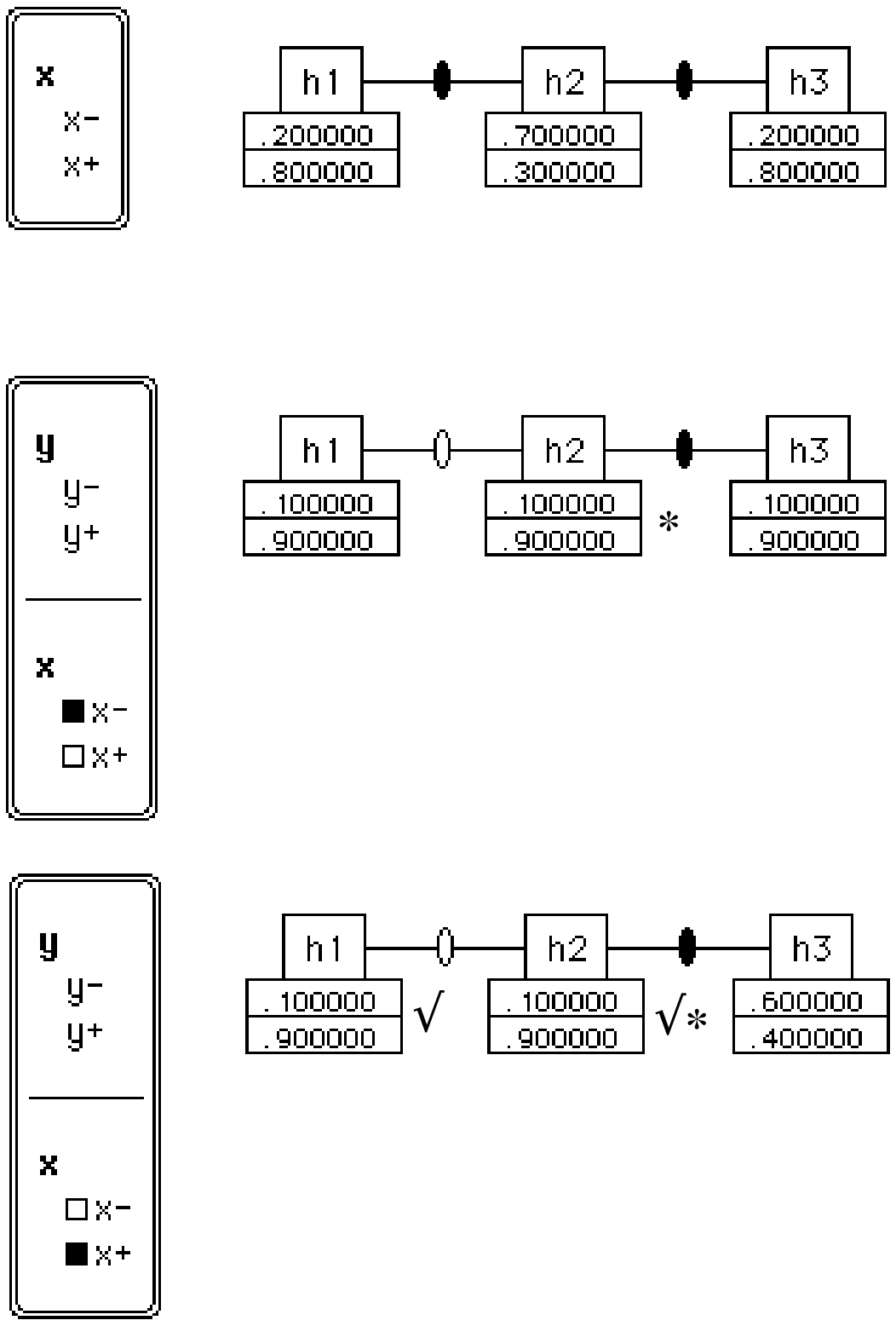

**Figure 3.15:** Using a similarity network for assessment.
The figure shows how the ordinary similarity network in Figure 3.11 is used to assess the joint distribution for the variables in the network. The check marks and asterisks indicate the assessments that are avoided through the use of hypothesis-specific independence and subset independence, respectively.

Now let us extend this process to include partitions. Figure 3.16 illustrates how partitions can be used to simplify further the construction of the joint distribution in the previous example. Because $p(x|h_1, \xi) = p(x|h_3, \xi)$, we can place hypotheses $h_1$ and $h_3$ in a common set within the partition for $x$. Consequently, we need to assess only two distributions for $x$. Because the distribution $p(y|x_-, h_i, \xi)$ is the same for all $h_i$, we need to provide only one distribution in this case. Finally, the two distributions $p(y|x_+, \{h_1, h_2\}, \xi)$ and $p(y|x_+, h_3, \xi)$ are required. Because $x$ and $y$ are independent in $h_1$ and $h_2$, the assessment of $p(y|x_+, \{h_1, h_2\}, \xi)$ is avoided (see the check mark in Figure 3.16).

In general, the partitions composed by an expert may be inconsistent with the assertions of hypothesis-specific independence embodied in a similarity network. For example, if the the partition for $y$, given $x_+$, contained the set "$h_2$ or $h_3$," then an inconsistency would exist, because $x$ and $y$ must be independent given $h_2$, and dependent given $h_3$. Thus, if domains are encountered in which hypothesis-specific independence can significantly simplify assessments, the development of procedures for testing the mutual consistency of partitions and similarity networks will be useful.

## 3.11 Summary

In Section 3.3, we defined formally the hypothesis-specific, comprehensive, and ordinary similarity network, and the comprehensive and ordinary global knowledge map. We saw that all representations except the hypothesis-specific similarity network could be constructed from another representation among this collection. In Section 3.4, we saw that each construction maps a more detailed representation into a less detailed one and, consequently, that each construction is sound for strictly positive distributions. In Section 3.5, we examined hypothesis-specific networks and minimal comprehensive networks and showed that they can be inconsistent. We showed that a minimal comprehensive similarity network is consistent for some strictly positive joint distribution if and only if it can be constructed from a consistent hypothesis-specific similarity network. We employed this result to derive necessary and sufficient conditions for a comprehensive network to be consistent and, based on these conditions, derived an algorithm for testing the consistency of such networks. In Section 3.6, we used these necessary and sufficient conditions to show that the c-global and o-global maps constructed from a consistent comprehensive network are identical. From this result, we proved that the construction of an o-global map from an ordinary similarity network is sound. In Section 3.7, we modified the algorithm for testing consistency, developed in Section 3.5, to accommodate ordinary similarity networks. We saw that the algorithm was tractable and could be modified further to assist users in the resolution of inconsistencies. In Section 3.8, we examined an alternative, more useful definition of ordinary local knowledge maps. In Section 3.9, we proved that the construction of an o-global map from an ordinary similarity network is exhaustive. Finally, in Section 3.10, we considered the two asymmetric forms of conditional independence encoded by a similarity network—subset independence and hypothesis-specific independence—and examined an approach for assessment that takes advantage of both forms of independence.

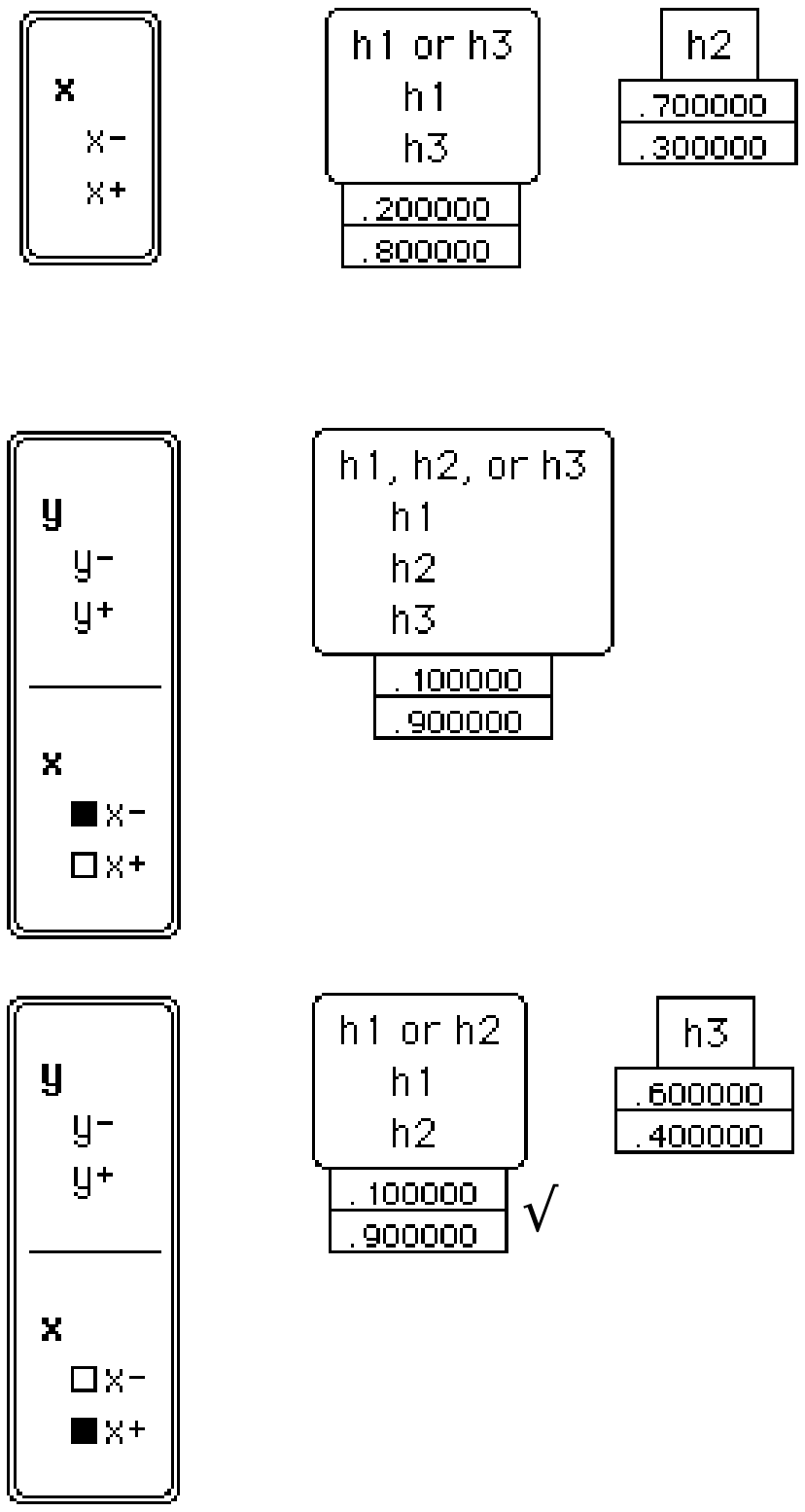

**Figure 3.16:** Using partitions for assessment.
The figure shows how partitions are used to simplify the assessments shown in Figure 3.15. The check mark indicates the assessment that we avoid by taking advantage of hypothesis-specific independence.

# 4 Pathfinder: A Case Study

In this chapter, we examine similarity networks and partitions from a practical perspective. In particular, we examine highlights of the composition and assessment of the Pathfinder knowledge map using these representations. We begin with a brief history of the Pathfinder project. Next, we look at representative portions of the similarity network and partitions that combine to form the joint probability distribution for Pathfinder. We then discuss, in quantitative terms, the amount of effort that went into the creation of the knowledge base. After this discussion, we examine important insights about the new representations that the expert and I gained by building Pathfinder. Finally, we look at the Pathfinder inference algorithm.

## 4.1 History of Pathfinder

The latest generation of Pathfinder, composed with the similarity-network and partition representations, is the fourth implementation of the expert system. The Pathfinder project began in 1983 as a joint project among researchers at Stanford University and the University of Southern California, including Larry Fagan, Eric Horvitz, Bharat Nathwani, and me (Heckerman et al., 1985). The earliest version of the Pathfinder, called Pathfinder I, was a rule-based system written in the Meta-Level Representation System (MRS) (Genesereth, 1983). This version of Pathfinder had two major problems: (1) it did not incorporate any mechanism for uncertain reasoning, and (2) it recommended features for observation based on a fixed depth-first traversal through the rule-base graph, rather than on an analysis of the current differential diagnosis.

The Pathfinder team modeled the second version of Pathfinder, called Pathfinder II, after INTERNIST-1, a diagnostic program for internal medicine that was the precursor of Quick Medical Reference (QMR) (Miller et al., 1982; Miller et al., 1986). This version of Pathfinder, implemented in LISP on a DEC-2060, addressed both problems encountered with the first version. Specifically, we incorporated into Pathfinder the hypothetico-deductive approach, discussed in Chapter 1, so that the system could base its recommendations for additional observations on the current differential diagnosis. Furthermore, the expert and I experimented with several methods for uncertain reasoning, including the Mycin certainty-factor (CF) model (Shortliffe and Buchanan, 1975), the Dempster–Shafer theory of belief (Shafer, 1976), and the simple Bayes model, which we examined in Chapter 1.

In an experiment similar to the one described in the next chapter, I showed that the diagnostic accuracy of the simple Bayes model was superior to that of the other two models (Heckerman, 1988). The experiments convinced us to use probability theory as a representation for uncertainty. These same experiments, however, revealed two flaws in knowledge-base composition that were more serious than the failure to represent the dependencies among features. First, the expert was too cavalier when assigning 0 to the probability of many events. In preliminary evaluations of Pathfinder, we found that over 10 percent of the cases were diagnosed incorrectly, because the correct disease was ruled out by a feature that was unlikely (but not impossible) to be seen in that disease. Second, the expert was not comfortable with many of the probability assessments that he provided for the second version of the program. Specifically, he assessed the probability matrix required by the simple Bayes model, $p\,(\text{feature}|\text{disease}, \xi)$ for all diseases and features, by fixing a disease and assessing probabilities across all features. In an analysis that followed the completion of the knowledge base, we found that he strongly preferred making assessments by fixing a feature and assessing probabilities of that feature across all diseases. Thus, the expert reassessed the entire probability matrix for the simple Bayes model with this new ordering, paying close attention to unlikely events. The result constituted the next version of Pathfinder, called Pathfinder III. The program was implemented first in LISP on an HP-9836 workstation, and later using MacApp on the Macintosh II.

Finally, we tackled the problem of representing dependencies among features. As mentioned in Chapter 1, the expert and I could not compose the knowledge map for the lymph-node domain using available techniques. With the aid of similarity networks and partitions, however, we were able to complete the construction of such a knowledge map. This knowledge base and the associated inference algorithms, discussed in Section 4.5, constituted Pathfinder IV. This version of the expert system was implemented in MacApp on the Macintosh II.

From this history, we see that a comparison of Pathfinder III and IV reveals the advantages and disadvantages of the new representations developed in this book. In particular, the expert was comfortable with probability assessment throughout the construction of both systems; the single difference between the two knowledge bases is that we used the similarity-network and partition representations to construct only Pathfinder IV. In the remainder of this chapter, and in Chapter 5, we compare Pathfinder III and Pathfinder IV.

## 4.2 Highlights of Knowledge-Base Composition

We created the similarity network and partitions for Pathfinder IV using the SimNet program. An important product of our efforts—the global knowledge map for Pathfinder IV—is shown in Figure 4.1. In all phases of the construction, the expert and I worked together. In the early phases of development, I demonstrated the various capabilities of the program to the expert, and asked pointed questions to evoke portions of his knowledge about the lymph-node domain. As our work progressed, however, the expert became familiar with the program and with the strategies I was using to elicit knowledge, and my role as knowledge engineer diminished.

### 4.2.1 Similarity Network

Figure 4.2 shows the similarity graph for the lymph-node domain. The expert had no trouble creating the graph; in fact, he completed its composition in approximately 3 hours. We began building the graph by identifying the most similar sets of diseases from the list of diseases found in Pathfinder III. Then, we connected these diseases, and moved them close to one another on the computer screen. We continued, connecting diseases that were less and less similar, until there was a path from every disease in the graph to every other disease. In several cases, we composed more than one path between diseases (see, for example, the nodes AIDS EARLY, RHEUMATOID ARTHRITIS, and GLH PLASMA CELL TYPE in Figure 4.2). At the time of composition, the expert was thinking only about how similar one disease was to another—that is, how likely it was that he or another pathologist would confuse one disease with another. He was not thinking about whether he could build a local knowledge map for each pair of diseases connected in the graph.

We composed the local knowledge maps using the two-phase approach described in Section 3.8. Specifically, in phase 1, for each pair of diseases connected in the similarity graph, the expert identified those features that were relevant to that disease pair in some context (see Definitions 2.1 and 3.17). We consulted the list of features from Pathfinder III, during this process. In phase 2, he represented the dependencies among the features. As I mentioned in Chapter 3, the expert found the added decomposition afforded by this approach to be extremely useful.

Figure 4.3 shows the local knowledge map for L&H DIFFUSE HD (lymphocytic and histiocytic diffuse Hodgkin's disease) and MIXED-CELLULARITY HD (mixed-cellularity Hodgkin's disease). The knowledge map expresses the expert's assertion that only MUMMY (mummy cells), L&H SR (lymphocytic and histiocytic variants of Sternberg–Reed cells), MONONUCLEAR SR (mononuclear variants of Sternberg–Reed cells and), and CLASSIC SR (classic Sternberg–Reed cells) are relevant to the pair of diseases represented by the local knowledge map. In addition, the knowledge map—asserted to be minimal by the expert—represents independencies and dependencies among these features. For example, the arcs among the features reflect the expert's assertion that CLASSIC SR depends on MONONUCLEAR SR, and that MONONUCLEAR SR depends on MUMMY. In contrast, the lack of an arc from MUMMY to CLASSIC SR represents his assertion that CLASSIC

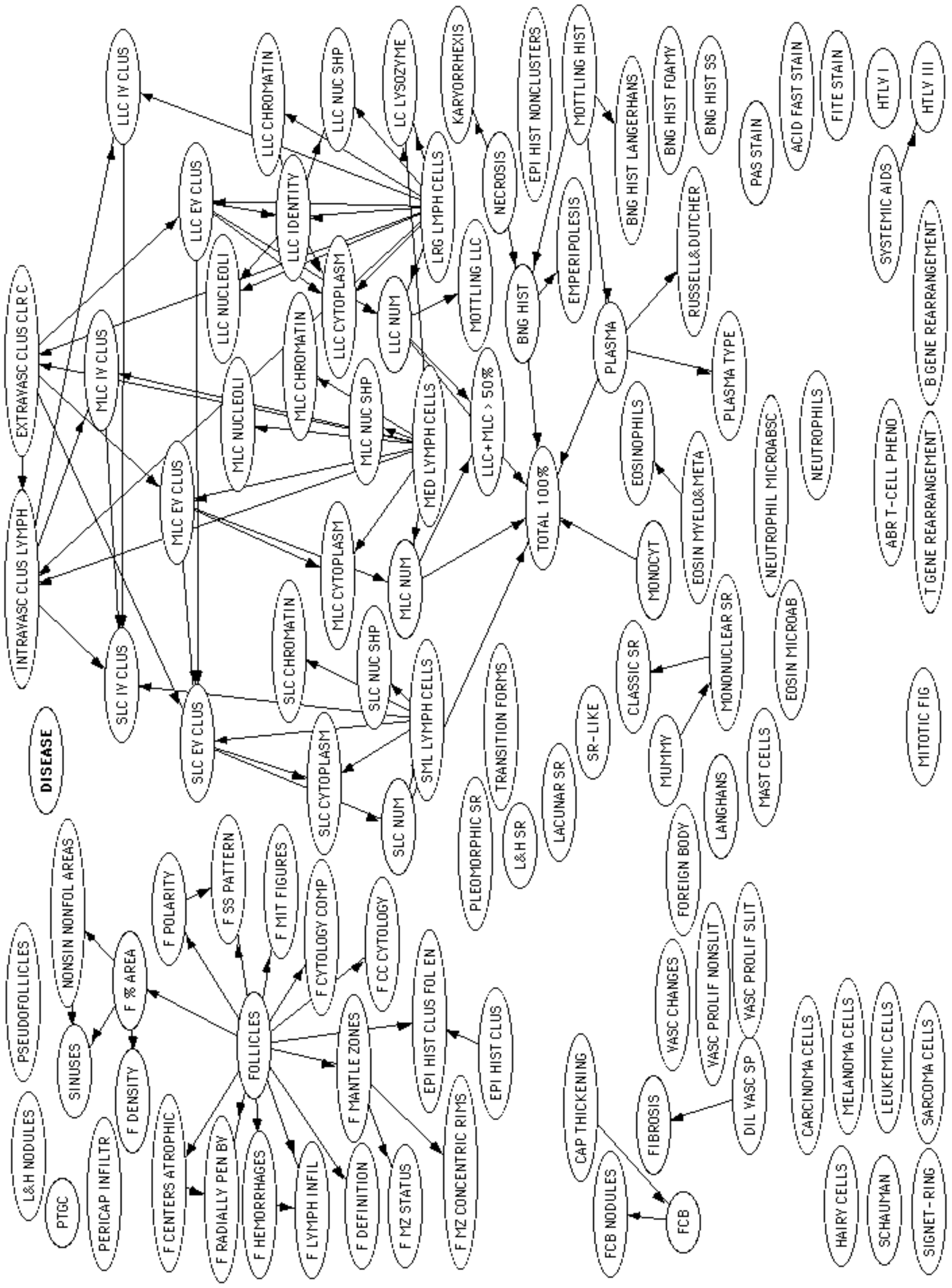

**Figure 4.1:** The global knowledge map for Pathfinder.
This figure is identical to Figure 1.10 on page 15, except that all conditioning arcs from DISEASE to other nodes are not shown so that the conditional dependencies among features are highlighted. Appendix C contains a key to the feature abbreviations.

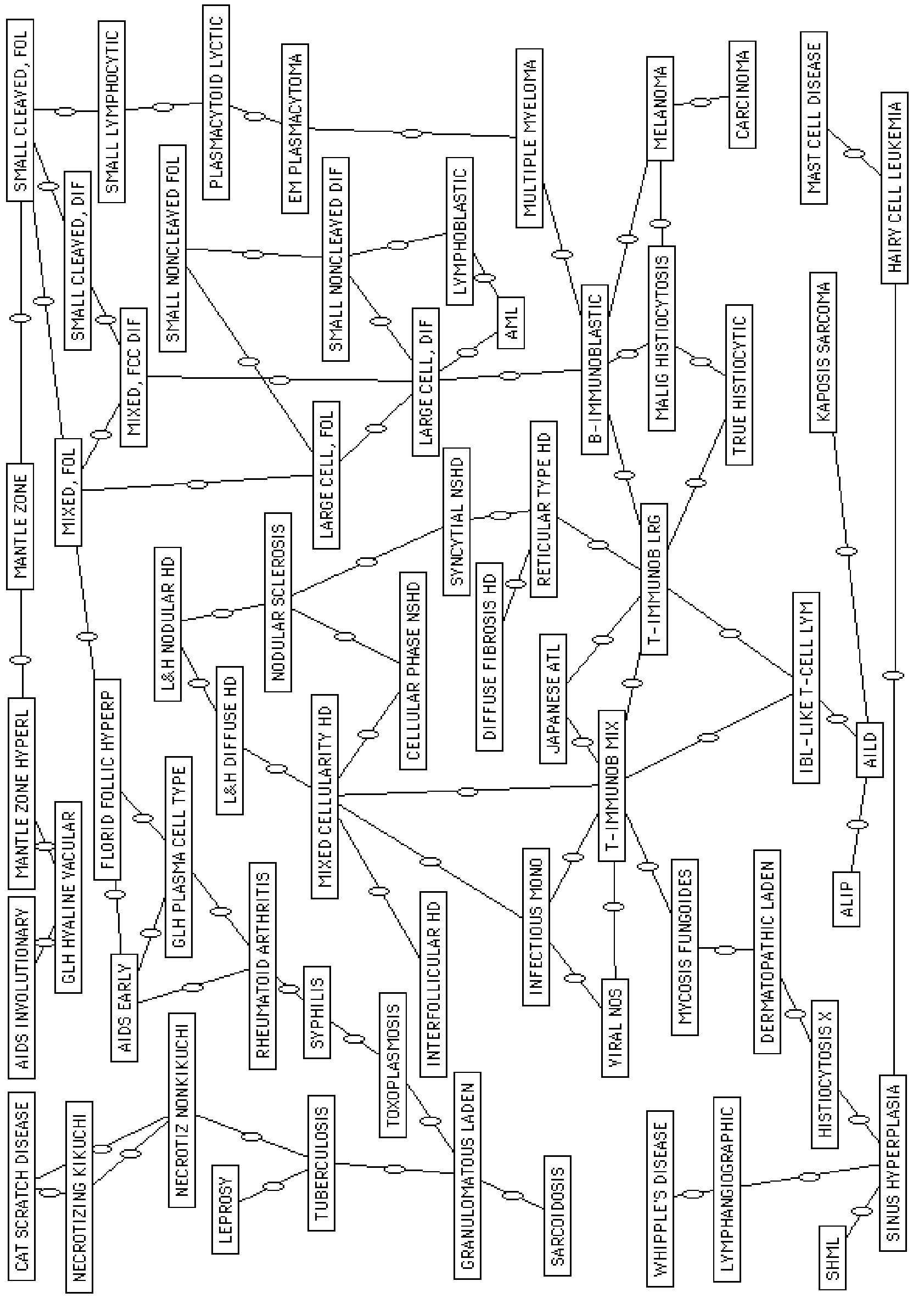

**Figure 4.2:** The similarity graph for Pathfinder.
Nodes in the graph represent the mutually exclusive diseases that can manifest in a lymph node. Edges in the graph connect diseases that are similar. The graph is multiply connected. Appendix C contains a key to the disease abbreviations.

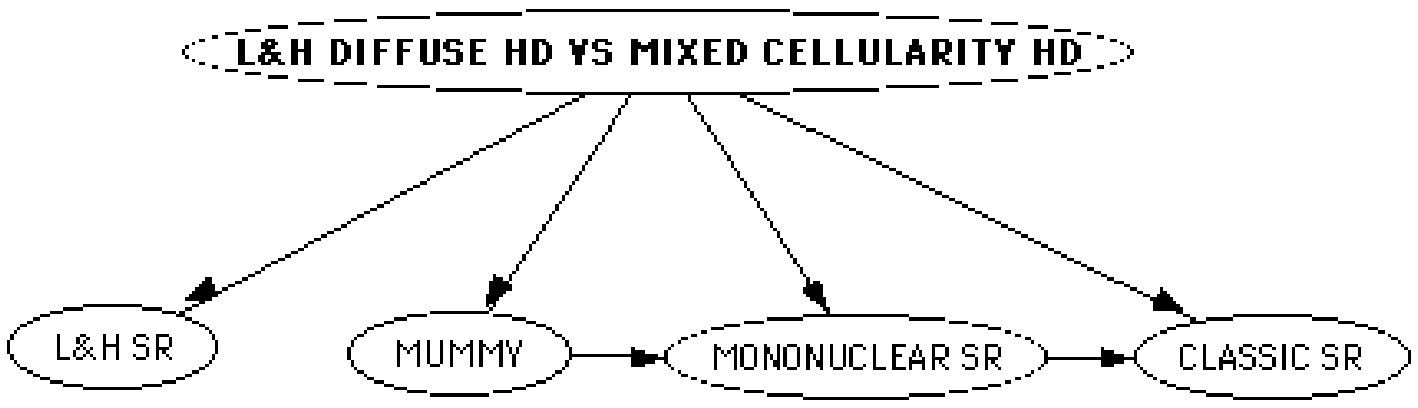

**Figure 4.3:** A small local knowledge map.
This knowledge map is the local knowledge map for the diseases L&H DIFFUSE HD (lymphocytic and histiocytic diffuse Hodgkin's disease) and MIXED-CELLULARITY HD (mixed-cellularity Hodgkin's disease). The knowledge map represents the assertions that only the features MUMMY (mummy cells), L&H SR (lymphocytic and histiocytic variants of Sternberg–Reed cells), MONONUCLEAR SR (mononuclear variants of Sternberg–Reed cells and), and CLASSIC SR (classic Sternberg–Reed cells) are relevant to the disease pair. The knowledge map also represents assertions of conditional independence and dependence among these features.

SR is independent of MUMMY if MONONUCLEAR SR is known. The lack of other arcs represents the expert's assertion that the feature L&H SR is conditionally independent of the other features.

Most of the local knowledge maps in the similarity network are small, as is the one shown in Figure 4.3. Several of the local knowledge maps, however, are extremely large. Figure 4.4 shows the largest local knowledge map in the similarity network. The map represents the problem of distinguishing the diseases T-IMMUNOB LRG (T-immunoblastic lymphoma, large-cell type) and IBL-LIKE T-CELL LYM (immunoblastic lymphadenopathy-like T-cell lymphoma). For clarity, several features that are not directly relevant to the disease pair and all arcs from the disease node to the feature nodes, are omitted from the figure.

The knowledge map is complex for the following reason. In IBL-like T-cell lymphoma, we see clusters of lymphoid cells with clear cytoplasm. These clusters can occur inside blood vessels (intravascular) or outside blood vessels (extravascular). In addition, the lymphoid cells in these clusters may be small (less than 10 microns in diameter), of medium size (10 to 20 microns in diameter), large (greater than 20 microns in diameter), or they may be a combination of sizes. The presence of these clusters is an important clue for discriminating IBL-like T-cell lymphoma from T-immunoblastic lymphoma. In fact, *by definition*, such clusters must be seen in IBL-like T-cell lymphoma, and they may or may not be seen in T-immunoblastic lymphoma. Pathologists say that the presence of such clusters is a *criterion* for the disease IBL-like T-cell lymphoma.

The size of the cells in the clusters is not directly relevant to the discrimination of the disease pair. In contrast, the number of small, medium-sized, and large lymphoid cells in the lymph-node section as a whole is directly relevant to the discrimination of the two

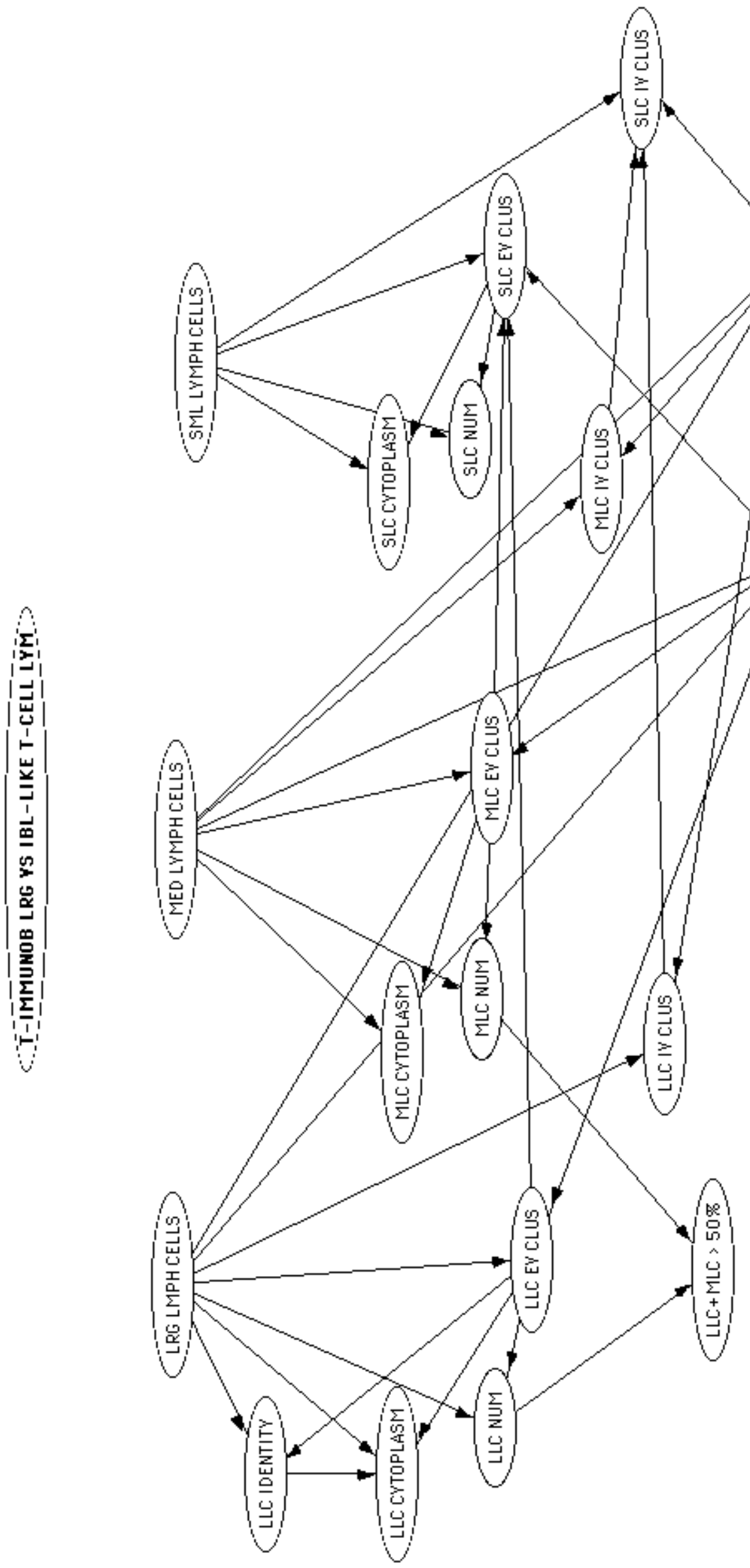

**Figure 4.4:** A large local knowledge map.

This knowledge map represents the problem of discriminating T-IMMUNOB LRG (T-immunoblastic lymphoma, large-cell type) and IBL-LIKE T-CELL LYM (IBL-like T-cell lymphoma). Arcs from the distinguished node to other nodes are not shown.

diseases.[1] In fact, if the number of medium and large cells (both in and not in clusters) is less than 50 percent of the total cell population, then the disease T-immunoblastic lymphoma, large-cell type is ruled out. That is, the predominance of medium and large lymphoid cells is a criterion for this disease. Such predominance, however, is not a criterion for IBL-like T-cell lymphoma. In addition, the amounts of cytoplasm in the majority of small, medium-sized, and large cells are criteria for other diseases in the lymph-node domain. Thus, we must include nodes in the Pathfinder knowledge map that represent the size and cytoplasm of the total lymphoid-cell population, the presence of clusters of clear cells (which contribute to the total population of lymphoid cells with clear cytoplasm), and the dependencies among these features.

In Figure 4.4, the nodes EXTRAVASC CLUS CLEAR C and INTRAVASC CLUS CLEAR C represent the number of cells found in extravascular and intravascular clusters, regardless of the distribution of cell size within the clusters. The nodes SLC NUM, MLC NUM, and LLC NUM denote the total number of small, medium-sized, and large cells that make up the lymph-node section, regardless of whether or not the cells occur in clusters. Similarly, the nodes SLC CYTOPLASM, MLC CYTOPLASM, and LLC CYTOPLASM reflect the amount of cytoplasm in the majority of small, medium-sized, and large cells in the lymph-node section, regardless of whether or not the cells occur in clusters. The features SLC IV CLUS, MLC IV CLUS, MLC IV CLUS, SLC EV CLUS, MLC EV CLUS, and MLC IV CLUS represent the number of cells in intravascular and extravascular clusters for each cell size. The inclusion of these features break many of the dependencies among the primary features. For example, if we did not include the features SLC EV CLUS, MLC EV CLUS, and MLC IV CLUS in the knowledge map, seeing extravascular clusters of clear cells would fail to render the features SLC CYTOPLASM, MLC CYTOPLASM, and LLC CYTOPLASM independent, given disease. That is, if we see extravascular clusters and observe no small cells with clear cytoplasm, the chances that we will see medium-sized and large cells with clear cytoplasm increase. In contrast, if we see extravascular clusters and we know how many small, medium-sized, and large cells are in the clusters, then the lack of small cells with clear cytoplasm does not alter the chances that we will see medium-sized and large cells with clear cytoplasm.

There are two nodes in Figure 4.4 that we have not yet discussed. The node LLC+MLC>50% in Figure 4.4 represents the event that the combined number of the large and medium-sized cells in the lymph-node section exceeds 50 percent. As mentioned previously, this event is a criterion for T-immunoblastic lymphoma, large-cell type. A special node for this feature is included in the knowledge map, because the instances for large and medium-size lymphoid cells are discretized too coarsely to capture the threshold of 50 percent. The node LLC IDENTITY denotes the identity of the majority of large lymphoid cells. This feature is included because it helps to break the dependencies among features describing the cytoplasm, nuclear shape, nucleoli, and chromatin of the large lymphoid cells. Although these features are not directly relevant to the discrimination of IBL-like T-cell lymphoma and T-immunoblastic lymphoma, they are criteria for other diseases and must be included in the Pathfinder knowledge map. The feature LLC IDENTITY is included in the knowledge map of Figure 4.4 because the presence of clear-cell clusters increases the chances that a large lymphoid cell of a particular type will predominate.

[1] In lymph-node pathology, the number of cells of a particular cell type is expressed in terms of percent of the total cell population in a lymph-node section. In Pathfinder, for example, the feature SLC NUM (number of small lymphoid cells) has the instances 0 percent, 1 to 10 percent, 11 to 50 percent, 51 to 90 percent, and 91 to 100 percent of the total cell population.

In closing this section, let us examine two techniques that we employed to represent the dependencies among features. We captured most of the dependencies among features in the Pathfinder similarity network by representing each feature explicitly, and by drawing arcs among those features. We represented dependencies for about 10 groups of features, however, using a technique that Pearl calls *clustering* (Pearl, 1988, pages 195–197). To understand clustering, let us consider the three features that describe the nucleoli in large lymphoid cells: the number of nucleoli (0, 1, or more than 1), the size of the nucleoli (small or large), and the location of the nucleoli within the nucleus (central or peripheral). These features are mutually dependent, given disease. Furthermore, these dependencies are asymmetric. In particular, if there are no nucleoli, then the size and shape of nucleoli are features that do not apply to the current case; if there is only one nucleolus and it is small, then its location is irrelevant to diagnosis; if there is only one nucleolus and it is prominent, then its location is necessarily central; finally, if there are two or more nucleoli, their size are irrelevant to diagnosis. Figure 4.5(a) shows a knowledge map that describes these features and the dependencies among them. As illustrated in the figure, each feature is associated with three instances. Alternatively, we can represent these features by clustering them into a single variable, as shown in Figure 4.5(b). Usually, when we cluster a set of variables, the number of instances of the clustered variable is equal to the product of the number of instances over each feature that we cluster. Due to the asymmetries in the dependencies among the nucleolar features, however, the total number is far less (5 instances, instead of $3^3 = 27$ instances). In general, it is prudent to cluster features when they are mutually dependent and are easily observed in combination, and when those features exhibit asymmetrical dependencies that lead to a reduction in the number of instances for the feature that is formed by their clustering.

Also, to simplify our efforts, we approximated several of the dependencies in the lymph-node domain. One approximation that we employed is illustrated in a portion of the global knowledge map shown in Figure 4.6. The nodes in the middle row of the figure represent the six cell types that can contribute significantly to the total cell population of a lymph node. We have already mentioned the nodes SLC NUM, MLC NUM, and LLC NUM. The nodes BNG HIST, PLASMA, and MONOCYT represent the cell types of benign histiocytes (nonfoamy, non–starry sky, and nonlangherhans), plasma cells, and monocytoid cells, respectively. The feature for each cell type is associated with a set of instances that denote ranges for the percent number of cells of that type that are seen in the lymph-node section (the ranges differ from one feature to the next). These six features are dependent, and, in principle, should be represented by a fully connected knowledge map. In Pathfinder, however, we approximate these dependencies as follows. First, we assume that the six features are conditionally independent, given disease. Then, we define a deterministic node called TOTAL 100% that represents the fact that the number of cells of each type must sum to 100 percent. This node is unlike an ordinary node in a knowledge map, in that it has only one preinstantiated instance. Next, we condition the node TOTAL 100% on the six features in the middle row. Finally, we approximate the summation constraint using the probability matrix for TOTAL 100% shown in Table 4.1. In the table, an instance of the six features is assigned a probability of 1 if and only if that instance is consistent with a total of 100 percent. As we see in the following chapter, this approximation, and others that we employed in the construction of the knowledge base, do not lead to poor performance of the system.

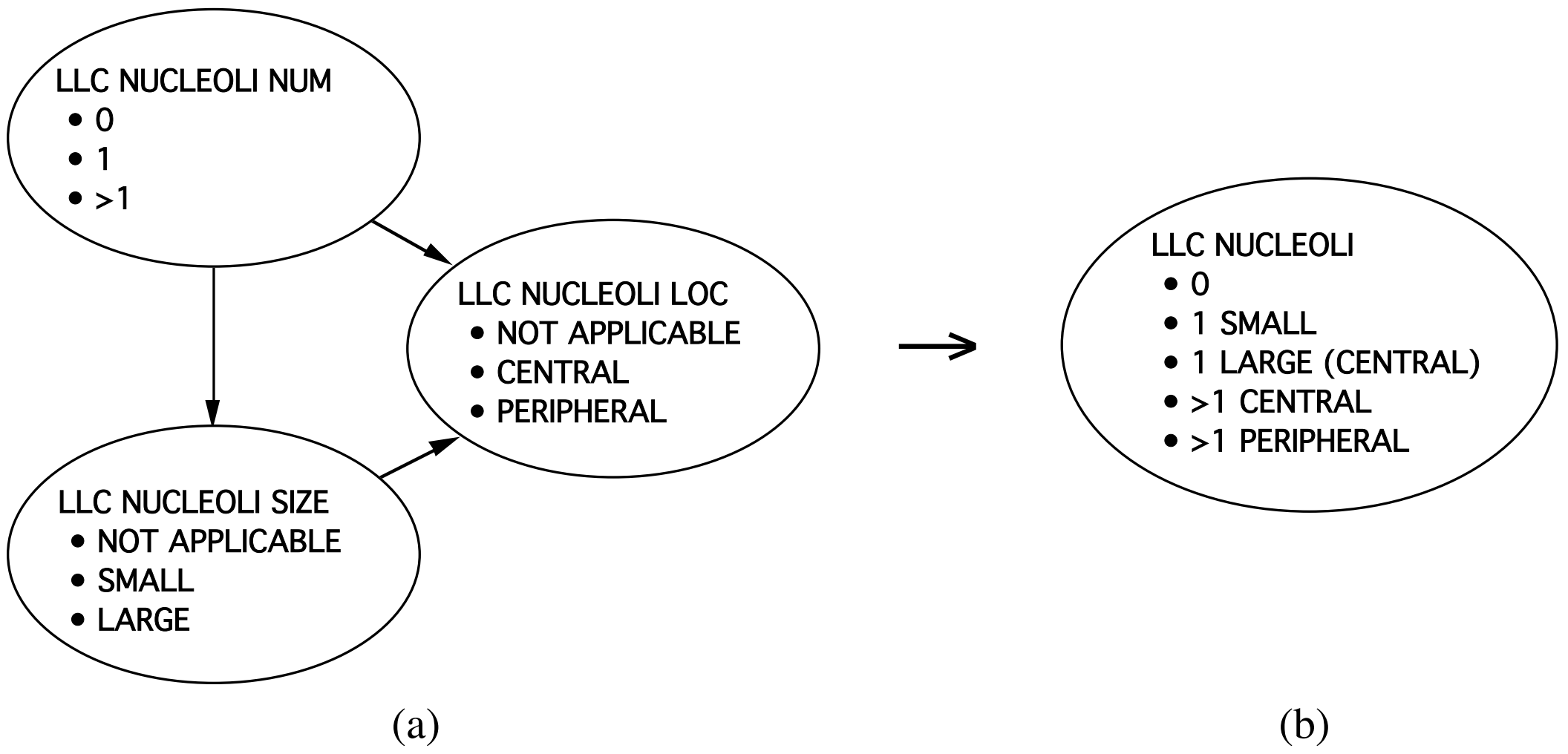

**Figure 4.5:** The clustering of nodes in Pathfinder.
(a) The knowledge map for the features LLC NUCLELOI NUM (number of nucleoli), LLC NUCLEOLI SIZE (size of the nucleoli), and LLC NUCLEOLI LOC (location of the nucleoli within the nucleus), where each feature is represented by a single node. (b) The knowledge map for these same features when they are clustered into a single variable.

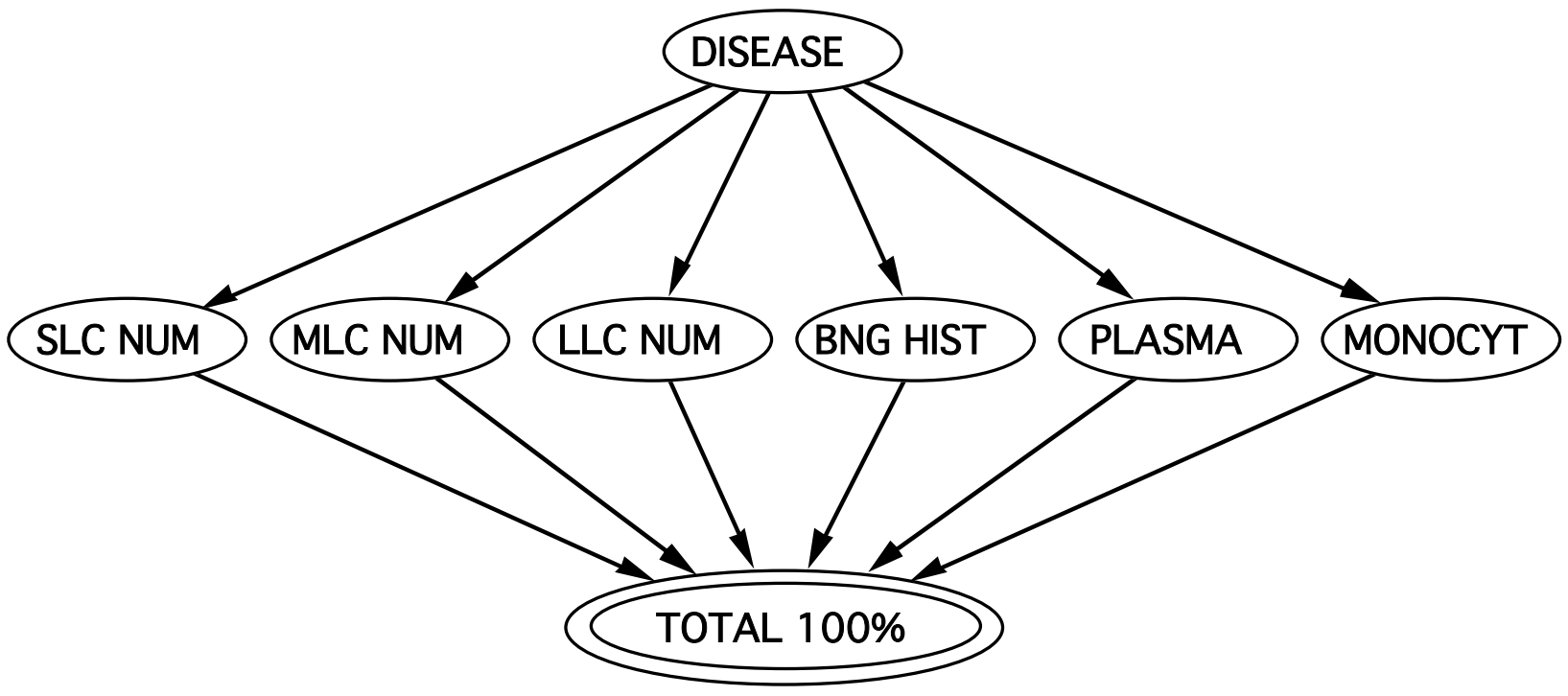

**Figure 4.6:** An approximation of dependencies.
The portion of the Pathfinder global knowledge map for the features SLC NUM (small lymphoid cell number), MLC NUM (medium-sized lymphoid cell number), LLC NUM (large lymphoid cell number), BEN HIST (benign histiocyte number), PLASMA (plasma cell number), and MONOCYT (monocytoid cell number). The presence of the deterministic node TOTAL 100% approximates the mutual dependency among the feature nodes.

**Table 4.1:** The probability distribution for the node TOTAL 100%.

| Conditioning Events (percent of total cell population) | | | | | | |
|---|---|---|---|---|---|---|
| SLC NUM | MLC NUM | LLC NUM | BNG HIST | PLASMA | MONOCYT | $p$ |
| 0 | 0 | 0 | 0–4 | 0 | 0 | 0 |
| 0 | 0 | 0 | 0–4 | 0 | 1–4 | 0 |
| 0 | 0 | 0 | 0–4 | 0 | 5–50 | 0 |
| 0 | 0 | 0 | 0–4 | 0 | 51–100 | 1 |
| | | | ⋮ | | | ⋮ |
| 11–50 | 11–50 | 11–50 | 5–50 | 21–50 | 5–50 | 1 |
| 11–50 | 11–50 | 11–50 | 5–50 | 21–50 | 51–100 | 0 |
| 11–50 | 11–50 | 11–50 | 5–50 | 51–90 | 0 | 1 |
| 11–50 | 11–50 | 11–50 | 5–50 | 51–90 | 1–4 | 1 |
| | | | ⋮ | | | ⋮ |
| 91–100 | 91–100 | 91–100 | 51–100 | 91–100 | 5–50 | 0 |
| 91–100 | 91–100 | 91–100 | 51–100 | 91–100 | 51–100 | 0 |

### 4.2.2 Partitions

We assessed probabilities for the global knowledge map for Pathfinder using the partition representation. We began the assessment phase of knowledge-base construction by composing the partition shown in Figure 4.7. This partition was modeled after a classification hierarchy (see Section 2.2.3) for lymph-node pathology used commonly by hematopathologists. We next identified features or distinctions that were created by research pathologists to rule in a specific disease or a specific group of diseases, and composed the partitions for those features. These partitions tended to contain a small number of sets. In the process of composing these structures, the expert became familiar with the partition concept and with the mechanisms in SimNet for copying and modifying partitions. We then composed the partitions for the remaining features. The expert provided the probabilities for each partition before composing the next partition.

More than 40 percent of the features are associated with partitions that are small (i.e., contain less than 10 sets). Figure 4.8 shows one such partition for the feature progressively transformed germinal centers (PTGC). This partition contains only five sets: MOST DISEASES (majority of diseases), MOST BENIGN (majority of benign diseases), OTHER BENIGN (a small set of benign diseases for which the probability of seeing progressively transformed germinal centers is greater than for most benign diseases), and the singletons FLORID FOLLIC HYPERP (follicular hyperplasia) and L&H NODULAR HD (lymphocytic and

- BENIGN
  - AIDS
    - AIDS EARLY
    - AIDS INVOLUTIONARY
  - ALIP
  - CAT SCRATCH DISEASE
  - DERMATOPATHIC LADEN
  - FLORID FOLLIC HYPERP
  - GLH
    - GLH HYALINE VACULAR
    - GLH PLASMA CELL TYPE
  - GRANULOMATOUS LADEN
  - HISTIOCYTOSIS X
  - LEPROSY
  - LYMPHANGIOGRAPHIC
  - MANTLE ZONE HYPERL
  - NECR LYMPHADENITIS
    - NECROTIZ NONKIKUCHI
    - NECROTIZING KIKUCHI
  - RHEUMATOID ARTHRITIS
  - SARCOIDOSIS
  - SHML
  - SINUS HYPERPLASIA
  - SYPHILIS
  - TOXOPLASMOSIS
  - TUBERCULOSIS
  - VIRAL
    - INFECTIOUS MONO
    - VIRAL NOS
  - WHIPPLE'S DISEASE

- HODGKIN'S
  - LYMPH PREDOM
    - L&H DIFFUSE HD
    - L&H NODULAR HD
  - LYMPHOCYTE DEPLETION
    - DIFFUSE FIBROSIS HD
    - RETICULAR TYPE HD
  - MIXED CELL TYPE
    - INTERFOLLICULAR HD
    - MIXED CELLULARITY HD
  - NODULAR SCLEROSIS
    - CELLULAR PHASE NSHD
    - NODULAR SCLEROSIS
    - SYNCYTIAL NSHD

- NONHODGKIN'S
  - FOLLICULAR
    - LARGE CELL, FOL
    - MIXED, FOL
    - SMALL CLEAVED, FOL
    - SMALL NONCLEAVED FOL
  - MANTLE CELL
    - MANTLE ZONE
  - NONFOLLICULAR
    - B-IMMUNOBLASTIC
    - IBL-LIKE T-CELL LYM
    - JAPANESE ATL
    - LARGE CELL, DIF
    - LYMPHOBLASTIC
    - MALIG HISTIOCYTOSIS
    - MIXED, FCC DIF
    - PLASMACYTOID LYCTIC
    - SMALL CLEAVED, DIF
    - SMALL LYMPHOCYTIC
    - SMALL NONCLEAVED DIF
    - T-IMMUNOB LRG
    - T-IMMUNOB MIX
    - TRUE HISTIOCYTIC

- METASTATIC
  - CARCINOMA
  - LEUKEMIAS
    - AML
    - HAIRY CELL LEUKEMIA
  - LYMPHOID
    - MULTIPLE MYELOMA
    - MYCOSIS FUNGOIDES
  - MELANOMA

- OTHER
  - AILD
  - EM PLASMACYTOMA
  - KAPOSIS SARCOMA
  - MAST CELL DISEASE

**Figure 4.7:** A template for many of Pathfinder's partitions.
We derived partitions for many of Pathfinder's features from this partition. The partition reflects a classification hierarchy used commonly by expert hematopathologists. Diseases are grouped into 5 sets that represent BENIGN, HODGKIN'S, NONHODGKIN'S, METASTATIC, and OTHER diseases; these sets reflect the uppermost branch of the classification hierarchy. Each set, in turn, is organized as a tree (note the indentation in each set); these trees represent the lower branches of the classification hierarchy.

histiocytic nodular sclerosing Hodgkin's). The structure of this partition is closely related to that of the partition shown in Figure 4.7. Using SimNet, we can exploit this fact, and can compose the partition for PTGC from the partition in Figure 4.7 with just a few mouse and keystroke commands.

Figure 4.9 shows an equivalent partition for the feature PTGC. In the figure, the substructure of some of the sets in the partition are hidden from view. Using SimNet, we can hide or reveal the contents of any set or component of a set (such as HODGKIN'S) with a single mouse command. This feature of SimNet immensely helped the expert to manage the complexity of the probability-assessment task.

Figure 4.10 shows the partition for the feature NECROSIS. Obviously, this partition is more complex than is the partition for PTGC. Nonetheless, much of the structure found in the partition for PTGC is preserved in the partition for NECROSIS. For example, most of the benign and metastatic diseases remain clustered in the set labeled MOST DISEASES. Again, using SimNet, we can exploit this fact to compose this partition with ease.

## 4.3 Construction Statistics

Table 4.2 summarizes the statistics for the construction of Pathfinder III and IV. Pathfinder IV contains three more diseases than does Pathfinder III. The reason for this difference is that three diseases in Pathfinder III—AIDS, necrotizing lymphadenitis, and T-immunoblastic lymphoma—are each split into two subtypes in Pathfinder IV. In Section 5.4.1, we discuss the significance of adding these distinctions. Also, Pathfinder IV contains four fewer features than does Pathfinder III. Two factors that we have already discussed account for the difference in number. First, additional features are included in the Pathfinder IV knowledge base to break dependencies among particular sets of features. Second, several groups of features in Pathfinder III are clustered into single features in Pathfinder IV. The latter factor dominated the former by a small amount.

We created the structure of the Pathfinder III knowledge base in approximately 8 hours. Because we employed the simple Bayes model to build the knowledge base, we did not compose a knowledge map explicitly. Instead, we simply enumerated all lymph-node diseases and all features and instances relevant to the diagnosis of those diseases. In contrast, the expert and I spent approximately 35 hours constructing the global knowledge map for Pathfinder IV. As shown in Table 4.2, we spent about 3 hours building the similarity graph, and the remainder of the time composing the local knowledge maps. We consulted the list of diseases and features from Pathfinder III during the construction of Pathfinder IV. Therefore, to make the comparison of the two systems fair, we add 8 hours to the construction time for the Pathfinder IV knowledge map.

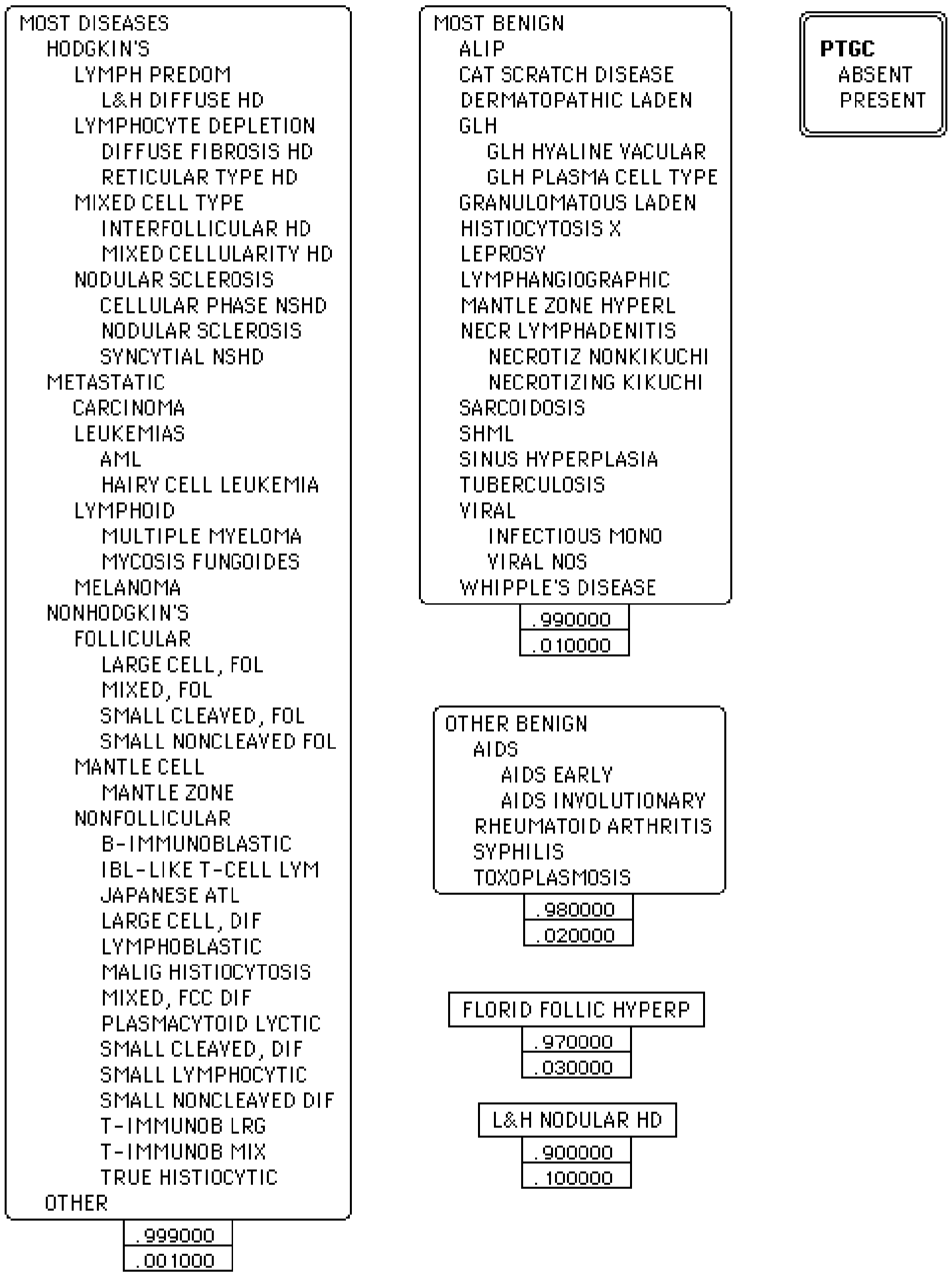

**Figure 4.8:** A partition for the feature PTGC.

For any two diseases in the same set, the feature PTGC is not relevant to that disease pair. Thus, for each set, we require only a single probability distribution for PTGC given disease. These distributions are shown below each set.

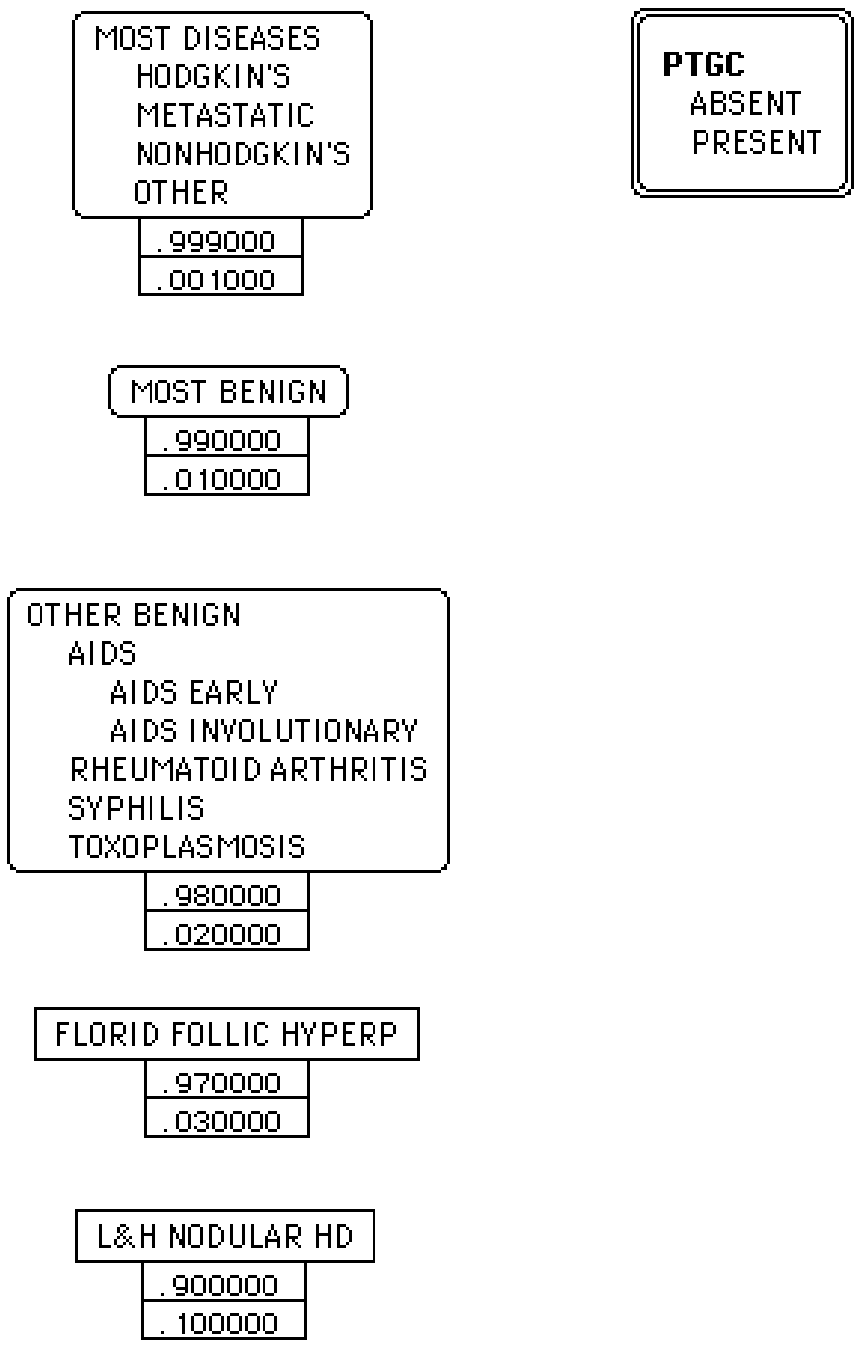

**Figure 4.9:** A partition for PTGC equivalent to that in Figure 4.8.
In this partition for PTGC, the substructure of MOST BENIGN DISEASES and most of the substructure of MOST DISEASES are hidden from view. Using SimNet, we can easily transform the structure of this partition to that of the partition shown in Figure 4.8, and vice-versa.

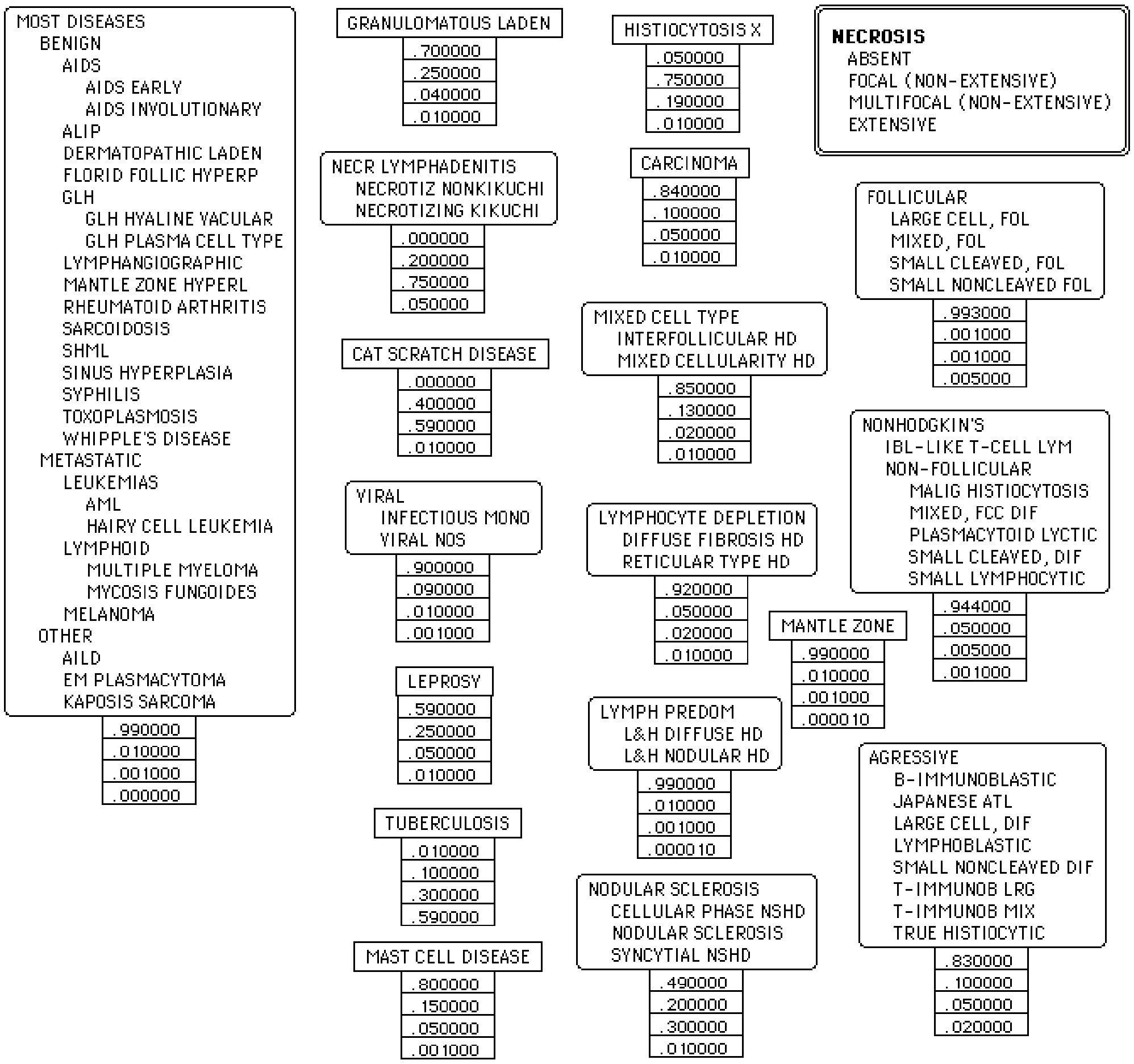

**Figure 4.10:** A partition for NECROSIS.
The structure of this partition is more complex than is that of the partition for germinal centers. Nonetheless, portions of this partition resemble the simpler partition.

**Table 4.2:** Statistics for the construction of Pathfinder III and IV.

| Statistic | Knowledge Base | |
|---|---|---|
| | Pathfinder III | Pathfinder IV |
| Number of diseases | 60 | 63 |
| Number of features | 112 | 108 |
| Time to define distinctions (hours) | 8 | 8 |
| Time to build similarity graph (hours) | - | 3 |
| Time to build knowledge map (hours) | - | 32 |
| Assessments required by system | 16,620 | 74,854 |
| Assessments derived from partitions | - | 61,118 |
| Assessments made by expert | 16,620 | 13,736 |
| Time to assess probabilities (hours) | 35–40 | 39 |

As shown in Table 4.2, Pathfinder IV contains approximately 4.5 times as many assessments as does Pathfinder III.[2] The figure of 74,854 assessments for Pathfinder IV, however, does not take into account the decrease in the number of assessments resulting from the use of partitions. If we count only the number of actual assessments (one distribution for every set in a partition), Pathfinder IV contains only 13,736 assessments—5.4 times fewer assessments than without partitions. Overall, Pathfinder IV contains slightly fewer probability assessments than did Pathfinder III. Furthermore, the time we spent on probability assessment for the two versions of the system are comparable (35 to 40 hours for Pathfinder III versus 39 hours for Pathfinder IV). When building Pathfinder IV, we spent only a small fraction of assessment time (less than 20 percent) composing partitions.

The statistics in Table 4.2 do not reflect our efforts in testing and refining the knowledge maps for the two systems. Testing consisted mostly of observing differential diagnoses generated from imagined cases. When the expert was dissatisfied with a differential diagnosis, it usually was easy to pinpoint the source of his dissatisfaction and to correct the problem. We have tested Pathfinder III—constructed over 3 years ago—on over 300 imagined cases, whereas we have tested Pathfinder IV on only about 70 imagined cases. Also, Pathfinder III was tested on 24 real cases in a formal evaluation (Heckerman, 1988).

[2] These figures do not count the assessments that are determined by the fact that the probabilities for the instances of a given feature must sum to 1.

In the following chapter, we compare the diagnostic accuracy of Pathfinder III and IV. In examining the results of the comparison, we should keep these facts in mind, because the greater efforts spent on testing and refining Pathfinder III bias the results in favor of this system.

Overall, the similarity-network and partition representations greatly facilitated the capture of dependencies among features in the lymph-node domain. A similarity network made possible the construction of the global knowledge map for the domain; partitions reduced the number of probability assessments (and time to assess) by more than a factor of five.

These observations apply to a single expert who was familiar with decision-theoretic concepts at the time we began the construction of Pathfinder IV. Nonetheless, I expect that even experts who are not accustomed to thinking in decision-theoretic terms will find these representations useful for knowledge acquisition. There are many techniques for helping people to structure decision-theoretic models and to provide accurate probability and utility assessments (Winkler, 1967a; Winkler, 1967b; Spetzler and Stael von Holstein, 1975; Howard, 1988a; Langlotz, 1989; Klein, 1989).[3] I used several of these techniques to train the Pathfinder expert prior to the construction of Pathfinder III. My experience, and the experience of other researchers, suggests that most people can adapt easily to decision-theoretic thinking.

## 4.4 Insights

In building the Pathfinder knowledge base, the expert and I developed important insights about the similarity-network and partition representations, and about probability assessment. In this section, we examine several of these insights.

### 4.4.1 Insights About Similarity Networks

Let us reexamine the sufficient conditions for the soundness result, and discuss how they affected the construction of the global knowledge map for Pathfinder. Recall, from Chapter 3, that the construction of a global knowledge map from a similarity network is sound whenever the following constraints are satisfied:

1. Hypotheses are mutually exclusive and exhaustive.
2. The similarity graph is connected.
3. The global knowledge map that is equal to the graph union of the local knowledge maps contains no directed cycles.

[3] Also, see Appendix A.

4. There are no arcs pointing to the distinguished node in any local knowledge map.
5. The joint distribution for the domain is *strictly positive* (i.e., there are no probabilities in the distribution that are equal to zero).

As we discussed in Chapter 1, constraint 1 does not pose a problem for the lymph-node domain, because diseases that co-occur in the same patient almost always manifest in different lymph nodes or in different sections of the same lymph node. In Chapter 6, we discuss a procedure for using the similarity-network representation even when diseases or hypotheses are not mutually exclusive. As a consequence of this procedure, the first constraint is of little concern for many domains.

In Section 4.2, we saw that the lymph-node expert had no trouble composing a connected similarity graph for Pathfinder. This observation suggests that constraint 2 will not be a serious impediment to the use of the similarity-network representation, in general. If we do come across a domain in which an expert cannot compose a connected similarity graph, we should consider the possibility that the domain is actually two or more well-isolated subdomains, and build separate expert systems for each cluster of connected hypotheses in the similarity graph.

Constraint 3 implies that the arcs in each knowledge map must flow in the same direction. More precisely, if there is an arc from nodes $x$ to $y$ in some local knowledge map, there can be no path from nodes $y$ to $x$ in any local knowledge map. Due to the particular implementation of the similarity-network representation within SimNet, this constraint was not a significant barrier to the construction of the global knowledge map. Specifically, when we added feature $x$ to a local knowledge map, then for all features $y$ already in the local knowledge map, SimNet automatically added an arc between $x$ and $y$ if that arc existed in the global knowledge map. It was up to the expert to remove an arc if, given the context of the local knowledge map, he wanted to assert the conditional independence constraints implied by that removal. Thus, it was impossible for the expert to draw an arc from features $x$ to $y$ in one local knowledge map and an arc in the opposite direction in another local knowledge map. Furthermore, in the rare circumstances when we created a directed cycle in the global knowledge map by adding an arc in a local knowledge map, SimNet notified us of this condition. By immediately inspecting the global knowledge map, we were able to remove the cycle we had created with little effort.

Constraint 4 also did not impose any barrier to the construction of the knowledge map for Pathfinder. That is, the expert did not introduce any features that were predecessors of the disease node. This situation, however, is not typical across medical domains. In many such domains, features can cause disease. For example, excessive alcohol intake tends to cause liver disease. In such cases, experts are almost always more comfortable drawing arcs from features to diseases. In Chapter 6, we discuss an approach for using the similarity-network representation in these situations.

Constraint 5 affected significantly the construction of the global knowledge map. Many of the distributions in Pathfinder contain probabilities equal to 0. To circumvent this constraint, the expert spent considerable time (over 8 hours) checking that the global knowledge map was valid by direct inspection. That is, he considered many conditional independence constraints implied by the global knowledge map, and, by introspection, determined that he was willing to assert those constraints. In general, this solution is not adequate. If an expert finds it difficult to compose a knowledge map for the entire domain, he should not be expected to validate that knowledge map once it is constructed. Indeed, in the words of the expert, this process of validating of the global knowledge map for Pathfinder was "quite painful." More theoretical work is needed to characterize those joint probability distributions for which the construction of the global knowledge map from the similarity network is sound. I suspect that the soundness result will hold for many types of nonpositive distributions. For example, as is suggested by the similarity network in Figure 3.4(c) on page 76, it appears that the soundness result for comprehensive similarity networks will apply to any distribution provided nodes that are nondeterministic in the global knowledge map do not become deterministic in any local knowledge map within the similarity network. If this or a similar result can be proved and can be extended to ordinary similarity networks, then direct validation of the global knowledge map for many domains will become unnecessary. In addition, we must extend the algorithm for checking the consistency of a similarity network to include nonpositive distributions.

Aside from evaluating the effect of theoretical constraints on the soundness of knowledge-map construction, we discovered several other properties of similarity networks during the construction of the Pathfinder knowledge map. Recall that, in phase 1 of the two-phase composition of a local knowledge map, the expert identified features that were relevant to disease pairs without considering dependencies among the features. This process helped to remind the expert of all features that were relevant to the lymph-node domain. In particular, during this phase, the expert discovered several features that were excluded from Pathfinder III, simply because he had not thought to include them. In addition, the results of phase 1 provided a reminder tool for probability assessment. Occasionally, during assessment, the expert forgot why a particular feature was useful for diagnosis. In this situation, we had SimNet highlight on the similarity graph those disease pairs that were discriminated by the feature. In these cases, the expert immediately recalled the value of the feature for diagnosis, and then proceeded with probability assessment.

In phase 2 of local-knowledge map composition, when we add a feature to a local knowledge map, SimNet automatically adds arcs between this feature and any other features in the local knowledge map whenever those arcs exist in the global knowledge map. This feature of SimNet greatly facilitated the capture of dependencies. After about one-third of the local knowledge maps were created, most of the arcs that eventually would be drawn in the global knowledge map were present in this knowledge map. Furthermore, as we discussed in Section 3.8, if two features were dependent in one local knowledge map, they were also dependent in almost all other maps. Thus, for the majority of local knowledge maps, the expert needed only to verify the dependency arcs that were automatically created by SimNet in those maps.

Also, during phase 2, the expert quickly appreciated how arcs drawn in a local knowledge map affected the global knowledge map. With this understanding, he occasionally assessed dependencies among features while viewing the global knowledge map directly. In particular, he would assume that only one of two diseases were present (as he did when building a local knowledge map), but draw the informational arcs appropriate to this background knowledge on the global knowledge map. At times, this mode of composition was useful, because the expert could see all the features on the computer screen simultaneously, and thereby could easily identify those features that were dependent in the local context.

### 4.4.2 Insights About Partitions

We made two important observations about partitions during the construction of the Pathfinder knowledge base. First, the expert found it exceptionally easy to compose the partitions. Of the 40 or so hours spent assessing probabilities for the knowledge map, we spent only about 4 hours composing the partitions. As we discussed in Chapter 2, construction was easy because (1) the expert could make judgments concerning subset irrelevance without first having to assess the underlying probability distributions, and (2) many partitions were identical or closely related from one feature to another.

Second, partitions improved the quality of the expert's probability assessments. In an informal experiment, the expert compared probability distributions for approximately 10 features from Pathfinder III with those distributions for identical features from Pathfinder IV. For each feature, the expert strongly preferred the distributions from Pathfinder IV. The experiment was not blinded, because the expert could easily identify the distributions taken from Pathfinder IV. Nonetheless, the expert had no incentive to favor one set of distributions over the other, and the results appear to be significant.

Several attributes of the partition representation contributed to the improvement. Often, the probability $p\,(\text{feature}|\text{disease}, \xi)$ does not depend on the disease itself. Instead, this probability depends on some abstract property of the disease. For example, consider the feature NECROSIS, which refers to the presence of dead cells. The degree of necrosis or cell death seen in a lymph-node section depends on the aggressiveness of the disease.

If a disease progresses rapidly, remnants of cells that have been killed by the spread of the disease are often present; if a disease progresses slowly, the immune system of the patient typically removes any such traces of the spread of the disease. In Pathfinder IV, we used the partition for NECROSIS (see Figure 4.10) to represent this knowledge explicitly. That is, in the partition, we grouped the diseases by their aggressiveness. In contrast, when constructing Pathfinder III without partitions, the expert assessed a probability distribution for NECROSIS given a particular class of aggressiveness many times. Each assessment produced a slightly different distribution. (The expert could rarely reproduce any probability assessment to within more than one significant figure.) Thus, these direct probability assessments indicated incorrectly that necrosis was relevant to many disease pairs. The partition representation provided us with a means to avoid the introduction of such *spurious relevancies*.

Another attribute of partitions that facilitated better assessments is illustrated by the partition for progressively transformed germinal centers shown in Figure 4.9. In this partition, the sets of diseases are arranged such that a set $S_2$ is below $S_1$ if and only if the probability of seeing germinal centers, given a disease in set $S_2$, is higher than the probability of seeing germinal centers, given a disease in set $S_1$. To assess the probabilities for PTGC, the expert first arranged the sets as described. He then used the graphical arrangement of these sets to avoid assessments that were inconsistent with his qualitative understanding of the relationship between PTGC and disease. In its basic form, a partition allows an expert to represent *equivalencies* among probability distributions. Here, we see that a partition, implemented in graphical form, allows an expert to represent *differences* among assessments as well. Such representation facilitates the comparison and, thereby, the evaluation of probability assessments.

### 4.4.3 An Insight About Probability Assessment

While assessing probabilities, the expert and I repeatedly encountered a problem that has a simple solution. I mention this difficulty and its solution here for other researchers who plan to construct large influence diagrams.

Let us consider the feature NECROSIS, which conditions the feature KARYORRHEXIS. From Figure 4.10, the probability that NECROSIS is ABSENT, given CAT SCRATCH DISEASE is 0. Thus, we do not need to assess a probability distribution for KARYORRHEXIS and CAT SCRATCH DISEASE, given that there is no necrosis in the tissue section. In SimNet, we represent distributions with impossible conditioning events, by placing 0 in each slot of the distribution. This convention is illustrated in the partition for KARYORRHEXIS, shown in Figure 4.11.

In the current implementation of SimNet, the user must make sure that the assessments obey this convention. Thus, there is the possibility of inconsistency. In particular, when assessing a feature that is dependent on other features, an expert might forget that a particular conditioning instance is impossible, and thereby needlessly assess a distribution. Conversely, the expert might sometimes believe incorrectly that a conditioning instance was impossible, and fail to assess a needed distribution. Avoiding inconsistencies becomes especially difficult during the phase of knowledge-base development in which the system is tested and the probabilities are modified. When a probability of seeing a feature is changed from 0 to another value, an expert easily can forget to assess the distributions that are conditioned by that feature.

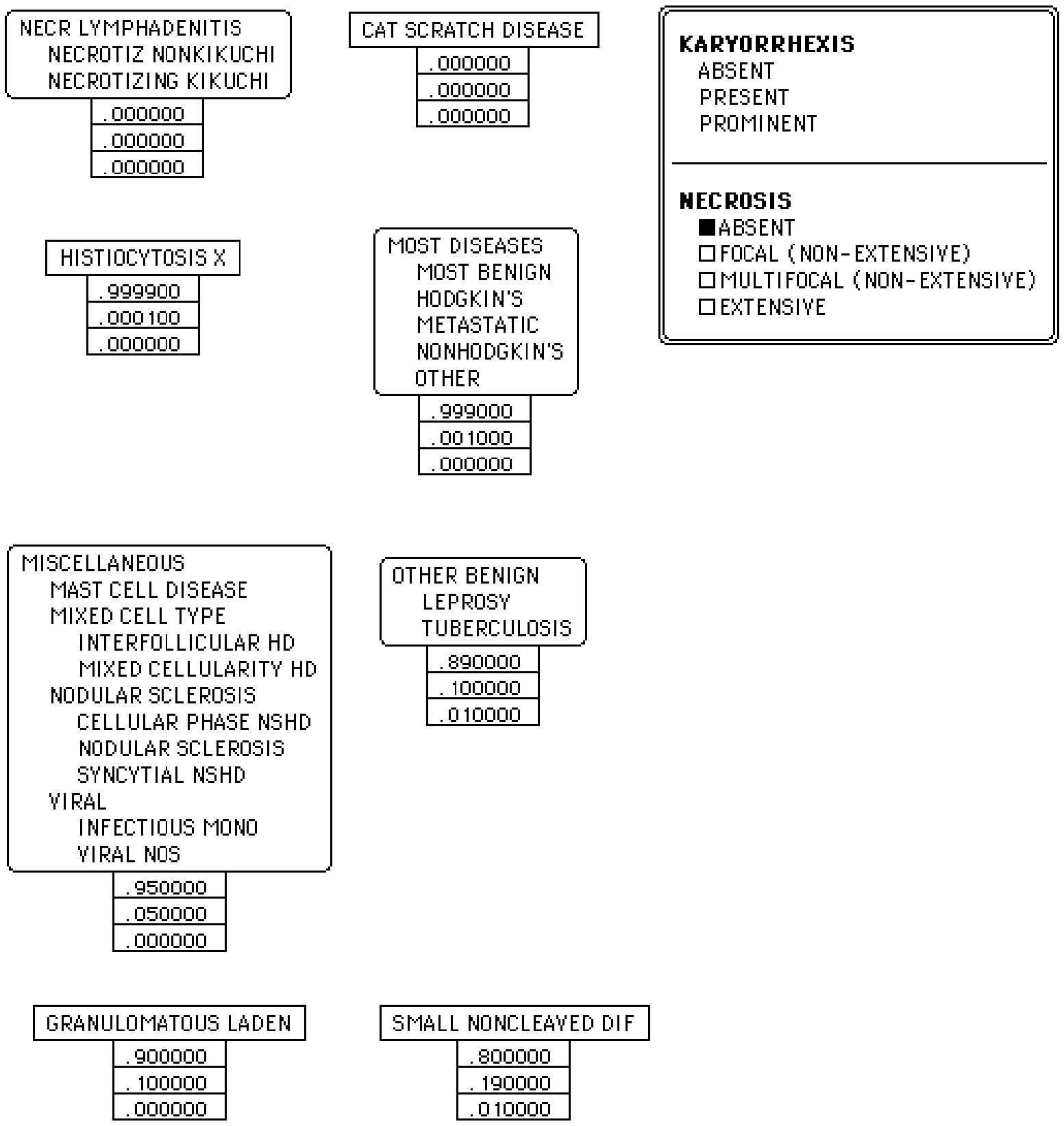

**Figure 4.11:** The probability assessments for the feature KARYORRHEXIS. The probability that NECROSIS is ABSENT, given CAT SCRATCH DISEASE, is 0. Thus, we do not assess a probability distribution for KARYORRHEXIS and CAT SCRATCH DISEASE, given the absence of necrosis. Similarly, we do not assess a probability distribution for either form of NECR LYMPHADENITIS (necrotizing lymphadenitis).

While assessing the probabilities for Pathfinder, the expert and I were careful to avoid these errors. Whenever there was a question as to the possibility of a conditioning instance, we checked the probability of that event, using the display facilities of SimNet. Nonetheless, the process of maintaining a consistent set of probability assessments was tedious and time consuming. Furthermore, despite our efforts, we made several errors when modifying probability distributions in the later stages of knowledge-map construction.

To avoid this problem, SimNet—and other influence-diagram programs—could identify graphically all probability distributions that have impossible conditioning events. For example, the boxes that surround such a probability distribution could be displayed in a shade of gray, rather than in black. Also, when a probability is changed from 0 to another value, these programs could inform the user that additional assessments are required, and direct the user to those assessments.

## 4.5 The Pathfinder Inference Algorithm

So far in this chapter, we have discussed exclusively the construction of the knowledge base for lymph-node pathology. In concluding this chapter, let us examine the algorithm for probabilistic inference that Pathfinder employs to manipulate this knowledge. (For a definition of probabilistic inference, see Appendix A, page 184).

The algorithm, designed and implemented by Jaap Suermondt, exploits the assertions of symmetric conditional independence in the global knowledge map. Specifically, the algorithm is based on the observation that the global knowledge map consists of relatively small clusters of features such that each cluster is conditionally independent of all others, given a disease.[4] This observation is illustrated schematically in Figure 4.12(a). The clusters $X^1$, $X^2$, ... $X^n$, which each contain one or more features, are conditionally independent, given an instance of the disease variable $d$.

[4]This observation is apparent in Figure 4.1. In the figure, all arcs from DISEASE to the feature nodes are omitted. Thus, a cluster is a set of features that is connected by an undirected path in the figure.

To understand the algorithm, suppose we observe one or more features $O^1$ in $X^2$ and one or more features $O^2$ in $X^2$. Let $O_i^1$ and $O_j^2$ denote the instances that we observe for $O^1$ and $O^2$, respectively. From Bayes' theorem and the conditional independence of clusters, we can compute the probability of each disease, given our observations.

$$p\left(d_m|O_i^1, O_j^2, \xi\right) = \frac{p\left(O_i^1|d_m\xi\right)\ p\left(O_j^2|d_m, \xi\right)\ p\left(d_m|\xi\right)}{\sum_{d_n}\ p\left(O_i^1|d_n\xi\right)\ p\left(O_j^2|d_n, \xi\right)\ p\left(d_n|\xi\right)} \tag{4.5.1}$$

In the current version of the algorithm, we calculate the terms $p\left(O_i^1|d_m, \xi\right)$ and $p\left(O_j^2|d_m, \xi\right)$ using a brute-force computation that does not exploit conditional independence within the clusters. For example, suppose cluster $X^i$ is composed of the three features $x$, $y$, and $z$, as illustrated in Figure 4.12(b). Further, suppose that we have observed the instance $z_k$ for $z$. We compute

$$p\left(z_k|d_m, \xi\right) = \sum_{x_i, y_j} p\left(z_k|y_j, d_m, \xi\right)\ p\left(y_j|x_i, d_m, \xi\right)\ p\left(x_i|d_m, \xi\right)\ p\left(d_m|\xi\right) \tag{4.5.2}$$

where the sum runs over all the instances of the features $x$ and $y$.

Given observations for the majority of features in Pathfinder, this inference algorithm requires less than 1 second to compute the posterior probability of all diseases. For some features in the large cluster containing lymphoid cells (see Figure 4.4), however, the algorithm requires almost 20 minutes to compute these probabilities. Thus, members of the Pathfinder team are currently investigating methods for increasing the efficiency of inference. In one approach, we could exploit conditional independence within the clusters by applying either the Lauritzen–Spiegelhalter algorithm (Lauritzen and Spiegelhalter, 1988) or Pearl cutset-conditioning algorithm (Pearl, 1988) to compute $p\left(O_i|d_k, \xi\right)$ for each cluster $X^i$. Either algorithm, for example, would simplify the computation in Equation 4.5.2, by exploiting the fact that $z$ is independent of $x$, given $d$ and $y$. In effect, either algorithm would move the summation over $x$ to the right of the first term to obtain

$$p\left(z_k|d_l, \xi\right) = \sum_{y_j} p\left(z_k|y_j, d_l, \xi\right) \sum_{x_i} p\left(y_j|x_i, d_l, \xi\right)\ p\left(x_i|d_l, \xi\right)\ p\left(d_l|\xi\right) \tag{4.5.3}$$

To simplify inference further, we could exploit the assertions of asymmetric conditional independence represented within the similarity network. We discuss this approach further in Section 6.2.1.

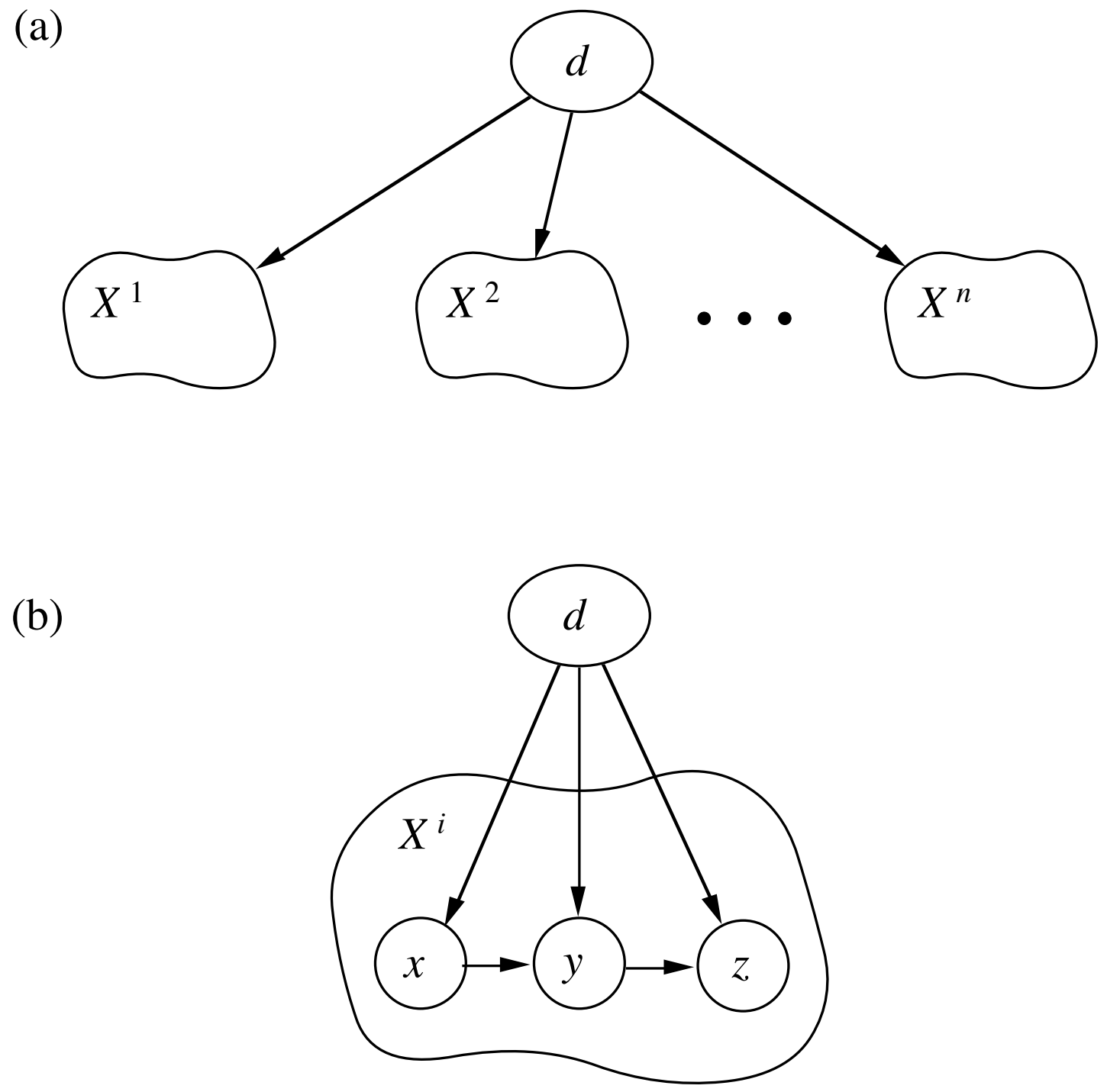

**Figure 4.12:** A schematic knowledge map for Pathfinder.
(a) The features in Pathfinder can be arranged into clusters of features $X^1$, $X^2$, ... $X^n$. The features within each cluster are dependent, but the clusters are conditionally independent, given the disease variable $d$. (b) A detailed view of the features $x$, $y$, and $z$ within cluster $X^i$.

# 5 An Evaluation of Pathfinder

In the previous chapter, we saw that similarity networks made possible the construction of a version of Pathfinder in which dependencies among features are represented, and that partitions significantly reduced the time that would otherwise be required to assess a knowledge map. In this chapter, we examine whether the knowledge base created through the use of these new representations is more accurate for diagnosis than is the original simple Bayes knowledge base.[1] Specifically, we compare Pathfinder III and IV in a three-phase experiment. In the first phase, we ask the question: Is the diagnostic accuracy of Pathfinder IV greater than Pathfinder III? In the second phase, we ask: What factors are responsible for improvement, if any? In the third phase, we ask: Are the improvements worth the effort of building the more sophisticated knowledge base?

## 5.1 Selection of Cases

A set of cases for the experiment was selected in sequence from a large library of cases that had been referred to Dr. Nathwani from community pathologists. Because such cases were referrals, they were likely to be at least as difficult as cases in which nonexperts would seek the help of a computer aid. Cases were rejected only if glass slides were unavailable or if the case did not involve lymph-node tissue. Sections that were poorly stained or improperly sliced were not excluded. Over 100 cases were selected; because of time constraints, however, the experiment was conducted on only the first 53 cases.

## 5.2 Entry of Features

A community pathologist entered features observed in each case into both Pathfinder III and Pathfinder IV. She entered only morphologic features (i.e., features observed through a microscope); she did not perform tests that were expensive or that would have caused significant delays in the experiment. She was allowed to see the recommendations for additional observations made by both systems if she was unsure about what feature to enter next. Also, if she was unsure about the identification of a feature, she was allowed to access a library of over 4000 video images that illustrates the morphologic features that can be reported to the two systems. For each case, the pathologist entered features until she believed that no additional observations were relevant to that case. In most instances, she stopped entering features when neither program had further features to recommend that she examine. For several cases, however, although one or both systems recommended that she examine additional features, she did not observe any of those features, because she believed that they would not have a significant effect on the differential diagnosis. For several other cases, where neither system recommended that she examine additional features, she identified on her own features that she thought might be relevant to the case, and entered those features.

[1] Several attributes of expert systems other than diagnostic accuracy are critical to the acceptance of such systems in clinical practice. These attributes include the usability of a system and the degree to which a system can improve the quality of physicians' decisions. Here, we concentrate on diagnostic accuracy, because this attribute of expert-system performance is most directly affected by the new representations described in this work.

**Table 5.1:** Expert ratings for Pathfinder III and IV.

| Knowledge Base | Expert Ratings (0-10 scale) mean | sd |
|---|---|---|
| Pathfinder III | 7.99 | 2.32 |
| Pathfinder IV | 8.94 | 1.51 |

The pathologist chosen for the experiment was recently a fellow in hematopathology with Dr. Nathwani. She was selected because she was familiar with the lymph-node domain and with most of the terminology used by Pathfinder.

## 5.3 Phase 1: An Expert-Rating Metric

In phase 1 of the experiment, we wanted to determine whether the diagnostic accuracy of Pathfinder IV was greater than that of Pathfinder III. For each case, our expert was shown the features reported by the nonexpert, as well as the probability distributions produced by the two versions of the system. The expert was blinded as to the identity of the distributions, and the distributions were displayed in random order. For each probability distribution, the expert was asked, "On a scale from zero to ten—zero being unacceptable and ten being perfect—how accurately does the distribution reflect your beliefs?"

The mean and standard deviation of the expert ratings for Pathfinder III and Pathfinder IV are shown in Table 5.1. The case-by-case results are shown in Appendix D. The experiment reveals a significant difference between the two systems. Specifically, a bootstrap permutation test (Diaconis and Efron, 1983) yields an achieved significance level (ASL) of 0.007. The permutation test indicates that there is only a 0.007 chance that a more extreme result would be obtained if data were drawn at random from the set union of the ratings for both Pathfinder III and Pathfinder IV.

**Table 5.2:** Factors of Pathfinder IV that affected its diagnostic accuracy relative to Pathfinder III.

Factors that Increased Diagnostic Accuracy

| Number of Cases Affected | Factor |
|---|---|
| 5 | Conditioning produced better assessments |
| 3 | Dependencies existed among observed features |
| 2 | Partitions reduced spurious relevance in probability assessments |
| 2 | Comparisons afforded by partitions produced better assessments |
| 2 | Expert's knowledge improved since construction of Pathfinder III |
| 1 | Disease subdistinction introduced |
| 2 | Probability assessments improved for unknown reasons |

Factors that Decreased Diagnostic Accuracy

| Number of Cases Affected | Factor |
|---|---|
| 1 | Failure to maintain consistency of knowledge base |
| 2 | Probability assessment worsened for unknown reasons |

## 5.4 Phase 2: A Case-by-Case Analysis

The experiment described in the previous sections shows that there is a difference between Pathfinder III and Pathfinder IV, but it does not identify those aspects of the two knowledge bases that are responsible for these differences. To discern the causes for the observed differences, I examined each patient case where the difference between the expert ratings for Pathfinder III and IV exceeded 1.5.

There were 12 cases in which the expert rating for Pathfinder IV exceeded that for Pathfinder III by this threshold. In nine cases, a single factor was responsible for the increased performance; in two cases, two factors were responsible; and in one case, four factors were responsible. In contrast, there were only three cases in which the expert rating for Pathfinder III exceeded that for Pathfinder IV by 1.5. In all three cases, a single factor of the knowledge base was responsible for this decrease in diagnostic accuracy. Table 5.2 summarizes the factors of the Pathfinder IV knowledge base that increased or decreased its diagnostic accuracy relative to Pathfinder III. Many of these factors affected performance in more than one case. The tables show the number of times each attribute contributed to a difference in performance.

### 5.4.1 Causes of Increased Diagnostic Accuracy

In eight of the 12 cases where Pathfinder IV outperformed Pathfinder III, the representation of feature dependencies contributed to the superior performance of Pathfinder IV. In three of the cases, Pathfinder IV's increased accuracy was a direct consequence of the explicit encoding of dependencies. That is, in these three cases, the community pathologist observed features that were dependent. In the remaining five cases, however, the source of the improvement was indirect. In particular, by conditioning probability assessments for a feature on other features, the expert provided probabilities of higher quality. For example, let us consider the assessment of the probability distribution for LLC CYTOPLASM (color of large-lymphoid-cell cytoplasm), given DISEASE. In Pathfinder III, the expert provided these assessments directly. In Pathfinder IV, however, the expert conditioned these assessments on LLC IDENTITY (identity of large lymphoid cells). That is, he assessed a probability distribution for LLC IDENTITY, given DISEASE, and probability distributions for LLC CYTOPLASM, given DISEASE and LLC IDENTITY. This technique for decomposing the assessment of a probability distribution is called *extending the conversation*. Using this technique, an expert can avoid having to average over a set of distributions in his head, and thereby can produce better assessments. For a detailed discussion of this technique and the conditions under which it is useful, see Tribus (1969, Chapter 3), de Finetti (1977), and Heckerman and Jimison (1987).

The use of partitions led to increased performance in four of the 12 cases. In Section 4.4.2, we discussed two attributes of the partition representation that facilitated probability assessment. Specifically, partitions reduced the introduction of *spurious relevancies*, and partitions facilitated the comparison of probability assessments. Both of these attributes produced improvements in diagnostic accuracy in two of the 12 cases.

Another source of increased accuracy was that the expert's knowledge improved since the construction of Pathfinder III. For example, in a previous evaluation of Pathfinder III (Heckerman, 1988), the system performed poorly in many cases because the probabilities assessed for the feature epithelioid clusters of histiocytes were contradicted by data. That is, the expert said that these clusters were never seen in most diseases, yet, in the process of evaluating the system, he saw small numbers of these clusters in unexpected settings. During the year since that experiment, the expert paid close attention to these clusters in his daily diagnostic workups. Thus, the probability distributions for this feature that he provided later were significantly more informed than were those he provided for Pathfinder III.

In one case, the diagnostic accuracy of Pathfinder IV was superior to that of Pathfinder III because we introduced disease subtypes into the latter system. To see how the failure to include disease subtypes can decrease diagnostic accuracy, let us consider the disease necrotizing lymphadenitis, which has subtypes Kikuchi's and nonKikuchi's. In nonKikuchi's necrotizing lymphadenitis, we always see necrosis, and we sometimes see large numbers of plasma cells. On the other hand, in Kikuchi's necrotizing lymphadenitis, we may not see necrosis, and we never see large numbers of plasma cells. Furthermore, these two features are conditionally independent, given disease. Thus, if we fail to observe necrosis in a given lymph node, and if we see a large number of plasma cells in that same node, then both subtypes of necrotizing lymphadenitis should be ruled out. Suppose, however, that we construct an expert system that does not distinguish the two subtypes of disease, and retains the assertion of conditional independence. In this case, if we observe no necrosis and abundant plasma cells in a given lymph node, the expert system incorrectly reports that necrotizing lymphadenitis is a possible contender for the diagnosis of that node. Here, when we combine the two subtypes of necrotizing lymphadenitis, necrosis and plasma cells become conditionally dependent, given disease. Consequently, the diagnostic accuracy of such a system is less than that of a system that includes the distinction.[2]

Finally, in two of the 12 cases, we traced the improvements to differences between the systems in specific probability assessments. We could not, however, identify the underlying cause of the improvements.

### 5.4.2 Causes of Decreased Diagnostic Accuracy

In three cases, Pathfinder III outperformed Pathfinder IV. In Section 4.4.3, we discussed the problem of maintaining consistent probability assessments in a large knowledge map associated with nonpositive distributions. This difficulty was the source of Pathfinder IV's poor performance in one of these three cases. In the other two cases, we traced the decrement in accuracy to differences in specific probability assessments, but we could not determine the source of these differences.

## 5.5 Phase 3: A Decision-Theoretic Metric

The two approaches for evaluation that we have examined are easy to apply. Furthermore, they readily expose differences between the diagnostic accuracy of Pathfinder III and IV and the causes of these differences. Unfortunately, it is difficult to infer the *importance* of differences based on these experiments. Specifically, in Chapter 4, we saw that the construction of Pathfinder IV required approximately 40 more hours of effort than did the construction of Pathfinder III. Neither the difference between the average expert ratings of approximately 1.0 on a scale from 0 to 10 nor the identification of system factors responsible for the improvement, however, can tell us whether this additional effort was worth the improvement in diagnostic accuracy. In this section, we use an evaluation procedure, based on decision theory, that can address this tradeoff. The approach described here is similar to previous evaluations of several medical expert systems (Smets et al., 1975; Asselain et al., 1977; Habbema and Hilden, 1981).

[2] In principle, we could avoid the introduction of disease subtypes by representing the feature dependencies that result from such a representation. Usually, however, the number of induced dependencies is large, and this approach is impractical.

We compare Pathfinder III and IV by computing a quantity called *inferential loss* for both versions of the program and for each of the 53 test cases. The inferential loss associated with a version of Pathfinder and a given case is the decrease in expected utility that results from using a distribution produced by that version of the program, rather than the correct or *gold-standard* probability distribution associated with that case. To compute inferential loss, we require (1) gold-standard probability distributions for each case, and (2) the utility of every possible correct and incorrect diagnosis, given every disease that a patient might have. In Sections 5.5.1 and 5.5.2, we examine these components of the computation; in Section 5.5.3, we discuss the computation in detail.

### 5.5.1 Gold-Standard Distributions

It is difficult to produce an adequate gold standard in the domain of pathology. One approach, illustrated in Figure 5.1(a), is simply to use the *true* disease to construct the gold-standard distribution. That is, we assign a probability of 1 to the established diagnosis. In pathology, the disease that is manifested in a lymph node is determined (1) by an expert pathologist examining tissue sections under a microscope; (2) by expensive immunology, molecular biology, or cell-kinetics tests; (3) through observations of the time course of a patient's illness; or (4) by a combination of these approaches.

There are two problems with this gold standard. First, its use ignores the distinction between a good *decision* and a good *outcome*. For example, suppose the observations for a case suggest—say, through statistical data—that there is a 0.7 chance of Hodgkin's disease and a 0.3 chance of mononucleosis. Furthermore, suppose that mononucleosis is the true disease (not an unlikely event). In such a situation, an inference method that produces exactly this probability distribution for Hodgkin's disease and mononucleosis receives (unjustly) a lower rating than a distribution that produces a higher chance of mononucleosis. This problem with the approach, however, is not serious. We can attenuate differences between good decisions and good outcomes by considering a large number of cases.

A second, more serious, problem with this construction stems from details of how microscopic observations are made by experts and nonexperts. In my experience, when experts examine such biopsies, they typically see many features at once and come to a diagnosis immediately. When asked to identify specific features that appear in the biopsy, these pathologists report mostly features that confirm their diagnosis. Moreover, it is difficult to train these experts to do otherwise, and essentially impossible to determine whether or not such training is successful. Thus, when experts are used to identify features, both Pathfinder III and IV tend to perform well, and, in practice, it becomes impossible to identify significant differences from an experimental comparison. On the other hand, pathologists who do not specialize in the lymph-node domain misrecognize or fail to recognize some of features associated with diagnosis. It is unreasonable to compare the distributions produced by Pathfinder III and IV, derived from one set of observations, with the true disease, derived from a different set of observations. In fact, in a separate study, I showed that errors in diagnosis resulting from the misrecognition and lack of recognition of features by a nonexpert were sufficient to obscure completely the differences between the two versions of Pathfinder, when the true diagnosis was used as the gold standard (Heckerman et al., 1990).

An alternative procedure for constructing a gold standard is shown in Figure 5.1(b). In this procedure, an expert looks at only a list of observations for a case produced by another pathologist (expert or nonexpert), and assesses directly a probability distribution over the diseases. An associated drawback is that this construction ignores the possibility that one or both versions of Pathfinder might outperform the initial impressions of the expert. That is, if, for each case, our expert were to undergo a detailed decision analysis, costing thousands of dollars, the probability distributions determined by these analyses may be closer to the distributions produced by Pathfinder III or IV than to the expert's initial assessments of probability.

Thus, in the pathology domains, there appears to be no ideal gold standard. For this experiment, however, the construction of the gold-standard distribution using the true disease is unworkable, given the difficulties of feature observation associated with experts and nonexperts. Consequently, I employed the procedure illustrated in Figure 5.1(b). As mentioned in Section 5.2, a nonexpert pathologist identified features.

### 5.5.2 A Utility Model for Diagnosis

Pathfinder III and IV share the same utility model. The model is illustrated in the schematic influence diagram for Pathfinder IV shown in Figure 5.2. The chance node $d$ represents the set of all possible diseases. The decision node $dx$ represents all possible diagnoses, where a diagnosis is simply a statement of the form, "the patient has disease $d_j$." The node $u$ represents a patient's utility for all possible combinations of disease and diagnosis. We use the term $u_{d_i,d_j}$ to denote the utility of having disease $d_i$ and being diagnosed with disease $d_j$.

As we have discussed, a diagnosis in the domain of lymph-node pathology is a decision because clinicians base their treatment on the diagnoses rendered by pathologists. There are exceptions to this observation. For example, the treatment of Hodgkin's disease depends not only on the subtype of Hodgkin's disease, but also on the clinical stage of the disease (i.e., the degree to which the disease has spread throughout the body). In constructing the utility model, however, the expert and I addressed these exceptions by averaging over factors relevant to treatment that would be unknown at the time of diagnosis.

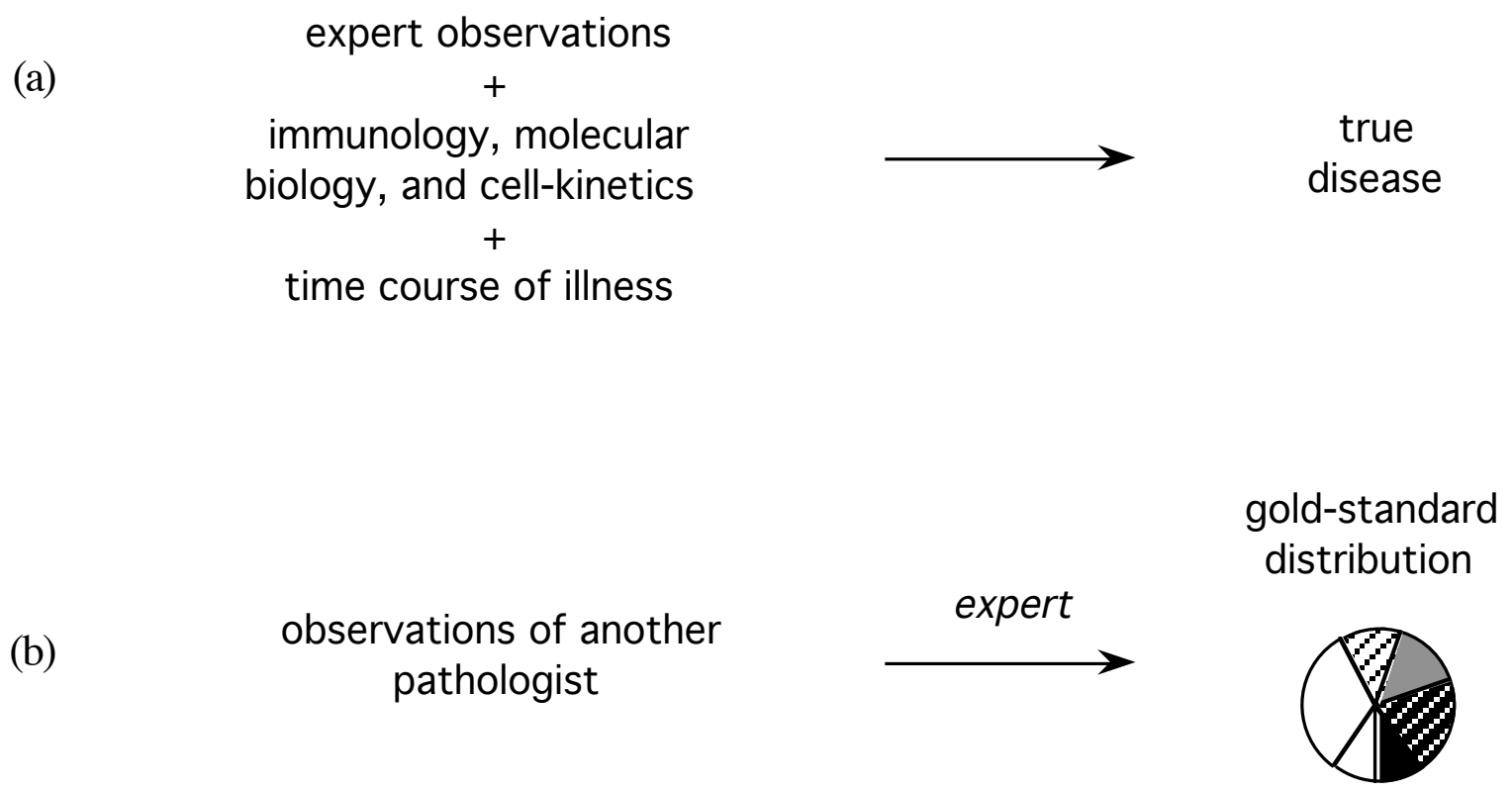

**Figure 5.1:** Alternative gold standards for a given case.
(a) From the morphologic observations of the expert, expensive immunology, molecular biology and cell-kinetics tests, and information about the time course of the patient's illness, we determine the true disease. (b) Given only the list of observations reported by another pathologist, the expert assesses a gold-standard probability distribution (represented by the shaded probability wheel).

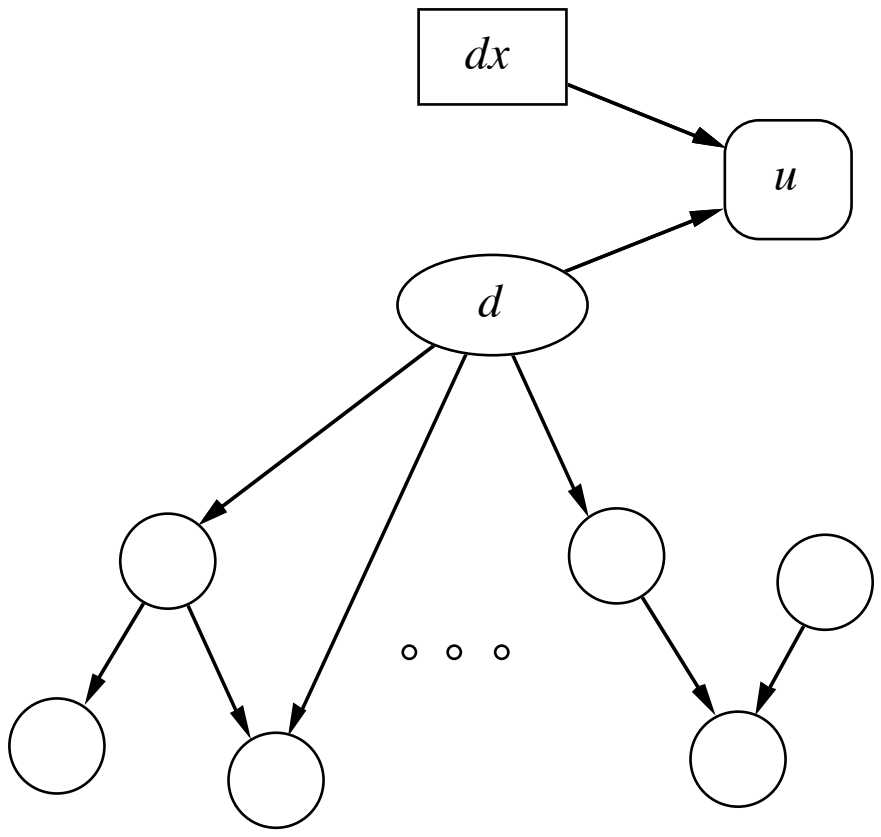

**Figure 5.2:** An influence diagram for Pathfinder IV.
The node $d$ represents a set of mutually exclusive and exhaustive diseases. The node $dx$ represents all possible diagnoses—that is, all possible statements of the form, "the patient has disease $d_i$." The node $u$ represents a patient's disutilities for all disease–diagnosis combinations. Pathfinder III uses the same utility model.

An important consideration in the assessment of diagnostic utilities is that preferences will vary from one decision maker to another. For example, the diagnostic utilities of a decision maker—the patient—faced with the results of a lymph-node biopsy are likely to be influenced by the person's age, gender, and state of health. Consequently, the inferential losses computed in this evaluation are meaningful to an individual only to the degree that the diagnostic utilities used in the evaluation match the diagnostic utilities of that individual.

For this experimental comparison, I used the utilities of the expert on the Pathfinder project. I choose the expert for two practical reasons. First, he was reasonably familiar with many of the ramifications of correct and incorrect diagnosis. Second, I had established a good working relationship with him during the construction of Pathfinder. The expert, because he is an expert, however, had biases that made his initial preferences deviate from those of a typical patient. For example, many sets of diseases of the lymph node currently have identical treatments and prognoses. Nonetheless, experts like to distinguish diseases within each of these sets, because doing so allows research in new treatments to progress. That is, experts often consider the value of their efforts to future patients. In addition, experts generally suffer professional embarrassment when their diagnoses are incorrect. Also, experts are concerned about the legal liability associated with misdiagnosis. In an effort to remove these biases, I asked the the expert to ignore specifically these attributes of utility. Further, I asked him to imagine that he himself had a particular disease, and to assess the diagnostic utilities accordingly.

Another important consideration in almost any medical decision problem is the wide range of severities associated with outcomes. For example, if a patient has a viral infection and is incorrectly diagnosed as having cat-scratch disease—a disease caused by an organism that is killed with antibiotics—the consequences are not severe. In fact, the only nonnegligible consequence is that the patient will take antibiotics unnecessarily for several weeks. If, however, a patient has Hodgkin's disease and is incorrectly diagnosed as having an insignificant benign disease such as a viral infection, the consequences are often lethal. If the diagnosis had been made correctly, the patient would have immediately undergone radio- and chemotherapy, with a 90-percent chance of a cure. If the patient is diagnosed incorrectly, however, and thus is not treated, the disease will progress. By the time its major symptoms appear and the patient once again seeks help, the cure rate with appropriate treatment will have dropped to less than 20 percent.

It is important for us to measure preferences across such a wide range, because sometimes we must balance a large chance of a small loss with a small chance of a large loss. For example, even though the probability that a patient has syphilis is small—say, 0.001—treatment with antibiotics may be appropriate, because the patient may prefer the harmful effects of antibiotics to the small chance of the harmful effects of untreated disease.

Early attempts to assess preferences for both minor and major outcomes in the same unit of measurement were fraught with paradoxes. For example, in a linear willingness-to-pay approach, a decision maker might be asked, "How much would you have to be paid in order to accept a one in ten-thousand chance of death?" If the decision maker answered, say, $1000, then the approach would dictate that he would be willing to be killed for $10 million. This inference is absurd.

Recently, Howard has constructed an approach that avoids many of the paradoxes of earlier models (Howard, 1980). Like several of its predecessors, the model determines what an individual is willing to pay to avoid a given chance of death, and what he is willing to be paid to assume a given chance of death. Also, like many of its predecessors, Howard's model shows that, for small risks of death (typically, $p < 0.001$), the amount someone is willing to pay to avoid, or is willing to be paid to to assume, such a risk is linear in $p$. That is, for small risks of death, an individual acts as would an expected-value decision maker with a finite value attached to his life. For significant risks of death, however, the model deviates strongly from linearity. For example, the model shows that there is a maximum probability of death, beyond which an individual will accept no amount of money to risk that chance of death. Most people find this result to be intuitive.[3]

In this book, the details of the model will not be presented; for a discussion of the approach see Howard (1980). Here, we need to assume only that willingness to buy or sell *small* risks of death is linear in the probability of death. Given this assumption, preferences for minor to major outcomes can be measured in a common unit, *the probability of immediate, painless death that a person is willing to accept to avoid a given outcome and to be once again healthy.* The undesirability of major outcomes can be assessed directly in these terms. For example, a decision maker might be asked, "If you have Hodgkin's disease and have been incorrectly diagnosed as having a viral infection, what probability of immediate, painless death would you be willing to accept to avoid the illness and incorrect diagnosis, and to be once again healthy?" At the other end of the spectrum, the undesirability of minor outcomes can be assessed by willingness-to-pay questions, and can be translated, via the linearity result, to the common unit of measurement. For example, a decision maker might be asked, "How much would you be willing to pay to avoid taking antibiotics for two weeks?" If he answered $100, and if his small-risk value of life were $20 million, then the answer could be translated to a utility of a 5 in 1 million chance of death.

[3]The result makes several assumptions, such as the decision maker is not suicidal and is not concerned about how his legacy will affect other people.

An important task in assessing the $u_{d_i,d_j}$ is the determination of the decision maker's small-risk value of life. Howard proposes a model by which this value can be computed from other assessments (Howard, 1980). A simple version of the model requires a decision maker to trade off the amount of resources he consumes during his lifetime and the length of his lifetime, to characterize his ability to turn present cash into future income (summarized, for example, by an interest rate), and to establish his attitude toward risk. However, our expert did not find it difficult to assess the small-risk value of life directly.[4] When asked what dollar amount he would be willing to pay to avoid chances of death ranging from 1 in 20 to 1 in 1000, he was consistent with the linear model to within a factor of 2, with a median small-risk value of life equal to $20 million.

Note that, with this utility model, the inferential losses computed for Pathfinder III and IV will have units "probability of death." In many cases, we shall see that the losses are small in these units (on the order of 0.0001). Consequently, it is useful to define a *micromort*, a one–in–1-million chance of death. In these units, for example, the expert, who has a small-risk value of life of $20 million, should be willing to buy and sell risks of death at the rate of $20 per micromort. This unit of measurement is also useful because it helps to emphasize that the linear relationship between risk of death and willingness to pay holds for only small probabilities of death. Howard (1989b) discusses in detail the use of the micromort for medical decision making.

Finally, an important consideration is the complexity of the utility-assessment procedure. There are approximately 60 diseases represented in Pathfinder. The direct measurement of the $u_{d_i,d_j}$ therefore requires about $60^2 = 3600$ assessments. Clearly, the measurement process would be tedious. Thus, several steps were taken to reduce the complexity of the task. For one, the expert established sets of diseases that have identical treatments and prognoses. The expert identified 36 such equivalence classes, reducing the number of direct utility assessments to $36^2 = 1296$. In addition, the expert and I decomposed many of the utilities into independent assessments such as the disutility of a disease when correctly treated, the disutility of delaying the appropriate treatment, and the disutility of the treatment in the absence of disease (e.g., the disutility of taking antibiotics or undergoing surgery). Through such decomposition, more than 80 percent of the direct assessments were avoided. In total, the construction of the utility model took approximately 60 hours.

[4] Howard also has observed that the small-risk value of life can be assessed directly (Howard, 1990).

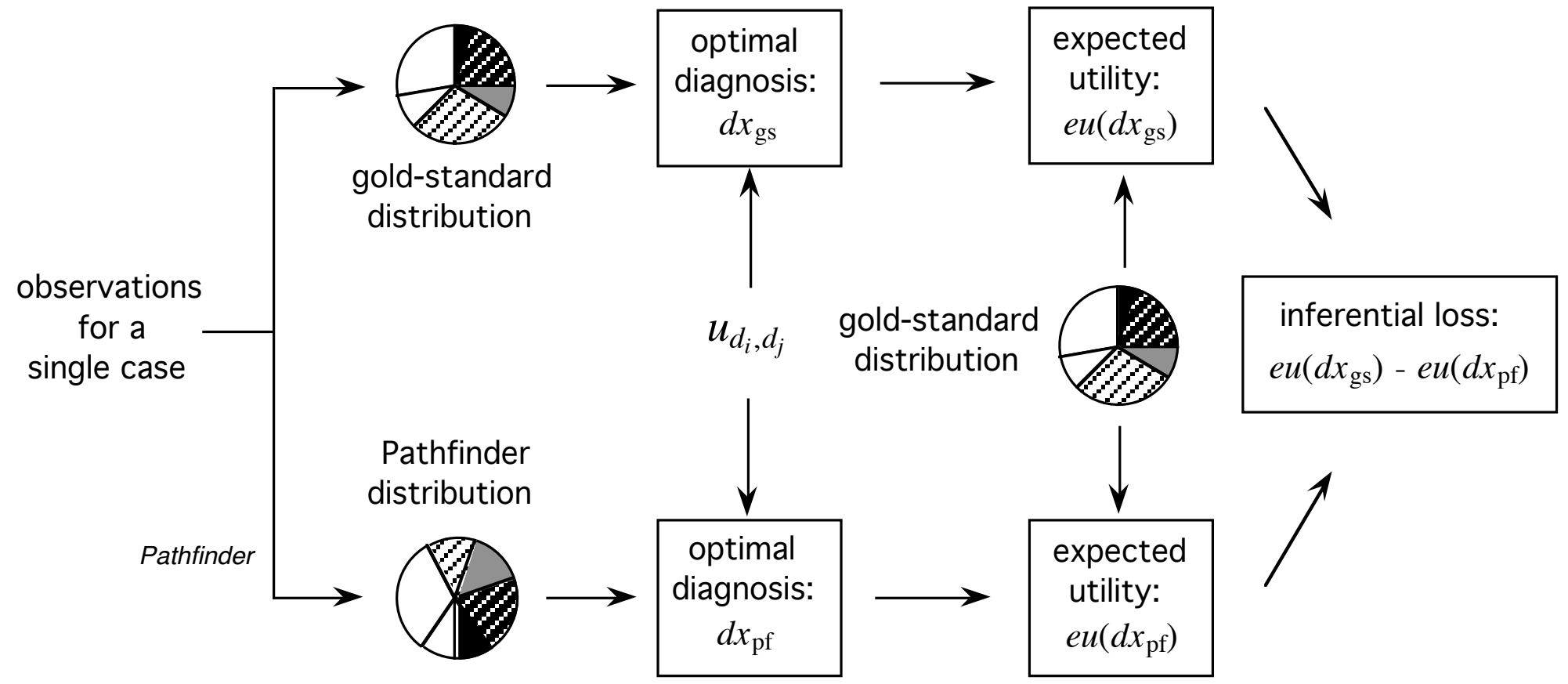

**Figure 5.3:** The computation of inferential loss.
Based on the features reported, Pathfinder (III or IV) produces a probability distribution over diseases (represented by the shaded probability wheel). Based on these same features, we also determine a gold-standard distribution. Given theses distributions, we identify the optimal diagnoses associated with them, denoted $dx_{\mathrm{pf}}$ and $dx_{\mathrm{gs}}$, using the principle of maximum expected utility. We compute the expected utility of the two diagnoses with respect to the gold-standard distribution. We then calculate the inferential loss associated with the Pathfinder distribution by subtracting the expected utility of the Pathfinder distribution from the expected utility of the gold-standard distribution.

### 5.5.3 The Computation of Inferential Loss

The procedure for computing inferential loss is identical for both versions of Pathfinder and is illustrated in Figure 5.3. First, based on the features reported for a given case, Pathfinder produces a probability distribution over diseases. Next, based on these same features, we determine a gold-standard distribution as described in Section 5.5.1.

Then, we determine the optimal diagnosis associated with a Pathfinder distribution, denoted $dx_{\mathrm{pf}}$, by identifying the diagnosis that maximizes the expected utility of the patient given that distribution. Similarly, we determine the optimal diagnosis associated with the gold-standard distribution, denoted $dx_{\mathrm{gs}}$. Formally, we compute

$$dx_{\mathrm{pf}} = \mathrm{argmax}_{d_j} \left[ \sum_{d_i} p_{\mathrm{pf}}(d_i)\ u_{d_i,d_j} \right]$$

$$dx_{\mathrm{gs}} = \mathrm{argmax}_{d_j} \left[ \sum_{d_i} p_{\mathrm{gs}}(d_i)\ u_{d_i,d_j} \right]$$

where $p_{\mathrm{pf}}(d_i)$ and $p_{\mathrm{gs}}(d_i)$ represent the probability of the $i$th disease under the Pathfinder and gold-standard probability distributions, respectively.

Next, we compute the expected utility of $dx_{\mathrm{pf}}$ and $dx_{\mathrm{gs}}$, denoted $eu(dx_{\mathrm{pf}})$ and $eu(dx_{\mathrm{gs}})$, respectively. When computing expected utility, we use the gold-standard distribution, which reflects the assumed best distribution. That is, we compute

$$eu(dx_{\mathrm{pf}}) = \sum_{d_i} p_{\mathrm{gs}}(d_i)\ u_{d_i,dx_{\mathrm{pf}}}$$

$$eu(dx_{\mathrm{gs}}) = \sum_{d_i} p_{\mathrm{gs}}(d_i)\ u_{d_i,dx_{\mathrm{gs}}}$$

Finally, we determine inferential loss, denoted IL, for the Pathfinder distribution, by subtracting the expected utility of the Pathfinder diagnosis from the expected utility of gold-standard diagnosis. That is,

$$\mathrm{IL} = eu(dx_{\mathrm{gs}}) - eu(dx_{\mathrm{pf}})$$

By construction, IL is always a nonnegative quantity. If both a Pathfinder distribution and the gold-standard distribution imply the same diagnosis, then the inferential loss for that Pathfinder distribution is zero, a perfect score. Note that the units of inferential loss are the same as those for the diagnostic utilities $u_{d_i,d_j}$—namely, micromorts.

### 5.5.4 Results

The mean and standard deviation of inferential loss for the two versions of Pathfinder are shown in Table 5.3. The case-by-case results are shown in Appendix D. Unlike the difference of 0.95 produced by the expert-rating metric, these results clearly reflect the increase in value of Pathfinder IV as a result of this system's superior diagnostic accuracy. Specifically, assuming that a patient is willing to convert micromorts to dollars at a rate of \$20 per micromort[5], as our expert was, the results in this metric show that it is worth approximately \$6000 *per case* to the patient to have the more sophisticated Pathfinder knowledge be used instead of the earlier knowledge base that assumed global independence among features. As was mentioned earlier, it took approximately 40 hours longer to construct Pathfinder IV than it did to construct Pathfinder III. Thus, assuming a combined hourly rate of \$400 for the expert and myself, the additional effort would more than pay for itself after only three cases had been run. If we include the time required to construct the utility model (60 hours), then the additional effort would more than pay for itself after seven cases had been run.

[5]The value of \$20 per micromort applies to the expert when he is healthy. We use this value to approximate his small-risk value of life in situations where he is ill.

**Table 5.3:** Inferential losses for Pathfinder III and IV.

| Knowledge Base | Inferential Loss (micromorts) mean | sd |
|---|---|---|
| Pathfinder III | 340 | 1684 |
| Pathfinder IV | 16 | 104 |

The standard deviations for inferential loss are quite large relative to the means. The reason for such large variances is easily appreciated. For many of the cases, the optimal diagnosis associated with the distributions produced by both versions of Pathfinder are identical to the optimal diagnosis associated with the gold standard. In particular, the optimal diagnoses for Pathfinder III agreed with the gold-standard diagnoses in 47 of the 53 cases; those for Pathfinder IV agreed in 50 of the 53 cases. In these cases, inferential loss is zero. In the remaining cases, the approaches determine a diagnosis that differs from the gold standard. Most of these nonoptimal diagnoses are associated with expected utilities that are significantly lower than is the expected utility associated with the gold-standard diagnosis. Thus, inferential losses fluctuate from zero in most cases to large values in the remainder.

Despite the large standard deviations for inferential loss, a Bootstrap permutation test suggests that the results are not due only to chance. In particular, the test yields an ASL of 0.08. Again, this means that there is only an 8 percent chance that the difference in diagnostic accuracy would be more extreme than what the current results show, if inferential losses were drawn at random for the set union of inferential losses for Pathfinder III and IV.

## 5.6 Discussion

All phases of this experiment provided useful results. The expert-rating approach was easy to implement, and it showed that there were differences in diagnostic accuracy between the two versions of Pathfinder. Thereby, it suggested that pursuing each of the second two phases would be worthwhile. The second phase showed that several factors were responsible for the superior performance of Pathfinder IV, the most frequent being that improvements in probability assessments resulted from conditioning the assessments on other events. Finally, the third phase showed that the additional work required to construct Pathfinder IV was well worth the effort.

The expert-rating metric used in phase 1 was more sensitive to differences in the two systems than was the decision-theoretic metric used in phase 3. This observation is not surprising because experts, who have their integrity at stake, tend to be hypersensitive to errors in diagnosis, regardless of the degree to which such errors matter to the patient. Of course, the decision-theoretic metric can be modified to be more sensitive. Considerations of integrity or liability, for example, can be incorporated into the diagnostic utilities. Indeed, the fact that components of preference can be made explicit and are under the direct control of the expert is one advantage of the decision-theoretic approach.

This evaluation has concentrated on an analysis of differences in diagnostic accuracy between Pathfinder III and IV that arise from differences in the probabilistic knowledge represented within these systems. This focus is important for evaluating the usefulness of similarity networks and partitions. Nonetheless, we can use the experimental design described in this chapter to investigate other facets of Pathfinder performance. For example, as I mentioned, I used this methodology to measure the decrease in diagnostic accuracy that aries from the misrecognition and lack of recognition of morphologic features by a nonexpert (Heckerman et al., 1990). Also, using this approach, the Pathfinder group plans to compare the diagnostic accuracy of community pathologists who have access to Pathfinder to that of pathologists who do not have such access. In addition, the dependencies among lymphoid cells and clusters of clear cells make the Pathfinder inference algorithm sluggish (see Section 4.5). Of course, there is a much room for improvement in the algorithm. Nonetheless, we can use the decision-theoretic metric to determine the negative value of ignoring these dependencies altogether, and thereby trade off the value of representing the dependencies with the cost of improving the algorithm. In yet another study, we can measure the sensitivity of diagnostic accuracy to changes in the joint probability distribution of Pathfinder IV. In general, we can use this approach to evaluate, in clear terms, a wide variety of issues related to the building of real-world expert systems.

# 6 Conclusions and Future Work

In this chapter, we consider possible extensions to the similarity-network and partition representations. We then examine conclusions that we can draw from the work presented in this book.

## 6.1 Weaker Conditions for Soundness

In Chapter 4, we discussed the conditions that are sufficient to guarantee the soundness of the global-knowledge-map construction. There, we examined the effect of each condition on the construction of the Pathfinder knowledge map, and outlined briefly the work we require to relax the positivity condition. In this section, we examine approaches for relaxing several of the other soundness conditions.

### 6.1.1 Local Knowledge Maps for More Than Two Diseases

Similarity-network theory and implementation, in their current form, require that we compose local knowledge maps only for pairs of hypotheses. This restriction lessened the usefulness of the representation for building the Pathfinder knowledge map. For example, Figure 6.1(a) contains a portion of Pathfinder's similarity graph where the diseases AIDS EARLY, RHEUMATOID ARTHRITIS, and GLH PLASMA CELL TYPE form a clique. Given this representation, the expert had to assess a local knowledge map for each of the three edges in the graph. The expert, however, preferred to assess one local knowledge map for the disease triplet.

In general, we can extend the similarity-network representation to include local knowledge maps for hypothesis sets of arbitrary size. In such an extension, we replace the similarity graph with a similarity hypergraph. A *hypergraph* consists of nodes and *hyperedges* that connect sets of nodes. We then compose one local knowledge map for each hyperedge. For example, we can replace the graph in Figure 6.1(a) with the hypergraph in Figure 6.1(b). In this hypergraph, the small oval and the three lines represent a hyperedge that connects the disease triplet. Given this hypergraph, the Pathfinder expert needs to compose only one local knowledge map.

To ensure that the global knowledge map constructed from such a network is sound, we must replace only the constraint that the similarity graph be connected, using instead the constraint that the similarity hypergraph be connected. We can generalize the proof of soundness in Chapter 3 in a straightforward fashion to demonstrate this observation.

### 6.1.2 Distinguished Node with Predecessors

One of the sufficient conditions for soundness is that the distinguished node can have no predecessors. We can eliminate this condition using the following steps. First, we build a similarity network for only those nodes that are not (direct or indirect) predecessors of the distinguished node. Second, we construct the global knowledge map from this similarity network. Third, we compose the remainder of the global knowledge map directly. This procedure will be useful in those domains where only a few nodes are predecessors of the distinguished node. In more complicated situations, we must look for extensions to the theory.

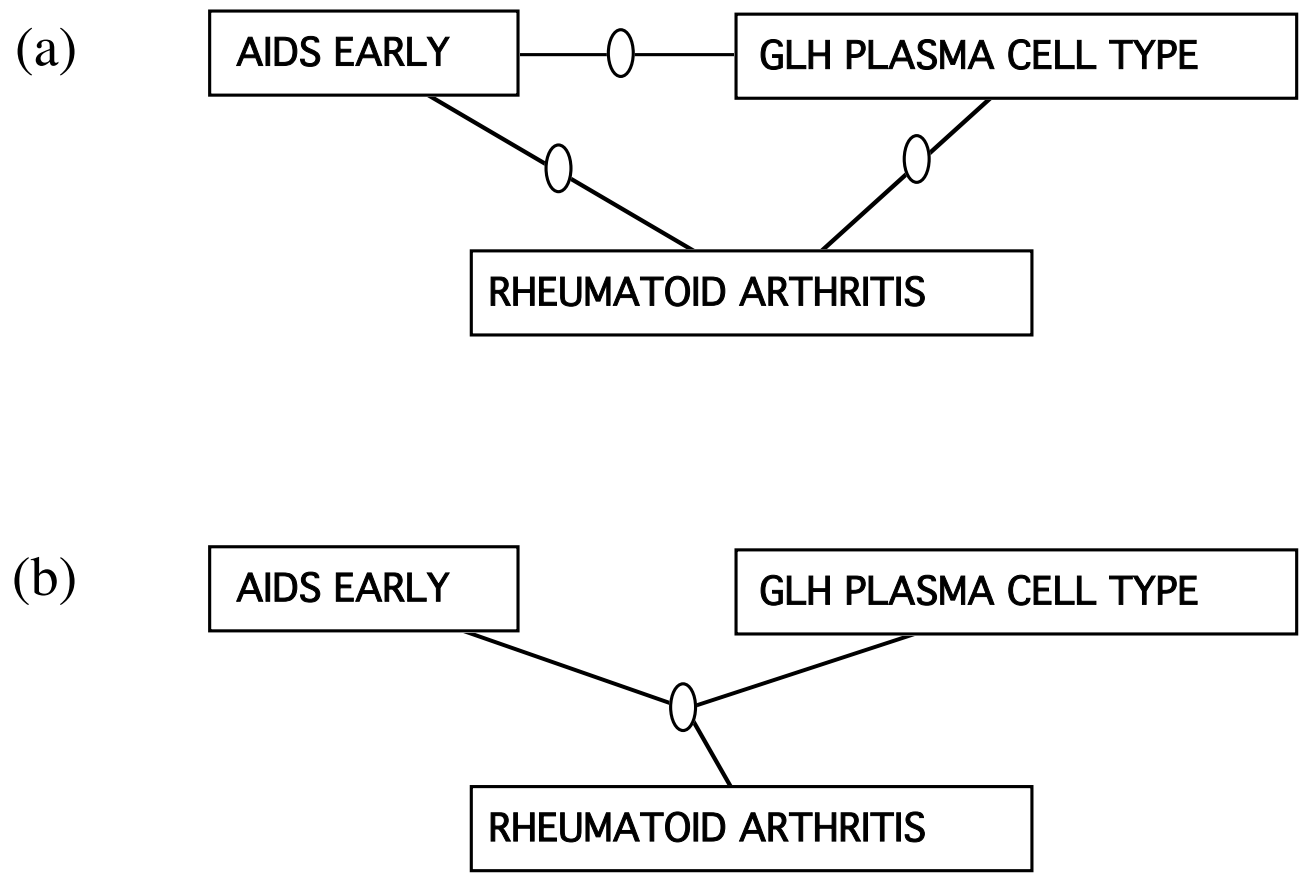

**Figure 6.1:** A similarity graph and its corresponding similarity hypergraph. (a) A portion of the Pathfinder similarity graph (see the upper-right corner of Figure 4.2 on page 109). The graph is associated with three local knowledge maps. (b) A similarity hypergraph that corresponds to the graph in (a). The similarity hypergraph is associated with one local knowledge map for the disease triplet.

### 6.1.3 Multiple Hypotheses

Another sufficient condition for soundness is that the hypotheses in a similarity network must be mutually exclusive. In many domains, however, hypotheses are not mutually exclusive. Patients admitted to the internal-medicine ward of a hospital, for example, often present with four to five coexisting diseases. In this section, we examine how we can use the similarity-network and partition representations to facilitate the construction of knowledge maps for the diagnosis of multiple hypotheses.

Figure 6.2 contains a small portion of a knowledge map for internal medicine. In the examples we have examined previously, we have represented diseases as instances of a single variable, under the assumption that these diseases are mutually exclusive. In Figure 6.2, however, the node APPI represents the absence or presence of unruptured acute appendicitis. Similarly, the node RUPTURED ECTOPIC represents the absence or presence of acute ruptured ectopic pregnancy. Thus, this knowledge map does not exclude the possibility that both diseases can manifest in the same patient.

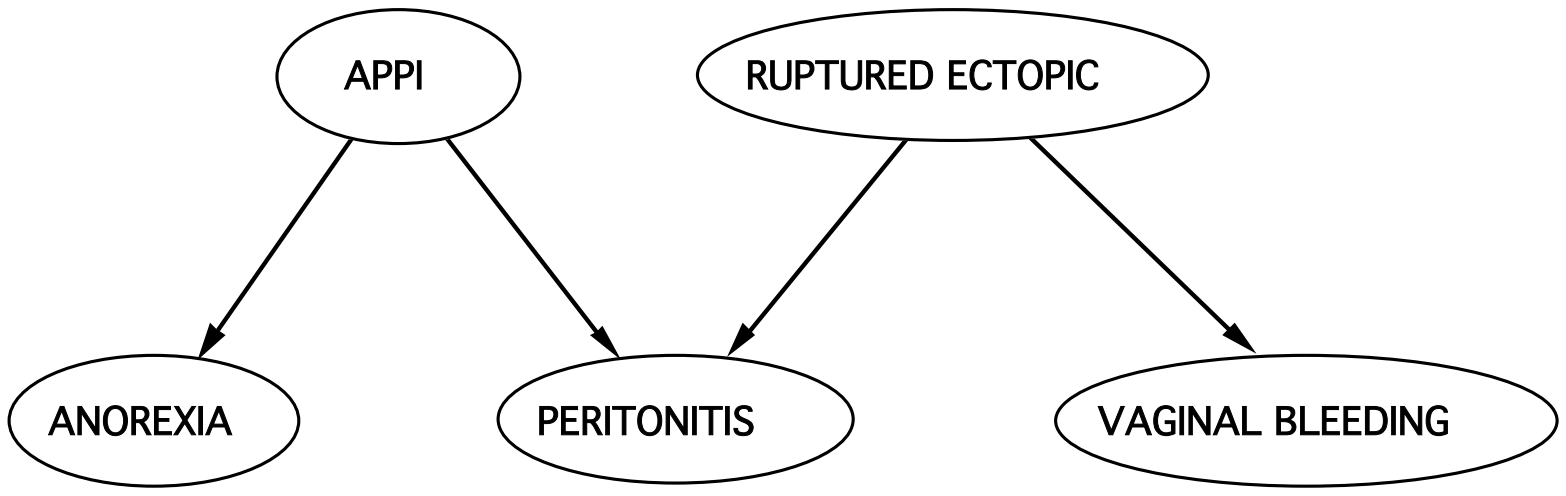

**Figure 6.2:** A knowledge map for the diagnosis of multiple diseases.
The nodes APPI and RUPTURED ECTOPIC represent the disorders unruptured acute appendicitis and acute ruptured ectopic pregnancy, respectively. Each disease may be absent or present. The nodes ANOREXIA, PERITONITIS, and VAGINAL BLEEDING represent patient findings relevant to the diagnosis of these diseases.

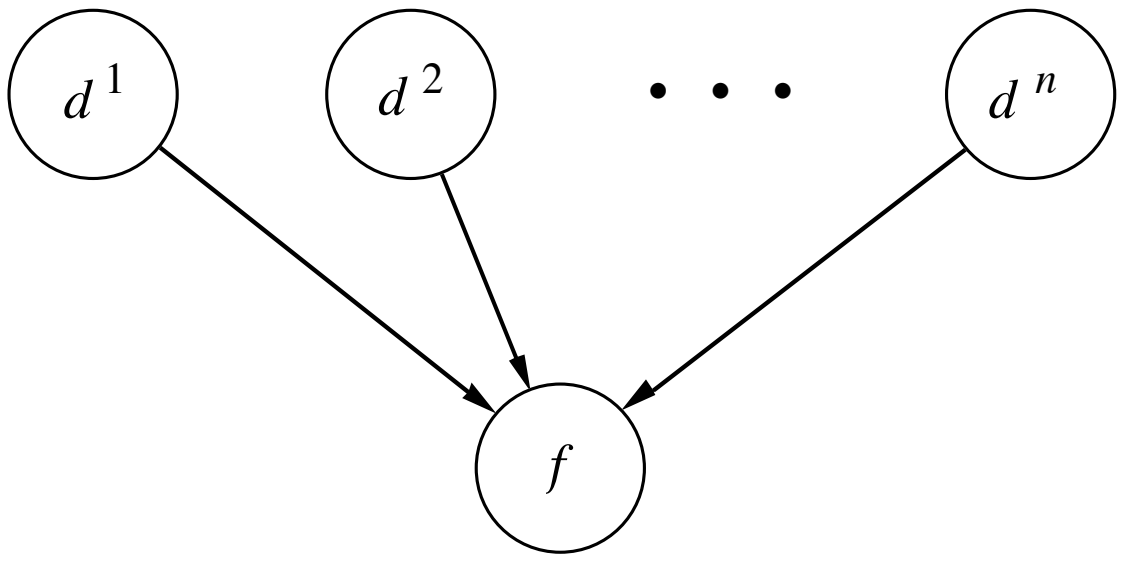

**Figure 6.3:** Multiple causes of the same finding.
Each disease $d^i$ may cause the finding $f$ to be present. If the diseases and the finding are binary, then we require $2^n$ probability assessments to quantify the interaction.

This example illustrates a difficulty that arises typically in situations where multiple hypotheses are possible. In particular, both diseases in Figure 6.2 condition the node PERITONITIS, which represents the absence or presence of an inflammatory response in the peritoneum (the lining of the abdominal cavity). Thus, without any additional information, we would have to assess four probability distributions for this finding. More generally, we can have the situation, illustrated in Figure 6.3, where diseases $d^1, d^2, \ldots d^n$ can each cause finding $f$ to appear. Here, the node $f$ is associated with $2^n$ probability distributions.

We can reduce dramatically the number of probability assessments for node $f$ by making an additional assertion of conditional independence, called *causal independence*. In the context of Figure 6.3, let $p^i$ denote the probability that a patient, initially without disease $d^i$ and without finding $f$, will develop finding $f$ when getting disease $d^i$. When we assert causal independence in this situation, we state that probability $p^i$ does not depend on whether or not the patient has any other diseases before he has $d^i$, and that the finding $f$ cannot disappear when the disease $d^i$ manifests in the patient.

Now let $D_a$ denote an arbitrary instance of the set of variables $D = \{d^1, d^2, \ldots, d^n\}$. That is, let $D_a$ denote some assignment of absent or present to each disease $d^i$. In addition, let $D_-$ denote the particular instance of $D$ where all diseases are absent. Given the assertion of causal independence, the finding $f$ will be absent in a patient only if two conditions are met: (1) the finding $f$ cannot be present in the patient initially, and (2) none of the patient's diseases can act to cause $f$ to appear. Thus, we obtain

$$p\left(f_-|D_a, \xi\right) = p\left(f_-|D_-, \xi\right) \prod_{i \in \mathcal{I}_a} \left[1 - p^i\right] \tag{6.1.1}$$

where $\mathcal{I}_a$ is the set of indices $i$ such that $d^i$ is present in $D_a$. Applying the sum rule to Equation 6.1.1, we obtain

$$p\left(f_+|D_a, \xi\right) = 1 \; - \; \left[1 - p\left(f_+|D_-, \xi\right)\right] \prod_{i \in \mathcal{I}_a} \left[1 - p^i\right] \tag{6.1.2}$$

If the patient has only disease $d^i$, Equation 6.1.2 becomes

$$p\left(f_+|\text{only } d^i_+, \xi\right) = 1 \; - \; \left[1 - p\left(f_+|D_-, \xi\right)\right] \left[1 - p^i\right] \tag{6.1.3}$$

Solving for $p^i$ in Equation 6.1.3, and substituting the result in Equation 6.1.2, we obtain

$$p\left(f_+|D_a, \xi\right) = 1 - \left[1 - p\left(f_+|D_-, \xi\right)\right] \prod_{i \in \mathcal{I}_a} \left[\frac{1 - p\left(f_+|\text{only } d^i_+, \xi\right)}{1 - p\left(f_+|D_-, \xi\right)}\right] \tag{6.1.4}$$

Thus, with the assertion of causal independence, we can determine all the probability distributions associated with the node $f$ in Figure 6.3, from only the probabilities $p\left(f_+|D_-, \xi\right)$ and $p\left(f_+|\text{only } d^i_+, \xi\right)$, $i = 1, 2, \ldots, n$.

Good and other theorists have described various forms of the causal-independence assertion (Good, 1961a; Suppes, 1970; Pearl, 1988). Pearl refers to the particular form we have discussed, where diseases and findings are binary, as a *noisy* OR*-gate* (Pearl, 1988). (We consider the origin of this name in the following paragraph.) Several researchers have noted that we can apply the noisy OR-gate and more general forms of causal independence to numerous situations within domains ranging from medicine to motorcycle repair (Habbema, 1976; Heckerman, 1987; Henrion and Cooley, 1987; Henrion, 1987).

The model of the noisy OR-gate, as we have examined it so far, makes reference to the appearance of diseases over time. We can also represent the model in an influence diagram, without this temporal reference, as illustrated in Figure 6.4. In the figure, the node labeled $d^i$–causes–$f$ represents the absence or presence of an intermediate event through which $d^i$ causes finding $f$ to be present with certainty. As is indicated by the label OR above the deterministic node $f$, if any of these intermediate events occur, then the finding $f$ will appear for certain (hence the name noisy OR-gate). The arc from $d^i$ to $d^i$–causes–$f$ reflects the assertion that the absence or presence of $d^i$ influences the probability distribution for the variable $d^i$–causes–$f$.[1] In particular, we assume that, if $d^i$ is absent, then the disease cannot act to cause $f$, whereas if $d^i$ is present, then it causes $f$ to be present with some probability greater than 0. This probability corresponds to $p^i$ in the temporal formulation of the model. The lack of arcs between nodes in the upper two rows of the knowledge map reflects the assertion of causal independence. In particular, the missing arcs represent the statement that the probability distribution for the variable $d^i$–causes–$f$ depends neither on the absence or presence of any other disease nor on the absence or presence of any other event leading to the occurrence of $f$. We require the node $d^0$–causes–$f$ to capture the possibility that finding $f$ will appear when all diseases are absent.

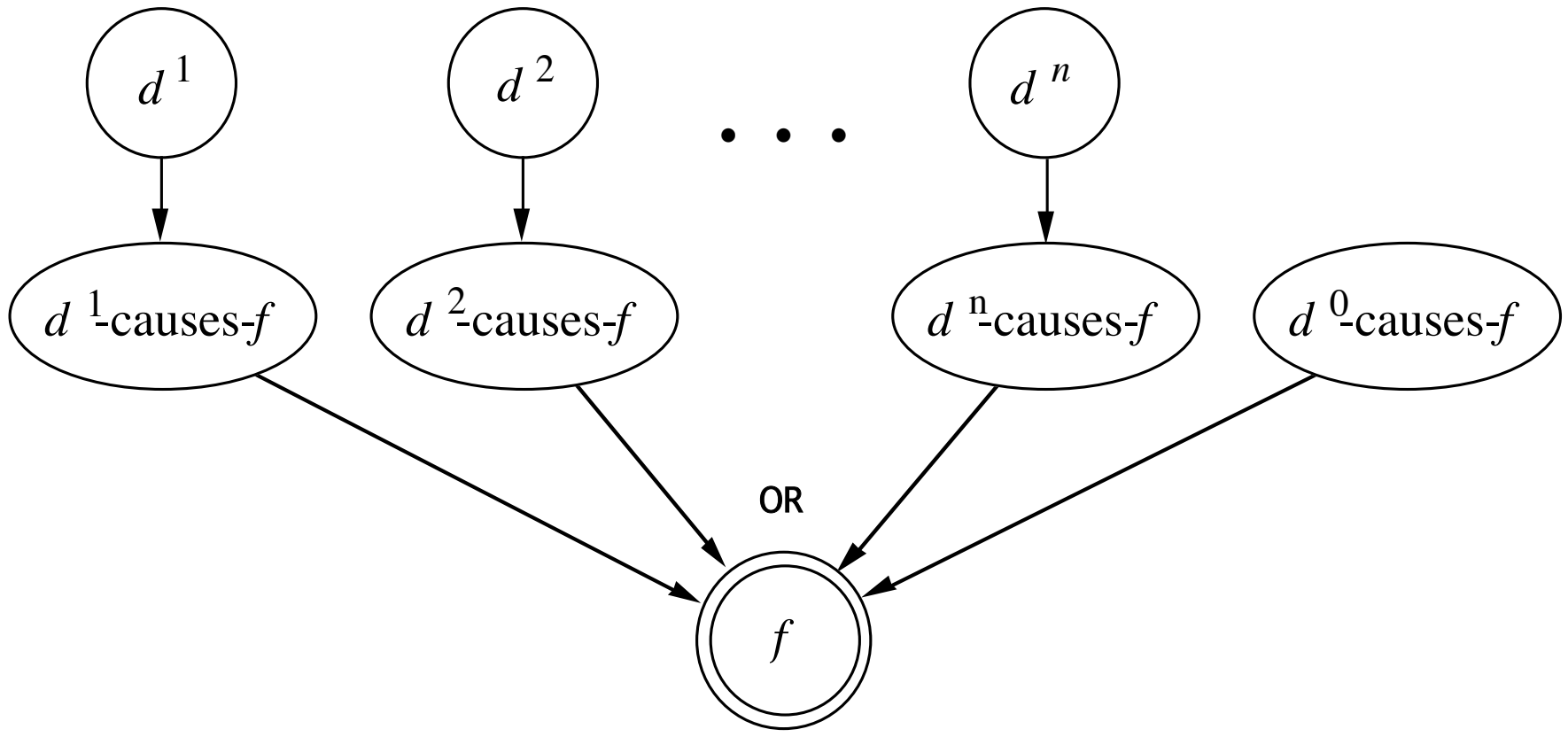

**Figure 6.4:** An atemporal representation of causal independence.
The node $d^i$ represents the absence or presence of disease $d^i$ in a given patient. The node $d^i$–causes–$f$ represents the absence or presence of an intermediate event through which $d^i$ causes finding $f$ to be present with certainty. The deterministic node $f$ is the disjunction of its parents. Thus, if any of these intermediate events occur, then the finding $f$ will appear in the patient for certain. The lack of arcs between nodes in the upper two rows of the influence diagram reflects an assertion of causal independence.

To make this model more concrete, let us consider the simple medical example in Figure 6.2. In this example, acute ruptured ectopic pregnancy can cause peritonitis, because blood from the rupture of a fallopian tube can collect in the peritoneal cavity, and thereby irritate the peritoneum. In contrast, the presence of unruptured acute appendicitis is associated with the release of substances that mediate the inflammatory response within the appendix. These substances can leak out of the appendix, and thereby cause an inflammatory response in the nearby peritoneum. Thus, the variable RUPTURED ECTOPIC–causes–PERITONITIS refers to the absence or presence of blood in the peritoneal cavity, whereas the variable APPI–causes–PERITONITIS refers to the absence or presence of inflammatory triggers of appendiceal origin in the peritoneum. To a good approximation, the probability that blood will collect in the peritoneal cavity is influenced neither by the presence of an unruptured acute appendicitis nor by the presence of inflammatory triggers of appendiceal origin in the peritoneum. Conversely, the probability that inflammatory triggers from the appendix will reach the peritoneum is influenced neither by the presence of an acute ruptured ectopic pregnancy nor by the presence of blood in the peritoneal cavity. Thus, we can assert causal independence for the interaction among these variables.

[1]Here, we assume that the knowledge map is minimal.

We can derive Equation 6.1.4 from both the temporal and atemporal models for causal independence. The atemporal model is somewhat problematic, because we often cannot define events of the form $d^i$–causes–$f$ precisely. Nonetheless, most people find this model easy to understand. In addition, we can use the framework to extend causal independence to situations where diseases and findings are not binary (Heckerman, 1987; Henrion, 1987).

Now let us examine how we can use assumptions of causal independence in conjunction with an assessed similarity network to construct a knowledge map for the diagnosis of multiple diseases (or hypotheses). The construction derives from Equation 6.1.4, which states that the only probability assessments we need to define the interaction illustrated in Figure 6.3 are those probabilities of the form $p\left(f_+|\text{only } d^i_+, \xi\right)$, $i = 1, 2, \ldots, n$ and the probability assessment $p\left(f_+|D_-, \xi\right)$. These probabilities are exactly those assessments that we can derive from a similarity network where we represent each disease as an instance of the distinguished node, and where we include the hypothesis NORMAL to represent the instance $D_-$.

With this observation in mind, let us consider the similarity network shown in Figure 6.5. From this similarity network, we can construct the multiple-disease knowledge map shown in Figure 6.2, in the following steps. First, we construct the global knowledge map from the similarity network, and transfer the findings ANOREXIA, PERITONITIS, and VAGINAL BLEEDING in the global knowledge map to the multiple-disease knowledge map. Also, if there were any arcs between these findings, we would transfer those arcs to the multiple-disease knowledge map. Second, for each node in the similarity graph, except NORMAL, we construct a binary chance node in the multiple-disease knowledge map. In particular, we construct the binary nodes APPI and RUPTURED ECTOPIC. Third, in the multiple-disease knowledge map, we draw an arc from APPI to ANOREXIA, and from APPI to PERITONITIS. Conversely, we do not draw an arc from APPI to VAGINAL BLEEDING. We can omit this arc because the local knowledge map for APPI and NORMAL states that the probability distribution for VAGINAL BLEEDING given APPI is equal to the distribution for VAGINAL BLEEDING given NORMAL, and because we assert causal independence. Similarly, we draw arcs from RUPTURED ECTOPIC to PERITONITIS and to VAGINAL BLEEDING, but we do not draw an arc from RUPTURED ECTOPIC to ANOREXIA. Fourth, we use the probability assessments associated with the similarity network in conjunction with the noisy-OR-gate model (Equation 6.1.4) to compute the probability distributions for each finding. Finally, we assert that APPI and RUPTURED ECTOPIC are marginally independent, and assess the prior probabilities for these variables.

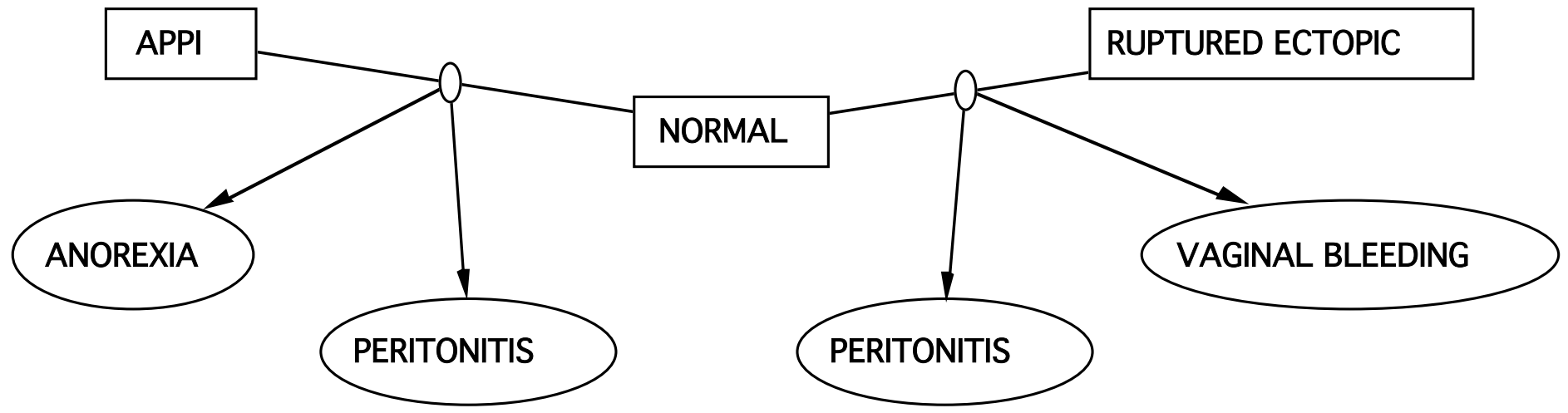

**Figure 6.5:** A similarity network for APPI and RUPTURED ECTOPIC.
The similarity network contains a local knowledge map for APPI and NORMAL and a local knowledge map for RUPTURED ECTOPIC and NORMAL. The former knowledge map contains the findings ANOREXIA and PERITONITIS, whereas the latter knowledge map contains the findings PERITONITIS and VAGINAL BLEEDING. The small ovals from which the arcs emanate represent the distinguished node in the local knowledge maps. (See Chapter 3, page 67, for a detailed description of this graphical shorthand for a similarity network.) In composing this network, we assume that APPI, NORMAL, and RUPTURED ECTOPIC are mutually exclusive hypotheses. From this similarity network and additional assertions of conditional independence, including assertions of causal independence, we can construct the multiple-disease knowledge map shown in Figure 6.2.

In transforming the similarity network to a multiple-disease knowledge map, we added several assertions of conditional independence. In particular, the similarity network implies only that the findings are conditionally independent given NORMAL, APPI alone, and RUPTURED ECTOPIC alone. The multiple-disease knowledge map, however, also encodes the assertion that the findings are independent given that both APPI and RUPTURED ECTOPIC are present in a patient. In general, when we apply the transformation described in the previous paragraph, we must verify that these additional assertions hold.

Also, in transforming the similarity network to a multiple-disease knowledge map, we used the fact that NORMAL was connected to each of the remaining hypotheses in the similarity graph. That is, we used the fact that the similarity graph had a *star topology*, with NORMAL as its center. To understand this observation, let us consider the similarity network in Figure 6.6. Here, APPI and NORMAL are not connected, and thus we cannot identify those findings that are conditioned by the chance node APPI in the multiple-disease knowledge map. Of course, we could add an arc from APPI to every finding in the multiple-disease knowledge map, but, in so doing, we would loose the assertions of conditional independence implied by the absence of the arc from APPI to ANOREXIA.

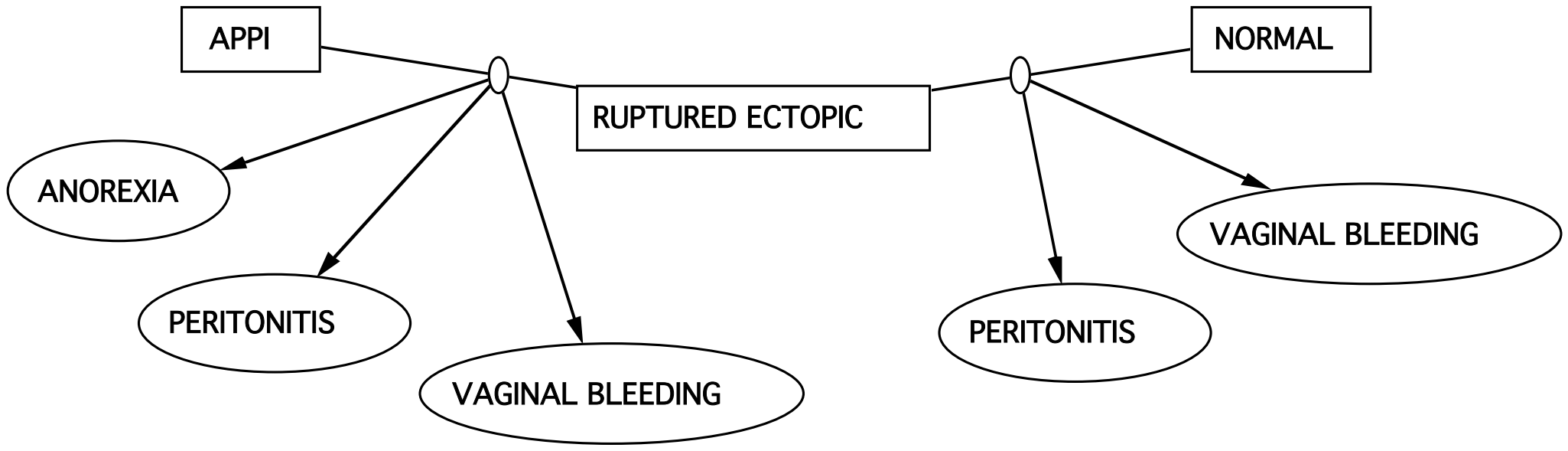

**Figure 6.6:** An alternative similarity network for APPI and RUPTURED ECTOPIC. In this similarity network, there is no local knowledge map for the hypothesis pair APPI and NORMAL, and thus we cannot determine that APPI (when considered a binary variable) and VAGINAL BLEEDING are conditionally independent. Consequently, we cannot construct the knowledge map shown in Figure 6.2 from this similarity network.

Although the transformation is facilitated by a similarity graph with a star topology, we should not require an expert to compose such graphs. Indeed, an expert might not be able to compose a local knowledge map for distinguishing a particular disease from NORMAL. Fortunately, however, we can transform any similarity network to one whose similarity graph has a star topology. Specifically, given any similarity network, we first construct and assess the global knowledge map associated with that similarity network. Then, for each hypothesis in the similarity graph (other than NORMAL), we construct a local knowledge map for discriminating that hypothesis with NORMAL, using the probability distributions from the global knowledge map, and any ordering over the nondistinguished variables that is consistent with the global knowledge map (see Theorem 3.3). Once we obtain this new similarity network, we can construct the multiple-hypothesis knowledge map from that similarity network as described previously.

In this section, we have outlined only one possible transformation procedure in the context of a simple example. We require a general transformation algorithm and a proof that the algorithm is sound and exhaustive under certain conditions. In Appendix E, we consider a candidate for such an algorithm.

## 6.2 Applications Other Than Coherent Knowledge Acquisition

In this section, we examine several possible uses of the similarity-network and partition representations other than the coherent construction and assessment of a knowledge map.

### 6.2.1 Inference and Explanation

We can use the asymmetric assertions of conditional independence encoded in a similarity network to simplify and sometimes to avoid inference and value-of-clairvoyance computations. For example, consider the local knowledge map for L&H DIFFUSE HD and MIXED CELLULARITY in Figure 4.3 on page 110. For a given patient case, if we acquired enough evidence to rule out all diseases except the two diseases in this local knowledge map, then the subsequent observation of any feature other than L&H SR, MUMMY, MONONUCLEAR SR, and CLASSIC SR would not change the probabilities of the two diseases. In addition, the value of clairvoyance for all features not in this local knowledge map must be 0. Consequently, in this situation, we can avoid potentially time-consuming inference and value-of-clairvoyance computations.[2] We can generalize this procedure for simplifying computations to cases where more than two hypotheses are possible.

We can also use a similarity network to explain how the observations for a given case affect the probability distribution over hypotheses, and to justify the recommendations for evidence gathering that the system provides to the user. For example, let us examine the current facility in Pathfinder for justifying evidence-gathering recommendations. In Chapter 1, we saw a sample dialog between Pathfinder and a user of the system. In that dialog, the user entered the feature–instance pairs F % AREA: >90%, F DENSITY: BACK-TO-BACK, and F POLARITY: PROMINENT, and then asked the system to identify features that were cost-effective for narrowing the differential diagnosis (see Figure 1.7 on page 11). A justification for one of Pathfinder's recommendations—MONOCYTOID CELLS—is illustrated in Figure 6.7.

The dollar amount at the bottom of the window reflects the monetary cost associated with observing the feature. In general, the user can elect to display several components of cost, including an estimate of the time it takes to observe a feature, and the degree of tedium associated with such a task.

The graph in the middle of the window reflects the benefits associated with observing the feature. In general, for each instance $f_i$ of the feature $f$ being justified, the system graphs (on a log scale) the quantity

[2] Other researchers are also investigating methods that exploit asymmetries for inference (Smith, 1990).

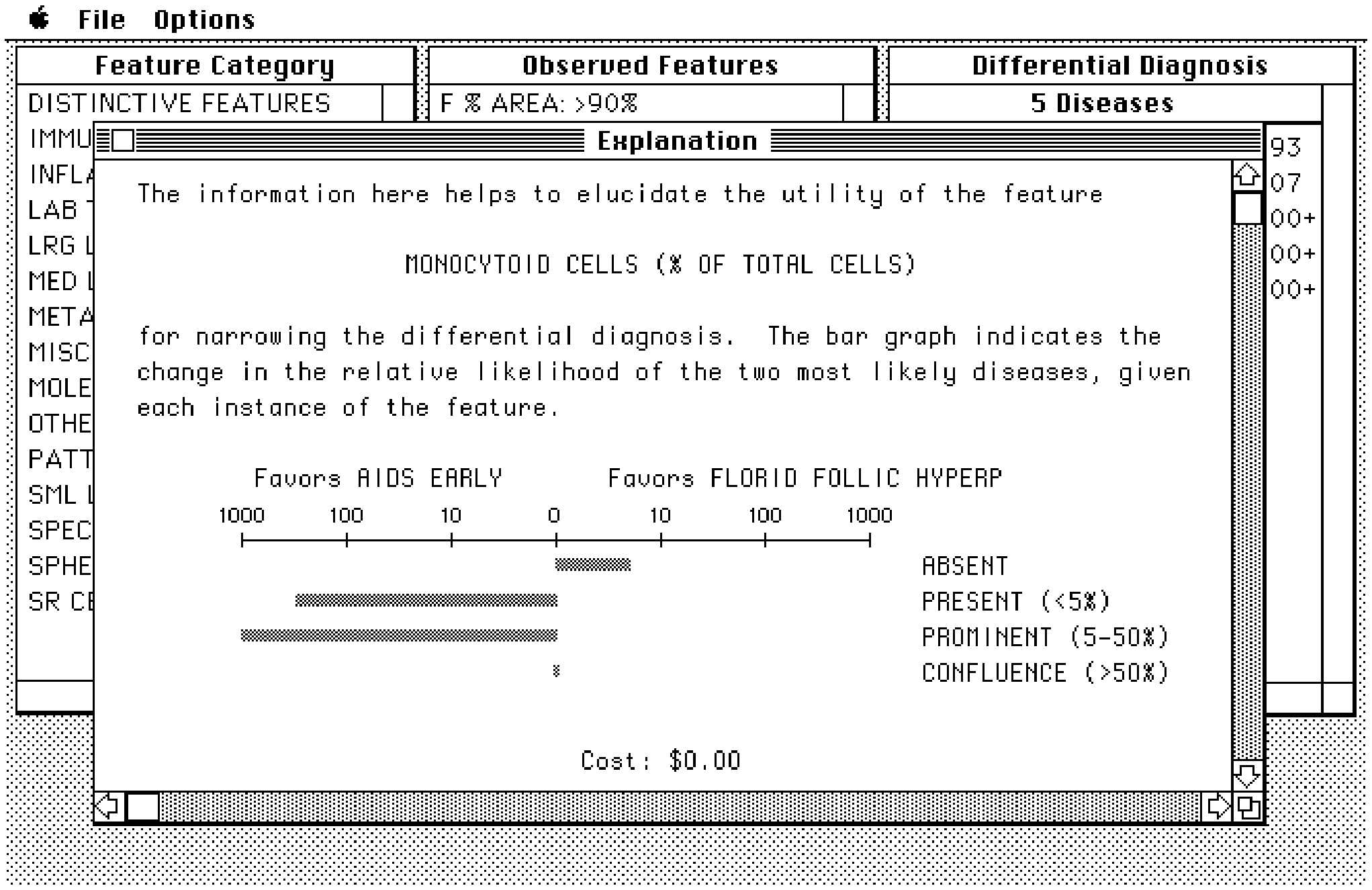

**Figure 6.7:** A justification for the recommendation of MONOCYTOID CELLS.
For each instance of the feature, the length and direction of a bar reflects the change in the probability of AIDS EARLY relative to the change in the probability of FLORID FOLLIC HYPERP, given the observation of that feature–instance pair. The justification also includes the monetary cost of observing the feature.

$$\frac{p\,(f_i|d_1,\xi)}{p\,(f_i|d_2,\xi)}$$

where $d_1$ and $d_2$ are the most probable and second most probable diseases, respectively, and where $\xi$ includes those features that the user has already observed. This quantity is called a *likelihood ratio*. The logarithm of this quantity is known as the *weight of evidence* for $f_i$ in favor of $d_1$ relative to that of $d_2$, given $\xi$ (Good, 1950). The likelihood ratio reflects the degree to which the probability of $d_1$ changes relative to $d_2$, when we observe $f_i$. Thus, according to the graph in Figure 6.7, if a user of the system observes that MONOCYTOID CELLS are ABSENT, then the probability of FLORID FOLLIC HYPER will increase relative to the probability of AIDS EARLY by almost a factor of 10. In contrast, if the user observes that MONOCYTOID CELLS are PRESENT, then the probability of AIDS EARLY will increase relative to FLORID FOLLIC HYPERP by more than a factor of 100.

The graph in Figure 6.7 gives an indication of whether or not MONOCYTOID CELLS is useful for narrowing the differential diagnosis. It does not however, describe the effect of observations on all the diseases. To overcome this drawback of the approach, the system can partition the diseases on the differential diagnosis into two sets. The system can then generate justifications like the one illustrated in Figure 6.7, replacing single diseases with disease groups. The Pathfinder research group has implemented this procedure (Heckerman et al., 1985; Heckerman et al., 1990). In another approach, we could make available the Pathfinder similarity graph to a user, who can then ask the system to generate justifications of the form shown in Figure 6.7 for one or more disease pairs defined by the similarity graph. Alternatively, the user may wish to see justifications generated based on a similarity graph of his own composition.

### 6.2.2 Heuristic Applications

For extremely large and complex domains, the similarity-network and partition representations might play a useful heuristic role in knowledge acquisition. In particular, when composing a similarity network, an expert may wish to include in a local knowledge map only those nondistinguished variables that are strong discriminators of the hypotheses associated with that local knowledge map. Similarly, when composing a partition for a given nondistinguished variable $x$, an expert may wish to allow two hypotheses $h_1, h_2$ to remain in the same set, even though $x$ is relevant to $\{h_1, h_2\}$, for several reasons. For example, $p\,(x|h_1,\xi)$ might be approximately equal to $p\,(x|h_2,\xi)$, or the disutilities associated with the misdiagnosis of $h_1$ for $h_2$ and of $h_2$ for $h_1$ might be small. Alternatively, the variable $x$ might be conditioned by other variables, and the conditioning events under which the expert is composing the partition might be extremely unlikely.

If we are to use similarity networks and partitions in this heuristic fashion, the quality of the knowledge bases we produce must exhibit *graceful degradation*. That is, if we decrease the amount of effort that we expend to compose a knowledge base by a small amount, then the diagnostic accuracy of that knowledge base must also decrease by only a small amount. The determination of the degradation characteristics of these representations requires both theoretical and empirical investigation.

### 6.2.3 Utility Assessment

The similarity-network representation might also facilitate the assessment of utilities. For example, a decision maker can compose a similarity graph (or hypergraph) where each node in the graph represents a decision outcome. For each edge or hyperedge in the graph, he can then identify attributes of preference (e.g., monetary losses and gains, pain, disability, and length of life) that discriminate the outcomes associated with that edge or hyperedge. This simple approach would serve to remind the decision maker of attributes that are important for his decision.

## 6.3 Conclusions

The computer-based transfer of knowledge through expert systems has helped many people who are confronted with confusing, important decisions. Furthermore, *normative* expert systems have the potential to deliver high-quality expertise that is free from many of the stereotypic errors in decision making made by both nonexperts and experts.

In this book, I have made several advances toward the goal of making the construction of normative expert systems practical. In particular, working with the Pathfinder expert, I have developed:

- *A general approach to the capture and representation of probabilistic knowledge.* In this approach, we identify assertions of conditional independence that an expert makes implicitly about his domain, and create a representation in which that expert can represent such assertions easily. This representation, in turn, facilitates the construction of an accurate probabilistic model for the expert's domain.
- *Similarity networks and partitions, two examples of the general approach.* In particular, the representations exploit subset independence and hypothesis-specific independence to facilitate the construction of knowledge maps for single- and multiple-fault diagnosis.
- *SimNet, an implementation of these representations on the Macintosh computer.*
- *Pathfinder, a normative expert system for lymph-node pathology.* The similarity-network and partition representations made the construction of Pathfinder not only tractable, but also cost effective.

The similarity-network and partition representation suffer from several weaknesses:

- *The assumption of strict positivity.* In proving that the construction of a global knowledge map from a similarity network is sound, I assumed that the underlying joint probability distribution of the network was strictly positive. This assumption is unrealistic for many domains, including Pathfinder's. Consequently, the expert and I had to expend additional effort to construct the global knowledge map for Pathfinder.
- *Limited experience:* I have demonstrated the usefulness of the similarity-network and partition representations for only one expert in a single domain. Whether we can apply these representations to diagnostic tasks in many other domains remains uncertain.[3]
- *Limited extensibility:* The representations will not be useful when there is no distinguished variable that can serve as the focus for attention during knowledge-map construction and assessment. Thus, the representations are less likely to facilitate knowledge acquisition for problems other than diagnosis.

Nonetheless, we can relax the assumption of strict positivity with additional theoretical work. In addition, given the extensions to similarity networks described in this chapter, the representation will probably make tractable the construction of a wide variety of normative expert systems for diagnosis. Furthermore, as we discussed in Chapter 4, most experts are likely to find the similarity-network and partition representations easy to use.

Most important, the general approach for making knowledge acquisition practical that I have described offers promise for problems that deviate from the model of diagnosis we have examined in this work. For such problems, a challenge lies in making forms of independence explicit and self-consistent, and in extending the influence-diagram representation to facilitate the expression of such forms of independence. If we can meet this challenge, then we will be able to construct normative expert systems for a wide variety of real-world domains.

[3]As of July, 1991, one year after the completion of this work, knowledge engineers have used the similarity-network and partition representations to construct expert systems for the diagnosis of (1) breast, intestine, ovary, skin, soft-tissue, testis, and thymus pathology, (2) sleep disorders, (3) eye diseases, (4) jet-engine failures, and (5) efficiency problems in gas turbines that generate electricity.

# A Background

In this book, we examine practical methods for using probability and decision theory to represent and manipulate knowledge within expert systems. This appendix provides an overview of the decision-theoretic concepts and techniques with which the reader must be familiar to understand this work. We begin with a discussion of the rules of probability and decision theory. We then examine decision analysis, the application of decision theory to real-world problems. Next, we consider alternative methodologies for reasoning under uncertainty, and discuss the advantages and disadvantages of these approaches with respect to decision theory. Finally, we examine the knowledge-map and influence-diagram representations, which are formal languages for plausible-inference and decision problems.

## A.1 Decision Theory and Decision Analysis

Decision theory is a tool for clearly describing and reasoning about a decision. The theory divides a decision into three fundamental components: what a decision maker *can do* (his alternatives), what he *knows* (his beliefs), and what he *wants* (his preferences). Within the theory, we use we use *probabilities* to describe a person's beliefs about whether or not various events will occur, and *utilities* to describe his preferences for each possible consequence of events.

### A.1.1 Uncertain Variables and Instances

A primary element of the language of probability is the uncertain variable. An *uncertain variable* represents a distinction about the world—that is, a set of mutually exclusive and exhaustive *instances* or *events*. An uncertain variable can represent a *binary* or *simple distinction*: an instance or event, and its negation. Alternatively, an uncertain variable can have more than two (possibly infinite) instances.

In this book, we denote uncertain variables with lower case letters, such as $x$, $y$, and $z$. We consider only variables with a finite number of instances; we subscript a variable to denote an instance or event for that variable. For example, $x_i$ denotes the $i$th instance of variable $x$. Also, we use $x_+$ and $x_-$ to refer to an instance or event and its negation.

### A.1.2 Probability as Personal Belief

The prevalent conception of the probability of some instance $x_i$ is that it is a measure of the frequency with which $x_i$ occurs, when we repeat many times an experiment with possible outcomes that correspond to the instances of $x$. A more general notion, however, is that the probability of $x_i$ represents the *degree of belief* held by a person that the event $x_i$ will occur in a single experiment. If a person assigns a probability of 1 to $x_i$, then he believes with certainty that $x_i$ will occur. If he assigns a probability of 0 to $x_i$, then he believes with certainty that $x_i$ will not happen. If he assigns a probability of between 0 and 1 to $x_i$, then he is to some degree unsure about whether or not $x_i$ will occur.

The interpretation of a probability as a frequency in a series of repeat experiments traditionally is referred to as the *objective* or *frequentist* interpretation. In contrast, the interpretation of a probability as a degree of belief is called the *subjective* or *Bayesian* interpretation, in honor of the Reverend Thomas Bayes, a scientist from the mid-1700s who helped to pioneer the theory of probabilistic inference (Bayes, 1958; Hacking, 1975).

In the Bayesian interpretation, a probability or belief will always depend on the state of knowledge of the person who provides that probability. For example, if we were to give someone a coin, he would likely assign a probability of 1/2 to the event that the coin would show heads on the next toss. If, however, we convinced that person that the coin was weighted in favor of heads, he would assign a higher probability to the event. Thus, we write the probability of $x_i$ as $p(x_i|\xi)$, which is read as the probability of $x_i$ *given* $\xi$. The symbol $\xi$ represents the state of knowledge or *background knowledge* of the person who provides the probability. Occasionally, when there is no ambiguity, we omit explicit mention of $\xi$.

Also, in this interpretation, a person can assess a probability based on information that he *assumes* to be true. For example, our coin tosser can assess the probability that the coin would show heads on the next toss, under the assumption that the same coin comes up heads on each of ten previous tosses. We write $p(x_+|y_+, \xi)$ to denote the probability of $x_+$ *given* that $y_+$ is true, and given background knowledge $\xi$.

The conception of probability as a measure of personal belief is central to research on the use of probability theory for representing and reasoning with expert knowledge in computer-based reasoning systems. There is usually no alternative to acquiring from experts the bulk of probabilistic information used in an expert system. Gathering a significant portion of frequencies through empirical study would entail much time and great expense. For example, in this book, we examine Pathfinder, an expert system for the diagnosis of lymph-node diseases. There are over 75 thousand probabilities in this expert system; some of these probabilities are on the order of $10^{-6}$. Furthermore, even when statistical studies have been performed in some domain, we often cannot employ the frequencies that these studies produce, because the specific distinctions used in an expert system for that domain may not match those distinctions used in the studies. Nonetheless, Bayesian probability theory provides for the gradual integration of appropriate statistical data into an expert system as those data become available (Howard, 1970a; Pearl, 1985; Spiegelhalter, 1986).

### A.1.3 Rules of Probability Theory

The framework of Bayesian probability theory consists of a set of rules that describes constraints among a collection of probabilities provided by a given person. We say that any set of beliefs that abide by these rules is *coherent*. There are many equivalent sets of rules. Here, we use the following set, where $x_+$ and $y_+$ are arbitrary events:

$$0 \leq p\,(x_+|\xi) \leq 1 \tag{A.1.1}$$

$$p\,(x_+ \text{ OR } x_-|\xi) = 1 \tag{A.1.2}$$

$$p\,(x_+|\xi) + p\,(x_-|\xi) = 1 \tag{A.1.3}$$

$$p\,(x_+, y_+|\xi) = p\,(x_+|y_+, \xi)\;\; p\,(y_+|\xi) \tag{A.1.4}$$

The disjunction of $x_+$ and $x_-$ in Equation A.1.2 is an event that is always true. The concatenation of $x_+$ and $y_-$ in Equation A.1.4 denotes the logical conjunction of the two events. Equations A.1.3 and A.1.4 are called the *sum rule* and *product rule*, respectively (Tribus, 1969; Jaynes, 1985).

Let us consider several consequences of these rules that we shall use frequently. By repeated application of the sum rule and product rule, we obtain

$$p\,(x_+ \text{ OR } y_+|\xi) = p\,(x_+|\xi) + p\,(y_+|\xi) - p\,(x_+, y_+|\xi) \tag{A.1.5}$$

Applying Equation A.1.5 to uncertain variable $z$ with mutually exclusive and exhaustive instances $z_1, z_2, \ldots, z_n$, we obtain the following more general version of the sum rule:

$$\sum_{z_i} p\,(z_i|\xi) \equiv p\,(z_1|\xi) + p\,(z_2|\xi) + \cdots + p\,(z_n|\xi) = 1 \tag{A.1.6}$$

Similarly, we have

$$\sum_{z_i} p\,(z_i, w_j|\xi) = p\,(w_j|\xi) \tag{A.1.7}$$

Applying the product rule to each term in the sum of Equation A.1.7, we obtain the *expansion rule* for probabilities, which tells us how to expand the probability of variable $w$ over variable $z$:

$$p\,(w_j|\xi) = \sum_{z_i} p\,(w_j|z_i, \xi)\;\; p\,(z_i|\xi) \tag{A.1.8}$$

Finally, if we divide both sides of the product rule by $p(y_+|\xi)$, we get

$$p(x_+|y_+,\xi) = \frac{p(x_+,y_+|\xi)}{p(y_+|\xi)} \tag{A.1.9}$$

Applying the product rule to $p(x_+,y_+,\xi)$ in Equation A.1.9 gives us

$$p(x_+|y_+,\xi) = \frac{p(y_+|x_+,\xi)\ p(x_+|\xi)}{p(y_+|\xi)} \tag{A.1.10}$$

which is *Bayes' theorem.* We sometimes refer to $p(x_+|\xi)$ and $p(x_+|y_+,\xi)$ as the *prior* and *posterior* probability of $x_+$, respectively.

### A.1.4 Proof of the Probability Rules

Within the frequentist interpretation, we can easily defend the rules of probability. A simple defense is not possible, however, within the Bayesian interpretation. For example, statisticians typically refer to Equation A.1.9 (omitting the reference to $\xi$) as the *definition* of a conditional probability. In the Bayesian interpretation, however, all probabilities are conditional. In particular, the probability $p(x_+|y_+,\xi)$ reflects a person's belief that $x_+$ will occur, given that he knows $y_+$, and given his background knowledge $\xi$. This person can assess $p(x_+|y_+,\xi)$ directly, as he can assess the other two probabilities in Equation A.1.9. Thus, in the Bayesian interpretation, Equation A.1.9—or the product rule, from which we derived Equation A.1.9—is a constraint among probabilities that we should prove. Similarly, we should prove the sum rule.

In the last 60 years, several researchers have derived the rules of probability (in one form or another) from fundamental axioms. For example, Ramsey and deFinetti have argued that anyone who is willing to bet in accordance with incoherent beliefs would be willing to accept a "Dutch book:" a combination of bets leading to a guaranteed loss under any circumstances (Ramsey, 1931; de Finetti, 1937).

The proof of the rules that I find most convincing was developed by the physicist Richard Cox. He was able to derive the probability rules without any mention of bets or payoffs. In particular, Cox identified a set of compelling axioms for a measure of belief. The axioms can be stated informally:

- *Clarity*: Events or variables should be well-defined.
- *Completeness*: A person can assign a degree of belief to any well-defined event.
- *Context dependency*: The degree of belief that a person assigns to an event can depend on the person's knowledge of other events. We denote the degree of belief in $x_+$, given background knowledge $\xi$, as $\phi(x_+|\xi)$.

- *Consistency*: If a person knows that events $x_+$ and $y_+$ are logically equivalent and that $z_+$ and $w_+$ are logically equivalent, then

  $$\phi(x_+|z_+,\xi) = \phi(y_+|w_+,\xi)$$

- *Complementarity*: For all simple distinctions $x$, there exists some function $f$ of $\phi(x_+|\xi)$ such that

  $$\phi(x_+|\xi) = f[\phi(x_-|\xi)]$$

  That is, we can compute the belief in the negation of an event from the belief in the event itself.

- *Hypothetical conditioning*: For all events $x_+$ and $y_+$, there exists some function $g$ of $\phi(x_+|y_+,\xi)$ and $\phi(y_+|\xi)$ such that

  $$\phi(x_+, y_+|\xi) = g[\phi(x_+|y_+,\xi),\ \phi(y_+|\xi)]$$

  where $g$ is nondecreasing in both of its arguments. That is, we can calculate the belief in a conjunction of events, from the belief in one event and the belief in the other event given that the first event is observed.

Cox showed that, given these axioms, the quantity $\phi(\cdot)$ must be a probability. That is, he proved that some monotonic transformation of $\phi(\cdot)$ must satisfy Equations A.1.1 through A.1.4. In deriving this result, Cox assumed that degrees of belief were represented by real numbers, and that the functions $f$ and $g$ were twice differentiable. Later, Aczel generalized the proof, showing that the functions $f$ and $g$ need only to be continuous (Aczel, 1966). Most recently, Aleliunas examined the case where degrees of belief are discrete. He showed that, provided we can multiply any degree of belief by itself enough times such that the product is not less than any other degree of belief, and provided that this measure of belief satisfies Cox's principles, then this measure must be isomorphic to a subalgebra of ordinary real-valued probabilities (Aleliunas, 1988).

Cox's axioms, with the exception of complementarity and hypothetical conditioning, require little if any justification. Tribus gives a detailed yet convincing argument for the axiom of hypothetical conditioning (Tribus, 1969). Here, we justify the axiom of complementarity. To many readers, it might seem obvious that, once a person provides a degree of belief in event $x_+$, then his degree of belief in the negation of that event is determined, because $x_+$ is the logical antithesis of $x_-$. Nonetheless, several artificial-intelligence researchers argue that, if our belief in the event $x_-$ is determined from our belief in the event $x_+$, then there is no room for us to express our degree of confidence about our belief in either event. There are, however, at least two mechanisms within the probabilistic framework for representing degrees of confidence. In one approach, we provide *bounds* on probabilities. Given a collection of probability bounds, we can infer bounds on other probabilities by applying the rules of probability on all possible point probabilities within those bounds. This approach was originally developed by Good (Good, 1962), and has attracted interest recently within the artificial-intelligence community (Nilsson, 1986). In another approach, we can equate a person's degree of confidence in his assessment of $p(x_+|\xi)$ with the degree to which that person's probability will change, when he obtains new information (de Finetti, 1977; Heckerman and Jimison, 1987; Pearl, 1987a; Pearl, 1988; Howard, 1988b). That is, if a persons holds a probability with a high degree of confidence, then he will not change that probability significantly, no matter what he learns about the world. In contrast, if he holds a probability with a low degree of confidence, then he is likely to change that probability, given new information.

Despite the arguments of Cox and other researchers for the rules of probability, there is still controversy in the artificial-intelligence field about the adequacy of probability for representing beliefs. We return to this discussion in Section A.2.

### A.1.5 The Maximum Expected Utility Principle

We can think of a decision as a choice among one or more *lotteries*. Given a lottery, we receive exactly one of a set of *prizes*. Associated with each prize is a chance (i.e., a probability) that we get that prize. In this section, we use decision trees to represent lotteries. For example, Figure A.1(a) illustrates a simple lottery where we receive a 2-week all-expenses-paid trip to Hawaii with probability 3/4, and nothing with probability 1/4. Figure A.1(b) shows a compound lottery with two stages. The number of prizes that we associate with a lottery can be infinite; to simplify the discussion, however, we assume that lotteries are finite.

People naturally ascribe degrees of preference to prizes. Let us represent a person's degree of preference for a prize by a real-valued quantity called the *utility* of that prize. Using upper-case letters (such as $A$, $B$, and $C$) to denote prizes, we let $u(A)$ represent the utility of prize $A$, for a given decision maker.

Suppose that we have a lottery with prizes $A_1$, $A_2$, ..., $A_n$, and that the probability of receiving $A_i$ is $p_i$, for $i = 1, 2, \ldots, n$.[1] The expected utility of that lottery is

$$\sum_{i=1}^{n} p_i \; u(A_i)$$

[1] In this and the following section, we deviate slightly from the notation described earlier, to simplify the presentation.

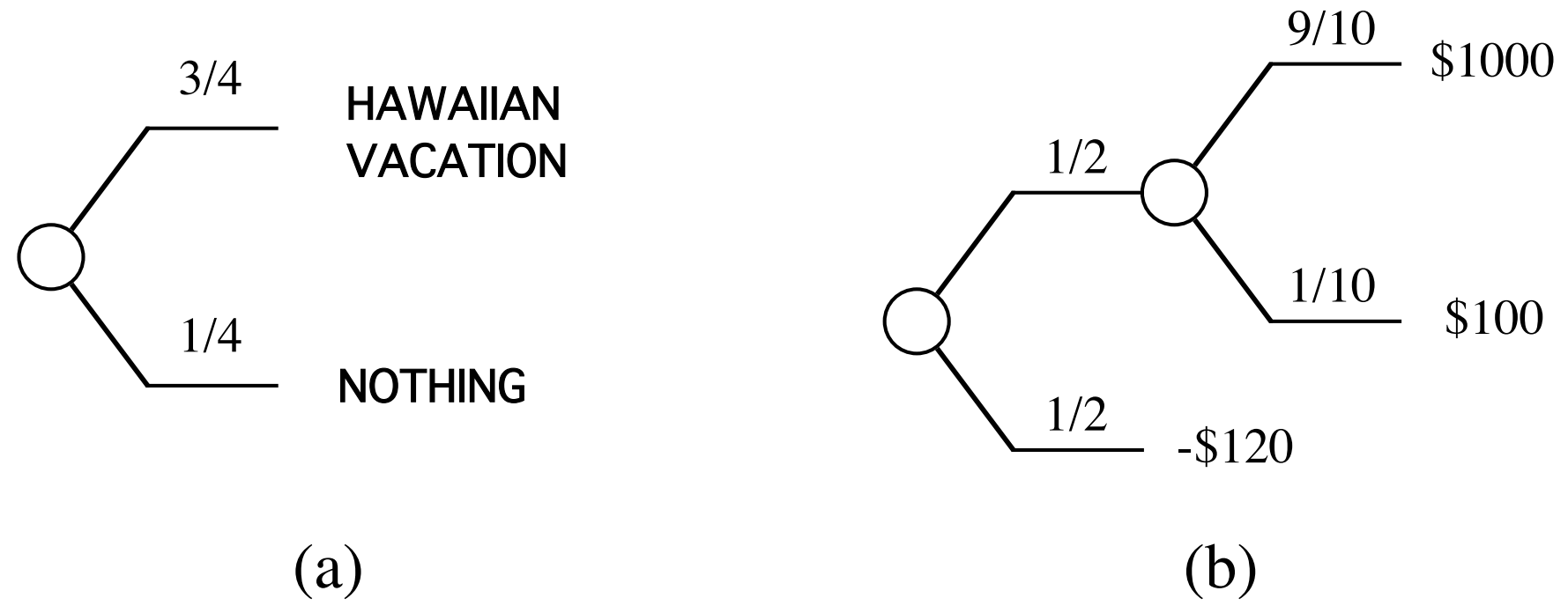

**Figure A.1:** Two lotteries.
(a) A simple lottery. The lottery offers a 3/4 chance of winning a 2-week all-expenses-paid trip to Hawaii, and a 1/4 chance of winning nothing. (b) A compound lottery. In the first round of the lottery, we face a 1/2 probability of losing \$120. If we do not lose in the first round, we get a 9/10 chance of winning \$1000 and a 1/10 chance of winning \$100.

Now suppose we are offered a choice between two lotteries. If we could choose between the same two lotteries many times, we would do best in the long run by always choosing the lottery with the greatest expected utility. In real life, however, we seldom are faced with a series of identical choices. So what do we do? The maximum expected utility (MEU) principle says that we should choose the lottery (i.e., alternative) with the maximum expected utility.

### A.1.6 Proof of the Maximum Expected Utility Principle

Pascal and his associates first suggested the MEU principle over 300 years ago (Hacking, 1975). It was not until 1947, however, that vonNeumann and Morgenstern provided the first formal argument for the principle (von Neumann and Morgenstern, 1947). In their argument, they identified five axioms that describe how a decision maker should choose among simple lotteries. They showed that if the person follows these axioms, then whenever he faces a choice among more complicated lotteries, he must act *as if* he had assigned a utility to each prize in the lottery, and then had selected the lottery with the greatest expected utility. These axioms (which include the rules of probability) form the basis of decision theory.

Let us examine a simple version of the argument presented by Howard (1970b). Von-Neumann and Morgenstern proposed the following axioms:

- *Orderability*: A decision maker must be able to state his preferences among the prizes of any lottery. That is, given any two prizes $A$ and $B$, he must be able to state whether he prefers $A$ to $B$, prefers $B$ to $A$, or is indifferent between $A$ and $B$. Furthermore, his preferences must be transitive. If a decision maker prefers $A$ to $B$, we write $A \succ B$; if he is indifferent between $A$ and $B$, we write $A \sim B$.

- *Continuity*: Consider the lottery shown in Figure A.2, in which a decision maker receives prize $A$ with probability $p$ and prize $C$ with probability $1-p$. If the decision maker has expressed the preferences $A \succ B \succ C$, then, for some probability $p$, the decision maker must be indifferent between receiving $B$ for certain, and receiving the lottery. We call the lottery where the decision maker receives $B$ with certainty the *certain equivalent* of the lottery involving $A$ and $C$.
- *Substitutability*: We can exchange a lottery with its certain equivalent without affecting preferences. For example, suppose a decision maker is indifferent between having $B$ for certain and having the lottery with prizes $A$ and $C$ illustrated in Figure A.2. Then he must be indifferent between the two lotteries in Figure A.3. The only difference between these two lotteries is that prize $B$ is substituted for the lottery with $A$ and $C$.
- *Monotonicity*: Let us suppose that a decision maker can choose between two lotteries $L_1$ and $L_2$, and that both lotteries have the same prizes $A$ and $B$. In addition, let us suppose that the probability of receiving prize $A$ is $p_1$ for lottery $L_1$, and is $p_2$ for lottery $L_2$. If the decision maker prefers $A$ to $B$, then he must prefer $L_1$ to $L_2$ if and only if $p_1 > p_2$. That is, the decision maker must prefer the lottery that offers the greater chance of receiving the better prize. This axiom is illustrated in Figure A.4.
- *Decomposability*: We can reduce compound lotteries to simple ones using the rules of probability. An example of such a reduction is shown in Figure A.5.

The axioms are compelling. For example, let us suppose that a person has the intransitive preferences $A \succ B \succ C \succ A$, and that he holds prize $A$. Because he prefers $C$ to $A$, he should be willing to exchange $A$ for $C$ and a small payment. Similarly, this person should be willing to exchange $C$ for $B$, and $A$ for $B$. As a result, we can extract payments from him, and yet leave him with the same prize. This imaginary device, called a *money pump*, provides a strong argument for the orderability axiom. We can defend each of the axioms with arguments like this one (Howard, 1970b).

Now let us examine the consequences of these axioms. Suppose that a decision maker must choose between two lotteries $L_1$ and $L_2$, each with prizes $A_1$, $A_2$, $\ldots A_n$. Using the orderability axiom, we can assume that $A_1 \succ A_2 \succ \cdots \succ A_n$. Further, suppose that the probability of receiving prize $A_i$, in lotteries $L_1$ and $L_2$, is $p_i^1$ and $p_i^2$, respectively, for $i = 1, 2, \ldots n$. Some of these probabilities might be equal to 0, so that all prizes may not be available in both lotteries.

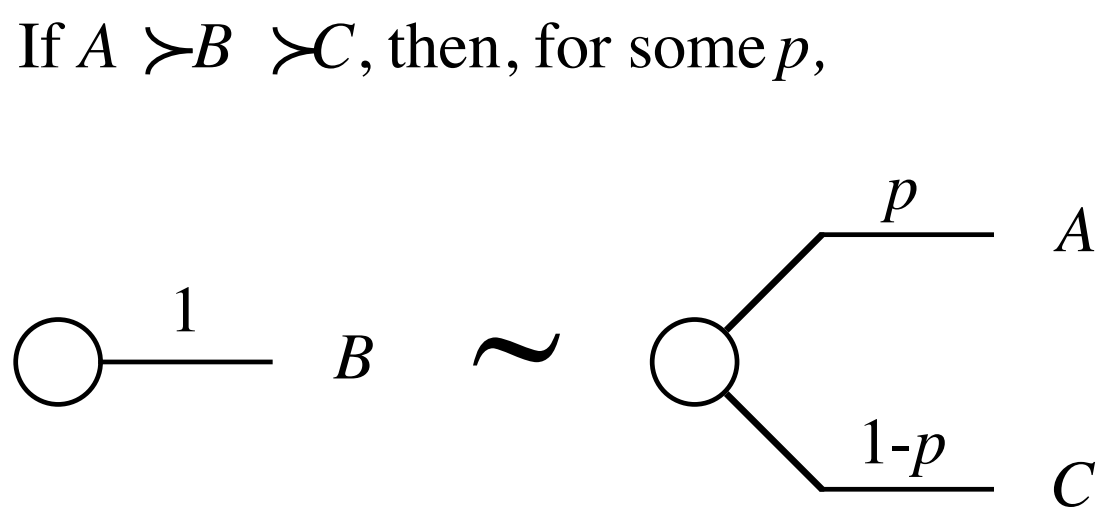

**Figure A.2:** The continuity axiom.
If a decision maker prefers $A$ to $B$, and $B$ to $C$, then for some probability $p$, he must be indifferent between having $B$ for certain, and having a lottery that rewards $A$ with probability $p$ and $C$ with probability $1 - p$.

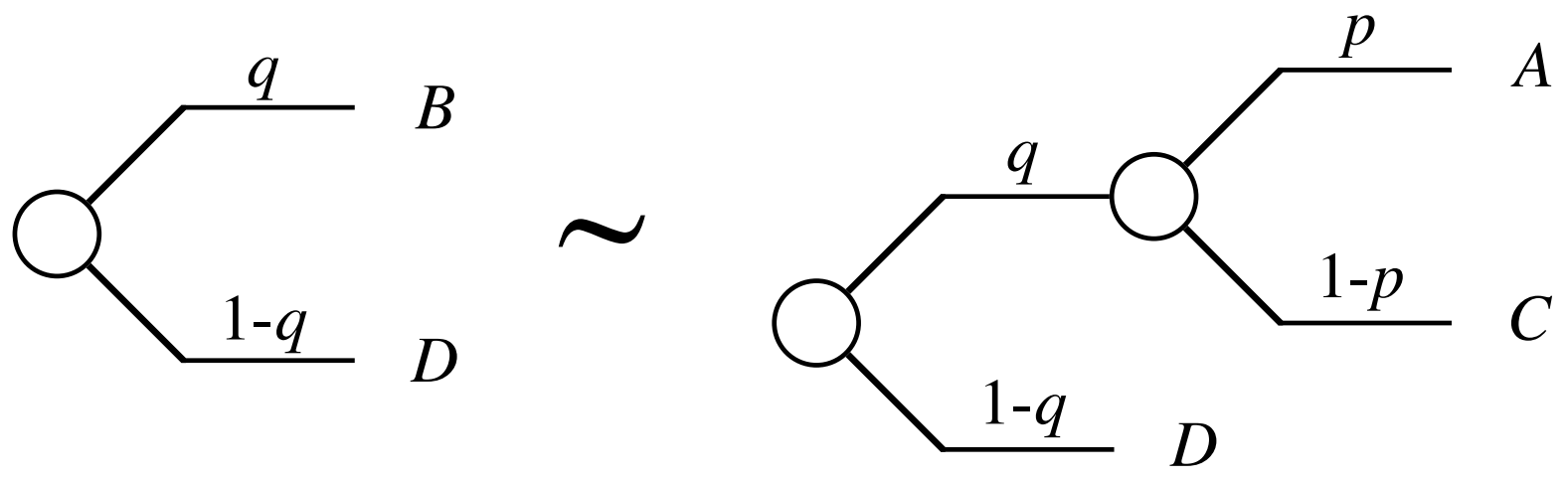

**Figure A.3:** An indifference implied by the substitutability axiom.
Given that a decision maker is indifferent between to two lotteries in Figure A.2, the substitutability axioms states that the decision maker must also be indifferent between the two lotteries shown here. In the lottery on the left of the figure, a person receives $B$ with probability $q$. In the lottery on the right of the figure, the person receives the lottery involving $A$ and $C$ with the same probability.

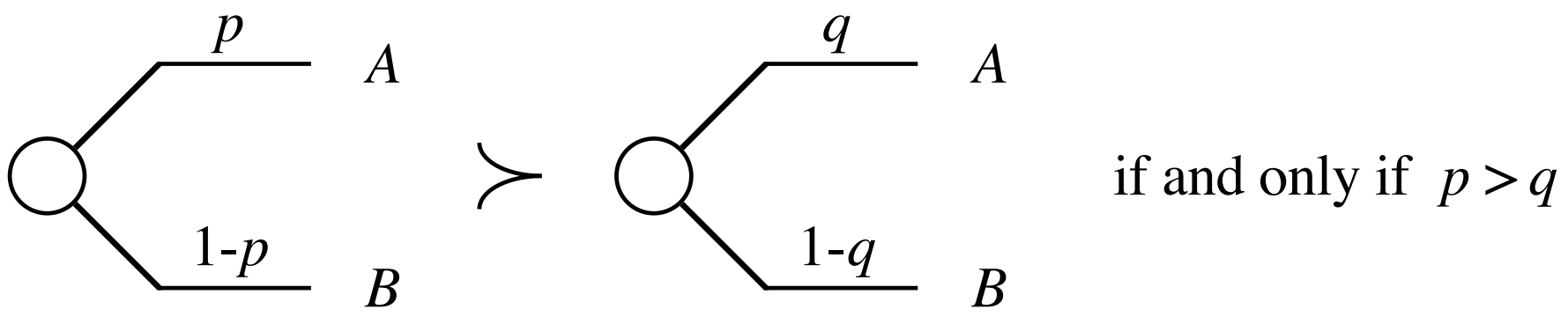

**Figure A.4:** The monotonicity axiom.
A decision maker should always prefer the lottery that offers the greater chance of winning the better prize ($A$ in this case).

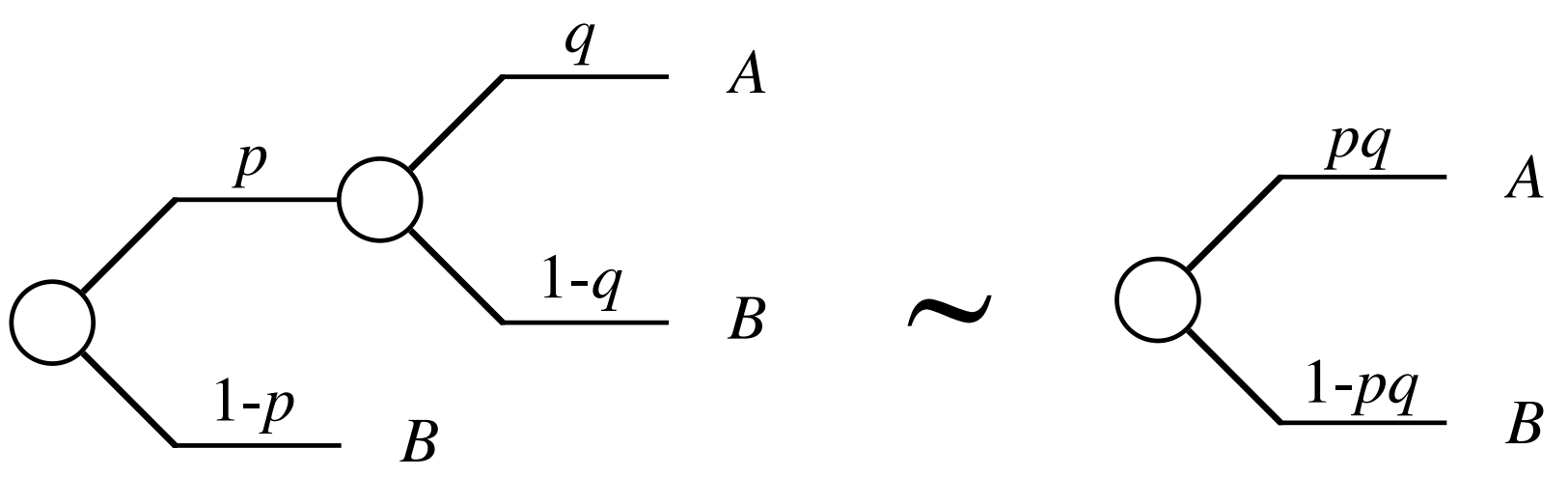

**Figure A.5:** An example of the decomposability axiom.
We apply the sum and product rules to the compound lottery on the left to obtain the simple lottery on the right.

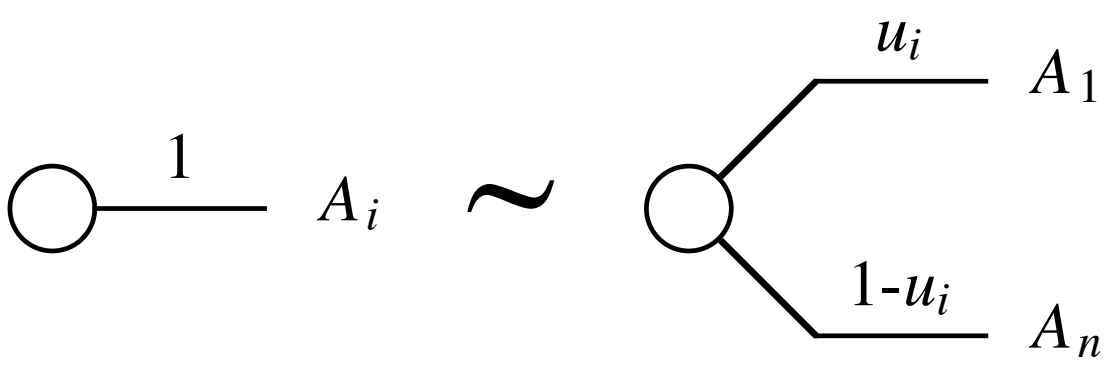

**Figure A.6:** A consequence of the continuity axiom.
For some probability $u_i$, a decision maker must be indifferent between getting prize $A_i$ for certain, and the lottery that awards $A_1$ with probability $u_i$ and $A_n$ with probability $1 - u_i$.

Given the continuity axiom, we know that, for every prize $A_i$, there exists a probability $u_i$, such that the decision maker is indifferent between receiving prize $A_i$ and the lottery that offers the most desirable prize, $A_1$, with probability $u_i$, and the least desirable prize, $A_n$, with probability $1 - u_i$. The equivalence is illustrated in Figure A.6. Using the substitutability axiom, we can replace each prize $A_i$ in lotteries $L_1$ and $L_2$ with its corresponding lottery derived from the continuity axiom. The substitution for lottery $L_1$ appears in Figure A.7. Applying the decomposability axiom, as shown in Figure A.8, we know that lottery $L_1$ is equivalent to a lottery that offers prize $A_1$ with probability $\sum_{i=1}^{n} u_i p_i^1$ and prize $A_n$ with probability $1 - \sum_{i=1}^{n} u_i p_i^1$. We can derive a similar equivalence for lottery $L_2$. Finally, given the monotonicity axiom, we know that the decision maker must prefer $L_1$ to $L_2$ if and only if

$$\sum_{i=1}^{n} u_i \; p_i^1 \; > \sum_{i=1}^{n} u_i \; p_i^2 \qquad \text{(A.1.11)}$$

This result is the MEU principle. We can identify $u_i$ as the utility of prize $A_i$, and the sums in Equation A.1.11 as the expected utilities of the lotteries.

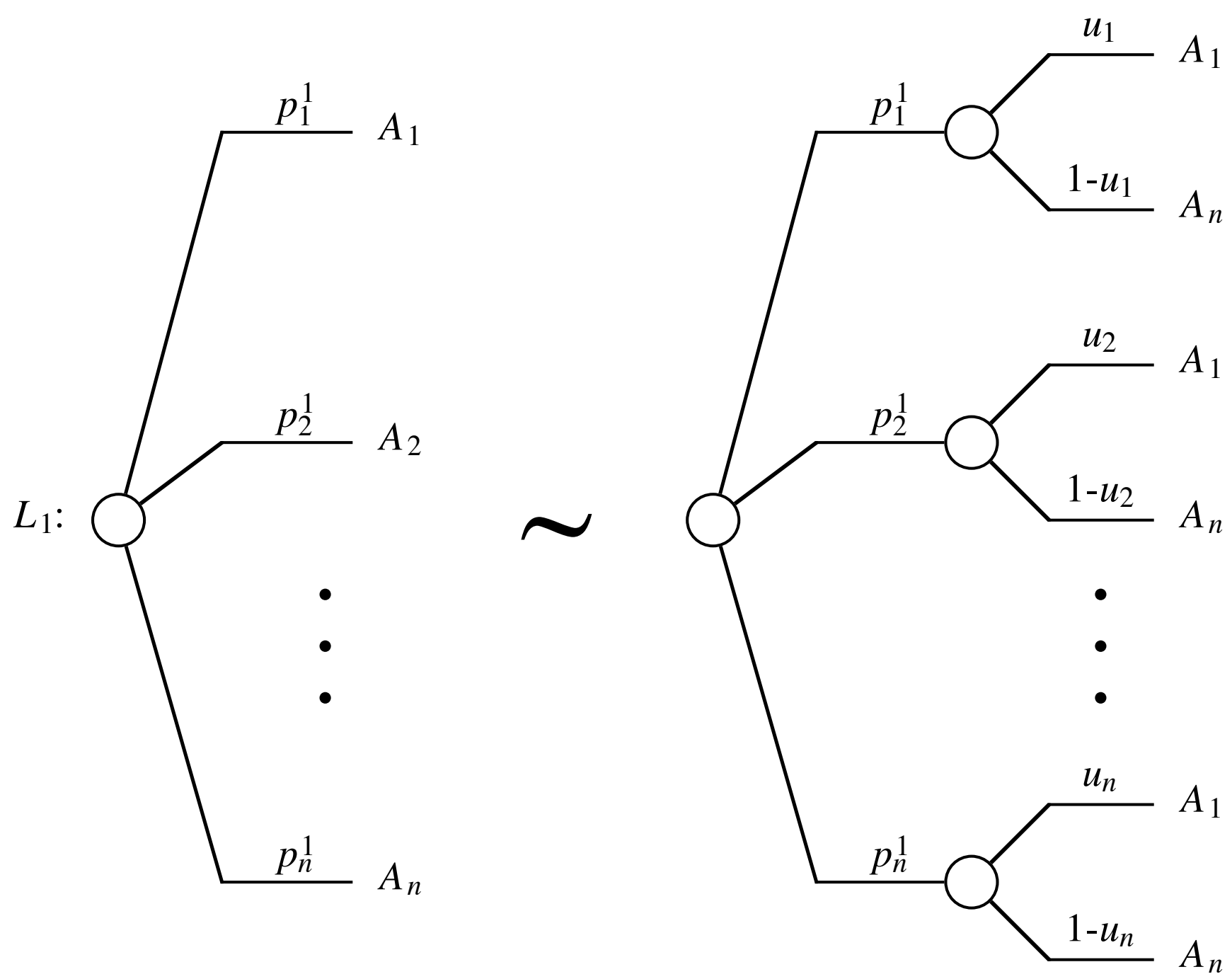

**Figure A.7:** A consequence of the substitutability axiom.
The decision maker must be indifferent between the simple lottery $L_1$ and the compound lottery where we replace each prize $A_i$ in $L_1$ by a lottery of the form shown in Figure A.6.

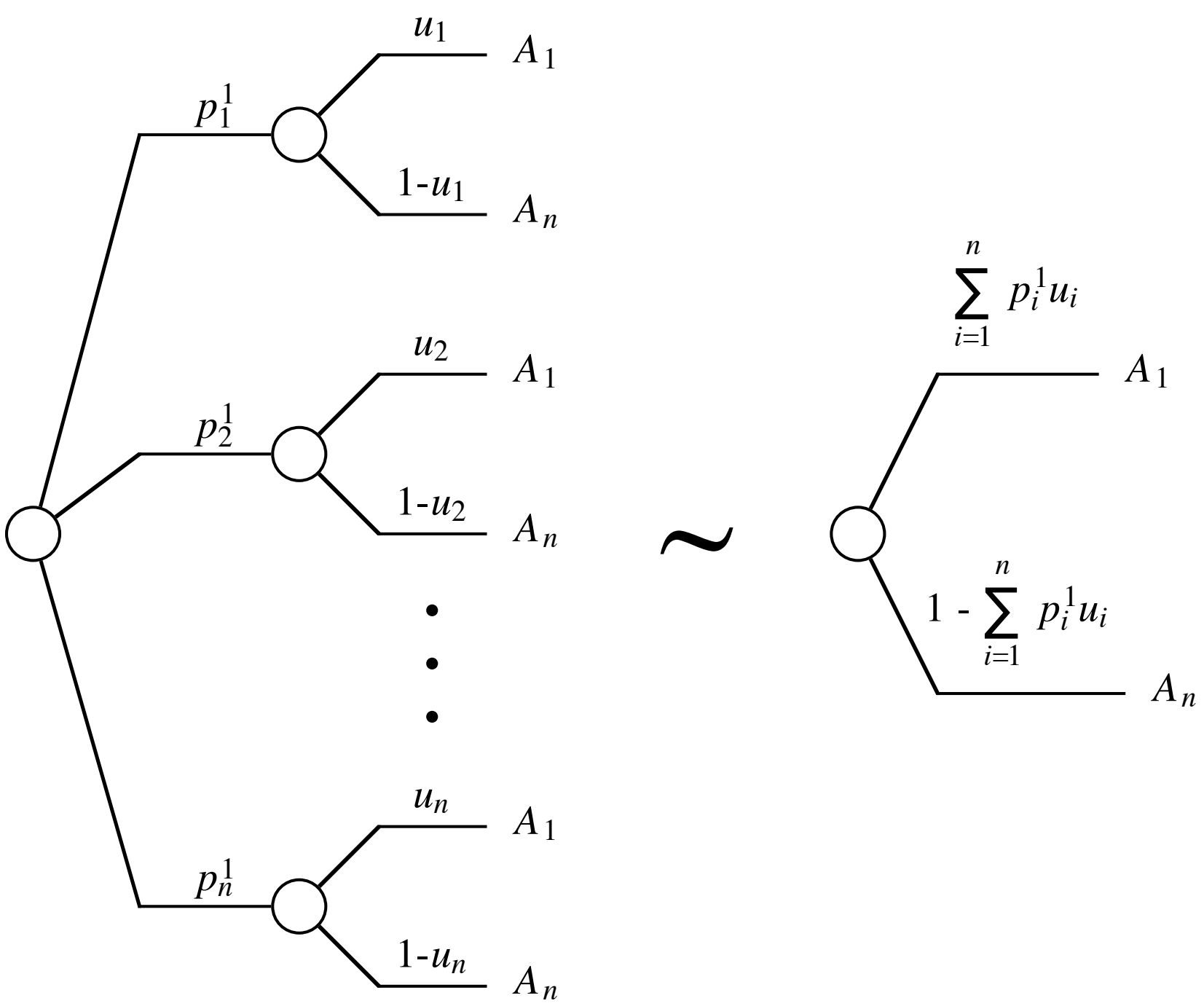

**Figure A.8:** A consequence of the decomposability axiom.
We derive this statement of indifference by applying the sum and product rules to the compound lottery on the left.

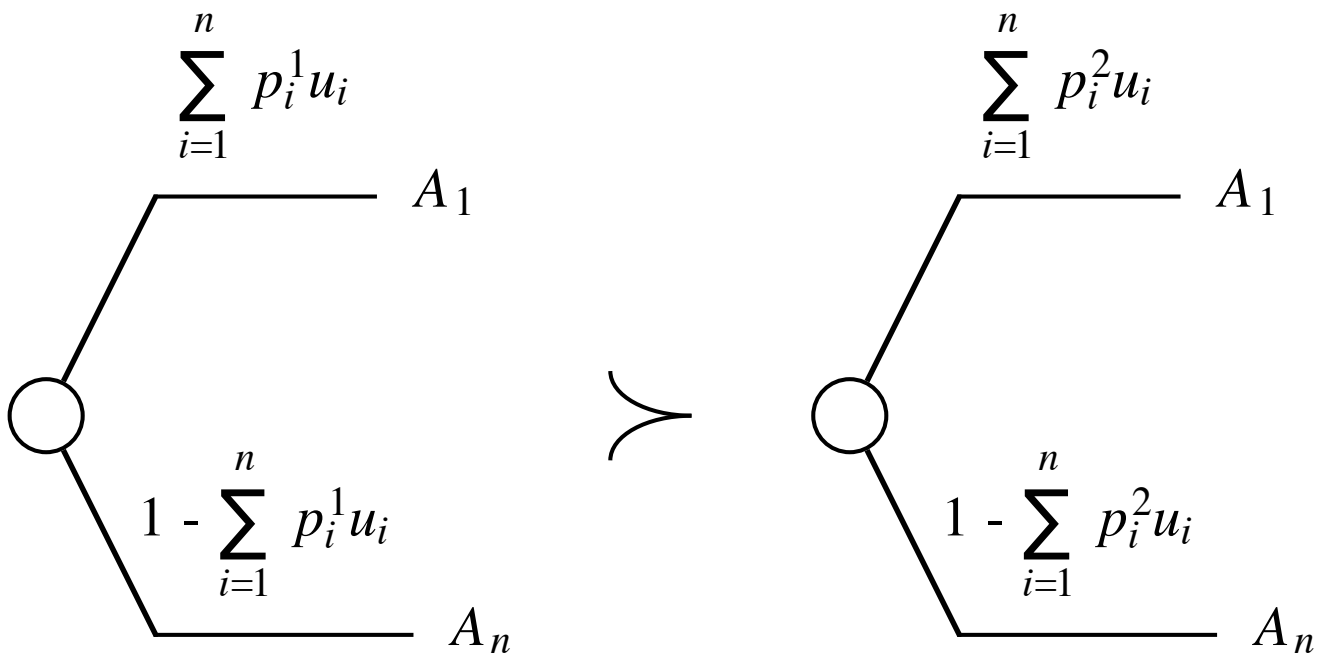

if and only if

$$\sum_{i=1}^{n} p_i^1 u_i > \sum_{i=1}^{n} p_i^2 u_i$$

**Figure A.9:** A consequence of the monotonicity axiom.
The MEU principle follows directly from this application of the monotonicity axiom. We can identify $u_i$ as the utility of prize $A_i$, and the sums as the expected utilities of the original lotteries $L_1$ and $L_2$.

### A.1.7 Normative Versus Descriptive Decision Making

In the proof of the MEU principle, we simply reduce a choice between two (or more) complex lotteries to a series of choices between simpler lotteries, using several desirable axioms. Although this approach is attractive from a theoretical perspective, many researchers have shown that people follow neither the MEU principle nor the axioms when they make decisions (Edwards, ed., 1956; Tversky and Kahneman, 1974; Kahneman et al., 1982). In fact, several researchers have demonstrated that people exhibit stereotypical deviations or *biases* from the axioms (Tversky and Kahneman, 1974; Kahneman et al., 1982). Thus, we distinguish between *normative* and *descriptive* decision making. That is, we distinguish between how we wish we could make decisions, and how we actually make them, when unaided by the axioms.

This distinction is important to people who build expert systems for medicine and for other domains where the stakes are high. Unaided physicians, whether they are nonspecialists or specialists, are not immune to errors in decision making (Elstein, 1976; Elstein et al., 1978). Thus, from the perspective of decision theory, these physicians are taking actions that sometimes incur great costs to their patients. *Normative expert systems*—computer-based systems that deliver expert knowledge in a decision-theoretic framework—have the potential to increase the quality of decisions made by physicians, and thereby to improve patient outcome dramatically. Such systems probably will be an improvement over traditional (nonnormative) expert systems, which faithfully reproduce the errors of experts. In addition, normative expert systems probably will allow specialists to benefit from their own knowledge—knowledge encoded with the aid of the axioms.

### A.1.8 Decision Analysis

Decision analysis is an engineering discipline that addresses practical issues concerning the application of decision theory to real-world decision problems. The discipline grew out of the fields of systems science and statistical decision theory in the mid-1960s (Howard, 1966; Howard, 1968). Since that time, decision analyses have been applied with great success to many domains, including those within the areas of government, industry, law, and medicine (Howard and Matheson, 1983).

Decision analysis augments decision theory with a set of representations and techniques that help a person confronted by a difficult decision. Representations include the *strategy-generation table*, a device for creating new alternatives, and the *influence diagram*, a graphical language that we examine in Section A.3.

Decision-analytic techniques include the *clarity test*, a procedure that helps a decision maker to define events and variables clearly (Howard, 1988a). This task is important, because it eliminates a component of uncertainty often associated with the assessment of beliefs and preferences. To apply the clarity test to an event, we ask ourselves whether or not a clairvoyant—an omniscient being able to see the past, present, and future with certainty—could tell us the outcome of that event. If the clairvoyant cannot determine the outcome, then we must work to make the definition of that event more precise. For example, using the clarity test, we could sharpen the definition of the event *it will rain at Stanford next Monday* to obtain the definition *the majority of people at the Stanford coffee house at 9pm next Monday (who are willing to vote) will say that it rained somewhere on the Stanford campus between sunrise and sunset that day.*

Other decision-analytic techniques include methods for correcting the biases in the probability and utility assessments of a decision maker. One such method is the *almanac game* (Raiffa, 1968; Howard, 1985). Decision analysts invented this game in response to the observation that most people believe they know more than they actually do know. In a simple version of this game, a decision analyst uses an almanac to generate a question whose answer is a continuous (or at least an ordered) value. For example, the decision maker might generate the question: "What is the height of the tallest building in Iowa?" Next, the decision maker is asked to provide two quantities LOW and HIGH, such that he believes the true answer to the question lies below LOW with probability 1/4 and above HIGH with the same probability. After answering a series of these questions, the decision analyst reveals the true answers to the questions. If the decision maker has accurately reflected his beliefs, then the true answers to the questions should lie between LOW and HIGH about one-half of the time. Typically, however, the true answers lie in that interval much less than one-half of the time. On observing this outcome, the decision maker can improve his probability assessments for the next round of play. As the game continues, the accuracy of the decision maker's probabilities increase.

Another decision-analytic technique is *sensitivity analysis.* The are several types of sensitivity analysis, including *deterministic sensitivity analysis* and several forms of *stochastic sensitivity analysis.* In applying a deterministic sensitivity analysis to a decision model, we sweep one or more variables in the model through their possible instances. We then observe whether or not such actions affect the utility of the resulting outcomes. We fix those variables that do not have a pronounced affect on utility at their nominal values, and we continue to model the remaining variables as such.

We can apply stochastic sensitivity analysis to decisions in a similar fashion. In one form of stochastic sensitivity analysis, we sweep one or more probabilities and utilities in a decision model through wide ranges of value, and determine whether or not such action affects the decision or decisions in the model. Both deterministic and stochastic sensitivity analyses help to direct a decision maker's attention to those components of his decision problem that are most worthy of consideration.

Besides representations and specific interview techniques, decision analysis provides a philosophy that emphasizes the insights that can be gained by a decision maker who goes through the decision-analytic process, rather than the results of MEU calculations. For example, the decision-analytic philosophy highlights the distinction between a good decision and a good outcome: a *good decision* is one that is consistent with the preferences and complete information of a decision maker; a *good outcome* is desirable. Sometimes, a good decision will, through a course of bad luck, lead to a bad outcome. Conversely, a bad decision will, with a great deal of good luck, lead to a good outcome. Nonetheless, the best way to achieve good outcomes in the long run, short of being all-knowing, is to make good decisions consistently. With this realization, people faced with confusing high-stakes decisions more easily can overcome feelings of helplessness that often paralyze their actions.

## A.2 Decision Theory Versus Other Formalisms for Decision Making

In the last two decades, several researchers have suggested that decision theory is not adequate for the representation and manipulation of uncertain knowledge. Investigators have cited both theoretic (Shafer, 1986) and practical (Gorry, 1973; Shortliffe and Buchanan, 1975) limitations of the theory. In response to such criticisms, researchers have developed alternative methods for uncertain reasoning. These approaches include the Dempster–Shafer theory of belief functions (Shafer, 1976; Shafer, 1981), fuzzy-set theory (Zadeh, 1983), and the Mycin certainty–factor (CF) model (Shortliffe and Buchanan, 1975). Although the developers of each of these approaches have concentrated on the representation of belief, they have suggested methods for decision making within their frameworks as well (Shafer, 1982; Zadeh, 1983; Kacprzyk and Orlovski, 1987; Buchanan and Shortliffe, 1984).

Despite such criticisms and the availability of these alternatives, I have employed the decision-theoretic framework exclusively in the work described in this book. I believe that decision theory is by far the most appropriate framework for encoding and reasoning with knowledge, in expert systems and elsewhere. Because this claim is controversial, I present my reasons for preferring decision theory, and point to other discussions of the relative merits of each approach.

There are five major advantages of using decision theory as a framework for expert systems. Not one of the alternative approaches shares all five of these advantages. First, a decision-theoretic framework requires that expertise encoded in a knowledge base be *self-consistent*. I have found that this consistency requirement exposes flaws in the reasoning of experts in much the same way that a computer-programming language exposes errors in human-designed algorithms. Revelations of this sort should improve the quality of knowledge stored in an expert system, and thereby improve the performance (e.g., diagnostic accuracy) of the system.

Second, assumptions that experts make during the process of encoding their knowledge in a decision-theoretic framework—for example, assumptions of conditional independence—are represented with clarity. The *unambiguous representation of assumptions* allows system builders and experts to address explicitly the inevitable tradeoffs that they encounter during the construction of an expert system, such as the tradeoff between the completeness of a knowledge base and the time and effort dedicated to knowledge-base construction. In addition, such clarity can simplify the evaluation and modification of an expert system. For example, if we find that a system's performance is poor, we first can examine and modify those assumptions that we believe to be least reasonable. Finally, the explicit representation of assumptions allows researchers to build on the work of other investigators.

Third, decision theory is *general*. That is, any aspect of rational thought related to decision making can be captured by decision theory. For example, despite claims to the contrary, we can express the degree of confidence a person has in a probability within a probabilistic framework (see Section A.1.4). Also, we can incorporate the results of statistical studies directly into a decision-theoretic model (Howard, 1970a; Pearl, 1985; Spiegelhalter, 1986).

Fourth, decision theory—especially probability theory—is *well developed*. That is, the theory has existed for several centuries. As a result of its maturity, probability theory is familiar to most researchers who have studied the representation of beliefs. Consequently, investigators can share work easily: researchers can build on the work of other people, and can avoid previous mistakes.

Fifth, and most important, decision theory is *normative*. That is, the axioms of decision theory are compelling, and psychologists have characterized people's deviations from these axioms, in detail. As a consequence of this characterization, decision analysts have developed techniques that help people to avoid mistakes in reasoning (see Section A.1.8).

The Dempster–Shafer theory of belief is neither general (it lacks a formal theory for decision making) nor well developed. In addition, many of the assumptions associated with the use of the theory are unclear (Pearl, 1990). Finally, with rare exceptions, most researchers do not believe that the theory is normative. That is, most investigators do not believe that people *should* act in accordance with the Dempster–Shafer theory. (For the two sides of this debate, see Lindley, 1982, and Shafer, 1986).

The most significant weakness of fuzzy-set theory is the ambiguity of the assumptions associated with the theory. In particular, the developers of the theory have not defined clearly the meanings of the quantities that represent degrees of belief and preference (Cheeseman, 1985). Furthermore, the theory is neither well developed nor normative (Cheeseman, 1985).

The CF model exhibits none of the advantages of decision theory that we have examined. For example, I have shown that the model allows—and sometimes encourages—the representation of inconsistent knowledge (Heckerman, 1985). In addition, the model makes strong assumptions of conditional independence that are not apparent to the user (Heckerman, 1985).

## A.3 Knowledge Maps and Influence Diagrams

One important advantage of the CF model and other ad hoc approaches is that these techniques are tractable. The main purpose of this book, however, is to demonstrate that we can make decision-theoretic methods tractable as well. The representations that we discuss in this section are a first step toward this goal.

### A.3.1 Knowledge Maps

A *knowledge map* is a graphical knowledge-representation language that encodes probabilistic dependencies among distinctions (Howard, 1989a). The representation rigorously describes probabilistic relationships, yet has a human-oriented qualitative structure that facilitates communication between the expert and the probabilistic model. In addition, the representation can represent any probabilistic-inference problem. Several researchers have developed and studied knowledge maps, although they have used various names for this representation such as *causal nets* (Good, 1961a; Good, 1961b), *probabilistic cause–effect models* (Rousseau, 1968), *Bayesian belief networks* and *causal networks* (Pearl, 1982; Pearl, 1988; Verma and Pearl, 1988; Geiger and Pearl, 1988; Lauritzen and Spiegelhalter, 1988), and *probabilistic causal networks* (Cooper, 1984). An *influence diagram* is an extension of the knowledge-map representation that can represent any decision problem (Howard and Matheson, 1981). In particular, an influence diagram can serve as a knowledge base for an expert system.

Let us first consider the knowledge-map representation, using a simple example taken from Kim and Pearl (1983):

> Mr. Holmes receives a telephone call from his neighbor, who notifies him that he has heard a burglar alarm sound from the direction of his home. As he is preparing to rush home, Mr. Holmes recalls that the previous sounding of his alarm was triggered by an earthquake. A moment later, he hears a radio newscast reporting an earthquake 200 miles from his house.

Figure A.10 shows a knowledge map for Mr. Holmes' situation. The knowledge map is a directed acyclic graph,[2] whose nodes represent the uncertain variables relevant to the problem, and whose arcs represent potential probabilistic dependencies among those variables. In the remainder of this discussion, we make no distinction between the variable $x$ and the node $x$ that represents that variable.

In a knowledge map, an arc from node $x$ to node $y$ reflects an assertion by the builder of that network that the probability distribution for $y$ may depend on the instance of the variable $x$. We say that $x$ *conditions* $y$. In the knowledge map for Mr. Holmes' situation, all variables are binary or simple distinctions. Thus—for example—the arc from ALARM to PHONE CALL in Figure A.10 represents Mr. Holmes' assertion that the probability of receiving the telephone call may depend on whether or not there was an alarm. Conversely, the lack of arcs in a knowledge map reflect assertions of *conditional independence*. For example, there is no arc from BURGLARY to PHONE CALL in Figure A.10. The lack of this arc encodes Mr. Holmes' belief that the probability of receiving the telephone call from his neighbor does not depend on whether or not there was a burglary, provided Mr. Holmes knows whether or not the alarm sounded.

In Chapter 3, we examine the formal relationship between conditional-independence assertions and the topology of a knowledge map. Here, it is important to recognize that, using knowledge maps, experts can *control* the assertions of conditional independence that are encoded in normative expert systems. Such control was not available to experts who constructed probabilistic expert systems in the 1960s (see Chapter 1).

Each node in a knowledge map is associated with a set of probability distributions. These distributions appear below the knowledge map in Figure A.10. In particular, a node has a probability distribution for every instance of its conditioning nodes. (An instance of a set of nodes is an assignment of an instance to each node in that set.) For example, in Figure A.10, ALARM is conditioned by both EARTHQUAKE and BURGLARY. Therefore, there are four probability distributions for ALARM, corresponding to the instances where both EARTHQUAKE and BURGLARY occur, BURGLARY occurs alone, EARTHQUAKE occurs alone, and neither EARTHQUAKE nor BURGLARY occurs. In contrast, RADIO NEWSCAST and PHONE CALL are each conditioned by only one node. Thus, there are two probability distributions for RADIO NEWSCAST and two probability distributions for PHONE CALL. Finally, EARTHQUAKE and BURGLARY do not have any conditioning nodes, and hence each node has only one—*marginal*—probability distribution.

[2]A directed acyclic graph contains no directed cycles. That is, in a directed acyclic graph, we cannot travel from a node and return to that same node along a nontrivial directed path.

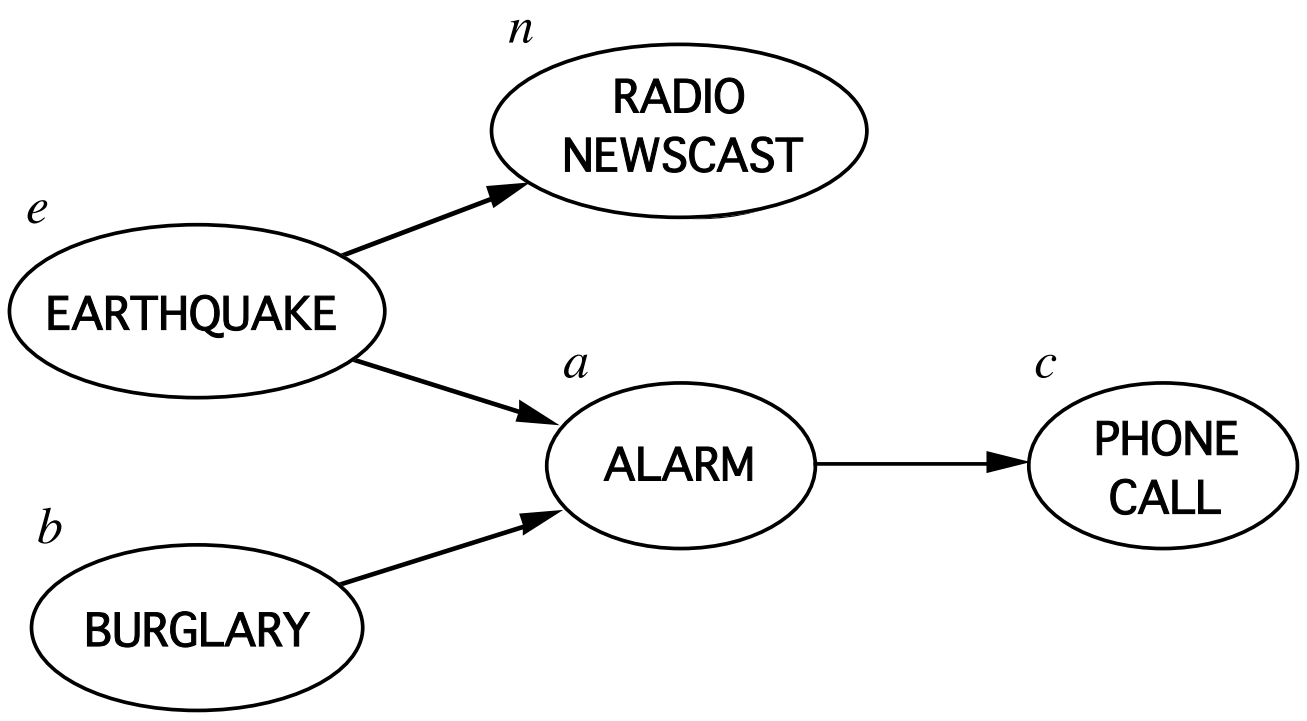

$$p(b_+ | \xi) = 0.003$$

$$p(e_+ | \xi) = 0.001$$

$$p(c_+ | a_-, \xi) = 0.05$$

$$p(c_+ | a_+, \xi) = 0.3$$

$$p(n_+ | e_-, \xi) = 0.00002$$

$$p(n_+ | e_+, \xi) = 0.2$$

$$p(a_+ | b_-, e_-, \xi) = 0.0003$$

$$p(a_+ | b_+, e_-, \xi) = 0.6$$

$$p(a_+ | b_-, e_+, \xi) = 0.5$$

$$p(a_+ | b_+, e_+, \xi) = 0.8$$

**Figure A.10:** A knowledge map for Mr. Holmes' situation.
The nodes in the knowledge map represent the uncertain variables relevant to Mr. Holmes' situation. The node PHONE CALL represents the event (and its negation) that Mr. Holmes received a telephone call from his neighbor reporting the alarm sound. The node ALARM and BURGLARY encode the events that the alarm sounded and that the burglary occurred, respectively. The node RADIO NEWSCAST corresponds to the event that Mr. Holmes heard a radio newscast reporting an earthquake, whereas the node EARTHQUAKE corresponds to the event that the earthquake itself occurred. The lack of arcs between nodes represent assertions of conditional independence. Each node in the knowledge map is associated with a set of probability distributions. These distributions appear below the graph. The variables in the probabilistic expressions correspond to the nodes that they label in the knowledge map. For example, $p\,(b_+|\xi)$ denotes the probability that a burglary has occurred. The figure does not display the probabilities that the events failed to occur. We can compute these probabilities by subtracting the probabilities shown from 1.0.

The probability distributions associated with each node are shown below the knowledge map in Figure A.10. For example, the probability that PHONE CALL occurs, given that ALARM occurs, and given Mr. Holmes' background knowledge $\xi$—denoted $p(c_+|a_+,\xi)$—is 0.3. Some texts refer to the combination of the graph and the probability distributions associated with the nodes in the graph as a knowledge map. In this work, however, we use the term *knowledge map* to refer to only the graph.

The *joint probability distribution* for a set of variables is the collection of probabilities for each instance of that set. A knowledge map determines a unique joint distribution over its variables. In particular, we can construct the joint distribution from the probability distributions associated with each node in the knowledge map, and from the assertions of conditional independence reflected by the lack of arcs in the knowledge map.

Let us again consider Mr. Holmes' situation. From repeated application of the product rule, we know that the probability that $e_+$, $b_+$, $a_+$, $n_+$, and $c_+$ will occur is given by

$$\begin{aligned} p(e_+,b_+,a_+,n_+,c_+|\xi) \quad = \quad & p(e_+|\xi)\cdot \\ & p(b_+|e_+,\xi)\cdot \\ & p(a_+|e_+,b_+,\xi)\cdot \\ & p(n_+|e_+,b_+,a_+,\xi)\cdot \\ & p(c_+|e_+,b_+,a_+,n_+,\xi) \end{aligned} \tag{A.3.12}$$

We obtain a similar equality for each instance of the five variables. In this book, we represent the collection of these equalities by omitting the subscripts on the variables. Thus, we write

$$p(e,b,a,n,c|\xi) = p(e|\xi)\ \ p(b|e,\xi)\ \ p(a|e,b,\xi)\ \ p(n|e,b,a,\xi)\ \ p(c|e,b,a,n,\xi) \tag{A.3.13}$$

The lack of an arc between EARTHQUAKE to BURGLARY implies that these two nodes are independent. Formally, we have

$$p(b|e,\xi) = p(b|\xi) \tag{A.3.14}$$

From the lack of other arcs in the knowledge map (see Chapter 3), we obtain

$$p(n|e,b,a,\xi) = p(n|e,\xi) \tag{A.3.15}$$

$$p(c|e,b,a,,\xi) = p(c|a,\xi) \tag{A.3.16}$$

Combining Equations A.3.13 through A.3.16, we have

$$p(e,b,a,n,c|\xi) = p(e|\xi)\ \ p(b|\xi)\ \ p(a|e,b,\xi)\ \ p(n|e,\xi)\ \ p(c|a,\xi) \tag{A.3.17}$$

The probability distributions on the right-hand side of Equation A.3.17 are exactly those distributions associated with the nodes in the knowledge map.

### A.3.2 Knowledge Maps and Probabilistic Inference

*Probabilistic inference* is the computation—via the rules of probability—of one set of probabilities from another set. Given a joint probability distribution over a set of variables, we can compute any conditional probability that involves those variables. For example, Mr. Holmes undoubtedly wants to determine the probability of BURGLARY given RADIO NEWSCAST and PHONE CALL. Applying the product rule and a generalization of the sum rule to the joint probability distribution for Mr. Holmes' situation, we obtain

$$p(b_+|n_+, c_+, \xi) = \frac{p(b_+, n_+, c_+|\xi)}{p(n_+, c_+|\xi)}$$

$$= \frac{\sum_{e_i, a_k} p(e_i, b_+, a_k, n_+, c_+|\xi)}{\sum_{e_i, b_j, a_k} p(e_i, b_j, a_k, n_+, c_+|\xi)}$$

where $e_i$, $b_j$, and $a_k$ denote arbitrary instances of the variables $e$, $b$, and $a$, respectively.

In the previous section, we saw that we can construct a joint distribution for a set of variables from the knowledge map for those variables. Thus, given a knowledge map for some domain, we can perform any probabilistic inference for that domain by constructing the joint distribution from the knowledge map, and by applying the rules of probability directly to this joint distribution.

We can also perform probabilistic inference directly within a knowledge map. In one such algorithm—developed by Howard, Matheson, Olmsted, and Shachter—we reverse arcs in the knowledge map (Howard and Matheson, 1981; Olmsted, 1983; Shachter, 1988). For example, let us consider the portion of the knowledge map for Mr. Holmes' situation shown in Figure A.11(a). The knowledge map contains a marginal distribution for EARTHQUAKE, and conditional distributions for RADIO NEWSCAST given EARTHQUAKE. We can transform this knowledge map into the one shown in Figure A.11(b) by reversing the arc between the two nodes. To transform the knowledge map, we use Bayes' theorem and the expansion rule. In particular, we obtain

$$p(e_+|n_+, \xi) = \frac{p(n_+|e_+, \xi)\ p(e_+|\xi)}{p(n_+|\xi)}$$

$$= \frac{p(n_+|e_+, \xi)\ p(e_+|\xi)}{p(n_+|e_+, \xi)\ p(e_+|\xi)\ +\ p(n_+|e_-, \xi)\ p(e_-|\xi)}$$

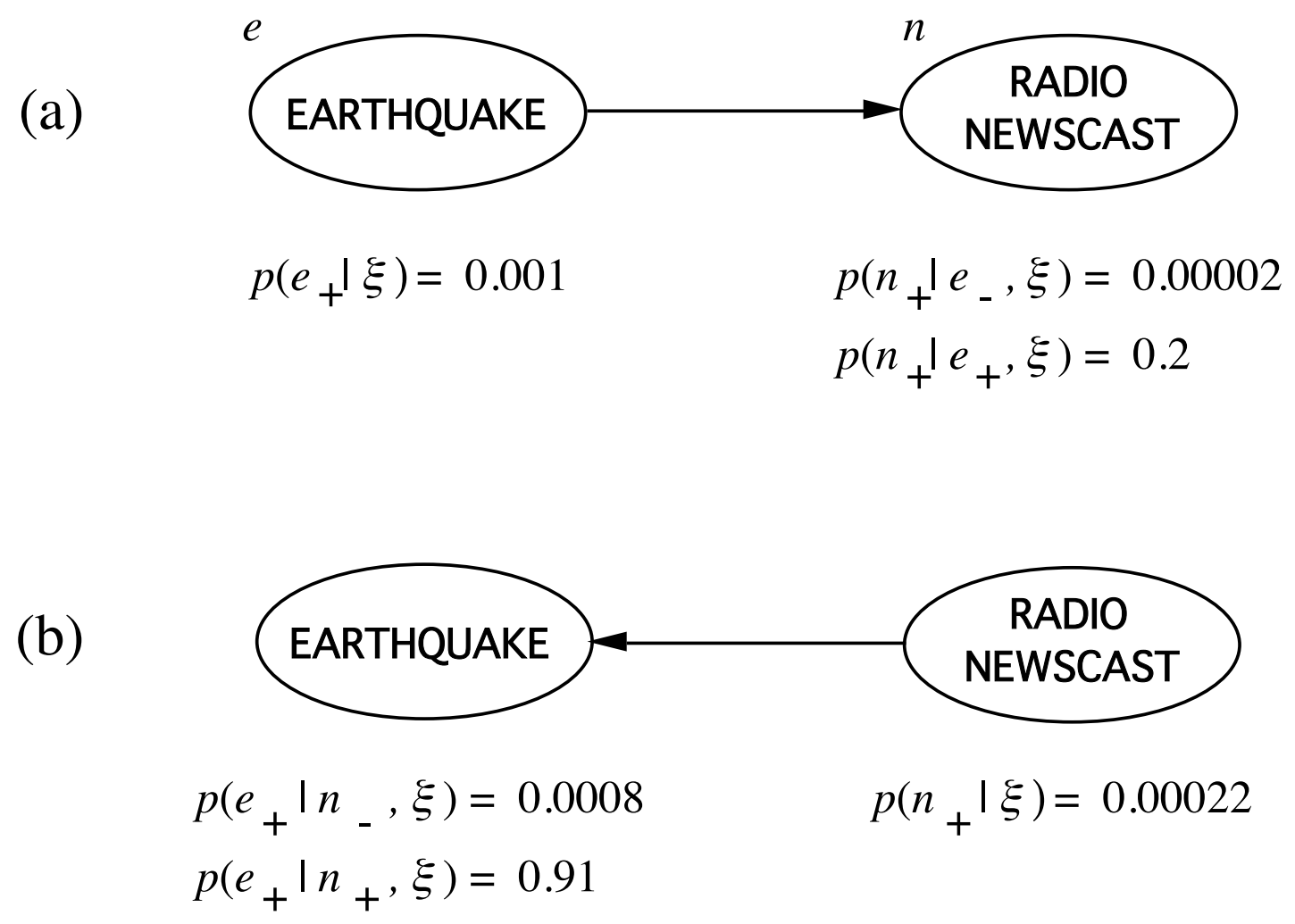

**Figure A.11:** Probabilistic inference as arc reversal in a knowledge map. (a) The arc between EARTHQUAKE and RADIO NEWSCAST is oriented as it is oriented in Figure A.10. (b) Using Bayes' theorem, we can reverse the arc so that it points from RADIO NEWSCAST to EARTHQUAKE. The probability distributions for the transformed knowledge map appear below the graph.

$$= \frac{(0.2)\ (0.001)}{(0.2)\ (0.001)\ +\ (0.00002)\ (0.999)} = 0.91$$

In general, an arc reversal corresponds to an application of Bayes' theorem. Shachter has shown that we can perform any probabilistic inference in a knowledge map with a series of arc reversals (Shachter, 1988). More important, he has shown that the arc-reversal transformation can exploit assertions of conditional independence that are encoded in a knowledge map.

Investigators have created other inference algorithms that exploit assertions of conditional independence in a knowledge map. In these algorithms, the topology of the directed graph remains fixed. For example, Pearl has developed a message-passing scheme that updates the probability distributions for each node in a knowledge map in response to observations of one or more variables (Pearl, 1986). Lauritzen and Spiegelhalter have created an algorithm that first builds an undirected graph from the knowledge map (Lauritzen and Spiegelhalter, 1988). The algorithm then exploits several mathematical properties of undirected graphs to perform probabilistic inference. Most recently, Cooper has developed an inference algorithm that recursively bisects a knowledge map, solves the inference subproblems, and reassembles the component solutions into a global solution (Cooper, 1990a).

Although we can exploit assertions of conditional independence in a knowledge map for probabilistic inference, such inference in an arbitrary knowledge map is an NP-hard task (Cooper, 1990b). Researchers are developing several techniques to circumvent this difficulty. For example, investigators are developing techniques for optimizing inference algorithms for a specific knowledge maps and for specific inference problems within a given knowledge map (Ramamurthi and Agogino, 1988; Shachter et al., 1990; Jensen and Andersen, 1990). In addition, researchers are creating hybrid algorithms that are custom-tailored to particular graph topologies (Heckerman, 1989; Suermondt and Cooper, 1991),[3] and are developing approximate algorithms that are based on stochastic-simulation techniques (Henrion, 1986; Pearl, 1987b; Shachter and Peot, 1989; Chavez and Cooper, 1990).

### A.3.3 Knowledge Maps for Knowledge Acquisition

A knowledge map simplifies knowledge acquisition—the capture and representation of knowledge—by exploiting a fundamental observation about the ability of people to assess probabilities. Namely, a knowledge map takes advantage of the fact that people can make assertions of conditional independence much more easily than they can assess numerical probabilities (Howard and Matheson, 1981; Pearl, 1986). In using a knowledge map, a person first builds the graph that reflects his assertions of conditional independence, and only then does he assess the probabilities underlying the graph. Thus, a knowledge map helps a person to *decompose* the construction of a joint probability distribution into the construction of a set of smaller probability distributions. In the main body of this work, we examine extensions to the knowledge-map representation that also profit from this observation about probability assessment.

Knowledge maps also facilitate the modification of probabilistic knowledge. This observation has important implications with respect to the construction of expert systems that must manage uncertainty. In the 1970s, researchers lauded the production-rule architecture for expert systems, because knowledge bases represented by this architecture were easy to modify. In particular, researchers cited the *modularity* of rules; they noted that rules could be added, deleted, or modified without the meaning of other rules in a knowledge base being affected. The production-rule architecture, however, was inspired by the properties of logical facts. In 1986, Horvitz and I showed that rules in a production system are not modular when they represent knowledge that is *uncertain*. In particular, we demonstrated that the addition, deletion, or modification of such rules within a knowledge base might necessitate the modification of all other rules in that knowledge base. We went on to establish that assertions of conditional independence represent a weaker form of modularity more appropriate to knowledge that is uncertain. Finally, we showed that knowledge maps, because they faithfully represent assertions of conditional independence, greatly simplify the process of modifying a probabilistic knowledge base (Heckerman and Horvitz, 1986; Heckerman and Horvitz, 1987).

[3]See also Section 4.5.

The example in Figure A.11 illustrates another feature of knowledge maps for knowledge acquisition. Namely, a knowledge provider can often choose the order in which he prefers to assess probability distributions. If Mr. Holmes wanted to specify the probability of EARTHQUAKE given RADIO NEWSCAST, he simply would create the knowledge map in Figure A.10, drawing instead an arc from RADIO NEWSCAST to EARTHQUAKE. Regardless of the direction in which Mr. Holmes assesses the conditional distributions, we can reverse arcs in the knowledge map to reveal the conditional probabilities of interest, if the need arises.

### A.3.4 Influence Diagrams

The influence diagram, an extension of the knowledge-map representation, represents the alternatives and preferences of a decision maker in addition to his beliefs. Let us consider the following embellishment to Mr. Holmes' situation, taken from Breese (1990):

> Mr. Holmes believes that, in the event of a burglary, he is more likely to recover his stolen goods if he reports the crime immediately. Therefore, if in fact a burglary has occurred, it is important that he return home as soon as possible. On the other hand, if he rushes home, he will miss an important sales meeting that could result in his earning a substantial commission.

The influence diagram for Mr. Holmes' decision is illustrated in Figure A.12. The oval nodes in the diagram are called *chance nodes*. Chance nodes are the type of nodes that we find in a knowledge map; they represent uncertain variables. The square node GO HOME? is called a *decision node*. This node represents Mr. Holmes' two mutually exclusive and exhaustive alternatives: go home (immediately), or remain at work. The rounded-square node UTILITY is a special type of chance node that represents Mr. Holmes' preferences regarding the possible outcomes of his decision. In general, an influence diagram can contain many decision nodes, but only one preference node.

The node ATTENDANCE AT MEETING in Figure A.12 is a special kind of chance node called a *deterministic node*. This node is distinguished from other chance nodes by its double-oval border. In general, the instance that we observe for a deterministic node $x$ is determined with certainty, given any instance of the nodes that condition $x$. The deterministic node in Figure A.12 reflects the fact that Mr. Holmes will attend the meeting if and only if he does not go home.

An influence diagram contains two types of arcs. Solid arcs are called *conditioning arcs*, and represent probabilistic dependence. For example, in Figure A.12, the arc from GO HOME? to CRIME REPORT represents Mr. Holmes' assertion that the probability of reporting the crime quickly depends on his decision to go home. In general, if an arc points to a chance node, then it must be a conditioning arc. In contrast, arcs that point to decision nodes are *informational arcs*. These arcs appear as dashed lines in Figure A.12. An informational arc from a chance or decision node $x$ to a decision node $y$ indicates that the decision maker knows $x$ at the time decision $y$ is made. The informational arcs in Figure A.12, for example, encode the fact that Mr. Holmes has received the telephone call, and has heard the radio newscast, at the time he must decide whether or not to go home.

We can represent any decision or series of decisions in an influence diagram that we can represent in a decision tree, provided the conditioning and informational arcs (1) form no directed cycles, and (2) reflect a complete ordering of the decisions over time. (Shachter, 1990, discusses the significance of these conditions.) An important advantage of the influence-diagram representation, however, is that we can represent assertions of conditional independence, and thereby make knowledge encoding and manipulation more tractable.

### A.3.5 Computations in Influence Diagrams

There are several computations that we perform commonly in an influence diagram. In this section, we examine two such computations that are relevant to the main body of this work. First, we can *solve the influence diagram*. That is, we can determine the alternative or alternatives in the diagram that maximize the expected utility of the decision maker, and, in the process, determine the expected utility of the decision maker. Several approaches exist for solving an influence diagram. In one approach, we convert the influence diagram to a decision tree, and then solve the tree (Howard and Matheson, 1981; Shachter, 1990). In another approach, we apply probabilistic-inference and expectation computations to the influence diagram directly (Shachter, 1986). Also, we can use any knowledge-map algorithm for probabilistic inference to solve an influence diagram (Cooper, 1988).

Second, we can compute the *value of clairvoyance* or *value of perfect information* for a variable. In general, the value of clairvoyance for $x$ is the largest amount (e.g., in dollars) that the decision maker would be willing to pay to observe $x$ with certainty. For example, let us return to Mr. Holmes' situation.

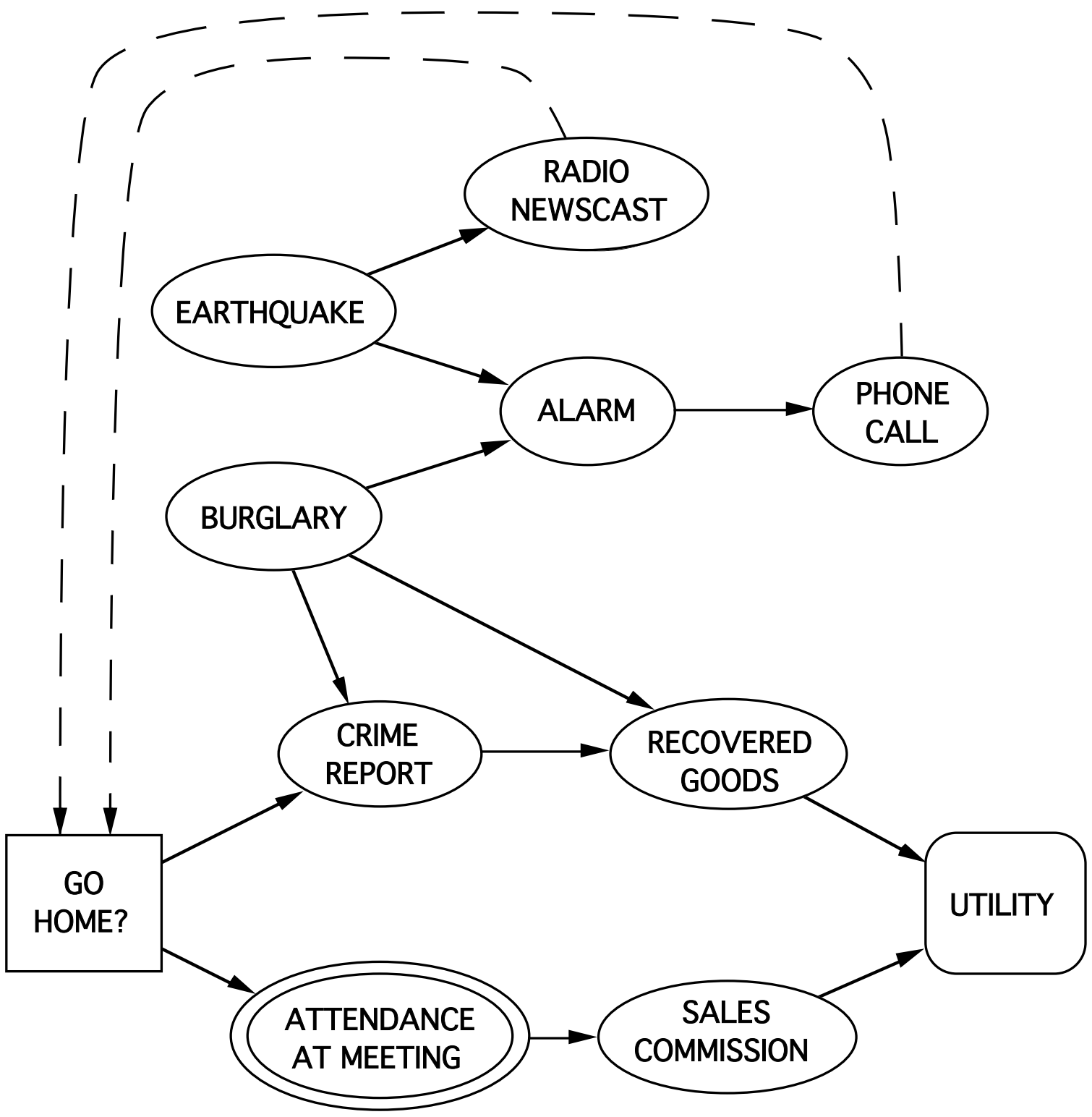

**Figure A.12:** An influence diagram for Mr. Holmes' decision.
The oval nodes, called *chance nodes*, represent the uncertain variables for Mr. Holmes' decision. Some of these nodes appear in Figure A.10. The node CRIME REPORT represents the various times at which Mr. Holmes might report the crime. The node RECOVERED GOODS represents the event (and its negation) that Mr. Holmes will recover his stolen property. ATTENDANCE AT MEETING and SALES COMMISSION correspond to the events that Mr. Holmes attends the meeting and obtains a sales commission, respectively. The node GO HOME?, called a *decision node*, represents Mr. Holmes' mutually exclusive and exhaustive alternatives: go home, or remain at work. The node UTILITY is a special type of chance node that encodes Mr. Holmes' preferences for the possible outcomes of his decision. The double-oval node ATTENDANCE AT MEETING is another special type of chance node called a *deterministic node*; the outcome of ATTENDANCE AT MEETING is a deterministic function of GO HOME?. The solid arcs are conditional arcs. These arcs represent probabilistic dependence, just as they do in a knowledge map. The dashed arcs are informational arcs. These arcs indicate what Mr. Holmes knows at the time that he must make his decision.

> Mr. Holmes remembers that his other next door neighbor, Mr. Watson, works as a security guard at night. He knows that Mr. Watson is always home at this time of day, and he imagines that, for suitable compensation, Mr. Watson would visit the house and determine whether it has been burglarized. Mr. Holmes need only to call Mr. Watson and wait for a return call.

The highest compensation that Mr. Holmes should offer Mr. Watson to check his house is the value of clairvoyance on the variable BURGLARY.

Using Mr. Holmes' influence diagram, we easily can approximate the value of clairvoyance for BURGLARY. First, we solve the influence diagram in Figure A.12 to determine Mr. Holmes' expected utility ($u_1$) if he does not call Mr. Watson. Then, we add an informational arc from BURGLARY to GO HOME? to indicate that Mr. Holmes' knows with certainty whether or not a burglary occurred. We solve this new diagram to determine Mr. Holmes' expected utility ($u_2$) if he does call Mr. Watson. Next, we invert Mr. Holmes' mapping from value to utility to determine the certain equivalents $c_1$ and $c_2$ from $u_1$ and $u_2$, respectively. Finally, we approximate the value of clairvoyance for BURGLARY by the difference between the two certain equivalents ($c_2 - c_1$).

In general, we can approximate the value of clairvoyance for node $x$ in an influence diagram with a single decision node by subtracting the certain equivalent of the current diagram from the certain equivalent of a modified diagram, in which we add an arc from $x$ to the decision node. If there is more than one decision node in the influence diagram, we add the arc from $x$ to the decision node that represents the point in time at which the decision maker is considering how much to pay for the observation of $x$.

The influence-diagram computation of value of clairvoyance is exact, provided the decision maker's utilities satisfy the *delta property.* In terms of lotteries and prizes, the delta property states that an increase in value of all prizes in a lottery by an amount $\triangle$ increases the certain equivalent of that lottery by $\triangle$. The property is illustrated in Figure A.13. Howard discusses the computation of value of clairvoyance when the delta property does not hold (Howard, 1985, page 27).

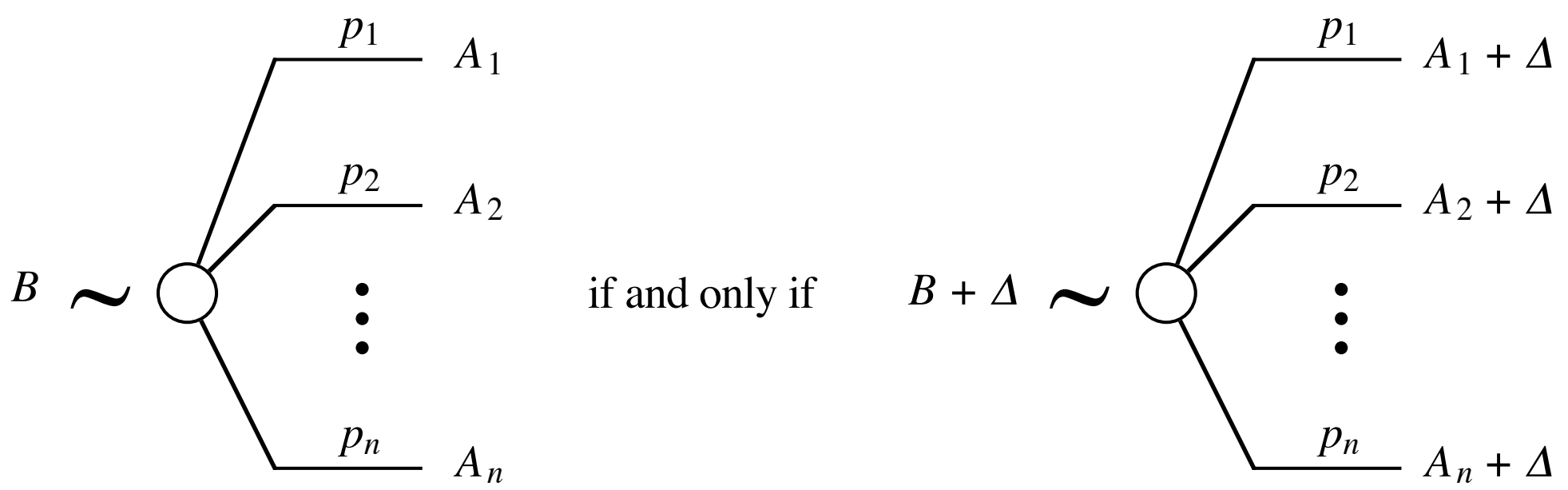

**Figure A.13:** The delta property.
The delta property holds if and only if an increase of all prizes in a lottery by an amount △ increases the certain equivalent of that lottery by △.

# B Proof of Theorems and Notation

## B.1 Proof of Theorem 2.1

As in Chapter 2, let $h$ denote a variable with mutually exclusive and exhaustive instances or hypotheses $h_1, h_2, \ldots h_n$, and let $h_{\subseteq}$ denote a subset of these hypotheses. Recall that, when used in the conditioning part of a probability, $h_{\subseteq}$ denotes the disjunction of its elements.

**Theorem 2.1** The variable $x$ is not relevant to the set of hypotheses $h_{\subseteq}$, given a state of knowledge $\xi$, if and only if

$$p(x|h_i, \xi) = p(x|h_j, \xi) \tag{B.1.1}$$

for all pairs $h_i, h_j \in h_{\subseteq}$.

**Proof:** From Bayes' theorem, we know that

$$p\left(x|h_i, h_{\subseteq}, \xi\right) = \frac{p\left(h_i|x, h_{\subseteq}, \xi\right)}{p\left(h_i|h_{\subseteq}, \xi\right)}\ p\left(x|h_{\subseteq}, \xi\right)$$

for all $h_i$. It follows from Definition 2.1 that $x$ is not relevant to the set $h_{\subseteq}$ if and only if

$$p\left(x|h_i, h_{\subseteq}, \xi\right) = p\left(x|h_{\subseteq}, \xi\right)$$

for all $h_i \in h_{\subseteq}$. Furthermore, because $h_i$ logically implies the disjunction of the elements in $h_{\subseteq}$, $x$ is not relevant to $h_{\subseteq}$ if and only if

$$p\left(x|h_i, \xi\right) = p\left(x|h_{\subseteq}, \xi\right) \tag{B.1.2}$$

for all $h_i \in h_{\subseteq}$. Consequently, if $x$ is not relevant to $h_{\subseteq}$, Equation B.1.2 applies to any pair of hypotheses $h_i, h_j \in h_{\subseteq}$, and Equation B.1.1 follows. Conversely, if Equation B.1.1 holds for every pair of hypotheses in $h_{\subseteq}$, it follows from the product and addition rules for probabilities that

$$\begin{aligned} p\left(x|h_{\subseteq}, \xi\right) &= \frac{p\left(x, h_{\subseteq}|\xi\right)}{p\left(h_{\subseteq}|\xi\right)} \\ &= \frac{\sum_{h_j} p\left(x|h_j, \xi\right)\ p\left(h_j|\xi\right)}{\sum_{h_j} p\left(h_j|\xi\right)} \end{aligned}$$

$$= \frac{p(x|h_i,\xi)\ \sum_{h_j} p(h_j|\xi)}{\sum_{h_j} p(h_j|\xi)}$$

$$= p(x|h_i,\xi)$$

for all $h_i \in h_{\subseteq}$, which implies Equation B.1.2. Hence, $x$ is not relevant to the set $h_{\subseteq}$. $\square$

## B.2 Proof of Theorem 3.7(a)

**Theorem 3.7(a)** The construction of a comprehensive similarity network from a hypothesis-specific similarity network is sound for strictly-positive distributions. The construction remains sound when both networks are minimal.

**Proof:** The proof for nonminimal networks is a special case of the proof for minimal networks. Here, we prove the more general result.

Let $\mathcal{HS}$ be the given hypothesis-specific similarity network, let $\mathcal{C}$ be the comprehensive similarity network constructed from $\mathcal{HS}$, and let $\mathcal{G}_c$ be the c-global map constructed from $\mathcal{C}$. Suppose we have assigned a strictly-positive joint distribution to the variables in $\mathcal{HS}$ that satisfy the assertions of conditional independence and dependence implied by the network, making $\mathcal{HS}$ consistent. (If no such distribution exists, then the constraints of $\mathcal{HS}$ imply trivially the constraints of $\mathcal{C}$.) Let $\leq_E$ be some expansion order that is consistent with $\mathcal{G}_c$. We show that, for all nondistinguished variables $y$ and all c-local maps $h_i$–$h_j$ in $\mathcal{C}$,

$$p\left(y|C^{ij}(y), \{h_i, h_j\}, \xi\right) = p\left(y| \leq_E(y), \{h_i, h_j\}, \xi\right) \tag{B.2.3}$$

for all instances of $C^{ij}(y)$. Furthermore, we show that, for all $x \in C^{ij}(y)$,

$$p\left(y|C^{ij}(y)\backslash x, \{h_i, h_j\}, \xi\right) \neq p\left(y|C^{ij}(y), \{h_i, h_j\}, \xi\right) \tag{B.2.4}$$

for some instance of $C^{ij}(y)$. By Theorems 3.3 and 3.5, it follows that the joint distribution satisfies the assertions of conditional independence and dependence implied by $\mathcal{C}$, and the theorem follows.

First, observe that $\leq_E$ must be consistent with every hs map, provided we omit $h$ from the ordering. Thus, we know that

$$\leq_E(y)\backslash h \subseteq \bar{S}^i(y) \tag{B.2.5}$$

for every hs map $\widehat{h_i}$.

Now, for node $y$ in c-local map $h_i$–$h_j$, consider the following two cases.

i. $h \notin C^{ij}(y)$. Because there is no arc from $h$ to $y$, it follows from the definition of construction $\mathcal{HS}\mapsto\mathcal{C}$ that

$$C^i(y) = C^j(y) \equiv C^{i/j}(y)$$

Also, from the definition of the construction, we know that

$$C^i(y) = C^j(y) = C^{ij}(y)$$

Since the joint distribution satisfies the constraints implied by the hs maps $\widehat{h_i}$ and $\widehat{h_j}$, it follows from Equation B.2.5 and Theorem 3.1 that

$$p\left(y|C^{ij}(y), h_i, \xi\right) = p\left(y| \leq_E(y)\backslash h, h_i, \xi\right) \tag{B.2.6}$$

$$p\left(y|C^{ij}(y), h_j, \xi\right) = p\left(y| \leq_E(y)\backslash h, h_j, \xi\right) \tag{B.2.7}$$

Also, from the relevance set $\mathcal{R}^{ij}$, we have the assertion

$$p\left(y|C^{ij}(y), h_i, \xi\right) = p\left(y|C^{ij}(y), h_j, \xi\right) \tag{B.2.8}$$

From equations B.2.6, B.2.7, and B.2.8 it follows that

$$p\left(y|C^{ij}(y), \{h_i, h_j\}, \xi\right) = p\left(y| \leq_E(y)\backslash h, h, \{h_i, h_j\}, \xi\right) \tag{B.2.9}$$

which is equivalent to Equation B.2.3.

Because the joint distribution satisfies the constraints of $\widehat{h_i}$ and $\widehat{h_j}$, we also know that

$$p\left(y|C^{ij}(y)\backslash x, x_r, h_i, \xi\right) \neq p\left(y|C^{ij}(y)\backslash x, x_s, h_i, \xi\right) \tag{B.2.10}$$

$$p\left(y|C^{ij}(y)\backslash x, x_r, h_j, \xi\right) \neq p\left(y|C^{ij}(y)\backslash x, x_s, h_j, \xi\right) \tag{B.2.11}$$

for some instance of $C^{ij}(y)\backslash x$ and for some instances of $x$, $x_r \neq x_s$. Given Equation B.2.8, however, the two terms on the left-hand sides of Equations B.2.10 and B.2.11 must be equal. Similarly, the two terms on the right-hand side of the equations are equal. Consequently,

$$p\left(y|C^{ij}(y)\backslash x, x_r, \{h_i, h_j\}, \xi\right) \neq p\left(y|C^{ij}(y)\backslash x, x_s, \{h_i, h_j\}, \xi\right) \tag{B.2.12}$$

which is equivalent to Equation B.2.4.

ii. $h \in C^{ij}(y)$. By construction, we have

$$C^{ij}(y) = C^i(y) \cup C^j(y) \cup h$$

Thus, given Equation B.2.5 and Theorem 3.1, we obtain

$$p\left(y|C^{ij}(y)\backslash h, h_i, \xi\right) = p\left(y| \leq_E(y)\backslash h, h_i, \xi\right) \quad \text{(B.2.13)}$$

$$p\left(y|C^{ij}(y)\backslash h, h_j, \xi\right) = p\left(y| \leq_E(y)\backslash h, h_j, \xi\right) \quad \text{(B.2.14)}$$

Together, Equations B.2.13 and B.2.14 imply Equation B.2.3.

To derive Equation B.2.4, consider the cases $x = h$ and $x \neq h$ separately.

1. $x = h$. If $C^i(y) = C^j(y) = C^{ij}(y)\backslash h$, then, by construction, the relevance set $R^{ij}$ must contain the assertion

   $$p\left(y|C^{ij}(y)\backslash h, h_i, \xi\right) \neq p\left(y|C^{ij}(y)\backslash h, h_j, \xi\right) \quad \text{(B.2.15)}$$

   Equation B.2.15 implies

   $$p\left(y|C^{ij}(y), \{h_i, h_j\}, \xi\right) \neq p\left(y|C^{ij}(y)\backslash h, \{h_i, h_j\}, \xi\right) \quad \text{(B.2.16)}$$

   which is equivalent Equation B.2.4.

   If $C^i(y) \neq C^j(y)$, choose node $z$ such that $z \in C^i(y)$ and $z \notin C^j(y)$. From Theorem 3.5, we obtain

   $$p\left(y|C^{ij}(y)\backslash\{z, h\}, z_r, h_i, \xi\right) \neq p\left(y|C^{ij}(y)\backslash\{z, h\}, z_s, h_i, \xi\right) \quad \text{(B.2.17)}$$

   for some instance of $C^{ij}(y)\backslash\{z, h\}$ and for some instances of $z$, $z_r \neq z_s$. However, since $z \notin C^j(y)$, we know from Theorem 3.1 that

   $$p\left(y|C^{ij}(y)\backslash\{z, h\}, z_r, h_j, \xi\right) = p\left(y|C^{ij}(y)\backslash\{z, h\}, z_s, h_j, \xi\right) \quad \text{(B.2.18)}$$

   for all instances of $C^{ij}(y)\backslash\{z, h\}$. Thus, either the two terms on the left-hand sides of Equations B.2.17 and B.2.18 are not equal, or the two terms on the right-hand sides of the equations are not equal. In either case, we get Equation B.2.4. This observation proves that if $C^i(y) \neq C^j(y)$, then $y$ must be relevant to $\{h_i, h_j\}$, given some instance of $C^i(y) \cup C^j(y)$. Thus, we do not include assertions of subset independence and dependence in a relevance set $\mathcal{R}^{ij}$, when $C^i(y) \neq C^j(y)$.

2. $x \neq h$. By construction, we know that $x \in C^i(y)$ or $x \in C^j(y)$. If $x \in C^i(y)$, Theorem 3.5 implies

   $$p\left(y|C^{ij}(y)\backslash\{x, h\}, h_i, \xi\right) \neq p\left(y|C^{ij}(y)\backslash h, h_i, \xi\right)$$

   for some instance of $C^{ij}(y)\backslash h$, which is equivalent to Equation B.2.4. Similarly, we obtain Equation B.2.4 when $x \in C^j(y)$. □

Note that the assumption of strict positivity was used in the proof only to show that the assertions of conditional *dependence* are preserved by the construction. Thus, for nonminimal networks, the construction of the c-local maps from the hs maps is sound for *all* distributions.

## B.3 Proof of Theorem 3.7(b)

**Theorem 3.7(b)** The construction of an c-global map from a comprehensive similarity network is sound for strictly-positive distributions. The construction remains sound when both representations are minimal.

**Proof:** Let $\mathcal{C}$ be the given comprehensive similarity network, and let $\mathcal{G}_c$ be the c-global map constructed from $\mathcal{C}$. Suppose we have assigned a strictly-positive joint distribution to the variables in $\mathcal{C}$ that satisfies the constraints of the c-local maps. (If no such distribution exists, then the constraints of $\mathcal{C}$ imply trivially the constraints of $\mathcal{G}_c$.) To prove the result, we must show that, for all nondistinguished variables $y$,

$$p\left(y|C^{\mathcal{G}_c}(y), \xi\right) = p\left(y|\bar{S}^{\mathcal{G}_c}(y), \xi\right) \tag{B.3.19}$$

for all instances of $C^{\mathcal{G}_c}(y)$. If the given network is minimal, we must also show that for every $x \in C^{ij}(y)$,

$$p\left(y|C^{\mathcal{G}_c}(y)\backslash x, \xi\right) \neq p\left(y|C^{\mathcal{G}_c}(y), \xi\right) \tag{B.3.20}$$

for some instance of $C^{\mathcal{G}_c}(y)$. We need to consider only nondistinguished variables, because $h$ has no nonsuccessors in the c-global map. (If there is a node completely disconnected from $h$ in the c-global map, it is irrelevant to the diagnosis of $h$, and it can be omitted from consideration.)

To begin, observe that, for each c-local map $h_i$–$h_j$,

$$C^{ij}(y) \subseteq C^{\mathcal{G}_c}(y) \quad \text{and} \quad \bar{S}^{\mathcal{G}_c}(y) \subseteq \bar{S}^{ij}(y) \tag{B.3.21}$$

by construction. Thus, because the joint distribution satisfies the constraints of each c-local map, it follows from Theorem 3.1 that

$$p\left(y|C^{\mathcal{G}_c}(y), \{h_i, h_j\}, \xi\right) = p\left(y|\bar{S}^{\mathcal{G}_c}(y), \{h_i, h_j\}, \xi\right) \tag{B.3.22}$$

for every pair of hypotheses $h_i$ and $h_j$ that is directly connected in the similarity graph. To see that this fact implies Equation B.3.19, consider the following two cases.

i. $h \in C^{\mathcal{G}_c}(y)$. In this case, Equation B.3.22 becomes

$$p\left(y|C^{\mathcal{G}_c}(y)\backslash h, h, \{h_i, h_j\}, \xi\right) = p\left(y|\bar{S}^{\mathcal{G}_c}(y)\backslash h, h, \{h_i, h_j\}, \xi\right) \tag{B.3.23}$$

Since $h_i$ or $h_j$ alone logically implies the disjunction of $h_i$ and $h_j$, we can rewrite Equation B.3.23 to be

$$p\left(y|C^{\mathcal{G}_c}(y)\backslash h, h, \xi\right) = p\left(y|\bar{S}^{\mathcal{G}_c}(y)\backslash h, h, \xi\right), \quad h = h_i, h_j \tag{B.3.24}$$

Because the similarity graph is connected, there is a c-local map associated with every hypothesis, and hence

$$p\left(y|C^{\mathcal{G}_c}(y)\backslash h, h_i, \xi\right) = p\left(y|\bar{S}^{\mathcal{G}_c}(y)\backslash h, h_i, \xi\right), \quad \forall\ h_i \tag{B.3.25}$$

which is equivalent to Equation B.3.19.

ii. $h \notin C^{\mathcal{G}_c}(y)$. From Equation B.3.22, we get

$$p\left(y|C^{\mathcal{G}_c}(y), \{h_i, h_j\}, \xi\right) = p\left(y|\bar{S}^{\mathcal{G}_c}(y)\backslash h, h, \{h_i, h_j\}, \xi\right) \tag{B.3.26}$$

However, because there is no arc from $h$ to $y$, it follows from Theorem 3.1 that

$$p\left(y|C^{\mathcal{G}_c}(y), h, \{h_i, h_j\}, \xi\right) = p\left(y|C^{\mathcal{G}_c}(y), \{h_i, h_j\}, \xi\right) \tag{B.3.27}$$

Consequently, we can rewrite Equation B.3.26 to obtain

$$p\left(y|C^{\mathcal{G}_c}(y), h, \{h_i, h_j\}, \xi\right) = p\left(y|\bar{S}^{\mathcal{G}_c}(y)\backslash h, h, \{h_i, h_j\}, \xi\right) \tag{B.3.28}$$

Furthermore, since $h_i$ or $h_j$ alone logically implies the disjunction of $h_i$ and $h_j$, Equation B.3.28 becomes

$$p\left(y|C^{\mathcal{G}_c}(y), h, \xi\right) = p\left(y|\bar{S}^{\mathcal{G}_c}(y)\backslash h, h, \xi\right), \quad h = h_i, h_j \tag{B.3.29}$$

and we obtain

$$p\left(y|C^{\mathcal{G}_c}(y), h_i, \xi\right) = p\left(y|\bar{S}^{\mathcal{G}_c}(y)\backslash h, h_i, \xi\right), \quad \forall\ h_i \tag{B.3.30}$$

because there is a c-local map associated with every hypothesis.

Now from Equation B.3.27, it follows that

$$p\left(y|C^{\mathcal{G}_c}(y), h_i, \xi\right) = p\left(y|C^{\mathcal{G}_c}(y), h_j, \xi\right) \tag{B.3.31}$$

for any hypothesis pair $h_i$ and $h_j$ that is directly connected in the similarity graph. However, Equation B.3.31 must also hold for *any* hypothesis pair in the network, whether or not that hypothesis pair is directly connected, because the similarity graph is connected and the joint distribution is strictly positive (see the discussion that follows the statement of the theorem in Chapter 3). Thus, we obtain

$$p\left(y|C^{\mathcal{G}_c}(y), h_i, \xi\right) = p\left(y|C^{\mathcal{G}_c}(y), \xi\right), \quad \forall\ h_i \tag{B.3.32}$$

Together, Equations B.3.30 and B.3.32 imply Equation B.3.19.

Now suppose the given comprehensive network is minimal. If there is an arc from $x$ to $y$ in the c-global map, where $x$ is either a nondistinguished node or the distinguished node $h$, then there must be a corresponding arc in some c-local map—say, $h_i$–$h_j$—by construction. In this case, because the joint distribution is strictly positive, it follows from Equation B.3.21 and Theorem 3.5 that

$$p\left(y|C^{\mathcal{G}_c}(y)\backslash x, \{h_i, h_j\}, \xi\right) \neq p\left(y|C^{\mathcal{G}_c}(y)\{h_i, h_j\}, \xi\right) \tag{B.3.33}$$

for some instance of $C^{\mathcal{G}_c}(y)$. If $x = h$, Equation B.3.33 becomes

$$p\left(y|C^{\mathcal{G}_c}(y)\backslash h, \{h_i, h_j\}, \xi\right) \neq p\left(y|C^{\mathcal{G}_c}(y), \{h_i, h_j\}, \xi\right) \tag{B.3.34}$$

for some instance of $C^{\mathcal{G}_c}(y)$ where $h = h_i$ or $h_j$. Consequently, it follows that

$$p\left(y|C^{\mathcal{G}_c}(y)\backslash h, h_i, \xi\right) \neq p\left(y|C^{\mathcal{G}_c}(y)\backslash h, h_j, \xi\right) \tag{B.3.35}$$

for some instance of $C^{\mathcal{G}_c}(y)\backslash h$. If $x \neq h$, Equation B.3.33 becomes

$$p\left(y|C^{\mathcal{G}_c}(y)\backslash \{x, h\}, h, \xi\right) \neq p\left(y|C^{\mathcal{G}_c}(y)\backslash h, h, \xi\right) \tag{B.3.36}$$

for some instance of $C^{\mathcal{G}_c}(y)\backslash h$ where $h = h_i$ or $h_j$. In either case, we obtain Equation B.3.20. □

## B.4 Proof of Theorem 3.8

**Theorem 3.8** A hypothesis-specific similarity network is consistent if and only if there is no cycle in the similarity graph such that, for any nondistinguished node $y$, the assertion

$$p\left(y|C^{i/j}(y), h_i, \xi\right) = p\left(y|C^{i/j}(y), h_j, \xi\right) \tag{B.4.37}$$

is in all but one relevance set $\mathcal{R}^{ij}$ in the cycle.

**Proof** (only if): Consider the cycle defined by the c-local maps $h_1$–$h_2$, $h_2$–$h_3$, $\ldots h_n$–$h_1$, where $n > 2$. Suppose the assertion

$$p\left(y|C^{i/i+1}(y), h_i, |\xi\right) = p\left(y|C^{i/i+1}(y), h_{i+1}, |\xi\right) \tag{B.4.38}$$

is in the relevance set $\mathcal{R}^{i,i+1}$ for only $i = 1, 2, \ldots n{-}1$. By definition of hypothesis-specific similarity networks, it follows that

$$C(y) = C^1(y) = C^2(y) = \cdots = C^n(y)$$

Consequently, the assertion

$$p\left(y|C^{1/n}(y), h_1, \xi\right) = p\left(y|C^{1/n}(y), h_n, \xi\right) \tag{B.4.39}$$

must be in the relevance set $\mathcal{R}^{1n}$, a contradiction.

**Proof** (if): The proof for nonminimal networks is a special case of the proof for minimal networks. So let us suppose that the given network is minimal. If there are no cycles in the similarity graph, we can always find a joint distribution over the variables in the network that satisfies the constraints in each hs map. First, choose any $h_i$ in the similarity graph. Assign distributions $p\left(y|C^i(y), h_i, \xi\right)$ to each node $y$ in the hs map $\widehat{h_i}$. Use Theorem 3.5 to be sure that the distributions imply that every arc from $x$ to $y$, $x \in C^i(y)$, is nonsuperfluous. Second, choose any $h_j$ that is directly connected to $h_i$ in the similarity graph. If $C^i(y) \neq C^j(y)$, choose distributions for $y$ in $\widehat{h_j}$ in the same manner as they were chosen for $\widehat{h_i}$. If $C^i(y) = C^j(y)$ and the assertion

$$p\left(y|C^{i/j}(y), h_i, \xi\right) \neq p\left(y|C^{i/j}(y), h_j, \xi\right)$$

is in the relevance set $\mathcal{R}^{ij}$, select distributions for $y$ that satisfy both this assertion and the constraints of Theorem 3.5. If $C^i(y) = C^j(y)$ and the assertion

$$p\left(y|C^{i/j}(y), h_i, \xi\right) = p\left(y|C^{i/j}(y), h_j, \xi\right)$$

is in the relevance set $\mathcal{R}^{ij}$, copy the distributions $p\left(y|C^{i/j}(y), h_i, \xi\right)$ to $p\left(y|C^{i/j}(y), h_j, \xi\right)$. Finally, repeat the second step of the assignment process, traversing each arc in the similarity graph.

If there are cycles in the similarity graph, add the following "look-ahead" procedure to the second step. Before assigning distributions to $y$ in $\widehat{h_j}$, determine whether there is a path from $h_j$ to some $h_k$ in the similarity graph such that the distributions for $y$ in $\widehat{h_k}$ have been assigned, and such that the assertion

$$p\left(y|C^{m/n}(y), h_m, \xi\right) = p\left(y|C^{m/n}(y), h_n, \xi\right)$$

is in every relevance set $\mathcal{R}^{mn}$ along the path. If such a path exists, copy $p\left(y|C^k(y), h_k, \xi\right)$ to $p\left(y|C^j(y), h_j, \xi\right)$ for every instance of $C^k(y) = C^j(y)$. Otherwise, select distributions for $y$ as described in the previous paragraph. Because Equation B.4.37 never holds for only one relevance set, no conflicting assignments can be generated. □

## B.5 Proof of Theorem 3.10

**Algorithm 3.1 (Consistency, comprehensive networks)**

1 For every pair of nondistinguished nodes $x$ and $y$ such that $x \longrightarrow y$ in the c-global map constructed from the given network do
2     For every c-local map $h_i$–$h_j$ such that $x \not\longrightarrow y$ do
3         Post the constraint "$x \not\longrightarrow y$" on $h_i$ and on $h_j$
4     For every c-local map $h_i$–$h_j$ such that $x \longrightarrow y$ do
5         If $h \longrightarrow y$ and "$x \not\longrightarrow y$" is posted on $h_i$ and on $h_j$ then
6             Return "inconsistent"
7         Else if $h \not\longrightarrow y$ and "$x \not\longrightarrow y$" is posted on $h_i$ or on $h_j$ then
8             Return "inconsistent"
9     For every hypothesis $h_i$ do
10         If the constraint "$x \not\longrightarrow y$" is not posted on $h_i$ then
11             Add $x \longrightarrow y$ to $\widehat{h_i}$
12 For every nondistinguished node $y$ in the c-global map do
13     For every c-local map $h_i$–$h_j$ where $h \longrightarrow y$ do
14         From the similarity graph, construct the edge-induced subgraph, $\mathcal{G}$, containing edge $(h_i, h_j)$, and edges $(h_k, h_l)$ such that $h \not\longrightarrow y$ in $h_k$–$h_l$
15         If the edge $(h_i, h_j)$ is in a cycle in $\mathcal{G}$
16             Return "inconsistent"
17 For every nondistinguished node $y$ in the c-global map do
18     For every c-local map $h_i$–$h_j$ such that $C^i(y) = C^j(y) \equiv C^{i/j}(y)$ do
19         If $h \not\longrightarrow y$ then
20             Add "$p\left(y|C^{i/j}(y), h_i, \xi\right) = p\left(y|C^{i/j}(y), h_j, \xi\right)$" to $\mathcal{R}^{ij}$
21         Else if $h \longrightarrow y$ then
22             Add "$p\left(y|C^{i/j}(y), h_i, \xi\right) \neq p\left(y|C^{i/j}(y), h_j, \xi\right)$" to $\mathcal{R}^{ij}$
23 Return "consistent"

**Theorem 3.10 (Consistency, comprehensive networks)** Algorithm 3.1 applied to a comprehensive similarity network returns "consistent" if and only if there is a strictly-positive joint distribution that makes the network consistent and minimal. Moreover, if Algorithm 3.1 returns "consistent," it generates the hypothesis-specific network that is the maximal constructor of the given network.

**Proof** (if): If the algorithm returns "inconsistent" within the for-loop beginning at line 1, it follows from Corollary 3.1 that the network is inconsistent for strictly-positive distributions. If the algorithm returns "inconsistent" within the for-loop beginning at line 12, then there is some cycle in the similarity graph such that there is an arc from $h$ to $y$ in exactly one c-local map of that cycle. Thus, by Corollary 3.2. the network is inconsistent.

**Proof** (only if): Suppose Algorithm 3.1 returns "consistent." Since the for-loop beginning at line 12 does not return "inconsistent," there is no cycle in the similarity graph such that there is an arc from $h$ to $y$ in exactly one c-local map of the cycle. Consequently, by Theorem 3.8, the hypothesis-specific network created by the algorithm is consistent. Thus, we need only to show that the given network can be constructed from this hypothesis-specific network. It then follows from soundness of the construction (Theorem 3.7a) that the comprehensive network is consistent and minimal.

The structure among nondistinguished nodes in each c-local map is correctly reproduced by the hs maps created by the algorithm. If there is an arc from $x$ to $y$ in the c-local map $h_i$–$h_j$, Algorithm 3.1 guarantees that there is such an arc in either $\widehat{h_i}$ or $\widehat{h_j}$. Therefore, the c-local map $h_i$–$h_j$ constructed from $\widehat{h_i}$ and $\widehat{h_j}$ contains the arc. Conversely, if there is no arc from $x$ to $y$ in the c-local map $h_i$–$h_j$, Algorithm 3.1 guarantees that there is no such arc in either $\widehat{h_i}$ or $\widehat{h_j}$. It follows that the c-local map $h_i$–$h_j$ constructed from $\widehat{h_i}$ and $\widehat{h_j}$ does not contain the arc. Also, by construction, the arcs and missing arcs from $h$ to nondistinguished nodes in the c-local maps are correctly reproduced by the relevance sets created by the algorithm.

**Proof** (maximal constructor): Algorithm 3.1 places an arc from $x$ to $y$ in the hs map $\widehat{h_i}$ whenever there is such an arc in the c-global map constructed from the given network and whenever there is no constraint derived from the c-local maps that makes the addition impossible. Since there can be no arc in any hs map unless there is a corresponding arc in the c-global map, it follows that the hypothesis-specific similarity network generated by the algorithm is a maximal constructor of the comprehensive network. □

## B.6 Proof of Theorem 3.12

**Algorithm 3.2 (Consistency, ordinary networks)**

For every pair of nondistinguished nodes $x$ and $y$ such that $x \longrightarrow y$ in the o-global map constructed from the given network do
    For every o-local map $h_i$–$h_j$ such that $x \not\longrightarrow y$ do
        Post the constraint "$x \not\longrightarrow y$" on $h_i$ and on $h_j$
    For every o-local map $h_i$–$h_j$ such that only one of $x$ and $y$ is on the map do
        Post the constraint "$x \not\longrightarrow y$" on $h_i$ and on $h_j$

    Mark all o-local maps as unvisited
    While there is an unvisited o-local map containing neither $x$ nor $y$ such that the constraint "$x \not\longrightarrow y$" is posted on $h_i$ or on $h_j$ do
        Post the constraint "$x \not\longrightarrow y$" on $h_i$ and on $h_j$
        Mark the o-local map as visited

    For every o-local map $h_i$–$h_j$ such that $x \longrightarrow y$ do
        If $h \longrightarrow y$ and "$x \not\longrightarrow y$" is posted on $h_i$ and on $h_j$ then
            Return "inconsistent"
        If $h \not\longrightarrow y$ and "$x \not\longrightarrow y$" is posted on $h_i$ or on $h_j$ then
            Return "inconsistent"

    For every hypothesis $h_i$ do
        If the constraint "$x \not\longrightarrow y$" is not posted on $h_i$ then
            Add $x \longrightarrow y$ to $\widehat{h_i}$

For every nondistinguished node $y$ in the c-global map do
    For every c-local map $h_i$–$h_j$ where $h \longrightarrow y$ do
        From the similarity graph, construct the edge-induced subgraph, $\mathcal{G}$, containing edge $(h_i, h_j)$, and edges $(h_k, h_l)$ such that $h \not\longrightarrow y$ in $h_k$–$h_l$
        If the edge $(h_i, h_j)$ is in a cycle in $\mathcal{G}$
            Return "inconsistent"

For every nondistinguished node $y$ in the c-global map do
    For every c-local map $h_i$–$h_j$ such that $C^i(y) = C^j(y) \equiv C^{i/j}(y)$ do

```
25          If $h \not\rightarrow y$ then
26              Add "$p\left(y|C^{i/j}(y), h_i, \xi\right) = p\left(y|C^{i/j}(y), h_j, \xi\right)$" to $\mathcal{R}^{ij}$
27          Else if $h \longrightarrow y$ then
28              Add "$p\left(y|C^{i/j}(y), h_i, \xi\right) \neq p\left(y|C^{i/j}(y), h_j, \xi\right)$" to $\mathcal{R}^{ij}$

29      Return "consistent"
```

**Theorem 3.12 (Consistency, ordinary networks)** Algorithm 3.2 applied to an ordinary similarity network returns "consistent" if and only if there is a strictly-positive distribution that makes the network consistent and minimal. Moreover, if Algorithm 3.2 returns "consistent," it generates the hypothesis-specific network that is the maximal constructor of the given network.

**Proof** (if): Let $\mathcal{O}$ be the given ordinary similarity network. Suppose some strictly-positive joint distribution satisfies the constraints implied by the o-local maps of $\mathcal{O}$. From the definition of ordinary similarity networks, we know that the joint distribution satisfies the constraints of some minimal comprehensive similarity network from which we can construct $\mathcal{O}$. Call this comprehensive network $\mathcal{C}$. Given Theorem 3.10, we know that Algorithm 3.1 applied to $\mathcal{C}$ returns "consistent." We use this fact to show that Algorithm 3.2 applied to $\mathcal{O}$ must also return "consistent."

First, observe that the o-global constructed from $\mathcal{O}$ and the c-global constructed from $\mathcal{C}$ are identical, because $\mathcal{O}$ is consistent. Thus, the main for-loops in Algorithms 3.1 and 3.2 iterate over the same set of arcs.

Second, observe that if the constraint "$x \not\rightarrow y$" is posted on $h_i$ by Algorithm 3.2 applied to $\mathcal{O}$, then the constraint is posted on $h_i$ by Algorithm 3.1 applied to $\mathcal{C}$. In particular, Algorithm 3.2 posts no more constraints than does Algorithm 3.1. To derive this observation, consider separately the case where the constraint is posted by line 3 or 5 of Algorithm 3.2 and the case where the constraint is posted by line 8 of the algorithm. In the first case, the observation follows immediately from the definition of o-local maps. In the second case, we can prove the observation using an inductive argument. Consider the first time within the while-loop of Algorithm 3.2 that the constraint "$x \not\rightarrow y$" is posted on some $h_i$ by line 8 of the algorithm. If the algorithm is checking the o-local map $h_i$–$h_j$, we know that the constraint "$x \not\rightarrow y$" is already posted on $h_j$ by both Algorithms 3.1 and 3.2. Furthermore, we know that neither $x$ nor $y$ are on the o-local map and, hence, that there is no arc from $h$ to $y$ on the c-local map $h_i$–$h_j$. It follows that there can be no arc from $x$ to $y$ on the c-local map and Algorithm 3.1 must post the constraint "$x \not\rightarrow y$" on $h_i$. Applying this argument to each subsequent constraint posted by line 8 of Algorithm 3.2, we find that an identical constraint must also be posted by Algorithm 3.1.

Third, observe that Algorithm 3.2 applied to $\mathcal{O}$ cannot return "inconsistent" at line 12 or 14. This observation follows from the previous observation, from the fact that Algorithm 3.1 applied to $\mathcal{C}$ returns "consistent," and from the fact that an arc in $\mathcal{C}$ from $x$ to $y$ is considered by Algorithm 3.1 in line 4 only if the corresponding arc in $\mathcal{O}$ is considered by Algorithm 3.2 in line 10.

Finally, observe that there is an arc from $h$ to $y$ in an o-local map if and only if there is such an arc in that o-local map's corresponding c-local map. Therefore, line 22 of Algorithm 3.2 cannot return "inconsistent" when applied to $\mathcal{O}$, because Algorithm 3.1 applied to $C$ returns "consistent."

**Proof** (only if): We know that Algorithm 3.2 applied to $\mathcal{O}$ returns "consistent." So let $\mathcal{HS}$ be the hypothesis-specific similarity network created by the algorithm, and let $\mathcal{C}$ be the comprehensive similarity network constructed from $\mathcal{HS}$. We show that Algorithm 3.1 applied to $\mathcal{C}$ returns "consistent." It then follows from Theorem 3.10 that $\mathcal{O}$ is consistent and minimal.

First, observe that, by construction, the c-global of $\mathcal{C}$ is identical in structure to the o-global of $\mathcal{O}$. Consequently, the main for-loops in Algorithms 3.2 and 3.1 iterate over the same set of arcs.

Second, observe that, if the constraint "$x \not\rightarrow y$" is posted on $h_i$ by Algorithm 3.1 applied to $\mathcal{C}$, then the constraint is posted on $h_i$ by Algorithm 3.2 applied to $\mathcal{O}$. To see this fact, let us suppose that there is an arc from $x$ to $y$ on the (c- or o-) global map, and that the constraint "$x \not\rightarrow y$" is not posted on $h_i$ by Algorithm 3.2 applied to $\mathcal{O}$. In this case, the for-loop at line 15 of Algorithm 3.2 guarantees that there is an arc from $x$ to $y$ on the hs map $\widehat{h_i}$. Consequently, there is an arc from $x$ to $y$ on all c-local maps that are bordered by $h_i$, and the constraint "$x \not\rightarrow y$" is not posted by Algorithm 3.1 when applied to $\mathcal{C}$.

Third, observe that neither line 6 nor line 8 of Algorithm 3.1 returns "inconsistent" when applied to $\mathcal{C}$. If $x$ and $y$ are both connected to $h$ in the c-local map $h_i$–$h_j$, this observation follows from the previous observation and from the fact that Algorithm 3.2 applied to $\mathcal{O}$ returns "consistent." If only one of $x$ and $y$ is connected to $h$ in $h_i$–$h_j$, there can be no arc from $x$ to $y$. In this case, it is impossible for Algorithm 3.1 to return "inconsistent." If both $x$ and $y$ are disconnected from $h$ in the c-local map $h_i$–$h_j$, Algorithm 3.2 guarantees that there is an arc from $x$ to $y$ in $\widehat{h_i}$ or $\widehat{h_j}$ only if the constraint "$x \not\rightarrow y$" is not posted on either $h_i$ or $h_j$. Thus, Algorithm 3.1 returns "inconsistent" when considering this arc.

Finally, observe that line 16 of Algorithm 3.1 does not return "inconsistent" when applied to $\mathcal{C}$ because Algorithm 3.2 applied to $\mathcal{O}$ returns "consistent."

**Proof** (maximal constructor): By construction, Algorithm 3.2 places an arc from $x$ to $y$ in the hs map $\widehat{h_i}$ whenever there is such an arc in the c-global map and whenever there is no constraint derived from the o-local maps that makes the addition impossible. It follows that the hypothesis-specific similarity network generated by the algorithm is a maximal constructor of the ordinary network. $\square$

# C Glossary of Pathfinder Terms

## C.1 Diseases of the Lymph Node

AIDS EARLY: AIDS, early phase

AIDS INVOLUTIONARY: AIDS, involutionary phase

AILD: Angio-immunoblastic lymphadenopathy

ALIP: Atypical lymphoplasmacytic and immunoblastic proliferation

AML: Acute myeloid leukemia

B-IMMUNOBLASTIC: Immunoblastic plasmacytoid diffuse lymphoma

CARCINOMA: Carcinoma

CAT SCRATCH DISEASE: Cat-scratch disease

CELLULAR PHASE NSHD: Cellular phase of nodular sclerosis Hodgkin's disease

DERMATOPATHIC LADEN: Dermatopathic lymphadenitis

DIFFUSE FIBROSIS HD: Diffuse fibrosis Hodgkin's disease

EM PLASMACYTOMA: Extramedullary plasmacytoma

FLORID FOLLIC HYPERP: Florid reactive follicular hyperperplasia

GLH HYALINE VACULAR: Giant lymph-node hyperplasia, hyaline vacular type

GLH PLASMA CELL TYPE: Giant lymph-node hyperplasia, plasma-cell type

GRANULOMATOUS LADEN: Granulomatous lymphadenitis

HAIRY CELL LEUKEMIA: Hairy cell leukemia

HISTIOCYTOSIS X: Histiocytosis x

IBL-LIKE T-CELL LYM: Immunoblastic lymphadenopathy-like T-cell lymphoma

INFECTIOUS MONO: Infectious mononucleosis

INTERFOLLICULAR HD: Interfollicular Hodgkin's disease

JAPANESE ATL: Japanese adult T-cell lymphoma

KAPOSIS SARCOMA: Kaposis sarcoma

L&H DIFFUSE HD: Lymphocytic and histiocytic diffuse Hodgkin's disease

L&H NODULAR HD: Lymphocytic and histiocytic nodular Hodgkin's disease

LARGE CELL, DIF: Large cell diffuse lymphoma

LARGE CELL, FOL: Large cell follicular lymphoma

LEPROSY: Leprosy

LYMPHANGIOGRAPHIC: Lymphangiography effect

LYMPHOBLASTIC: Lymphoblastic lymphoma

MALIG HISTIOCYTOSIS: Malignant histiocytosis

MANTLE ZONE: Mantle-zone lymphoma

MANTLE ZONE HYPERL: Mantle-zone hyperplasia

MAST CELL DISEASE: Mast-cell disease

MELANOMA: Melanoma

MIXED CELLULARITY HD: Mixed-cellularity Hodgkin's disease

MIXED, FCC DIF: Mixed (follicular center cell type) diffuse lymphoma

MIXED, FOL: Mixed (follicular center cell type) follicular lymphoma

MULTIPLE MYELOMA: Multiple myeloma

MYCOSIS FUNGOIDES: Mycosis fungoides

NECROTIZ NONKIKUCHI: NonKikuchi's necrotizing lymphadenitis

NECROTIZING KIKUCHI: Kikuchi's necrotizing lymphadenitis

NODULAR SCLEROSIS HD: Nodular sclerosis Hodgkin's disease

PLASMACYTOID LYCTIC: Small lymphocytic diffuse lymphoma with plasmacytoid features

RETICULAR TYPE HD: Reticular type Hodgkin's disease

RHEUMATOID ARTHRITIS: Rheumatoid arthritis

SARCOIDOSIS: Sarcoidosis

SHML: Sinus histiocytosis with massive lymphadenopathy

SINUS HYPERPLASIA: Sinus hyperplasia

SMALL CLEAVED, DIF: Small cleaved diffuse lymphoma

SMALL CLEAVED, FOL: Small cleaved follicular lymphoma

SMALL LYMPHOCYTIC: Small lymphocytic lymphoma

SMALL NONCLEAVED DIF: Small noncleaved diffuse lymphoma

SMALL NONCLEAVED FOL: Small noncleaved follicular lymphoma

SYNCYTIAL NSHD: Syncytial nodular sclerosis Hodgkin's disease

SYPHILIS: Syphilis

T-IMMUNOB LRG: Peripheral T-cell lymphoma, large-cell type

T-IMMUNOB MIX: Peripheral T-cell lymphoma, mixed-cell type

TOXOPLASMOSIS: Toxoplasmosis

TRUE HISTIOCYTIC: True histiocytic lymphoma

TUBERCULOSIS: Tuberculosis

VIRAL NOS: Viral lymphadenitis, not otherwise specified

WHIPPLE'S DISEASE: Whipple's disease

## C.2 Features of the Lymph Node

ABR T-CELL PHENO: Abberrant T-cell phenotype in medium-sized or large lymphoid cells

ACID FAST STAIN: Acid fast stain

B GENE REARRANGEMENT: Immunoglobulin gene rearrangement

BNG HIST: Benign histiocytes not otherwise specified in the nonfollicular areas

BNG HIST FOAMY: Foamy benign histiocytes in the nonfollicular areas that do not contribute to mottling

BNG HIST LANGERHANS: Langerhans benign histiocytes in the nonfollicular areas

BNG HIST SS: Starry-sky benign histiocytes in the nonfollicular areas

CAP THICKENING: Capsule thickening (number of lymphocytes thick)

CARCINOMA CELLS: Carcinoma cells

CLASSIC SR: Classic Sternberg–Reed cells (number per 4-square-centimeter

DIL VASC SP: Vascular spaces dilated by red blood cells

EMPERIPOLESIS: Number of histiocytes showing emperipolesis

EOSIN MICROAB: Eosinophil microabscessess

EOSIN MYELO&META: Eosinophilic myelocytes and metamyelocytes

EOSINOPHILS: Eosinophils (not in microabcesses)

EPI HIST CLUS: Epithelioid histiocyte clusters

EPI HIST CLUS FOL EN: Epitheliod histiocyte clusters encroaching and/or within follicles

EPI HIST NONCLUSTERS: Epitheliod histiocyte nonclusters (percent of total cell population)

EXTRAVASC CLUS CLR C: Extravascular clusters of clear lymphoid cells

F % AREA: Percent area occupied by follicles

F CC CYTOLOGY: Cytology of follicular center cells in most follicles

F CENTERS ATROPHIC: Atrophic centers in any follicles

F CYTOLOGY COMP: Similar cells inside and outside of most follicles

F DEFINITION: Definition of follicles

F DENSITY: Follicle density

F HEMORRHAGES: Hemmorrhages in any of the follicles

F LYMPH INFIL: Lymphocyte infiltration of any follicles

F MANTLE ZONES: Follicle mantle zones in any follicles

F MIT FIGURES: Follicle mitotic figures in 10 high-power fields

F MZ CONCENTRIC RIMS: Mantle zone concentric rims in any follicles

F MZ STATUS: Follicle mantle zones

F POLARITY: Prominent polarity in any follicle

F RADIALLY PEN BV: Number of follicles showing radially penetrating blood vessels

F SS PATTERN: Follicle starry-sky histiocytes (average number in one 10X objective power)

FCB: Fibrocollagenous bands or sclerosis

FCB NODULES: Nodules formed by fibrocollagenous bands

FIBROSIS: Prominent fibrosis

FITE STAIN: Fite stain

FOLLICLES: Follicles

FOREIGN BODY: Foreign body (number in 4-square-centimeter section)

HAIRY CELLS: Hairy cells

HTLV I: HTLV I antibody test

HTLV III: HTLV III antibody test

INTRAVASC CLUS LYMPH: Intravascular clusters of lymphoid cells

KARYORRHEXIS: Karyorrhexis

L&H NODULES: Lymphocytic and hitiocytic nodules

L&H SR: Lymphocytic and hitiocytic variants of Sternberg–Reed cells (number in 4-square-centimeter section)

LACUNAR SR: Lacunar variants of Sternberg–Reed cells (number in 4-square-centimeter section)

LANGHANS: Langhans cells (number in $4\text{cm}^2$ section)

LC LYSOZYME: Lysozyme positivity in medium-sized and/or large lymphoid cells

LEUKEMIC CELLS: Leukemic cells

LLC CHROMATIN: Chromatin of most large lymphoid cells

LLC CYTOPLASM: Cytoplasm of most large lymphoid cells

LLC EV CLUS: Large lymphoid cells in extravascular clusters of clear cells

LLC IDENTITY: Identity of most large lymphoid cells

LLC IV CLUS: Large lymphoid cells in intravascular clusters

LLC NUC SHP: Nuclear shape of most large lymphoid cells

LLC NUCLEOLI: Nucleolar features of most large lymphoid cells

LLC NUM: Number of large lymphoid cells in the nonfollicular areas (percent of total cell population)

LLC+MLC > 50%: Number of medium-sized and large lymphoid cells in the nonfollicular areas exceeds 50 percent of total cell population

LRG LMPH CELLS: Large lymphoid cells

MAST CELLS: Mast cells (number in $4\text{cm}^2$ section)

MED LYMPH CELLS: Medium-sized lymphoid cells

MELANOMA CELLS: Melanoma cells

MITOTIC FIG: Mitotic figures in 10 high-power fields (nonfollicular areas)

MLC CHROMATIN: Chromatin structure of most medium-sized lymphoid cells

MLC CYTOPLASM: Cytoplasm of most medium-sized lymphoid cells

MLC EV CLUS: Medium-sized lymphoid cells in extravascular clusters of clear cells

MLC IV CLUS: Medium-sized lymphoid cells in intravascular clusters

MLC NUC SHP: Nuclear shape of most medium-sized lymphoid cells

MLC NUCLEOLI: Nucleolar features of most medium-sized lymphoid cells

MLC NUM: Number of Medium-sized lymphoid cells in the nonfollicular areas (percent of total cell population)

MONOCYT: Monocytoid cells (percent of total cell population)

MONONUCLEAR SR: Mononuclear variants of Sternberg–Reed cells (number in 4-square-centimeter section)

MOTTLING HIST: Mottling by langerhans or other histiocytes

MOTTLING LLC: Mottling by large lymphoid cells

MUMMY: Large mummified cells (number in 4-square-centimeter section)

NECROSIS: Necrosis

NEUTROPHIL MICROABSC: Neutrophil microabcessess

NEUTROPHILS: Neutrophils (not in microabcesses)

NONSIN NONFOL AREAS: Nonsinus nonfollicular areas

PAS STAIN: Strong PAS positivity in the histiocytes

PERICAP INFILTR: Pericapsular infiltration

PLASMA: Plasma cells in the nonfollicular areas (percent of total cell population)

PLASMA TYPE: Plasma cell type

PLEOMORPHIC SR: Pleomorphic variants of Sternberg–Reed cells (number in 4-square-centimeter section)

PSEUDOFOLLICLES: Pseudofollicles

PTGC: Progressively transformed germinal centers

RUSSELL&DUTCHER: Russell and/or Dutcher bodies

SARCOMA CELLS: Sarcoma cells

SCHAUMAN: Schauman cells

SIGNET-RING: Signet-ring cells

SINUSES: Sinuses

SLC CHROMATIN: Chromatin structure of most small lymphoid cells

SLC CYTOPLASM: Cytoplasm of most small lymphoid cells

SLC EV CLUS: Small lymphoid cells in extravascular clusters of clear cells

SLC IV CLUS: Small lymphoid cells in intravascular clusters

SLC NUC SHP: Nuclear shape of most small lymphoid cells

SLC NUM: Number of small lymphoid cells in the nonfollicular areas (percent of total cell population)

SML LYMPH CELLS: Small lymphoid cells

SR-LIKE: Sternberg–Reed-like cells (number in $4cm^2$ section)

SYSTEMIC AIDS: Systemic AIDS

T GENE REARRANGEMENT: T-cell receptor gene rearrangement

TRANSITION FORMS: Transition forms (lymphoid cells having sizes other than the sizes of small, medium-sized, or large cells) in the nonfollicular areas

VASC CHANGES: Endarteritis or periarteritis

VASC PROLIF NONSLIT: Vascular proliferation (nonslitlike)

VASC PROLIF SLIT: Vascular proliferation slitlike

# D Evaluation Results

**Table D.1:** Expert ratings and inferential losses for Pathfinder III and IV.

| | Expert Rating | | Inferential Loss (micromorts) | |
|---|---|---|---|---|
| Case | Pathfinder III | Pathfinder IV | Pathfinder III | Pathfinder IV |
| bnn1001 | 7.0 | 7.0 | 0 | 0 |
| bnn1002 | 10.0 | 10.0 | 0 | 0 |
| bnn1003 | 10.0 | 10.0 | 0 | 0 |
| bnn1005 | 1.5 | 9.5 | 10 | 0 |
| bnn1008 | 9.5 | 8.5 | 0 | 0 |
| bnn1013 | 9.5 | 9.8 | 0 | 0 |
| bnn1015 | 9.0 | 9.0 | 0 | 0 |
| bnn1016 | 8.0 | 4.0 | 0 | 0 |
| bnn1018 | 8.0 | 7.0 | 0 | 70 |
| bnn1021 | 9.5 | 9.5 | 0 | 0 |
| bnn1022 | 9.0 | 10.0 | 0 | 0 |
| bnn1024 | 8.5 | 10.0 | 3269 | 0 |
| bnn1027 | 8.5 | 9.0 | 0 | 0 |
| bnn1028 | 9.0 | 9.0 | 0 | 0 |
| bnn1030 | 9.5 | 10.0 | 0 | 0 |
| bnn1034 | 9.5 | 9.5 | 0 | 0 |
| bnn1044 | 9.5 | 9.5 | 0 | 0 |
| bnn1050 | 9.0 | 8.0 | 0 | 0 |
| bnn1055 | 8.5 | 9.5 | 0 | 0 |
| bnn1059 | 8.0 | 4.0 | 0 | 0 |
| bnn1063 | 10.0 | 10.0 | 0 | 0 |
| bnn1065 | 6.0 | 6.0 | 0 | 0 |
| bnn1067 | 10.0 | 10.0 | 0 | 0 |
| bnn1070 | 9.0 | 9.0 | 0 | 0 |
| bnn1071 | 9.5 | 9.5 | 0 | 0 |
| bnn1072 | 1.0 | 10.0 | 0 | 0 |
| bnn1078 | 10.0 | 4.0 | 0 | 0 |
| bnn1079 | 4.5 | 9.5 | 106 | 0 |

**Table D.2:** Expert ratings and inferential losses continued.

| Case | Expert Rating | | Inferential Loss (micromorts) | |
|---|---|---|---|---|
| | Pathfinder III | Pathfinder IV | Pathfinder III | Pathfinder IV |
| bnn1084 | 6.8 | 10.0 | 3269 | 0 |
| bnn1087 | 8.0 | 9.5 | 0 | 0 |
| bnn1088 | 10.0 | 10.0 | 0 | 0 |
| bnn1091 | 8.5 | 9.0 | 0 | 0 |
| bnn1095 | 9.0 | 10.0 | 0 | 0 |
| bnn1098 | 9.9 | 9.9 | 0 | 0 |
| bnn1102 | 5.0 | 9.5 | 11358 | 0 |
| bnn1106 | 10.0 | 10.0 | 0 | 0 |
| bnn1109 | 7.0 | 9.0 | 0 | 0 |
| bnn1111 | 10.0 | 10.0 | 0 | 0 |
| bnn1112 | 5.0 | 8.0 | 0 | 0 |
| bnn1113 | 4.0 | 8.0 | 0 | 752 |
| bnn1119 | 8.0 | 9.0 | 0 | 0 |
| bnn1123 | 10.0 | 10.0 | 0 | 0 |
| bnn1126 | 2.0 | 9.5 | 7 | 0 |
| bnn1129 | 5.0 | 8.5 | 0 | 0 |
| bnn1139 | 9.5 | 9.5 | 0 | 0 |
| bnn1152 | 10.0 | 10.0 | 0 | 0 |
| bnn1154 | 8.0 | 9.0 | 0 | 0 |
| bnn1159 | 5.8 | 9.8 | 0 | 0 |
| bnn1162 | 6.5 | 7.0 | 0 | 0 |
| bnn1165 | 10.0 | 10.0 | 0 | 0 |
| bnn1167 | 7.5 | 9.0 | 0 | 7 |
| bnn1171 | 9.5 | 10.0 | 0 | 0 |
| bnn1172 | 8.5 | 9.0 | 0 | 0 |

# E Transformation for Multiple Hypotheses

In this appendix, we examine an algorithm for transforming a similarity network $\mathcal{S}$ into a multiple-hypothesis knowledge map $\mathcal{M}$. Let $h_0, h_1, h_2, \ldots, h_n$ denote the instances of the distinguished variable $h$ in $\mathcal{S}$, where $h_0$ is NORMAL. In addition, let $h^i, i = 1, 2, \ldots, n$, denote the binary variable in $\mathcal{M}$ that corresponds to hypothesis $h_i$ in $\mathcal{S}$. Also, let $x$ and $y$ denote arbitrary nondistinguished variables in $\mathcal{S}$. Finally, let $\mathcal{S}'$ denote a similarity network constructed from $\mathcal{S}$ such that the similarity graph of $\mathcal{S}'$ has a star topology with center $h_0$, and let $\mathcal{G}_{\mathcal{S}'}$ denote the global knowledge map constructed from $\mathcal{S}'$. We transform $\mathcal{S}$ into $\mathcal{M}$ as follows:

Construct $\mathcal{S}'$ from $\mathcal{S}$
Construct $\mathcal{G}_{\mathcal{S}'}$ from $\mathcal{S}'$
For all $x$ in $\mathcal{G}_{\mathcal{S}'}$, place $x$ in $\mathcal{M}$
For all $x \longrightarrow y$ in $\mathcal{G}_{\mathcal{S}'}$, place $x \longrightarrow y$ in $\mathcal{M}$
For $h_i$ in $\mathcal{S}'$, $i = 1, 2, \ldots, n$, place $h^i$ in $\mathcal{M}$
For local knowledge map $h_{0i}$ in $\mathcal{S}'$, $i = 1, 2, \ldots, n$,
    For all $x$ in $\mathcal{G}_{\mathcal{S}'}$, if $x$ is in $h_{0i}$, then place $h^i \longrightarrow x$ in $\mathcal{M}$
For all $x$,
    Determine the probability distributions for $x$ in $\mathcal{M}$ from the distributions for $x$ in $\mathcal{G}_{\mathcal{S}'}$, using assertions of causal independence
Assess dependencies among the $h^i$ in $\mathcal{M}$
For $h^i$ in $\mathcal{M}$, $i = 1, 2, \ldots, n$, assess the probability distributions for $h^i$

Note that, in this transformation, the nondistinguished variables do not need to be binary, provided we generalize beyond the noisy OR-gate model expressed by Equation 6.1.4. In addition, we can generalize this algorithm to include situations where $h^i$ is nonbinary, by representing each instance of $h^i$ in the similarity graph, and by identifying forms of causal independence that can account for the interaction between $h^i$ and each nondistinguished node.

# Index

**V**

**W**

# Notation

| | |
|---|---|
| $x, y, \ldots$ | Uncertain variables or their corresponding nodes in a knowledge map |
| $X, Y, \ldots$ | Sets of variables or corresponding sets of nodes |
| $x_i$ | The $i$th instance of variable $x$ |
| $x_+, x_-$ | The two instances of the binary variable $x$ |
| $x_\subseteq$ | A subset of the instances of the variable $x$ |
| $X_i$ | The $i$th instance of set $X$ |
| $X \setminus Y$ | The variables in $X$ that are not in $Y$ |
| $C(x)$ | The conditional or direct predecessors of node $x$ |
| $S(x)$ | The successors of node $x$ |
| $\bar{S}(x)$ | The nonsuccessors of node $x$ |
| $p\left(x_i \mid X_j, \xi\right)$ | The probability of $x_i$ given $X_j$ and background knowledge $\xi$ |
| $p\left(x_k \mid \{h_i, h_j\}, \xi\right)$ | The probability of $x_k$ given background knowledge $\xi$ and the knowledge that either $h_i$ or $h_j$ is true |
| $\leq_C$ | A partial ordering on nodes in a knowledge map defined by $x \leq_C y$ if and only if there is a directed path from $x$ to $y$ |
| $\leq_E$ | A total ordering on the nodes in a knowledge map used to expand the joint distribution |
| $x \longrightarrow y$ | Abbreviation for "arc from $x$ to $y$" |
| $x \not\longrightarrow y$ | Abbreviation for "no arc from $x$ to $y$" |
| $\mathcal{HS}, \mathcal{C}, \mathcal{O}$ | A hypothesis-specific, comprehensive, and ordinary similarity network |
| $\mathcal{G}_c$ | A comprehensive global knowledge map (c-global map) |
| $\mathcal{G}_o$ | An ordinary global knowledge map (o-global map) |
| $\widehat{h_i}$ | The hypothesis-specific knowledge map (hs map) associated with $h_i$ |
| $(h_i, h_j)$ | The edge between $h_i$ and $h_j$ in a similarity graph |
| $h_i – h_j$ | The comprehensive local knowledge map (c-local map) or ordinary local knowledge map (o-local map) associated with the edge $(h_i, h_j)$ in a similarity graph |
| $C^i(x)$ | The conditional predecessors of node $x$ in the hs map $\widehat{h_i}$ |
| $C^{ij}(x)$ | The conditional predecessors of node $x$ in the c-local or o-local map $h_i – h_j$ |
| $C^{\mathcal{G}_c}(x)$ | The conditional predecessors of node $x$ in a c-global map |
| $\mathcal{R}^{ij}$ | The relevance set for $h_i$ and $h_j$ in a hypothesis-specific similarity network |
| $\mathcal{HS} \longmapsto \mathcal{C}$ | The construction of $\mathcal{C}$ from $\mathcal{HS}$ |
| $\mathcal{HS} \models_{\mathcal{P}} \mathcal{C}$ | Abbreviation for "$\mathcal{HS}$ logically implies $\mathcal{C}$" |